\RequirePackage{fix-cm}
\documentclass[smallextended]{svjour3}       

\usepackage{graphicx}
\usepackage{amssymb}
\usepackage{subcaption}
\graphicspath{{pics/}}
\usepackage{soul}
\usepackage{xcolor}
\usepackage{bm}
\usepackage{amsmath}
\usepackage[linesnumbered,ruled,vlined]{algorithm2e}
\usepackage[export]{adjustbox}
\usepackage{tabularx}
\usepackage[section]{placeins}

\usepackage[utf8]{inputenc}

\smartqed  
\begin{document}

\title{Traversing the noise of dynamic mini-batch sub-sampled loss functions: A visual guide
}



\author{Dominic Kakfa         \and
        Daniel N. Wilke 
}


\institute{Dominic Kafka \at
              Centre for Asset and Integrity Management (C-AIM), \\
              Department of Mechanical and Aeronautical Engineering, \\
              University of Pretoria, Pretoria, South Africa. \\
              \email{dominic.kafka@gmail.com}           
           \and
              Daniel N. Wilke \at
              Centre for Asset and Integrity Management (C-AIM), \\
			  Department of Mechanical and Aeronautical Engineering, \\
			  University of Pretoria, Pretoria, South Africa. \\
			  \email{wilkedn@gmail.com}           
}

\date{Received: date / Accepted: date}

\maketitle

\begin{abstract}
Mini-batch sub-sampling in neural network training is unavoidable, due to growing data demands, memory-limited computational resources such as graphical processing units (GPUs), and the dynamics of on-line learning. In this study we specifically distinguish between static mini-batch sub-sampled loss functions, where mini-batches are intermittently fixed during training, resulting in smooth but biased loss functions; and the dynamic sub-sampling equivalent, where new mini-batches are sampled at every loss evaluation, trading bias for variance in sampling induced discontinuities. These render automated optimization strategies such as minimization line searches ineffective, since critical points may not exist and function minimizers find spurious, discontinuity induced minima. 

This paper suggests recasting the optimization problem to find stochastic non-negative associated gradient projection points (SNN-GPPs). We demonstrate that the SNN-GPP optimality criterion is less susceptible to sub-sampling induced discontinuities than critical points or minimizers.
We conduct a visual investigation, comparing local minimum and SNN-GPP optimality criteria in the loss functions of a simple neural network training problem for a variety of popular activation functions. Since SNN-GPPs better approximate the location of true optima, particularly when using smooth activation functions with high curvature characteristics, we postulate that line searches locating SNN-GPPs can contribute significantly to automating neural network training.

\end{abstract}

\section{Introduction: The stochastic neural network optimization problem}

The training of neural networks centres around minimizing loss functions that commonly take the form of
\begin{eqnarray}
\mathcal{L}(\boldsymbol{x}) = \frac{1}{M} \sum_{b=1}^{M} \ell (\boldsymbol{x};\;\boldsymbol{t}_b),
\label{eq:loss}
\end{eqnarray}

where $T = \{\boldsymbol{t}_1,\dots,\boldsymbol{t}_M	\}$ is the training dataset of $M$ samples, and the model parameters are given by vector $\boldsymbol{x}\in \mathbb{R}^{p}$. The loss quantifying the adequacy of parameters $\boldsymbol{x}$ in terms of training data $T$ is given by $\mathcal{L}(\boldsymbol{x})$. The gradient of the  loss function with regards to parameters $\boldsymbol{x}$ is given by
\begin{eqnarray}
\nabla\mathcal{L}(\boldsymbol{x}) = \frac{1}{M} \sum_{b=1}^{M} \nabla\ell (\boldsymbol{x};\;\boldsymbol{t}_b),
\label{eq:lossgrad}
\end{eqnarray}
which is computed efficiently using backpropagation \cite{Werbos1994}. The loss surface of $ \mathcal{L}(\boldsymbol{x}) $ is continuous, while the continuity of $ \nabla\mathcal{L}(\boldsymbol{x}) $ depends on the continuity and smoothness of the activation function (AF) used.

\begin{figure}[h]
	\centering
	\begin{subfigure}{.65\textwidth}
		\centering 
		\includegraphics[width=0.99\linewidth]{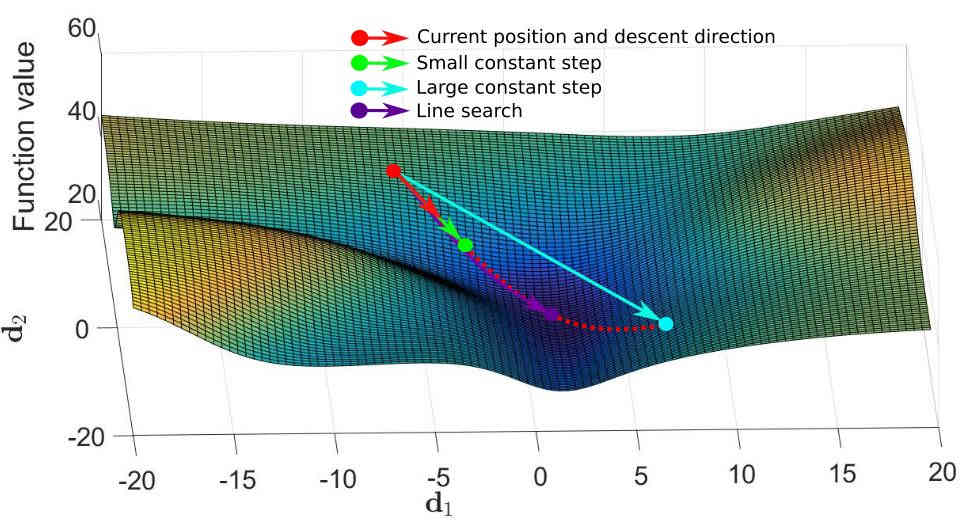}
		\caption{GD update in a neural network loss function.}
	\end{subfigure}%
	
	\begin{subfigure}{.6\textwidth}
		\centering 
		\includegraphics[width=0.99\linewidth]{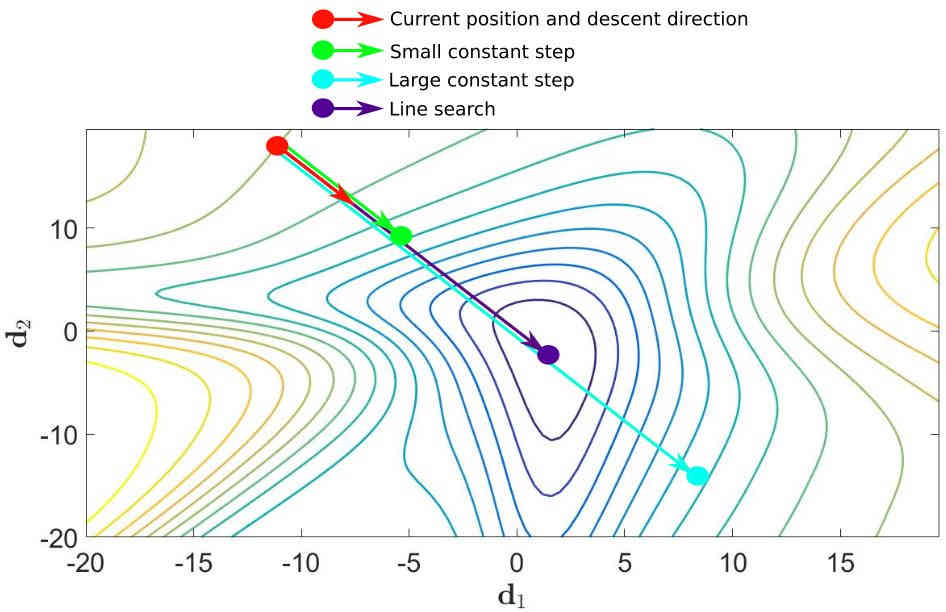}
		\caption{Contour plot of an GD update step.}
	\end{subfigure}%
	\caption{Potential gradient descent (GD) updates in a neural network loss function for the Iris \cite{Fisher1936} classification problem. }
	\label{fig_LScontext}
\end{figure}

A local minimum of a function $f(\boldsymbol{x})$ is defined as follows: 
\begin{definition}{Local Minimum:}
	Let $\boldsymbol{x}^*$ be a local minimum of $f(\boldsymbol{x})$, such that
	\begin{equation}
	\Delta  f(\boldsymbol{x}) = f(\boldsymbol{x}) - f(\boldsymbol{x}^*) \geq 0,
	\end{equation}
	for any point $ \boldsymbol{x} $ in the neighbourhood of $ \boldsymbol{x}^*$ \cite{Arora2011}.  
	\label{def_locmin}
\end{definition}

In order to find minima, consider the update step of the popular Gradient Descent (GD) algorithm, given as
\begin{equation}
\boldsymbol{x}_{n+1} = \boldsymbol{x}_{n} - \alpha \nabla \mathcal{L}(\boldsymbol{x}_n),
\label{eq_GD}
\end{equation}
at a given iteration, $n$. Here the update is a function of a search direction $\boldsymbol{d}_n = - \nabla  \mathcal{L}(\boldsymbol{x}_n)$, the gradient of the loss function at the current location, and an undetermined step size, scalar $\alpha$. A visual representation of selecting the step size along a gradient descent search direction is depicted in Figure~\ref{fig_LScontext}. If $\alpha$ is too small, insufficient progress is made along the descent direction, causing slow neural network training. If $\alpha$ is too large, the minimum along the search direction is overshot, and training can become unstable. Line searches are common methods employed in mathematical programming \cite{Nocedal1999,Wachter2005,Nie2006,Arora2011} to resolve the step size, $\alpha$, up to a desired accuracy, balancing training performance and stability. Importantly, line searches perform best when the full dataset is available to evaluate the loss function.

However, nowadays machine learning (ML) training is rarely conducted using full batches \cite{Krizhevsky2012,Csiba2018}. Growing data demands, memory limited efficient computational resources, such as graphical processing units (GPUs), the dynamic world of on-line learning \cite{Mahsereci2017a} and improved convergence characteristics  \cite{Saxe2013,Dauphin2014,Choromanska2015}, has cemented mini-batch sub-sampling as a {\emph de facto} standard in ML training. In particular, the use of a smaller number of samples,  i.e.  $\mathcal{B} \subset \{1,\dots,M\}$ with $|\mathcal{B}| \ll M$, have become common practice. The approximated loss and gradient functions are given by:

\begin{equation}
L(\boldsymbol{x}) = \frac{1}{|\mathcal{B}|} \sum_{b\in \mathcal{B}} \ell (\boldsymbol{x};\;\boldsymbol{t}_b),
\label{eq:lossgradbatch}
\end{equation}
and
\begin{equation}
\boldsymbol{g}(\boldsymbol{x}) = \frac{1}{|\mathcal{B}|} \sum_{b\in \mathcal{B}} \nabla\ell (\boldsymbol{x};\;\boldsymbol{t}_b).
\label{eq:g_lossgradbatch}
\end{equation}


For most sampling approaches, these approximations have expectation $ \mathbb{E} [ L(\boldsymbol{x}) ] = \mathcal{L}(\boldsymbol{x}) $  and $ \mathbb{E} [ \boldsymbol{g}(\boldsymbol{x}) ] = \nabla\mathcal{L}(\boldsymbol{x})$ \cite{Tong2005}. However, significant variations from the mean can be observed between expressions of individual batches, $\mathcal{B}$, which we call the {\it sampling error}. This study distinguishes between two mini-batch sub-sampling (MBSS) approaches  to compute a loss function, namely static and dynamic MBSS. These two MBSS approaches significantly affect the characteristics of the various computed loss functions. This study aims to highlight and demonstrate these implications, in addition to developing an intuition for interpreting the available information within the context of line search approaches to resolve learning rates.

Before we explore static and dynamic MBSS, it is important to explore adaptive sub-sampling methods, that primarily aims to resolve batches or batch sizes with desired characteristics. The aim might be to select a sub-sample such that $\boldsymbol{g}(\boldsymbol{x}) \approx \nabla   \mathcal{L}(\boldsymbol{x})$, or to ensure that descent directions computed by $\boldsymbol{d}_n = -\boldsymbol{g}(\boldsymbol{x})$ are indeed mostly descent directions \cite{Friedlander2011,Bollapragada2017}. A carefully selected mini-batch is usually kept constant over the minimum duration of a line search \cite{Martens2010,Friedlander2011,Byrd2011,Byrd2012,Bollapragada2017,Kungurtsev2018,Paquette2018,Bergou2018}, which can then also be used to conduct line searches to resolve learning rates. We call this or any other approach that keeps a mini-batch fixed along a search direction, $\boldsymbol{d}_n$, static MBSS. In this study, we denote loss and gradient approximations computed using static MBSS by $\bar{L}(\boldsymbol{x})$ and $\bar{\boldsymbol{g}}(\boldsymbol{x})$, respectively. Mini-batches sampled for static MBSS are denoted as $\mathcal{B}_n$.

\begin{figure}
	\centering
	\begin{subfigure}{.5\textwidth}
		\centering 
		\includegraphics[width=0.9\linewidth]{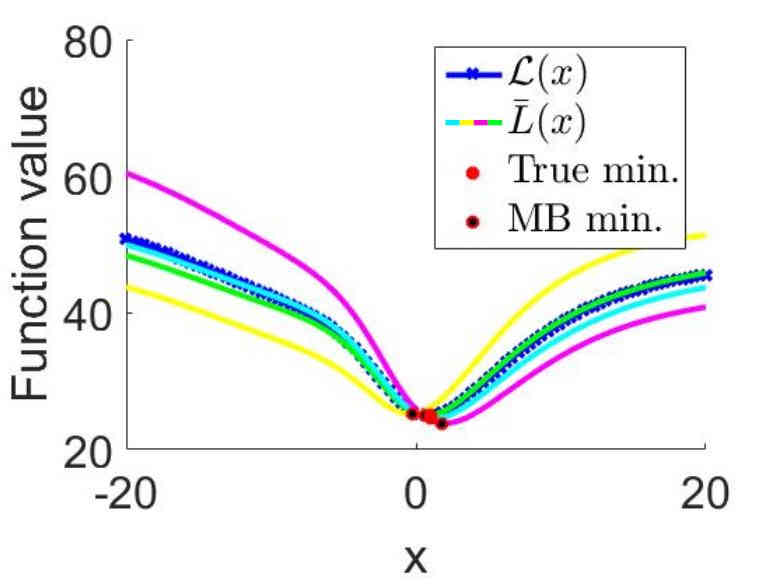}
		\caption{$\mathcal{L}(x)$ and 4 $\bar{L}(x)$ curves}
	\end{subfigure}%
	\begin{subfigure}{.5\textwidth}
		\centering 
		\includegraphics[width=0.9\linewidth]{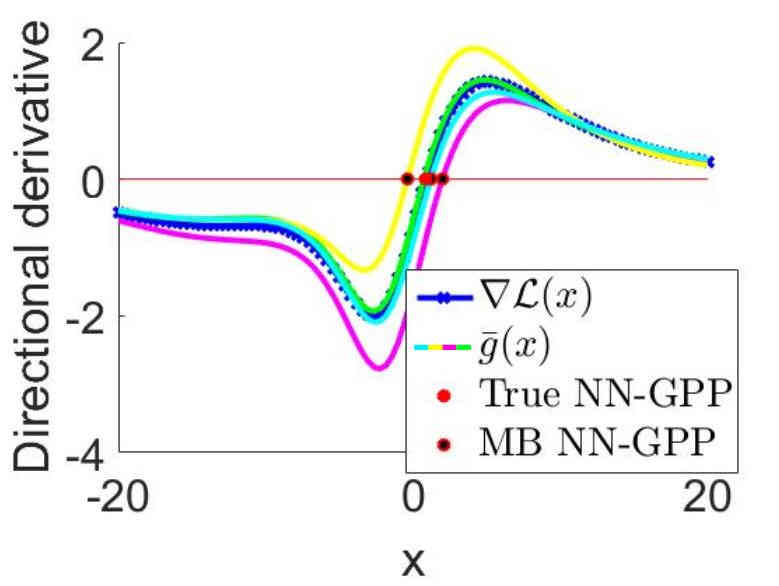}
		\caption{$\nabla \mathcal{L}(x)$ and 4 $\bar{g}(x)$ curves}
	\end{subfigure}%
	
	\begin{subfigure}{.5\textwidth}
		\centering
		\includegraphics[width=0.9\linewidth]{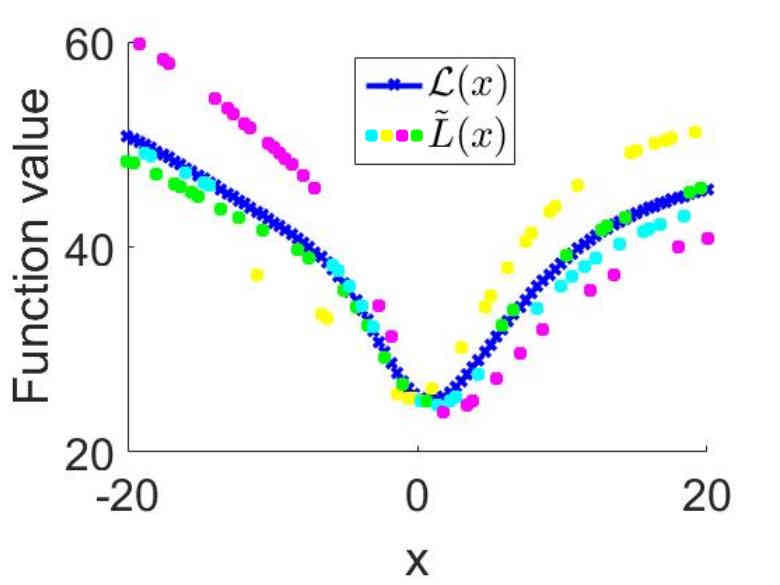}
		\caption{$\tilde{L}(x)$, with selected batches colored}
	\end{subfigure}%
	\begin{subfigure}{.5\textwidth}
		\centering
		\includegraphics[width=0.9\linewidth]{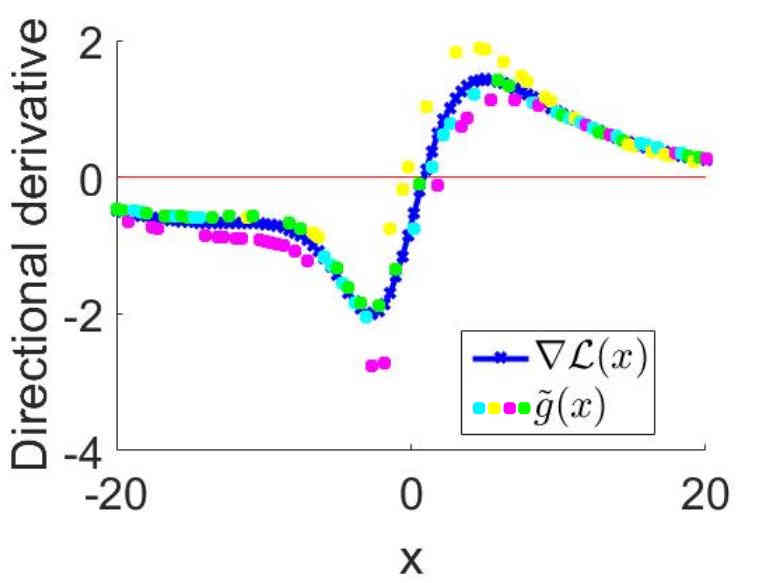}
		\caption{$\tilde{g}(x)$, with selected batches colored}
	\end{subfigure}%
	
	\begin{subfigure}{.5\textwidth}
		\centering
		\includegraphics[width=0.9\linewidth]{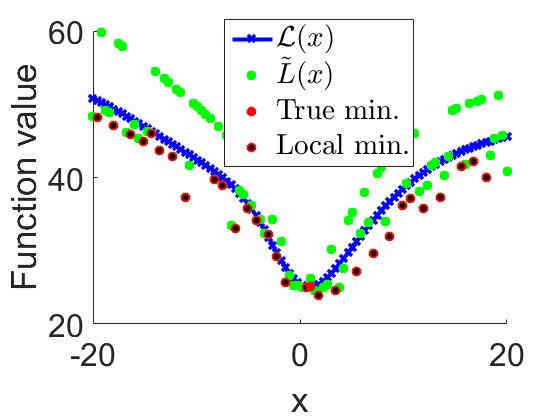}
		\caption{$\tilde{L}(x)$ as a single function}
	\end{subfigure}%
	\begin{subfigure}{.5\textwidth}
		\centering
		\includegraphics[width=0.9\linewidth]{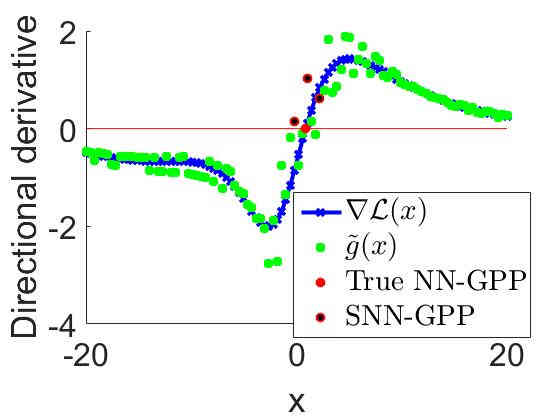}
		\caption{$\tilde{g}(x)$ as a single function}
	\end{subfigure}%
	\caption{Conceptual depiction of static and dynamic mini-batch sub-sampled loss functions. (a,b) The full batch expression (blue) is the average of 4 different batches (green, magenta, cyan, yellow), each construct own expressions of the loss function and derivative with different characteristics. (c,d) When sampling randomly between these 4 batches, the function value and corresponding derivative alternates between mini-batch loss function expressions. (e,f) The resulting loss function is discontinuous and non-smooth for function values and derivatives, which is readily interpreted as noise.}
	\label{fig_statVdyn1D}
\end{figure}

A 1-D loss function representation is illustrated in Figures~\ref{fig_statVdyn1D}(a) and (b). The training data is split into 4 equally sized, static mini-batches (green, magenta, cyan, yellow), each resulting in a continuous and smooth loss expression,  $\bar{L}(\boldsymbol{x})$, with own minimizer, $\boldsymbol{x}^{*\mathcal{B}_n}$, and associated local minimum. The blue curve denotes the true or full-batch loss, $\mathcal{L}(\boldsymbol{x})$. Evidently, the mini-batch minimizers, $\boldsymbol{x}^{*\mathcal{B}_n}$ denoted by the prefix "MB", are not equal to, but occur in a range around the true or full-batch minimizer, $\boldsymbol{x}^{*M}$ denoted by the prefix "True".

Suppose now that a new loss function is constructed by an oracle \cite{Agarwal2012}, that randomly selects one of the four batches for every increment, $i$, that the loss function or derivative is evaluated along $x$ as shown in Figures~\ref{fig_statVdyn1D}(c) and (d). We call this dynamic MBSS and differentiate loss function and gradient evaluations using this scheme by $\tilde{L}(\boldsymbol{x})$ and $\tilde{\boldsymbol{g}}(\boldsymbol{x})$, respectively. Mini-batches sampled using dynamic MBSS are denoted $\mathbf{B}_{n,i}$, as re-sampling occurs at every increment, $i$, within iteration, $n$, of a potential optimization algorithm. This results in a discontinuous loss function as the oracle randomly selects a loss function sampled by one of the four mini-batches. We show this explicitly in Figure~\ref{fig_statVdyn1D}(c) and (d), by retaining the colours of the contributing mini-batches towards the dynamic MBSS loss function. Here, the contribution of each batch is evident, whereas, plotting the computed dynamic MBSS loss function using a single color hinders identification of the contribution of a single batch, as shown in Figures~\ref{fig_statVdyn1D}(e) and (f). Note, that there are a finite number of combinations in which the discontinuities can occur for a dataset of fixed size. The number of combinations, in Figures~\ref{fig_statVdyn1D}(e) and (f), were limited to four, but sampling conducted with replacement, there are $K = \binom{M}{|\mathcal{B}_{n,i}|}$ combinations in which a mini-batch $\mathcal{B}_{n,i}$ can be assembled from the training set. Consequently, the number of combinations explode for large $M$ and small batch sizes, which results in these discontinuities being interpreted readily as stochastic noise \cite{Simsekli2019a}. Not surprisingly, these discontinuities hinder attempts to directly minimize dynamic MBSS loss functions using line searches. This is due to sampling induced discontinuities that manifest as local minimizers. Concurrently, critical points typically do not exist at individual samples $\tilde{L}(\boldsymbol{x}^*)$, although they might be present in expectation. This renders ineffective the predominant mathematical programming paradigm for defining optimal as the minimum loss function. 

In spite of this, Bottou \cite{Bottou2010} argues that approximate optimization methods, which implement dynamic MBSS, can benefit from being exposed to more information. In a recent study, the probabilistic line search developed by \cite{Mahsereci2017a} estimates step sizes for dynamic MBSS loss functions. Benefits of changing the mini-batch sub-sample at every evaluation include overcoming weaker local basins of attraction \cite{Kleinberg2018,Simsekli2019a}, in addition to acting as an effective regularizer to obtain solutions with better generalization properties \cite{Masters2018}. This motivated the use of adaptive and dynamic (recursive) sampling \cite{KASHYAP1970,Csiba2018}, as alluded to earlier, to reduce the impact of these discontinuities. 
Alternatively, subgradient methods \cite{Shor1985a}, originally developed for continuous non-smooth loss functions, utilizes fixed learning rates or fixed learning rate schedules. Considering a simple modification \cite{Wilke2011}, they are also able to optimize discontinuous loss functions, as has been done for years in ML training, albeit often unwittingly. In fact, Wilke \cite{Wilke2011} demonstrated that subgradient methods do not minimize discontinuous loss functions but rather resolve non-negative associated gradient projection points (NN-GPPs). NN-GPP were 
specifically designed to define optimality for discontinuous functions \cite{Wilke2013,Snyman2018}, and is given by:

\begin{definition}{NN-GPP:}
	A non-negative associated gradient projection point (NN-GPP) is defined as any point, $\boldsymbol{x}_{nngpp}$, for which there exists
	$\epsilon_{max} > 0$ such that 
	\begin{equation}
	\boldsymbol{u}\cdot \nabla f(\boldsymbol{x}_{nngpp} + \epsilon \boldsymbol{u} ) \geq 0,\;\;  \forall\; \boldsymbol{u} \in \left\{\boldsymbol{y}\in\mathbb{R}^p \;|\; \| \boldsymbol{y} \|_2 = 1\right\},\;\;\forall \;\epsilon \in (0,\epsilon_{max}].
	\end{equation}	
	\label{def_nngpp}
\end{definition}
This defines optimality solely based on the gradient of the loss function, without requiring the gradient to be zero at the solution and neither requiring additional second order information to be computed to ensure a local minimum solution.

NN-GPPs have been formalized on a rigorous mathematical foundation under gradient-only optimization, and proven to be equivalent to semi-positive definite local minimizers for twice continuously differentiable smooth functions \cite{Wilke2013}. Returning to our example in Figures~\ref{fig_statVdyn1D}(a) and (b), this means that both minimization and finding NN-GPPs can be applied to the full-batch and static MBSS loss functions, where they resolve exactly the same points for unimodal functions. A whole range of gradient-only line search optimizers to locate NN-GPP have been proposed \cite{Wilke2011,Wilke2013,Snyman2018}. In 1-D, to locate a NN-GPP along a descent direction, merely requires a sign change in the directional derivative from negative to positive to be identified. This study recognizes that gradient-only optimization may be effective in overcoming the difficulties associated with discontinuities induced by dynamic MBSS. 

%
%
%

However, when dynamic MBSS is employed to compute a loss function, minimization and gradient-only optimization may identify distinctly different solutions, although both definitions define the same optimal solution or true solution, $\boldsymbol{x}^{*M}$ when a unimodal full-batch sampled loss function is considered. Minimization may resolve sampling induced discontinuities local minima, whereas gradient-only optimization will largely ignore these sampling induced discontinuities unless they manifest as a sign change along a search direction. For dynamic MBSS loss functions, NN-GPP may appear and disappear stochastically in the vicinity of the true solution, $\boldsymbol{x}^{*M}$. Hence, sign changes in the directional derivative along a search direction may appear and disappear stochastically as the oracle updates the mini-batches. It is important to note that in addition to the sign change of each mini-batch loss function, additional sampling induced sign changes may manifest along a search direction. This occurs when the oracle switches between a negative and positive directional derivative for essentially the same step along a search direction. Similarly, in addition to the NN-GPP for each mini-batch loss function, additional sampling induced NN-GPP may appear and disappear in the vicinity of $\boldsymbol{x}^{*M}$. To accommodate these stochastic sampling induced NN-GPPs, the gradient-only optimality criterion has been extended to the Stochastic NN-GPP (SNN-GPP) as follows:

\begin{definition}{SNN-GPP:}
	A stochastic non-negative associated gradient projection point (SNN-GPP) is defined as any point, $\boldsymbol{x}_{snngpp}$, for which there exists
	$\epsilon_{max} > 0$ such that 
	\begin{equation}
	\boldsymbol{u}\cdot \boldsymbol{g}(\boldsymbol{x}_{snngpp} + \epsilon \boldsymbol{u} ) \geq 0,\;\;  \forall\; \boldsymbol{u} \in \left\{\boldsymbol{y}\in\mathbb{R}^p \;|\; \| \boldsymbol{y} \|_2 = 1\right\},\;\;\forall \;\epsilon \in (0,\epsilon_{max}],
	\end{equation}
	with probability greater than 0. \cite{Kafka2019jogo}
	\label{def_snngpp}
\end{definition}

There is a considerable difference in the quality of information presented in terms of proximity to true optimum when all minimizers are compared to all SNN-GPPs. Consider the red points in Figure~\ref{fig_statVdyn1D}, where we identify local minima using Definition \ref{def_locmin} in Figure~\ref{fig_statVdyn1D}(e), and SNN-GPPs by stepping over a directional derivative sign change from negative to positive. We choose the convention of highlighting the positive directional derivative at the sign change as the SNN-GPP, as shown in Figure~\ref{fig_statVdyn1D}(f).  Minimizers are largely unbounded along $x$, while SNN-GPPs are much more localized around the true solution ${x}^{*M}$. The majority of spurious minimizers in Figure~\ref{fig_statVdyn1D}(e), induced by dynamic MBSS, are completely ignored in Figure~\ref{fig_statVdyn1D}(f). In a recent study, a simple gradient-only line search was implemented to estimate the location of SNN-GPPs along a search direction \cite{Kafka2019}. This proved effective in automatically resolving the learning rate, and yielded competitive results when compared to probabilistic line searches \cite{Mahsereci2017a}. However, the aim of this study is not to propose yet another algorithm, but to rather gain insight into the implication of activation functions on the nature of the loss and gradient functions, which ultimately translates to an understanding of the potential for finding SNN-GPP as opposed to minimizers in ML training. 

Although many studies have been done on neural network loss functions in either a theoretical \cite{Nguyen2018,Liang2018} or visual context \cite{Goodfellow2015,Li2017,Im2016}, most studies concentrate on the convexity properties of the expected loss function or true loss function. Given that almost all ML optimizers ignore the loss function to rather flow with the gradients \cite{Robbins1951,Kingma2015,Duchi2011}, it is sensible to rather investigate the characteristics and properties of the gradient field or directional derivatives, as well as how they are influenced by activation functions (AFs) and dynamic MBSS sampling. In our study, we explicitly compare static and dynamic MBSS in Section \ref{sec_apply2sig}, and subsequently choose to focus our study on the affects of dynamic MBSS. To the best of our knowledge this is the first study to visually explore the qualitative characteristics of gradient or directional derivative information in dynamic MBSS loss functions of neural networks. We choose a visual approach to investigate the qualitative characteristics of function value and gradient information in dynamic MBSS, such that an intuitive understanding of the relationship between minima, SNN-GPPs and activation functions \cite{Serwa2017,Karlik2015,Laudani2015} can be developed. As we visually explore the characteristics of function values and directional derivatives for different activation functions under dynamic MBSS, it is important to continuously reassess the following:
{\it What is the quality of function value versus directional derivative information with regards to making informed decisions on candidate solutions in mini-batched sampled neural network training?}

\subsection{Our contribution}

The contribution of this work lies neither in the dataset, nor the network architecture we use. This paper is written from the perspective where an optimization problem has two key components, namely a loss function landscape, and an algorithm that searches for an optimum within this landscape. To take a step towards an optimum, many optimization algorithms conduct line searches, i.e. find a univariate optimum along a search direction. As discussed, mini-batch sub-sampling makes conducting line searches along search directions a non-trivial issue in loss functions with the form of Equation (\ref{eq:lossgradbatch}). Therefore, this work focusses on optimality criteria and the nature of information that is useful in static and dynamic MBSS losses. Subsequently, we wish build an intuitive and visual understanding of function minimizers and {\it stochastic non-negative associated gradient projection points} (SNN-GPPs) when dealing with dynamic mini-batch sub-sampled loss functions. Therefore, we restrict this study to an elementary neural network classification problem in order to highlight concepts with absolute clarity. The concepts explored in this paper are not restricted to the example problem we choose, as the optimality criteria investigated apply to all loss functions, including those of state of the art deep neural networks explored by \cite{Li2017} and \cite{Goodfellow2015}. We show that function minimization in stochastic loss functions is not effective. Instead, gradient-only optimization that finds SNN-GPPs allows for a much improved representation of full-batch optima. We demonstrate this for neural networks loss functions using the Sigmoid, Tanh, Softsign, ReLU, leaky ReLU and ELU activation functions. We also highlight some key differences between the features of the loss functions with the different activations in the context of SNN-GPPs. We show, that the characteristics of activation function's derivatives affect the localization of SNN-GPPs in weight space.

\section{Application of concepts to a practical neural network problem}
\label{sec_SigsConcepts}

In our investigations we analyse the loss functions of the a single hidden layer neural network applied to the classic Iris dataset classification problem. We use a fully connected layer containing 10 hidden units and employ the Mean Squared Error (MSE) loss. Since we do not actually train the network in our experiments, but rather investigate the loss landscape characteristics, there is no need to split the problem's dataset into training and test sets. Instead, we make all the available data available for the construction of our visualizations. The computed loss functions and directional derivative surfaces are evaluated in a 100x100 grid with range of $[-20,20]$ units in the directions $\boldsymbol{d}_1$ and $\boldsymbol{d}_2$ around a central point, $\boldsymbol{x}_0$. This initial guess, $\boldsymbol{x}_0$, is generated from a uniform distribution with range $[-0.1,0.1]$. Until and including Section \ref{sec_global}, the surface plots' axes denote steps along two random, but perpendicular, unit directions \cite{Li2017}, $\boldsymbol{d}_1$ and $\boldsymbol{d}_2$ in $\mathbb{R}^{p}$, where $p=83$ is the number of weights in the network. Care should be exercised in interpreting these two-dimensional visualizations of an 83-dimensional problem, as what seems to be a local optimum in these two-dimensional surfaces may well have descent directions leading away from these in different dimensions not depicted. Nevertheless, these visualizations allow the general characteristics of loss functions to be investigated. We use this platform initially to demonstrate some key concepts with the use of the Sigmoid AF, whereas from Section \ref{sec_AltActs}, we expand this to various other activation functions.

\subsection{Full combinatorial static and dynamic mini-batch sub-sampling}
\label{sec_statvdyn}

\begin{figure}[h!]
	\centering
	\begin{subfigure}{.38\textwidth}
		\centering 
		\includegraphics[width=0.99\linewidth]{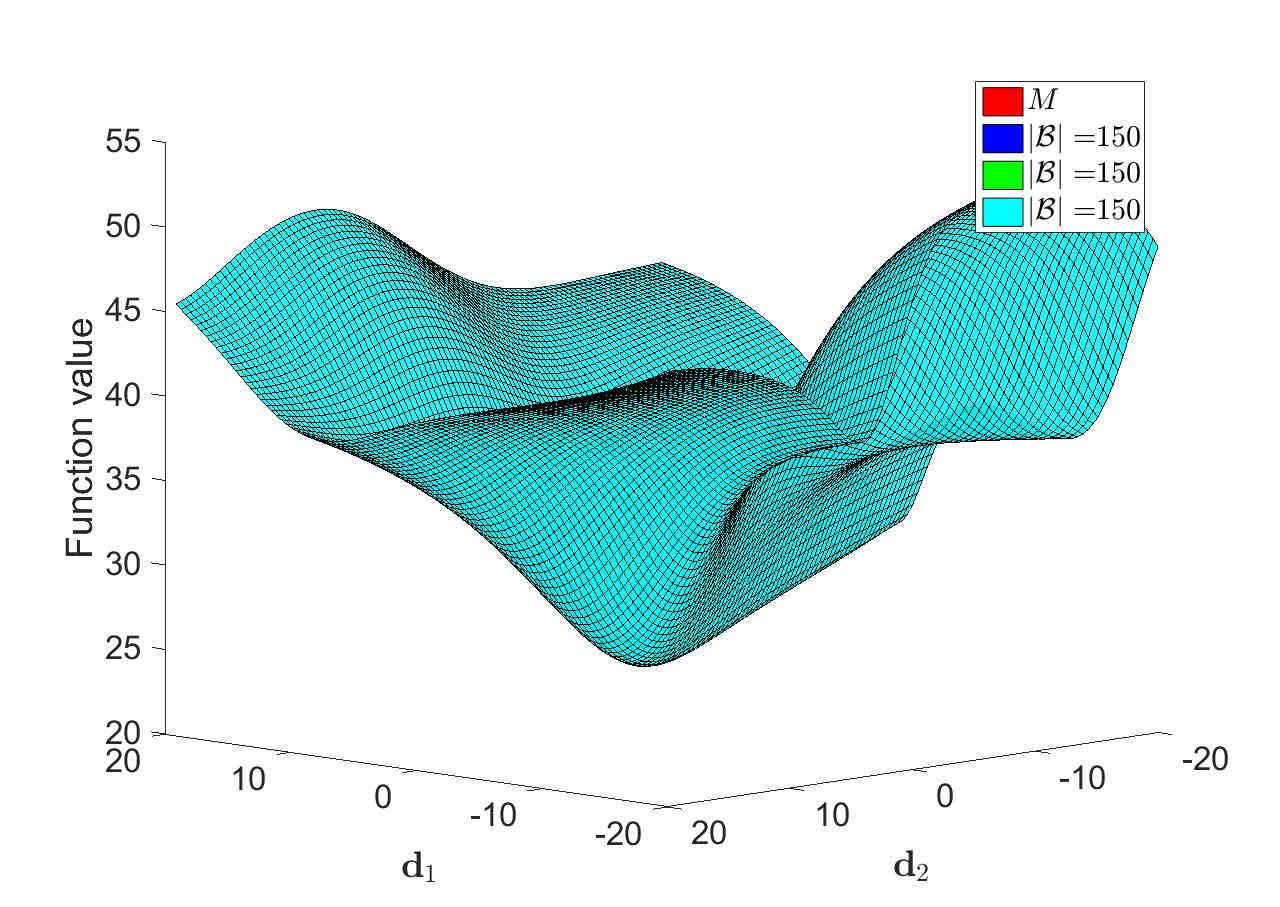}
		\caption{$\mathcal{L}(\boldsymbol{x})$, $M = 150$}
	\end{subfigure}%
	\begin{subfigure}{.38\textwidth}
		\centering
		\includegraphics[width=0.99\linewidth]{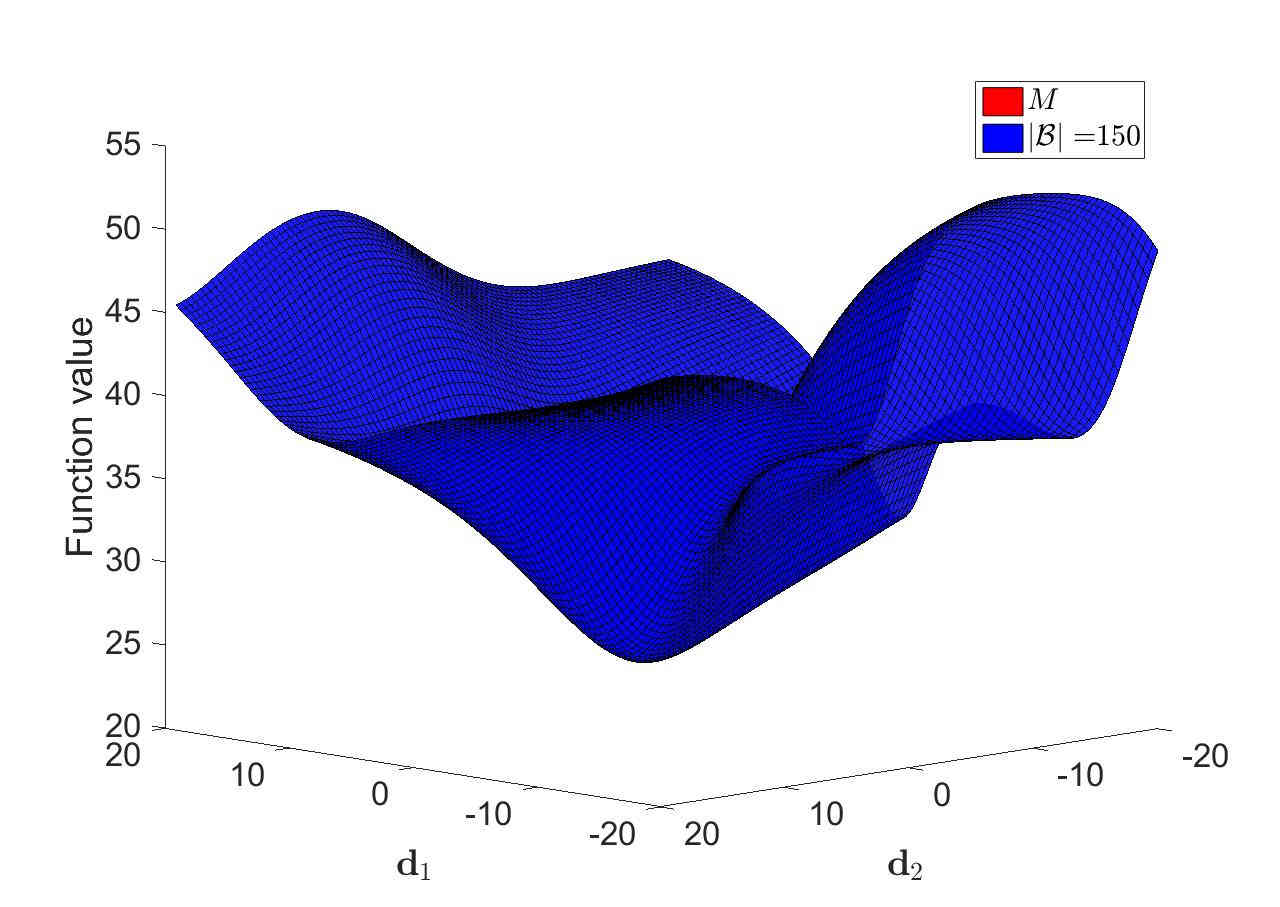}
		\caption{$\mathcal{L}(\boldsymbol{x})$, $M = 150$}
	\end{subfigure}%
	
	\begin{subfigure}{.38\textwidth}
		\centering 
		\includegraphics[width=0.99\linewidth]{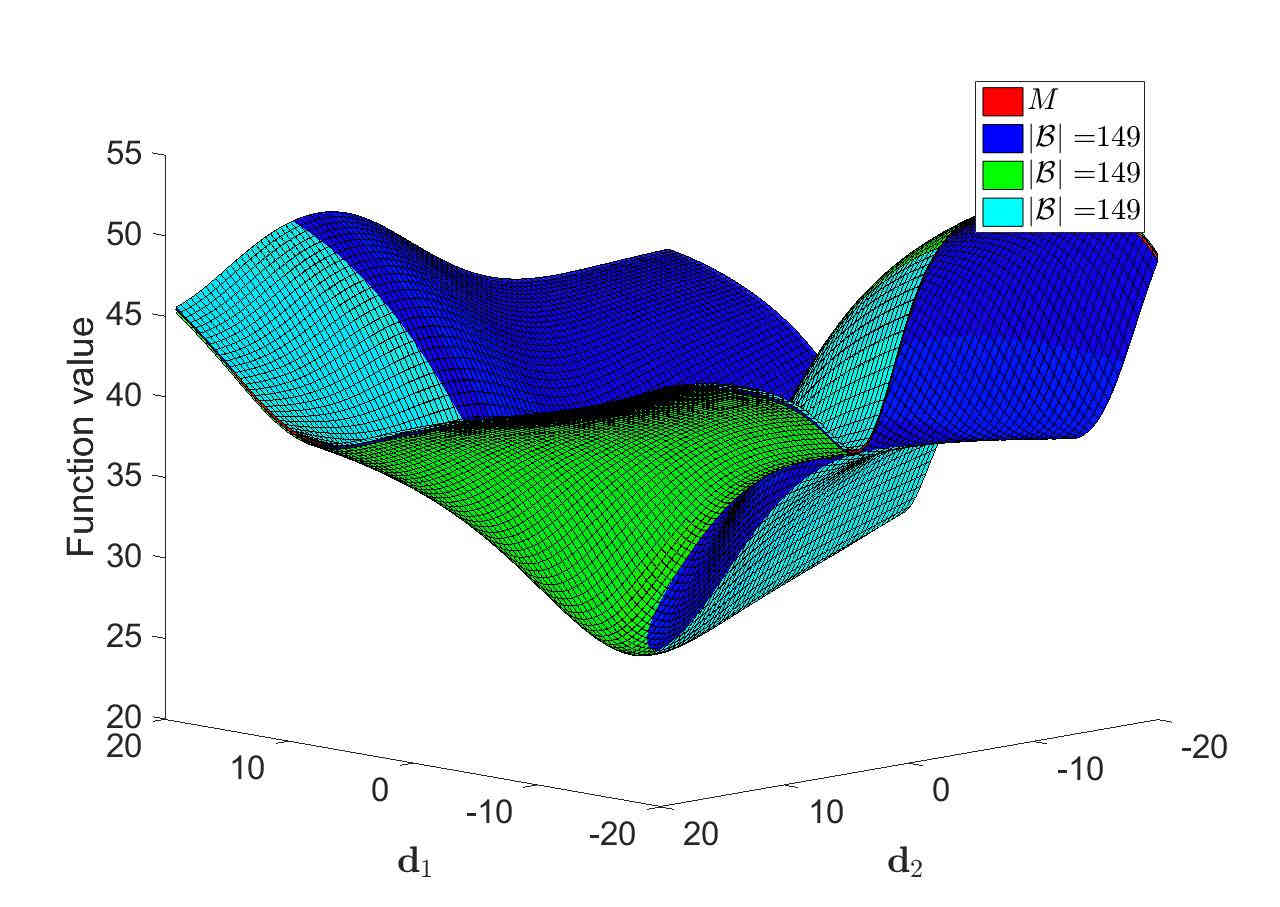}
		\caption{$\bar{L}(\boldsymbol{x})$, $|\mathcal{B}_{n}| = 149$}
	\end{subfigure}%
	\begin{subfigure}{.38\textwidth}
		\centering
		\includegraphics[width=0.99\linewidth]{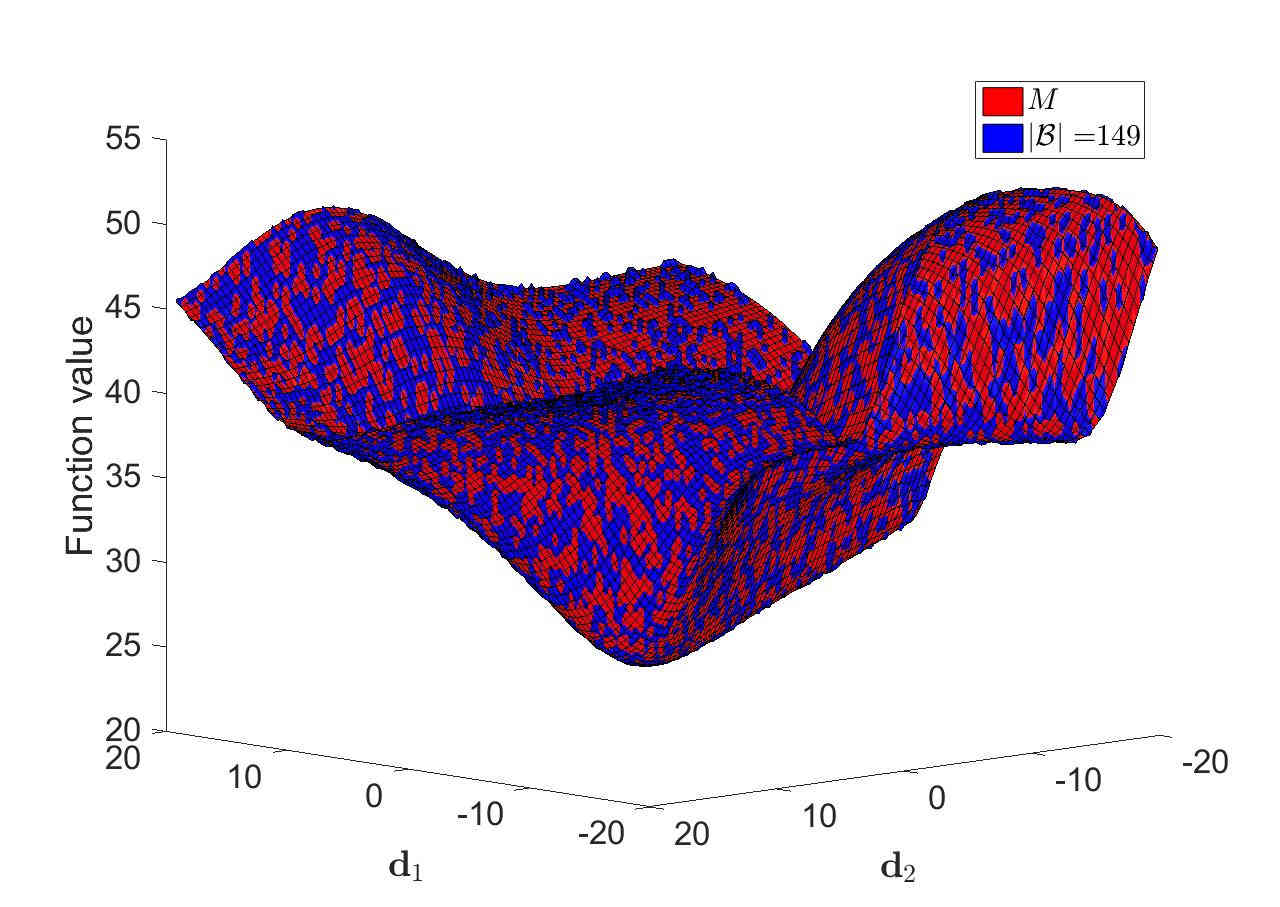}
		\caption{$\tilde{L}(\boldsymbol{x})$, $|\mathcal{B}_{n,i}| = 149$}
	\end{subfigure}%
	
	\begin{subfigure}{.38\textwidth}
		\centering 
		\includegraphics[width=0.99\linewidth]{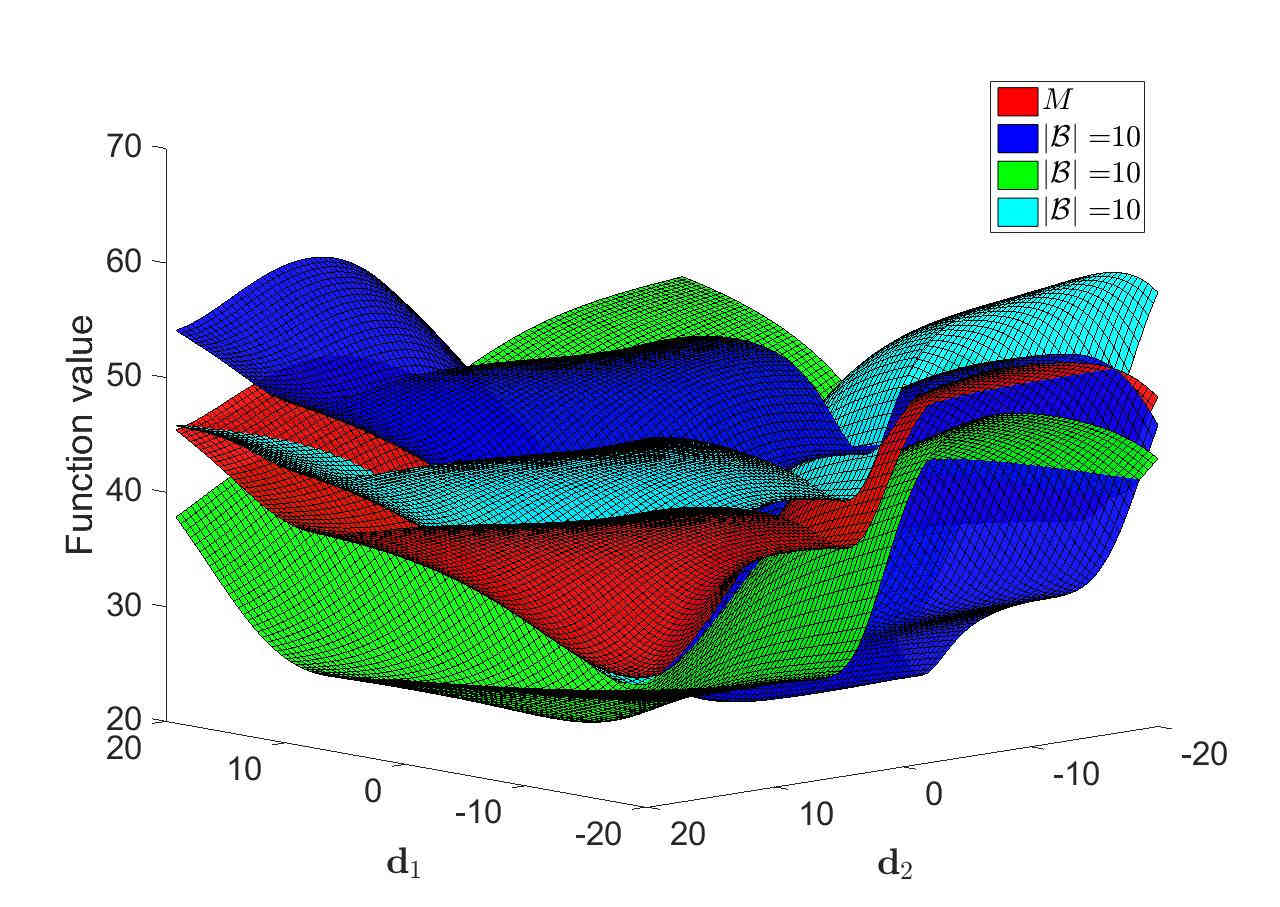}
		\caption{$\bar{L}(\boldsymbol{x})$, $|\mathcal{B}_{n}| = 10$}
	\end{subfigure}%
	\begin{subfigure}{.38\textwidth}
		\centering
		\includegraphics[width=0.99\linewidth]{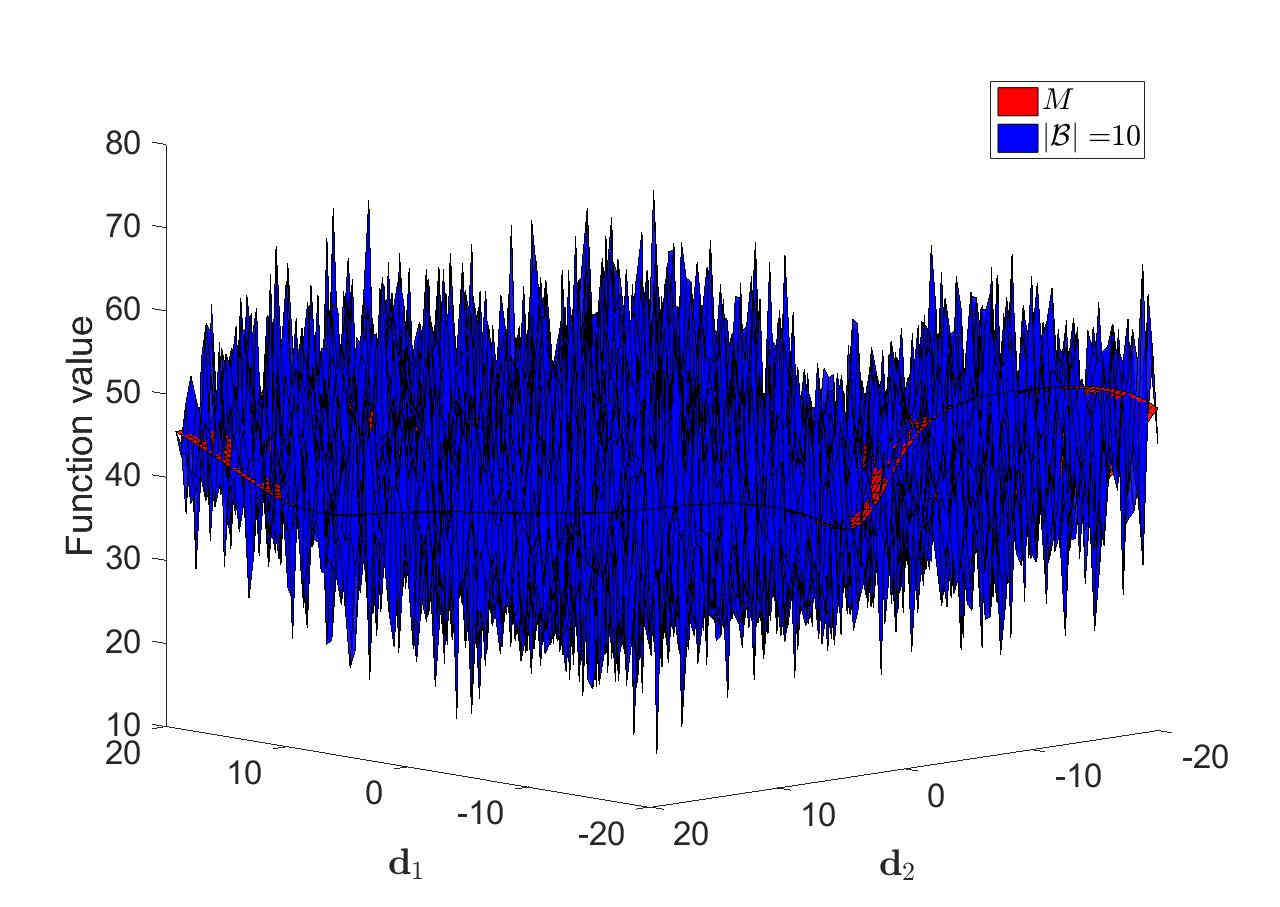}
		\caption{$\tilde{L}(\boldsymbol{x})$, $|\mathcal{B}_{n,i}| = 10$}
	\end{subfigure}%
	
	\begin{subfigure}{.38\textwidth}
		\centering 
		\includegraphics[width=0.99\linewidth]{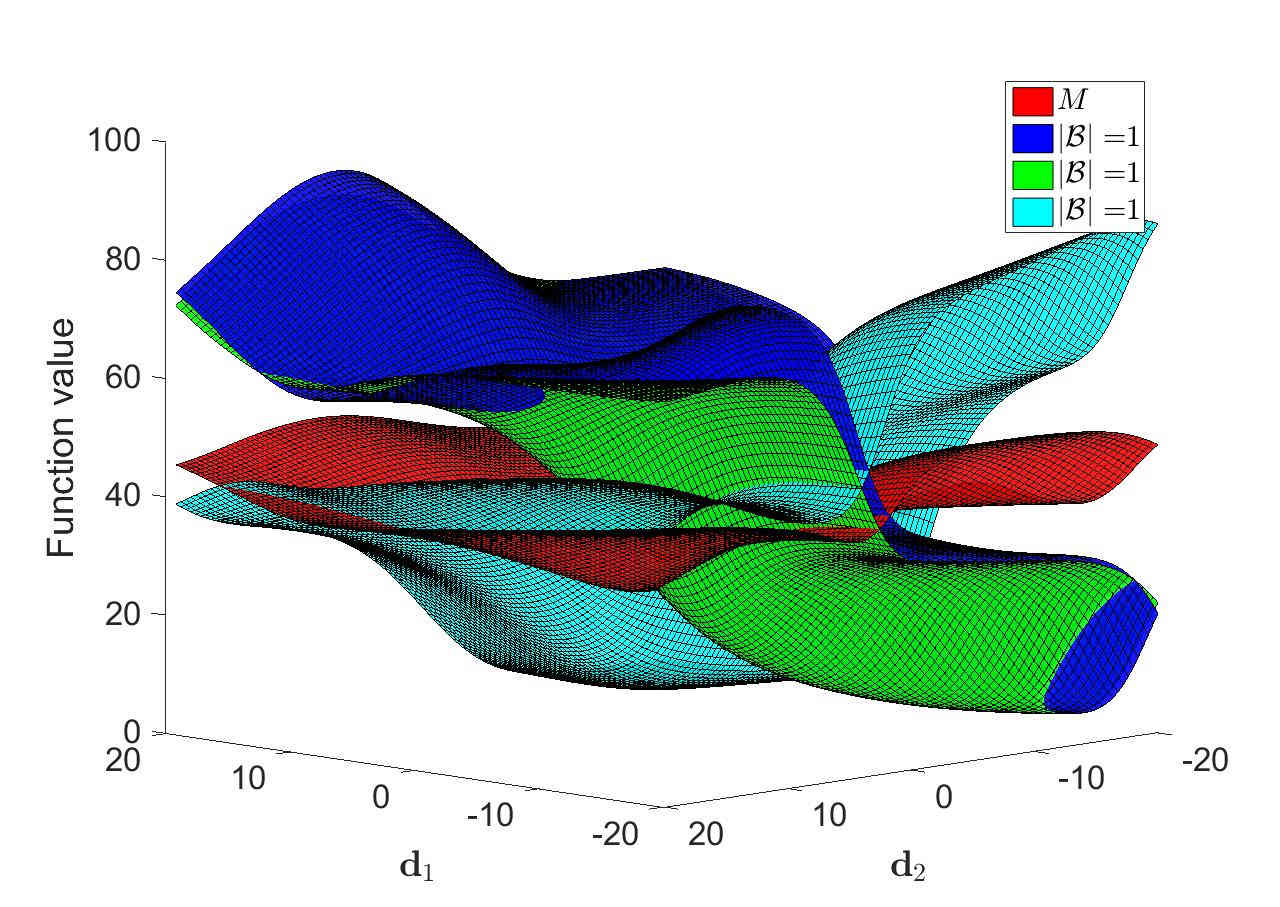}
		\caption{$\bar{L}(\boldsymbol{x})$, $|\mathcal{B}_{n}| = 1$}
	\end{subfigure}%
	\begin{subfigure}{.38\textwidth}
		\centering
		\includegraphics[width=0.99\linewidth]{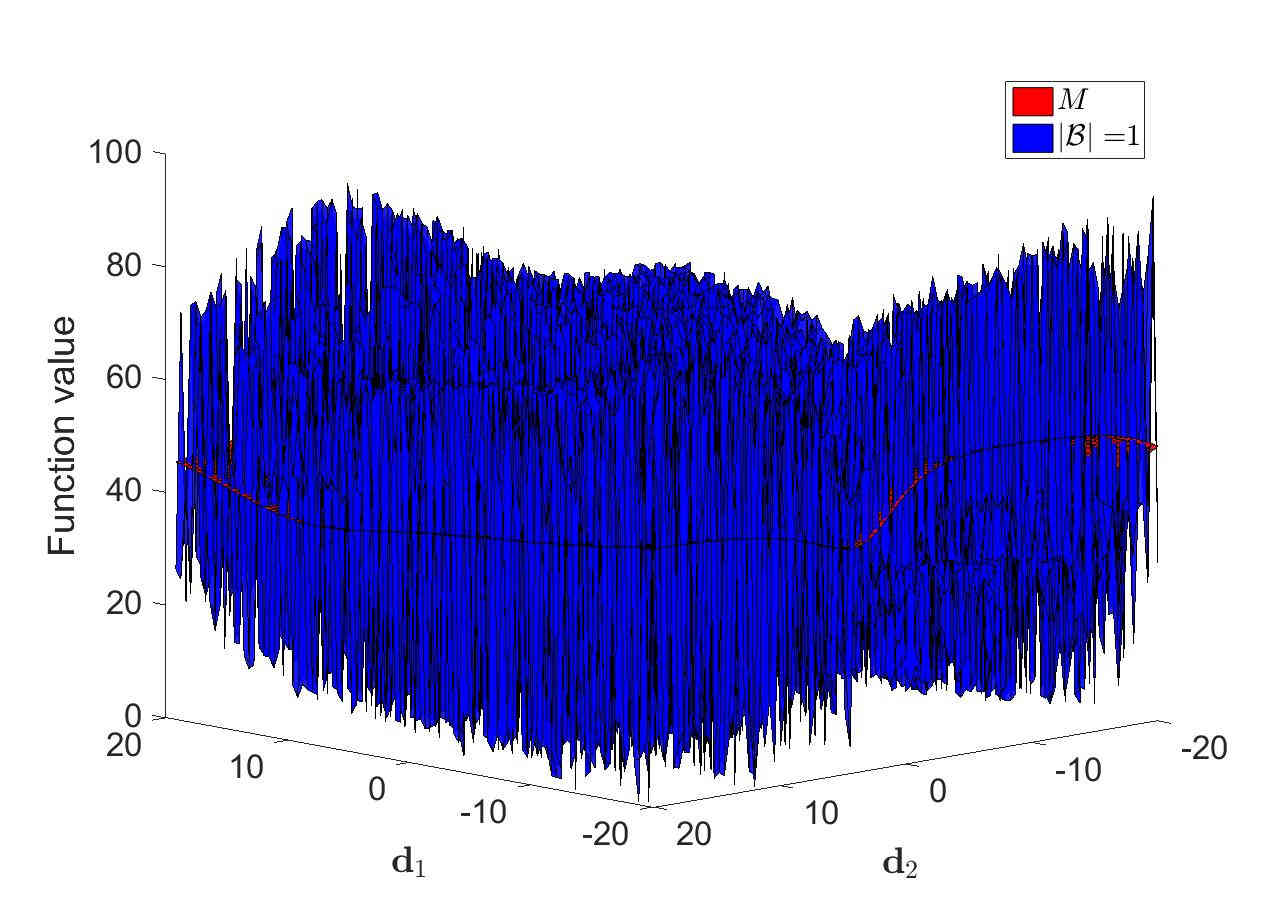}
		\caption{$\tilde{L}(\boldsymbol{x})$, $|\mathcal{B}_{n,i}| = 1$}
	\end{subfigure}%
	\caption{(a,c,e,g) Static and (b,d,f,h) dynamic mini-batch sub-sampling (MBSS) loss functions plotted along two random directions using batch sizes $M=|\mathcal{B}|=150$ and $ |\mathcal{B}| \in \{149, 10, 1\}$. Subscripts, $n$ and $i$, are omitted in plot legends in the interest of compactness. A bias-variance trade-off: Static MBSS produces continuous loss functions, which are biased to the sampling error in $\mathcal{B}_n$; dynamic MBSS represents $\mathcal{L}(\boldsymbol{x})$ on average, but introduces discontinuities by re-sampling $\mathcal{B}_{n,i}$ at every loss evaluation.}
	\label{fig_statVdyn}
\end{figure}

Firstly, we further explore the characteristics of static and dynamic MBSS in the context of sampling uniformly from the problem dataset with replacement. Therefore, this exposes the evaluation of $\tilde{L}(\boldsymbol{x})$ to the full $K = \binom{M}{|\mathcal{B}_{n,i}|}$ possible combinations of constructing mini-batches. In Figure~\ref{fig_statVdyn}, we plot the resulting loss in 2-D, to gain more intuition about the observed characteristics. For all plots the true loss function $\mathcal{L}(\boldsymbol{x})$, is given in red. For static MBSS we include three different loss function expressions of $\bar{L}(\boldsymbol{x})$, whereas for dynamic MBSS we add only one plot of $\tilde{L}(\boldsymbol{x})$ in the interest of visual simplicity. We plot the function value for both static and dynamic MBSS with batch sizes $M=150$ and $|\mathcal{B}_{n,i}| \in \{149,10,1 \}$.

In the case where the full batch is used, Figure~\ref{fig_statVdyn}(a) and (b), both sampling methods are equal, as both have access to the same information at every sampled point along the given axes. When one point is omitted, i.e. $ |\mathcal{B}_{n,i}| = 149 $ in Figure~\ref{fig_statVdyn}(c), a difference begins to emerge. The individual instances of $\bar{L}(\boldsymbol{x})$ have alternative loss function curvatures to $\mathcal{L}(\boldsymbol{x})$, resulting in different patches of the loss function being visible. This becomes increasingly pronounced as the batch size is decreased. In the case of $|\mathcal{B}_{n,i}| = 10$ in Figure~\ref{fig_statVdyn}(e), the optima of individual $\bar{L}(\boldsymbol{x})$ differ considerably and a sampling induced offset in loss function becomes evident. It is this offset, that causes large discontinuities when alternating between batches in dynamic MBSS. If the batch size is further reduced to $|\mathcal{B}_{n,i}| = 1$ in Figure~\ref{fig_statVdyn}(g), the individual $\bar{L}(\boldsymbol{x})$ are no longer reminiscent of $\mathcal{L}(\boldsymbol{x})$. Optimizing extensively in these individual loss functions would result in solutions which are not representative of $\mathcal{L}(\boldsymbol{x})$. 

Dynamic MBSS on the other hand has different properties: By only omitting one random sub-sample at every function evaluation, the surface of $\tilde{L}(\boldsymbol{x})$ becomes discontinuous, as is evidenced by the "spotted" look of Figure~\ref{fig_statVdyn}(d), where blue indicates $\tilde{L}(\boldsymbol{x})$ being larger than the true or full-batch surface $\mathcal{L}(\boldsymbol{x})$ depicted in red. The severity of these discontinuities increases as the batch size decreases. However, the important aspect of dynamic MBSS is that the mean of this discontinuous plot approximates the shape of $\mathcal{L}(\boldsymbol{x})$. This indicates that the comparison between static and dynamic MBSS is akin to the bias-variance trade-off. Static MBSS gives desired characteristics for effective optimization in its smooth landscape, but is not representative of the true problem unless a large, or according to adaptive sampling methods, a representative batch is chosen. Conversely dynamic MBSS results in low bias, but high variance in the resulting loss function. Optimization algorithms which are able to operate in this mode can make use of more information in the dataset, as the information in a fixed batch size is alternated at every function evaluation. However, they need to be be able to robustly deal with the sampling induced discontinuities.

\begin{figure}[h!]
	\centering
	\begin{subfigure}{.45\textwidth}
		\centering 
		\includegraphics[width=0.99\linewidth]{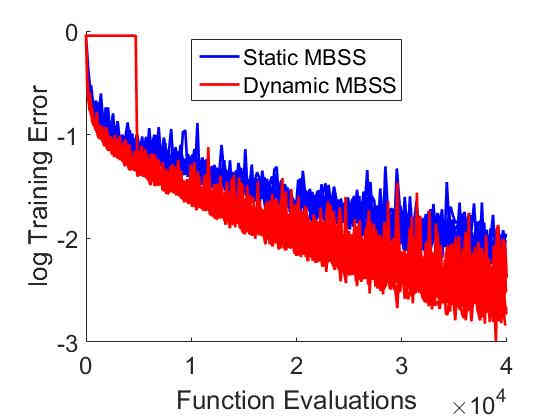}
		\caption{Training error}
	\end{subfigure}%
	\begin{subfigure}{.45\textwidth}
		\centering 
		\includegraphics[width=0.99\linewidth]{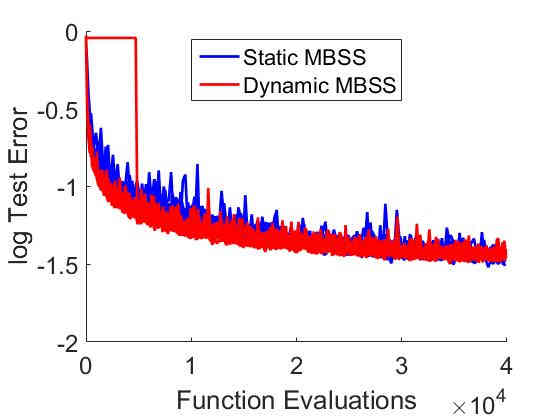}
		\caption{Test error}
	\end{subfigure}%
	\caption{A comparison between static and dynamic MBSS as applied in gradient-only line searches \cite{Kafka2019}, showing (a) training and (b) test classification error during training of the MNIST \cite{Lecun1998} dataset using the NetI \cite{Mahsereci2017a} network architecture.}
	\label{fig_svd_MNIST}
\end{figure}

A noteworthy attempt at optimizing using dynamic MBSS has been made with the application of probabilistic surrogates (Gaussian Processes) \cite{Mahsereci2017a} to conduct line searches in stochastic gradient descent (SGD). However, this method is hampered in its flexibility by having to use surrogates that have a bounded domain size for each iteration. An explicit study has been conducted by which static and dynamic MBSS were directly compared in the context of unbounded gradient-only line searches \cite{Kafka2019}. An example is given in Figure~\ref{fig_svd_MNIST}, where the well known MNIST \cite{Lecun1998} dataset is trained using the NetI \cite{Mahsereci2017a} network architecture. The faster decrease in both training and test classification error demonstrates that dynamic MBSS can increase the performance of training relative to computational cost. This is an example of where implementing dynamic MBSS in line searches during neural network training shows promise. In order to gain further insight, we wish to explore the landscapes of discontinuous loss functions with dynamic MBSS. Therefore, for the remainder of this study, dynamic MBSS is referred to when considering any form of sub-sampled loss function or directional derivative.

\subsection{Variance properties of function value and directional derivative information in the MSE loss using dynamic MBSS}
\label{sec_apply2sig}

\begin{figure}[h!]
	\centering
	\begin{subfigure}{.5\textwidth}
		\centering 
		\includegraphics[width=0.8\linewidth]{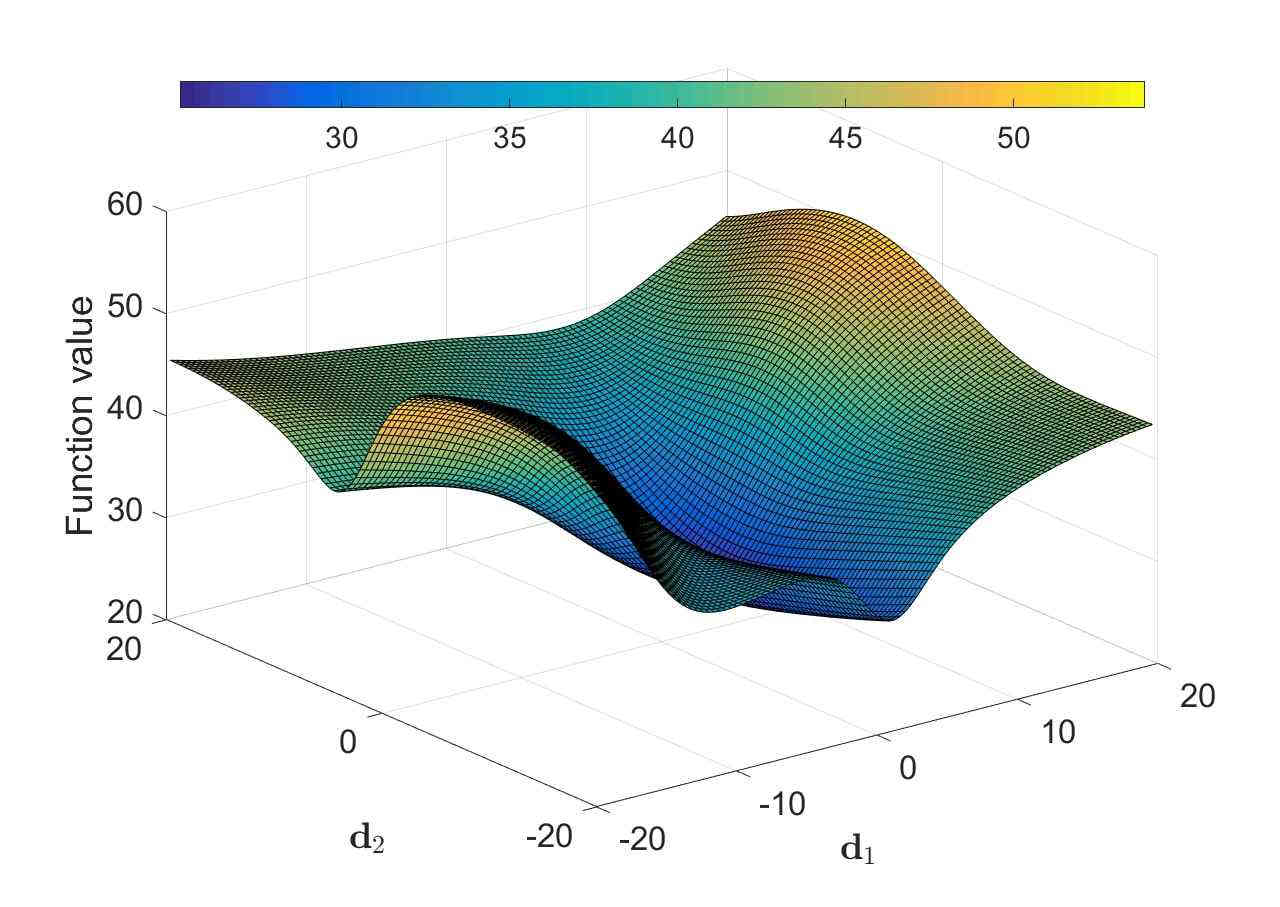}
		\caption{$\mathcal{L}(\boldsymbol{x})$, $M = 150$}
		\label{fig_sig_func_B}
	\end{subfigure}%
	\begin{subfigure}{.5\textwidth}
		\centering
		\includegraphics[width=0.8\linewidth]{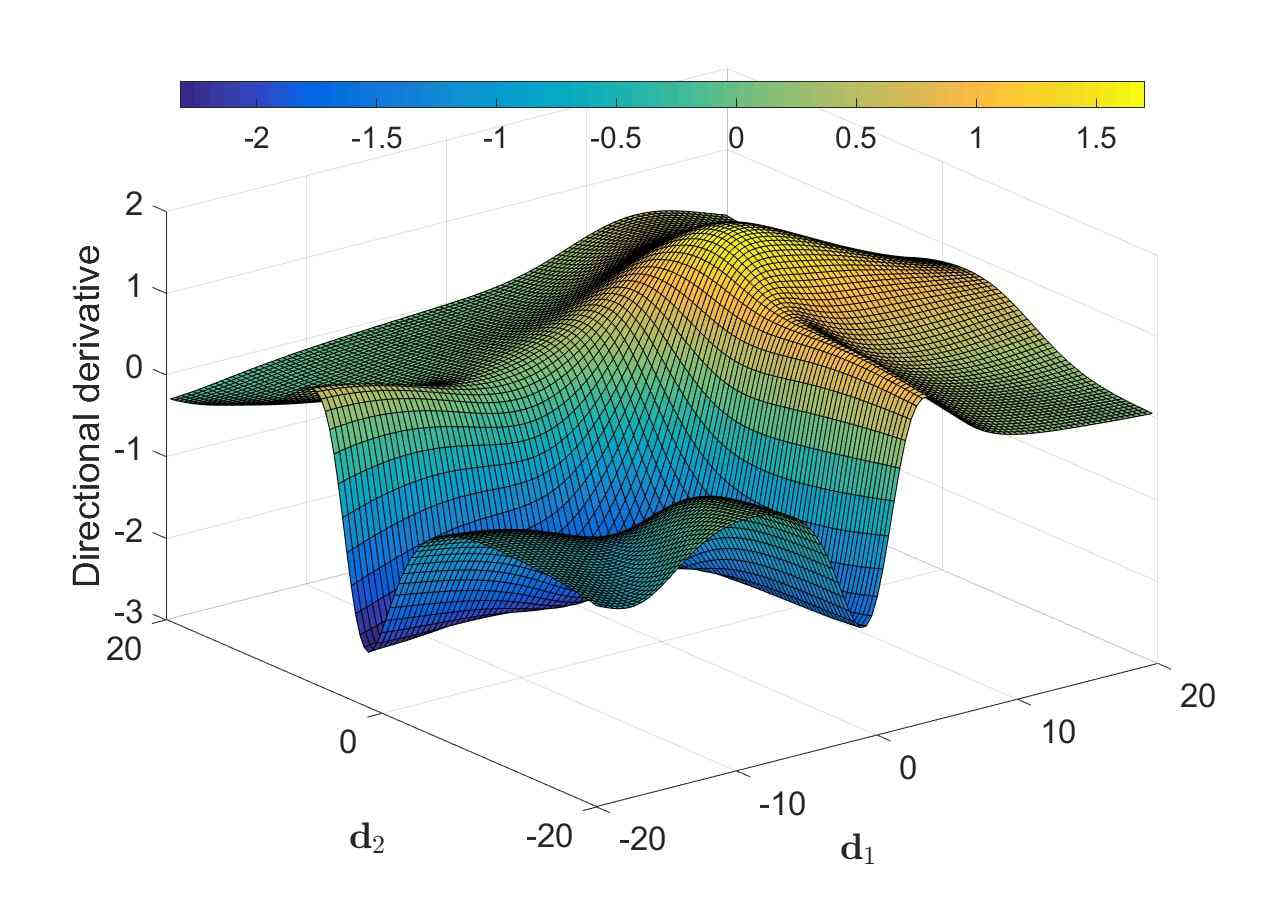}
		\caption{$\mathcal{D}_d(\boldsymbol{x})$, $M = 150$}
		\label{fig_sig_dd_B}
	\end{subfigure}%
	
	\begin{subfigure}{.5\textwidth}
		\centering 
		\includegraphics[width=0.8\linewidth]{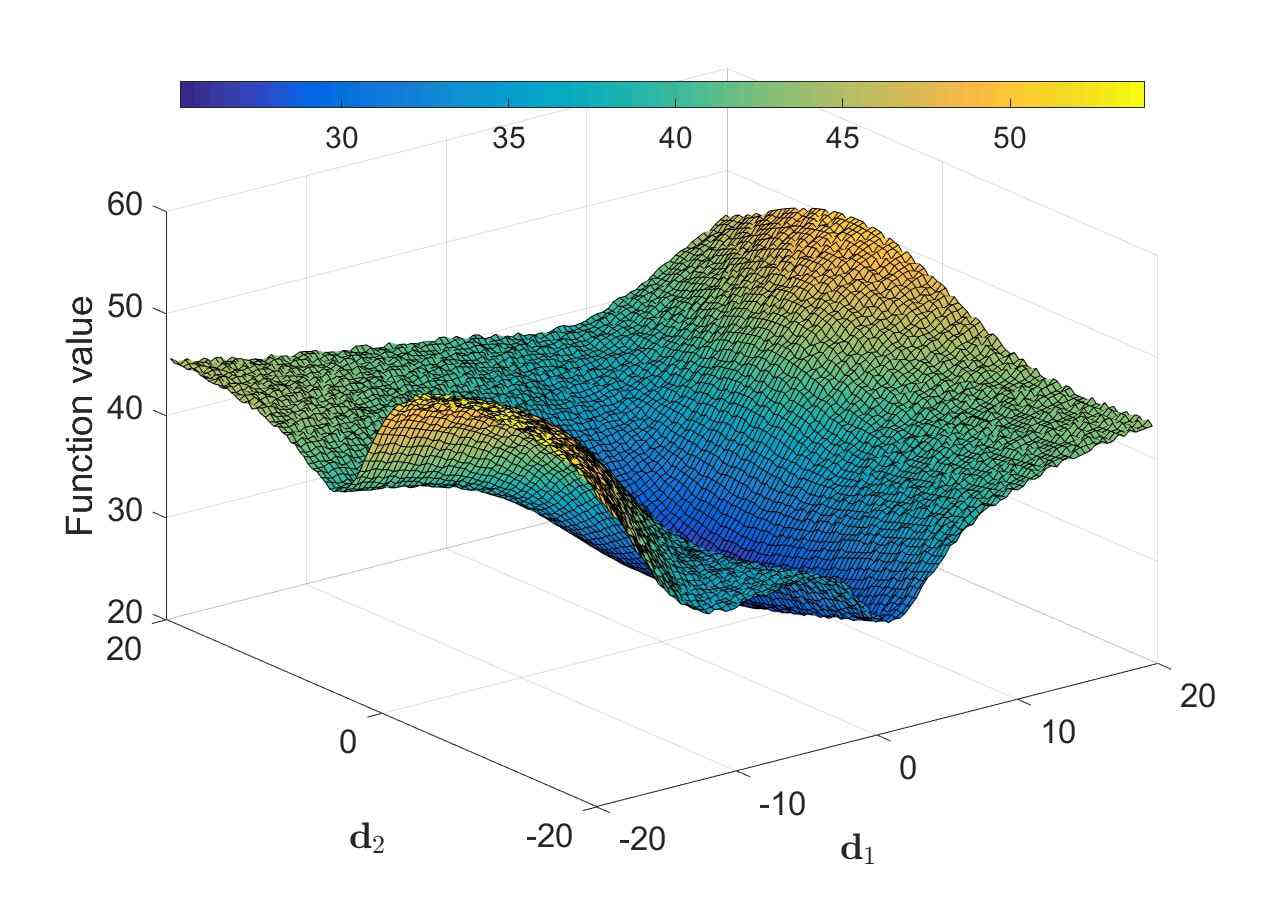}
		\caption{$\tilde{L}(\boldsymbol{x})$, $|\mathcal{B}_{n,i}| = 149$}
		\label{fig_sig_func_M4_149}
	\end{subfigure}%
	\begin{subfigure}{.5\textwidth}
		\centering
		\includegraphics[width=0.8\linewidth]{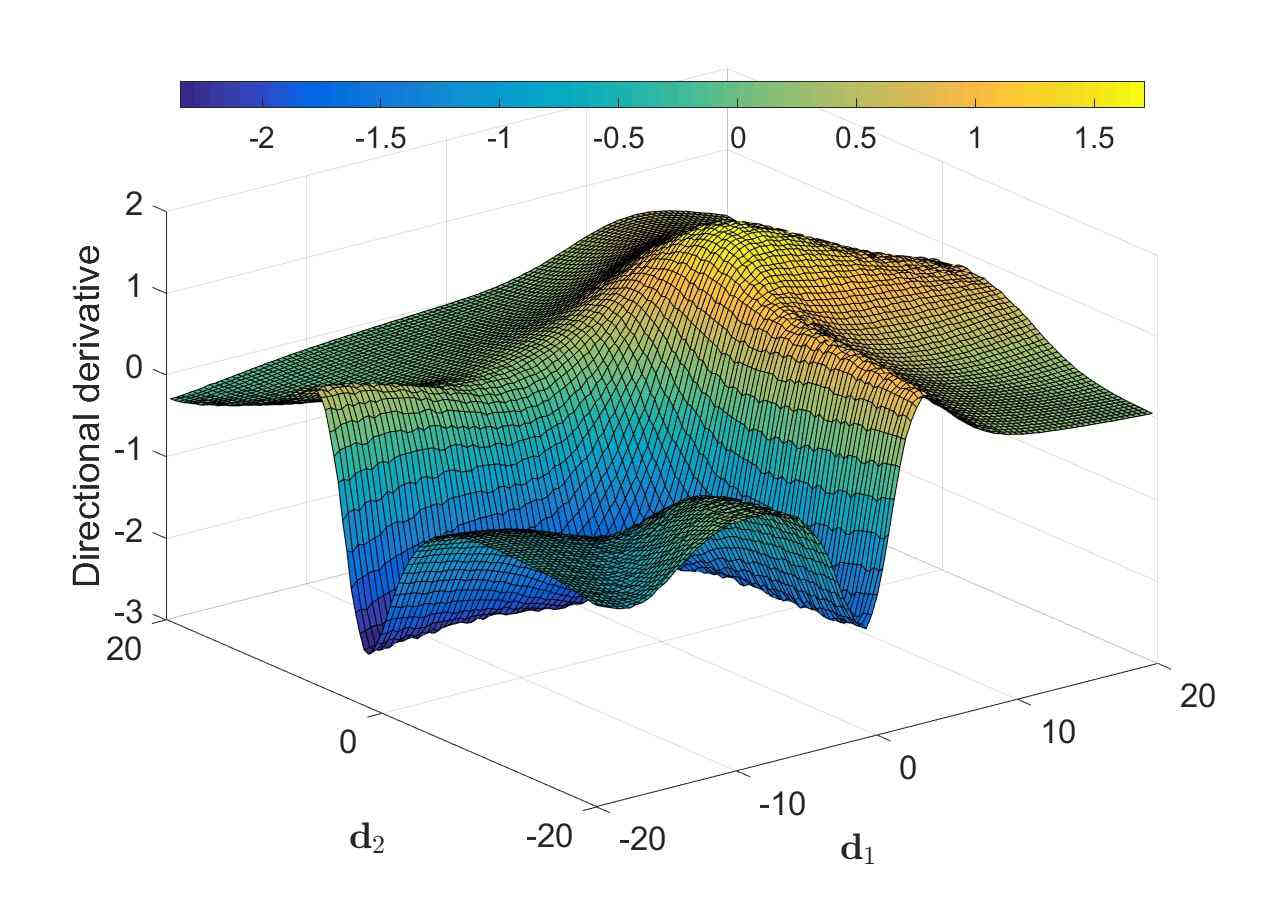}
		\caption{$\tilde{D}_d(\boldsymbol{x})$, $|\mathcal{B}_{n,i}| = 149$}
		\label{fig_sig_dd_M4_149}
	\end{subfigure}%
	
	\begin{subfigure}{.5\textwidth}
		\centering 
		\includegraphics[width=0.8\linewidth]{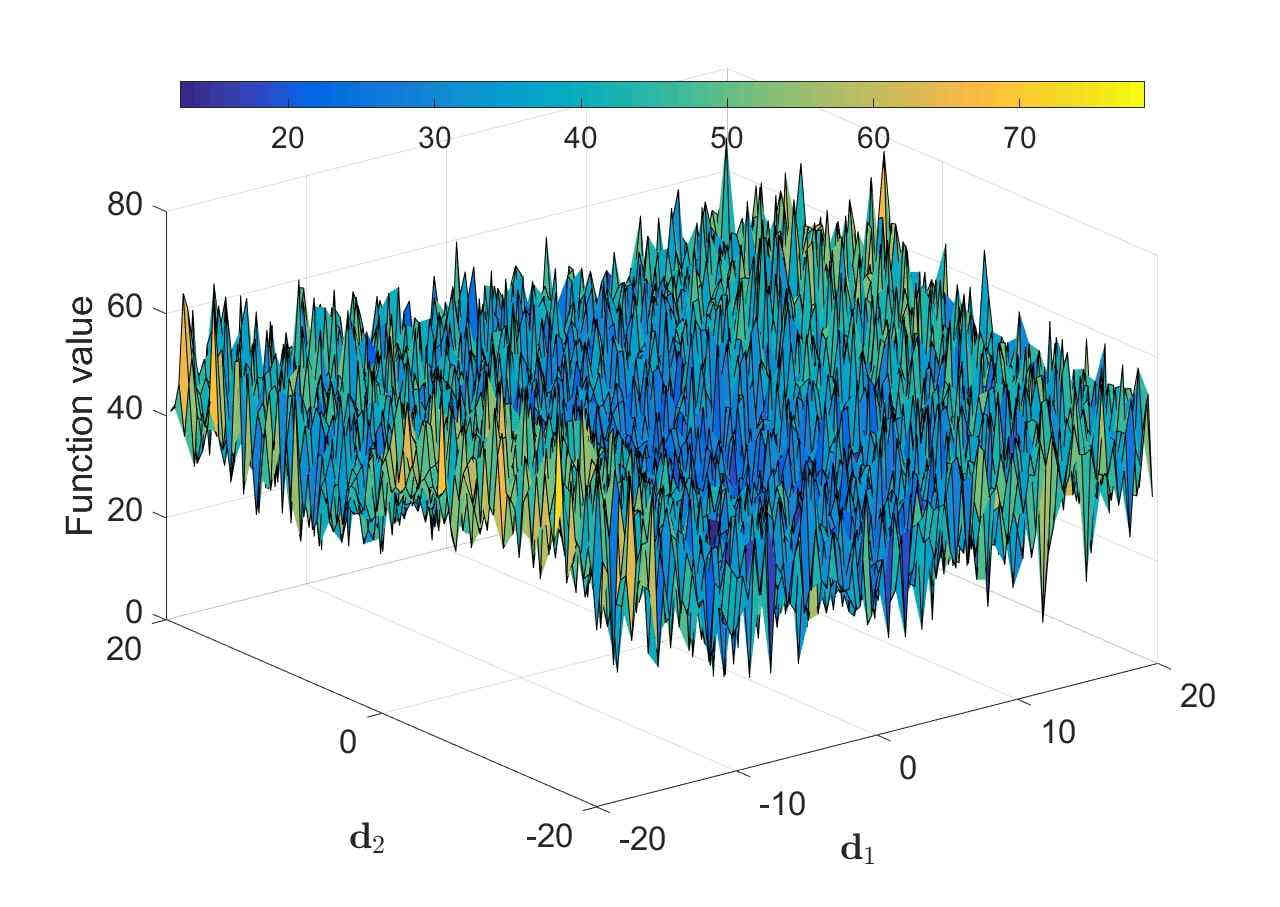}
		\caption{$\tilde{L}(\boldsymbol{x})$, $|\mathcal{B}_{n,i}| = 10$}
		\label{fig_sig_func_M}
	\end{subfigure}%
	\begin{subfigure}{.5\textwidth}
		\centering
		\includegraphics[width=0.8\linewidth]{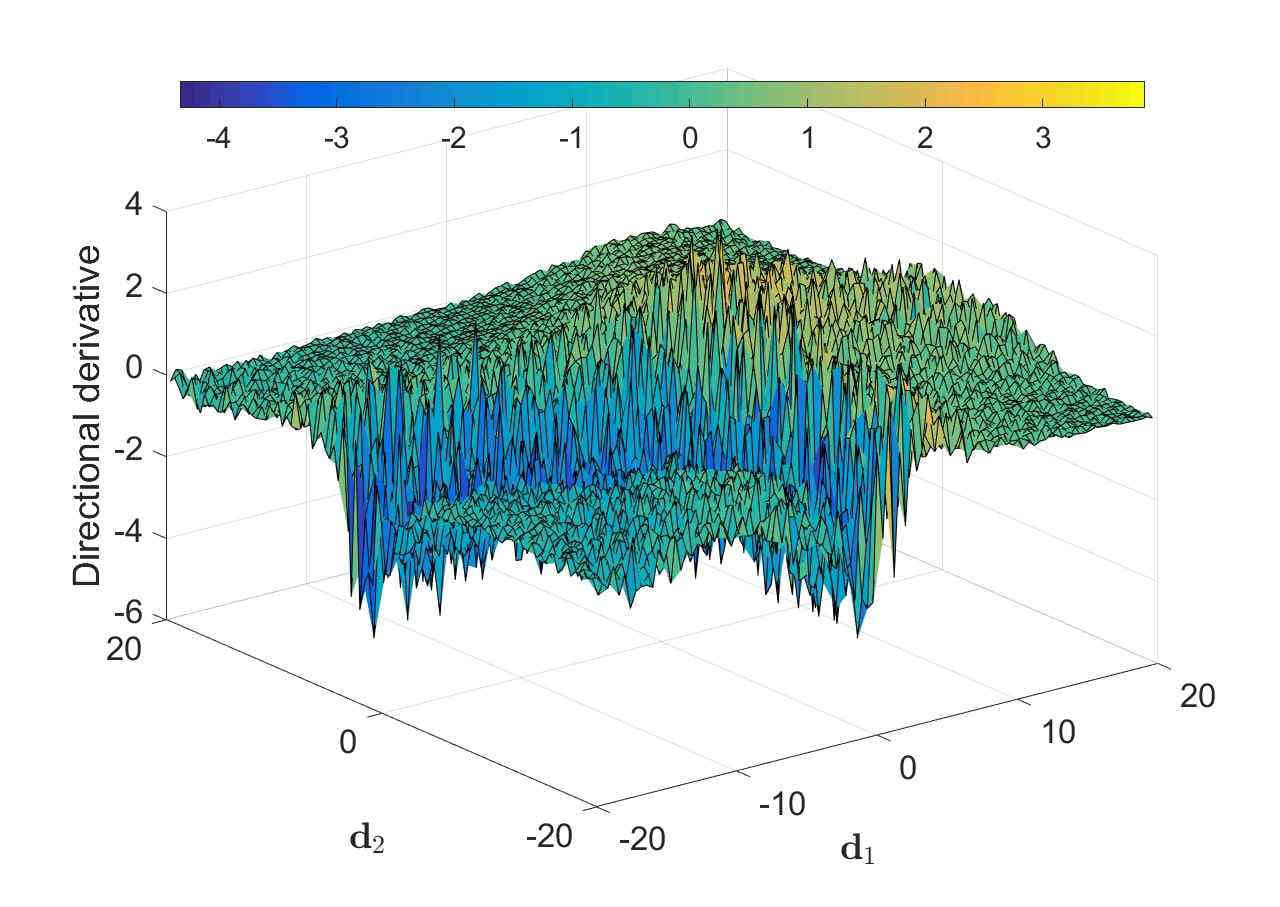}
		\caption{$\tilde{D}_d(\boldsymbol{x})$, $|\mathcal{B}_{n,i}| = 10$}
		\label{fig_sig_dd_M}
	\end{subfigure}%
	
	\begin{subfigure}{.5\textwidth}
		\centering 
		\includegraphics[width=0.8\linewidth]{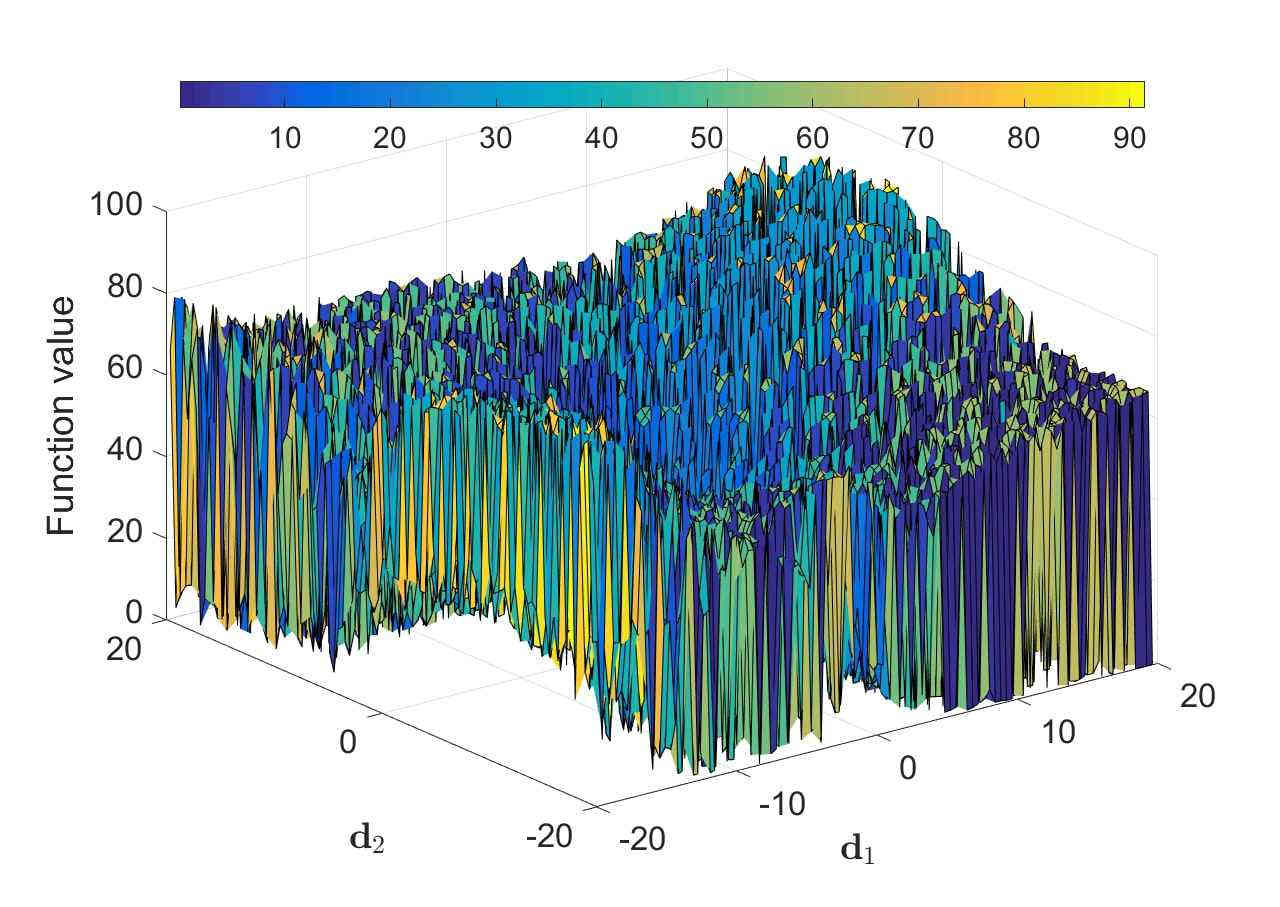}
		\caption{$\tilde{L}(\boldsymbol{x})$, $|\mathcal{B}_{n,i}| = 1$}
		\label{fig_sig_func_M4_1}
	\end{subfigure}%
	\begin{subfigure}{.5\textwidth}
		\centering
		\includegraphics[width=0.8\linewidth]{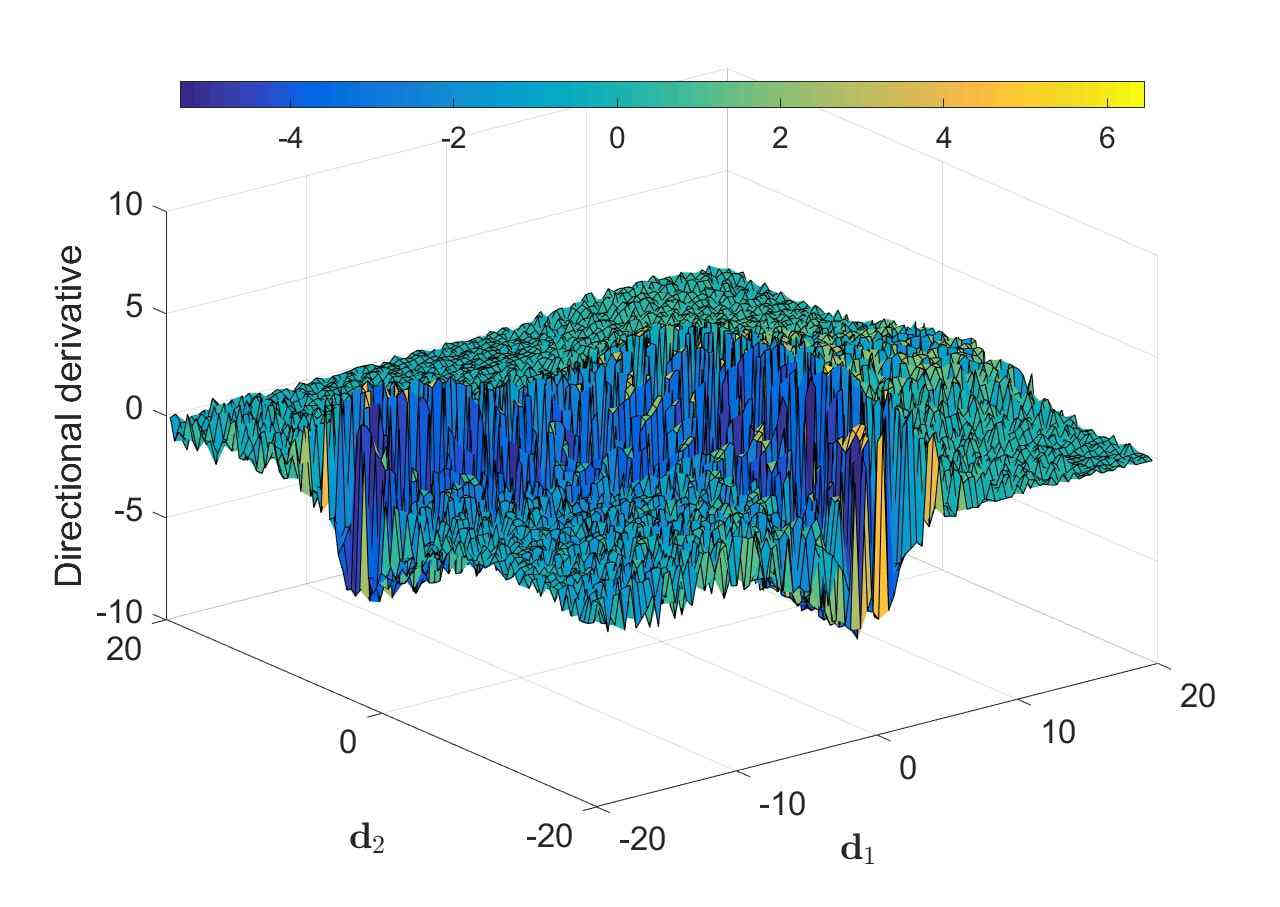}
		\caption{$\tilde{D}_d(\boldsymbol{x})$,  $|\mathcal{B}_{n,i}| = 1$}
		\label{fig_sig_dd_M4_1}
	\end{subfigure}%
	\caption{(a,c,e,g) Function value and (b,d,f,h) directional derivative plots along two orthogonal random directions, $\boldsymbol{d}_1$ and $\boldsymbol{d}_2$, using batch sizes $ |\mathcal{B}_{n,i}| \in \{150, 149, 10, 1\}$. Variance for both $\mathcal{L}(\boldsymbol{x})$ and $\tilde{D}_d(\boldsymbol{x})$ increases with decrease in batch size, with $\tilde{D}_d(\boldsymbol{x})$ being less affected than $\mathcal{L}(\boldsymbol{x})$.}
	\label{fig_sig}
\end{figure}

Subsequently, we explore the quality of function value and directional derivative information the context of our practical neural network problem with Sigmoid AFs. The directional derivatives were calculated using the positive diagonal, i.e. the sum of the two unit directions, given as follows:

\begin{equation}
\boldsymbol{d}_{dd} = \frac{(\boldsymbol{d}_1 + \boldsymbol{d}_2)}{|\boldsymbol{d}_1 + \boldsymbol{d}_2|}.
\label{eq_d_dd}
\end{equation}

The directional derivative is evaluated by projecting each computed gradient onto the normalized diagonal direction $\boldsymbol{d}_{dd}$, i.e. $\tilde{D}_d(\boldsymbol{x}) = \tilde{\boldsymbol{g}}(\boldsymbol{x})^T \cdot \boldsymbol{d}_{dd}$. This allows the scalar surface of $\tilde{D}_d(\boldsymbol{x})$ to be visualized over the same 100x100 grid used plot function values. We compare the characteristics of both function value and directional derivative surfaces in Figure~\ref{fig_sig} with mini-batch sizes $M=150$ and $|\mathcal{B}_{n,i}| \in \{149,10,1 \}$.

Note, that the apparent variance in the discontinuous directional derivative surface is much lower than the discontinuous function value surface. As the batch size is reduced to $\mathcal{B}_{n,i} = 10$, the representation of $\mathcal{L}(\boldsymbol{x})$ is largely hidden in the high variance of the discontinuities in $\tilde{L}(\boldsymbol{x})$. However, in some sections of the sampled domain, the directional derivative information remains considerably more consistent with its full-batch equivalent, compared to the function value plots. This trend becomes exaggerated, as the mini-batch size is reduced further to $\mathcal{B}_{n,i} = 1$.

We formalize this observation by considering the MSE loss: 
\begin{equation}
\ell  (\boldsymbol{x};\;\boldsymbol{t}_b) = \frac{1}{2}(\boldsymbol{t}_b^o-\hat{\boldsymbol{t}}_b^o(\boldsymbol{x},\boldsymbol{t}_b^i))^T (\boldsymbol{t}_b^o-\hat{\boldsymbol{t}}_b^o(\boldsymbol{x},\boldsymbol{t}_b^i)),
\label{eq_smallLoss}
\end{equation}
where $\boldsymbol{t}_b^i$ is the $b^{th}$ sample of the training data, $\boldsymbol{t}_b^o$ is the $b^{th}$ output sample and $\hat{\boldsymbol{t}}_b^o(\cdot,\cdot)$ is the neural network model estimation of output $\boldsymbol{t}_b^o$ as a function of the model parameters, $\boldsymbol{x}$. and training sample, $\boldsymbol{t}_b^i$.
We substitute Equation~(\ref{eq_smallLoss}) into Equation~(\ref{eq:loss}) to obtain:

\begin{equation}
\tilde{L}(\boldsymbol{x}) = \frac{1}{2 |\mathcal{B}_{n,i}|} \sum_{b\in \mathcal{B}_{n,i}} (\boldsymbol{t}_b^o -\hat{\boldsymbol{t}}_b^o(\boldsymbol{x},\boldsymbol{t}_b^i))^T (\boldsymbol{t}_b^o -\hat{\boldsymbol{t}}_b^o(\boldsymbol{x},\boldsymbol{t}_b^i)).
\label{eq_SubLoss}
\end{equation}

Therefore, the sampled gradient of the loss becomes
\begin{equation}
\tilde{\boldsymbol{g}}(\boldsymbol{x}) = \frac{-1}{|\mathcal{B}_{n,i}|} \sum_{b\in \mathcal{B}_{n,i}}
(\nabla \hat{\boldsymbol{t}}_b^o(\boldsymbol{x},\boldsymbol{t}_b^i)) (\boldsymbol{t}_b^o -\hat{\boldsymbol{t}}_b^o(\boldsymbol{x},\boldsymbol{t}_b^i)),
\label{eq_gradSubLoss}
\end{equation}

where the gradient of the model in terms of the model parameters is given by $ \nabla \hat{\boldsymbol{t}}_b^o(\boldsymbol{x},\boldsymbol{t}_b^i) $. To aid understanding of the different contributing factors, consider the notation $ \boldsymbol{e}_b = (\boldsymbol{t}_b^o -\hat{\boldsymbol{t}}_b^o(\boldsymbol{x},\boldsymbol{t}_b^i))$, which constitutes the ($q \times 1$) prediction error vector for observation $k$, where $q$ is the dimensionality of the output data and model output. The term $ \boldsymbol{C}_b = \nabla \hat{\boldsymbol{t}}_b^o(\boldsymbol{x},\boldsymbol{t}_b^i) $ denotes the ($p \times q$) gradient matrix of the model. This simplifies Equations~(\ref{eq_SubLoss}) and (\ref{eq_gradSubLoss}) to

\begin{equation}
\tilde{L}(\boldsymbol{x}) = \frac{1}{2 |\mathcal{B}_{n,i}|} \sum_{b\in \mathcal{B}_{n,i}} (\boldsymbol{e}_b^T \boldsymbol{e}_b),
\label{eq_lossAk}
\end{equation}
and 
\begin{equation}
\tilde{\boldsymbol{g}}(\boldsymbol{x}) = \frac{-1}{|\mathcal{B}_{n,i}|} \sum_{b\in \mathcal{B}_{n,i}} (\boldsymbol{C}_b \boldsymbol{e}_b).
\label{eq_dlossAk}
\end{equation}

In the loss function, Equation~\ref{eq_lossAk} depends only on the $ \boldsymbol{e}_b$ term, while the product of $\boldsymbol{C}_b \boldsymbol{e}_b$ determines the loss function gradient in Equation~\ref{eq_dlossAk}. According to the chain rule, the gradient of the model, $\boldsymbol{C}_b$, is a function of the weights, and the derivatives of the activation functions in the various layers of a neural network \cite{Webros1982}. For most activation functions, the derivative remains bounded. All the activation functions considered in this paper have derivatives $\leq 1$. This means that for a fixed batch, $\mathcal{B}_{n,i}$ and a fixed point $\boldsymbol{x} $ in space, which apply to both $\boldsymbol{e}_b$ and $ \boldsymbol{C}_b$ terms, the activation function derivatives are not going to increase the magnitude of information passing through the network. 

Let us now consider a fixed point in space, $ \boldsymbol{x} $, where we sample many mini-batches with a constant size $|\mathcal{B}_{n,i}|< M$. In this case both $\boldsymbol{e}_b$ and $\boldsymbol{C}_b$ vary only as a function of $\boldsymbol{t}_b$, which we can separate into expected values $\bar{\boldsymbol{e}}_b$ and $\bar{\boldsymbol{C}}_b$; with corresponding variance $\sigma(\boldsymbol{e}_b)$ and $\sigma(\boldsymbol{C}_b)$ to obtain

\begin{multline}
\tilde{L}(\boldsymbol{x}) = \frac{1}{2 |\mathcal{B}_{n,i}|} \sum_{b\in \mathcal{B}_{n,i}} (\bar{\boldsymbol{e}}_b + \sigma\left({\boldsymbol{e}_b}\right))^T (\bar{\boldsymbol{e}}_b + \sigma\left({\boldsymbol{e}_b}\right))
\\
= \frac{1}{2 |\mathcal{B}_{n,i}|} \sum_{b\in \mathcal{B}_{n,i}} (\bar{\boldsymbol{e}}_b^T \bar{\boldsymbol{e}}_b + 2 \bar{\boldsymbol{e}}_b^T \sigma\left({\boldsymbol{e}_b}\right) + \sigma\left({\boldsymbol{e}_b}\right)^T \sigma\left({\boldsymbol{e}_b}\right)) ,
\label{eq_lossAkvar}
\end{multline}
and 
\begin{multline}
\tilde{\boldsymbol{g}}(\boldsymbol{x}) = \frac{-1}{|\mathcal{B}_{n,i}|} \sum_{b\in \mathcal{B}_{n,i}} (\bar{\boldsymbol{C}}_b + \sigma(\boldsymbol{C}_b)) \left(\bar{\boldsymbol{e}}_b+\sigma(\boldsymbol{e}_b) \right)  
\\
=
\frac{-1}{|\mathcal{B}_{n,i}|} \sum_{b\in \mathcal{B}_{n,i}}
\left(\bar{\boldsymbol{C}}_b \bar{\boldsymbol{e}}_b + \sigma(\boldsymbol{C}_b) \bar{\boldsymbol{e}}_b
+ \bar{\boldsymbol{C}}_b \sigma(\boldsymbol{e}_b)
+
\sigma(\boldsymbol{C}_b) \sigma(\boldsymbol{e}_b) \right)
.
\label{eq_dlossAkvar}
\end{multline}

Hence, the variance in $\tilde{L}(\boldsymbol{x})$ is dictated by $2 \bar{\boldsymbol{e}}_b^T \sigma\left({\boldsymbol{e}_b}\right) + \sigma\left({\boldsymbol{e}_b}\right)^T \sigma\left({\boldsymbol{e}_b}\right)$, while the variance in $\tilde{\boldsymbol{g}}(\boldsymbol{x})$ is dictated by $\sigma(\boldsymbol{C}_b) \bar{\boldsymbol{e}}_b + \bar{\boldsymbol{C}}_b \sigma(\boldsymbol{e}_b) + \sigma(\boldsymbol{C}_b) \sigma(\boldsymbol{e}_b)$. This implies that should $ |\sigma(\boldsymbol{e}_b)|>> 1$, due to changes in the mini-batch, the variance in $\tilde{L}(\boldsymbol{x})$ will be larger than the variance in $\tilde{\boldsymbol{g}}(\boldsymbol{x})$. This is due to the term $\sigma(\boldsymbol{e}_b)^T \sigma(\boldsymbol{e}_b)$ dominating in  Equation~(\ref{eq_lossAkvar}), while Equation~(\ref{eq_dlossAkvar}) scales linearly in both $\sigma(\boldsymbol{e}_b)$ and $\sigma(\boldsymbol{C}_b)$ which is bounded $|\sigma(\boldsymbol{C}_b)| \leq 1$, depending on $\boldsymbol{x}$.

Alternatively, the variance in $\tilde{L}(\boldsymbol{x})$ is significantly reduced when $ |\sigma(\boldsymbol{e}_b)| << 1 $, since $\sigma\left({\boldsymbol{e}_b}\right)^T \sigma\left({\boldsymbol{e}_b}\right) \approx 0 $, while $2 \bar{\boldsymbol{e}}_b^T \sigma\left({\boldsymbol{e}_b}\right)$ depends on the magnitude of $\bar{\boldsymbol{e}}_b$. In turn, the variance in $\tilde{\boldsymbol{g}}(\boldsymbol{x})$ depends mainly on $\sigma(\boldsymbol{C}_b)$  which scales $\bar{\boldsymbol{e}}_b$ and $\sigma(\boldsymbol{e}_b)$, while $\bar{\boldsymbol{C}}_b \sigma(\boldsymbol{e}_b)$ also contributes. However, since $|\bar{\boldsymbol{C}}_b| \leq 1$ and $|\sigma(\boldsymbol{C}_b)| \leq 1$ for small weights $\boldsymbol{x}$, the variance in $\tilde{\boldsymbol{g}}(\boldsymbol{x})$ is bounded. Therefore, when $ |\sigma(\boldsymbol{e}_b)| << 1 $ the variance in $\tilde{\boldsymbol{g}}(\boldsymbol{x})$ can be larger than that of $\tilde{L}(\boldsymbol{x})$. Also, when the magnitude of $|\bar{\boldsymbol{e}}_b| << 1$ the variance in $\tilde{L}(\boldsymbol{x})$ is reduced as can be seen in the centre of the domains in Figures~\ref{fig_sig}(e) and (g). Lastly, it is evident that when $|\bar{\boldsymbol{C}}_b| \approx 0$, implying saturation of the activation function, the variance $|\sigma(\boldsymbol{C}_b)|\approx 0$, which in turn causes the variance of $\tilde{\boldsymbol{g}}(\boldsymbol{x})$ to diminish. The flat, low variance areas at the extremes of the sampled domains are particularly evident for the directional derivative plots sampled with small batches in Figure~\ref{fig_sig}(f) and (h). Here, the saturation of the activation functions drives down the variance in the gradient. 


\subsection{The influence of variance on minimum and SNN-GPP optimality criteria}

\begin{figure}
	\centering
	
	\begin{subfigure}{.45\textwidth}
		\centering 
		\includegraphics[width=1\linewidth]{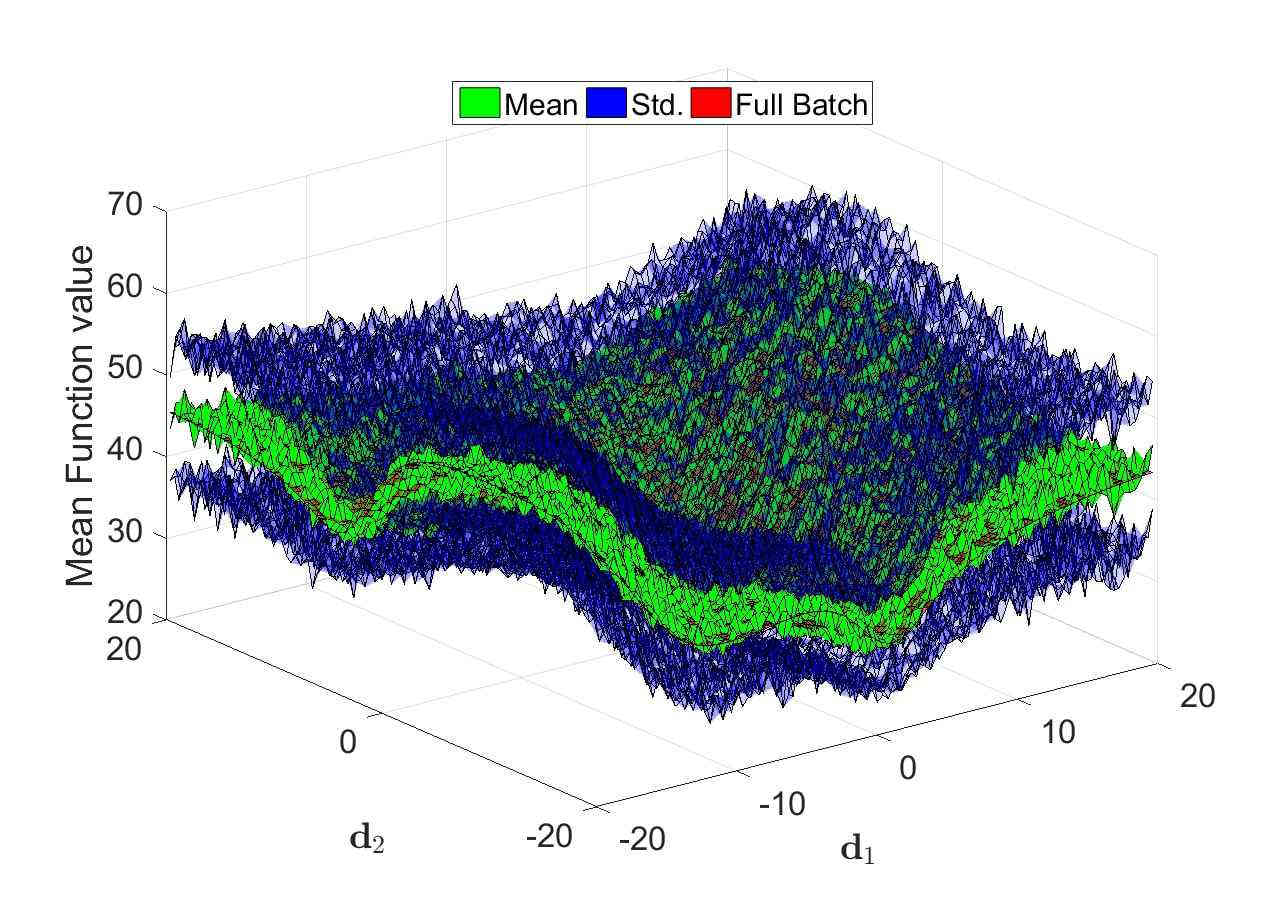}
		\caption{$\mathbb{E}[\tilde{L}(\boldsymbol{x})]$ and std.}
	\end{subfigure}%
	\begin{subfigure}{.45\textwidth}
		\centering
		\includegraphics[width=1\linewidth]{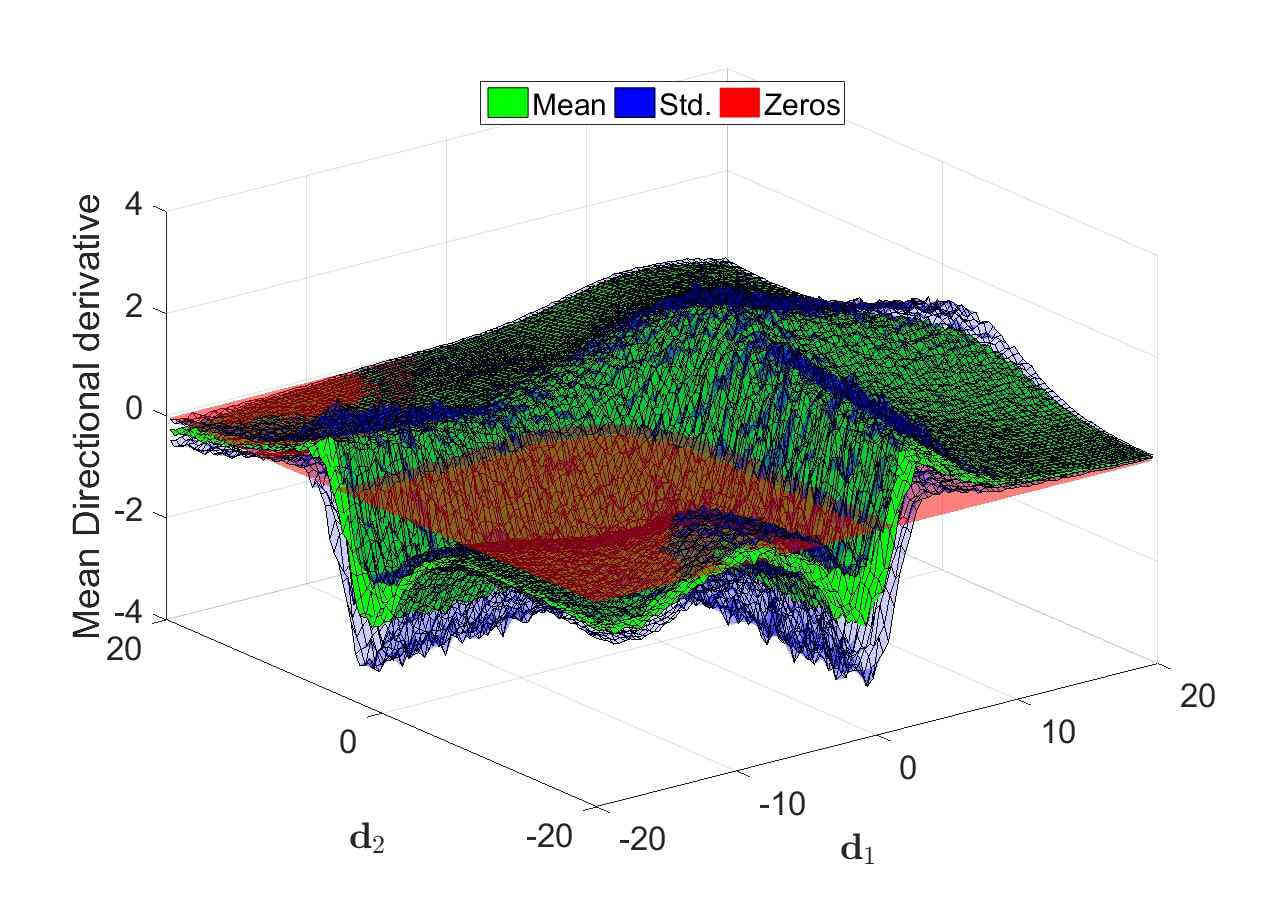}
		\caption{$\mathbb{E}[\tilde{D}_d(\boldsymbol{x})]$ and std.}
	\end{subfigure}%
	
	\begin{subfigure}{.45\textwidth}
		\centering 
		\includegraphics[width=1\linewidth]{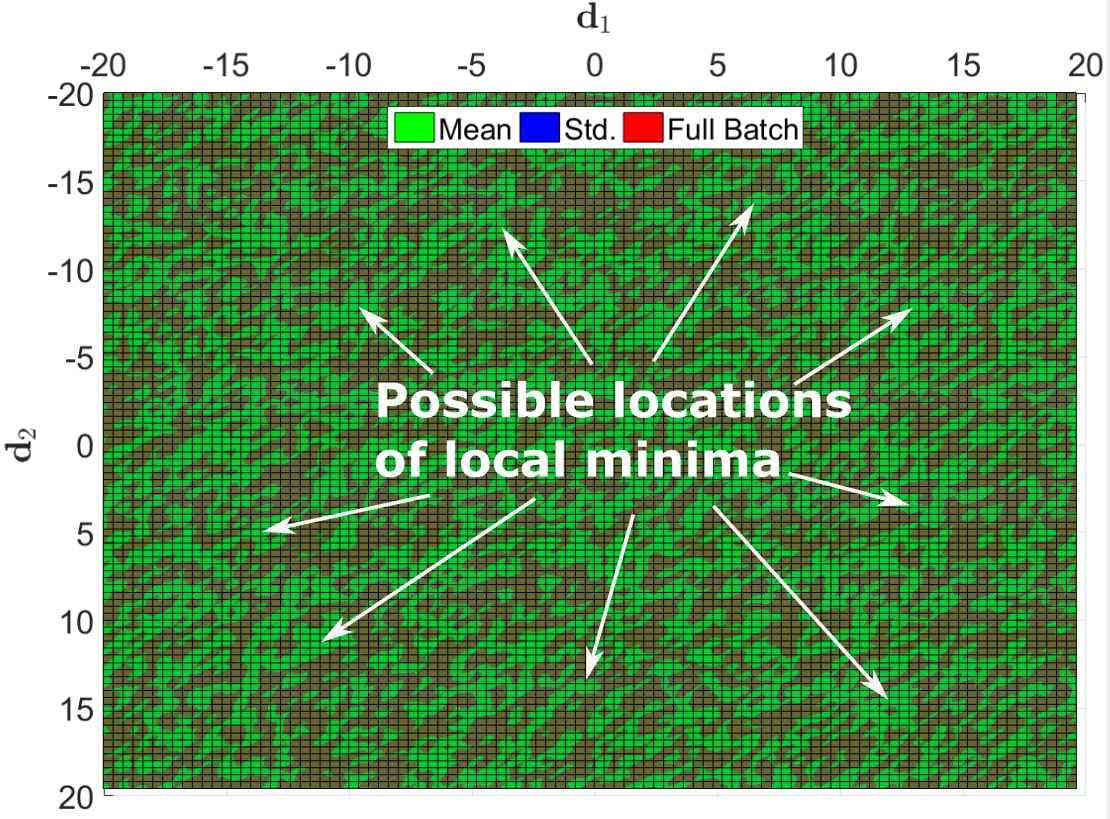}
		\caption{Spatial spread of local minima}
	\end{subfigure}%
	\begin{subfigure}{.45\textwidth}
		\centering
		\includegraphics[width=1\linewidth]{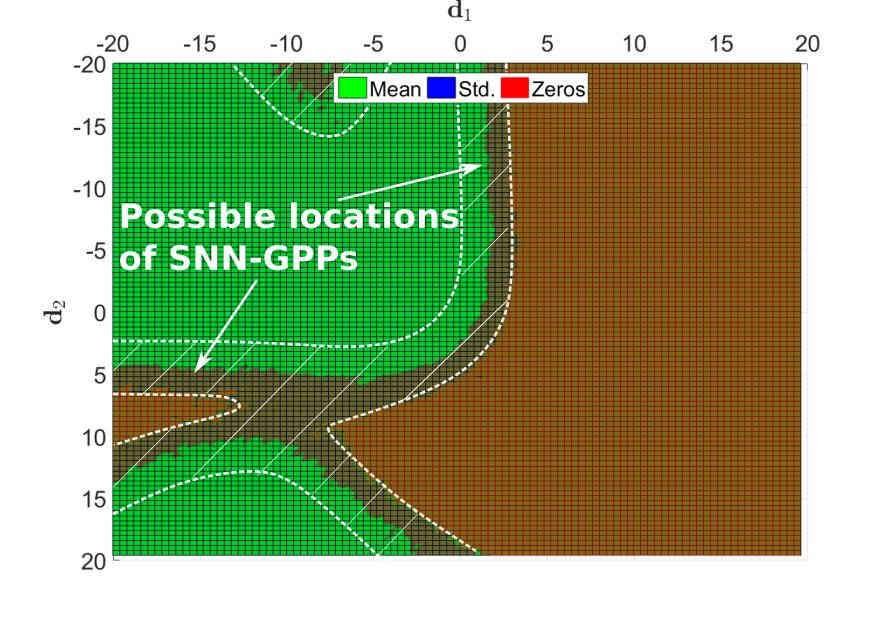}
		\caption{Spatial localization of SNN-GPPs}
	\end{subfigure}%
	
	\begin{subfigure}{.45\textwidth}
		\centering 
		\includegraphics[width=1\linewidth]{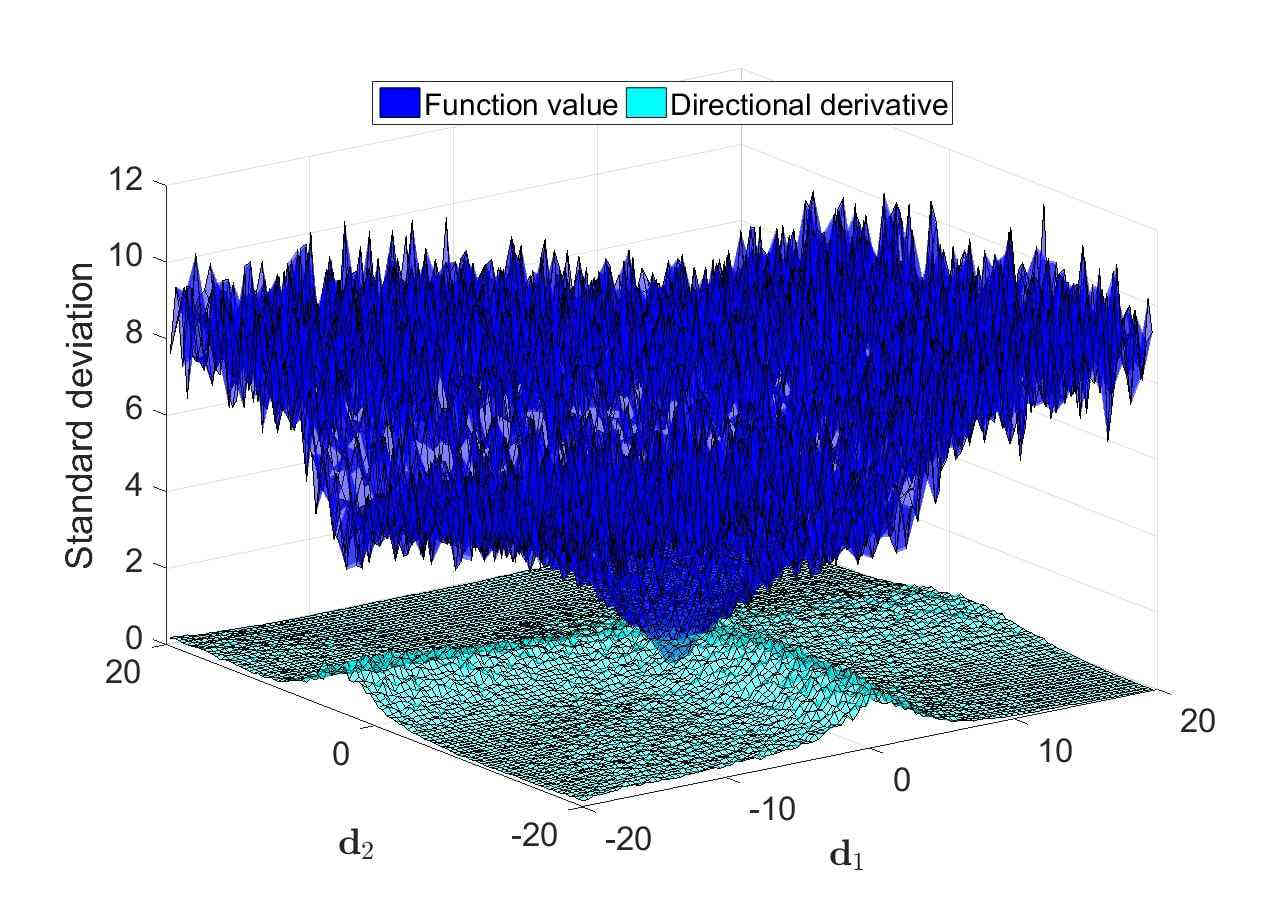}
		\caption{Side view of std. comparison}
	\end{subfigure}%
	\begin{subfigure}{.45\textwidth}
		\centering
		\includegraphics[width=1\linewidth]{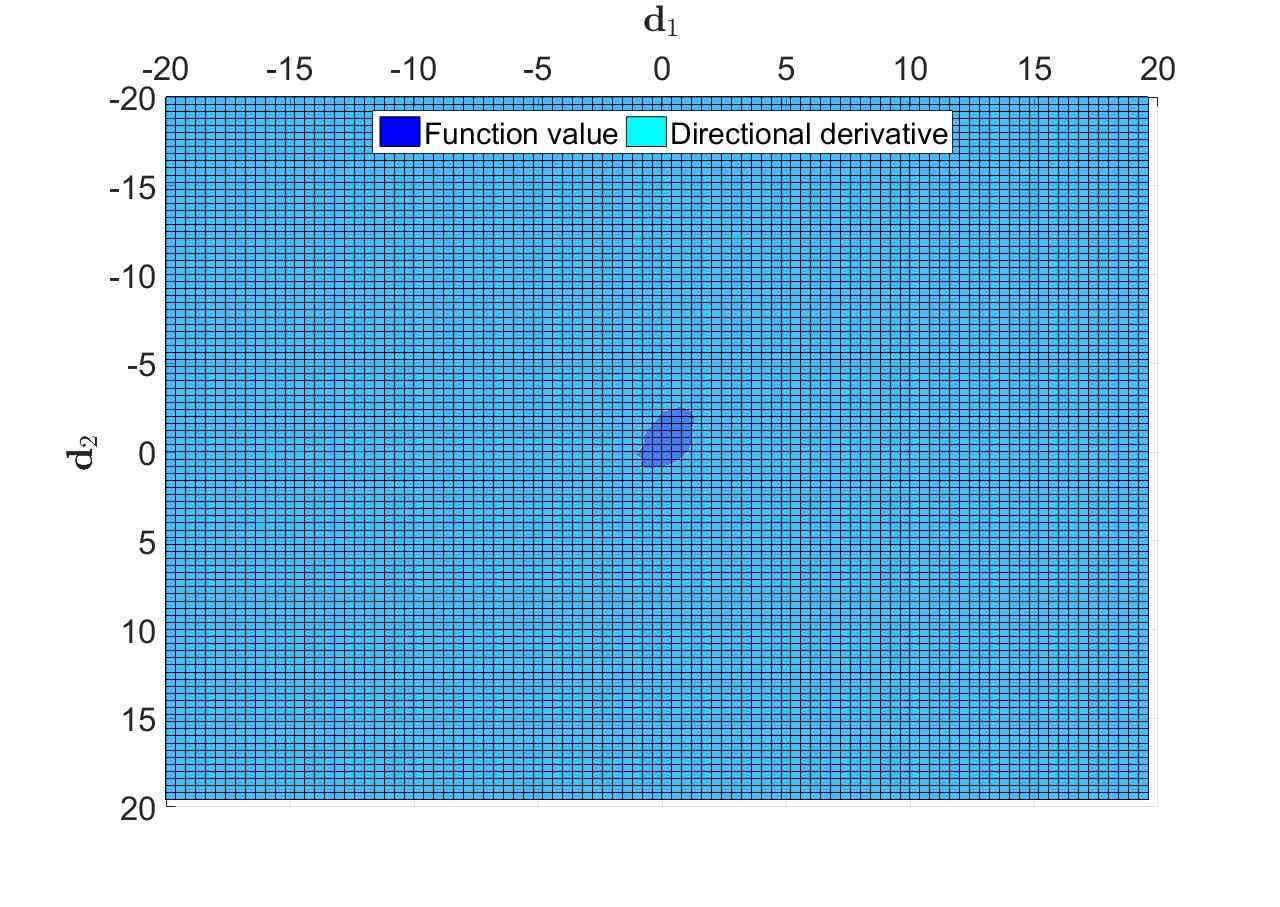}
		\caption{Lower view of std. comparison}
	\end{subfigure}%
	
	\caption{Estimated mean and standard deviation plots of (a) $\tilde{L}(\boldsymbol{x})$ and (b) $\tilde{\boldsymbol{g}}(\boldsymbol{x})$. In (a) $\mathcal{L}(\boldsymbol{x})$ is given in red, while in (b) the red plane denotes $d_d(\boldsymbol{x})=0$, to visually aid locating sign changes. Viewing these plots from below (c,d) illustrates the spatial spread of potential optima. Instances where $\mathbb{E}[\tilde{L}(\boldsymbol{x})]$ dips below $\mathcal{L}(\boldsymbol{x})$ (green areas) in (c) denote potential local minima. In (d) the green regions denote $\tilde{D}_d(\boldsymbol{x})<0$ and red areas $\tilde{D}_d(\boldsymbol{x})>0$ respectively. The white shaded areas in (d) are areas of uncertainty, where SNN-GPPs can occur. SNN-GPPs are highly localized, while minima are spread uniformly across the sampled domain. Isolating (d) $\sigma(\tilde{L}(\boldsymbol{x}))$ and (f) $\sigma(\tilde{D}_d(\boldsymbol{x}))$ surfaces, we observe a practical confirmation the discussion in Section \ref{sec_apply2sig}.}
	\label{fig_sig_vars}
\end{figure}

Subsequently, we explicitly evaluate the means and variances of function value and directional derivative surfaces by considering the same Iris dataset problem as discussed in Sections \ref{sec_statvdyn} and \ref{sec_apply2sig}. The mean and variance estimates were calculated using 50 independent draws of $|\mathcal{B}_{n,i}| = 10$ samples at every one of the 100x100 grid points. In Figure~\ref{fig_sig_vars}(a), the full batch function value surface, $\mathcal{L}(\boldsymbol{x})$, is shown in red, which allows for the comparison between $\mathcal{L}(\boldsymbol{x})$ and the estimated mean, $\mathbb{E}[\tilde{L}(\boldsymbol{x})]$ (green). If $\mathbb{E}[\tilde{L}(\boldsymbol{x})]$ (green) dips below $\mathcal{L}(\boldsymbol{x})$ (red), given the high variance, it is likely that a local minimum may occur at the given location. Conversely, if $\mathbb{E}[\tilde{L}(\boldsymbol{x})]$ lies above $\mathcal{L}(\boldsymbol{x})$, a spurious maximum is more likely to occur. 

We therefore show the function value plot from below in Figure~\ref{fig_sig_vars}(c), such that areas, where $\mathbb{E}[\tilde{L}(\boldsymbol{x})]$ (green) permeate through $\mathcal{L}(\boldsymbol{x})$ (red), can be easily identified. The result is a uniformly distributed mixture of red and green, indicating that local minima are spread over the entire sampled domain. Almost all of the indicated local minima misrepresent the true minimum of $\mathcal{L}(\boldsymbol{x})$. This highlights the challenges that a direct minimization strategy would face in dynamic MBSS loss functions.

When considering the directional derivative plot in Figures~\ref{fig_sig_vars}(b) and (d), we are concerned with localizing sign changes from negative to positive, in order to identify SNN-GPPs. Hence, we plot the zero plane in red as a visual aid for recognizing SNN-GPPs. We show Figure~\ref{fig_sig_vars}(b) from below (i.e. the negative domain) in Figure~\ref{fig_sig_vars}(d), where negative directional derivatives are presented as green and positive directional derivatives as red. The shaded domains between these two regions are uncertainty areas, where the zero plane intersects the standard deviation, but not yet the mean. Since we only depict the surface from below, the area of uncertainty is essentially double the size of that which is dark brown. Therefore, we extended this area visually with a white dotted line in Figure~\ref{fig_sig_vars}(d). The total encased area represents locations, where the probability of encountering a SNN-GPP is high. Immediately, there is a stark contrast between Figures~\ref{fig_sig_vars}(c) and (d): The areas of possible SNN-GPP occurrence are spatially localized, while the locations of possible local minima are uniformly distributed in the sampled domain. We remind the reader at this point, that both these plots represent the same problem, yet the ability to localize optima between the different optimization formulations is distinct. Searching for a solution in the form of a SNN-GPP is more representative of the true loss function, than minimizing the discontinuous loss function. The implications for optimization and line searches are clearer when considering the loss function along a 1-D search direction, as we will demonstrate in the next section.

\subsection{Optimality criteria along 1-D search directions}

As a number of optimization approaches used in neural network training make use of 1-D directional information \cite{Robbins1951,Nesterov1983,Mahsereci2017a}, we consider the behaviour of local minimum and SNN-GPP definitions along four selected hypothetical 1-D search directions in the 2-D sampled domain. This explores the applicability of these optimality criteria within line searches in dynamic MBSS loss functions. Note, that the euclidean norm, $\|. \|_2$ is indicated using $|.|$ from this section onwards in the interest of compactness.

\begin{figure}[h!]
	\centering
	\includegraphics[width=0.4\linewidth]{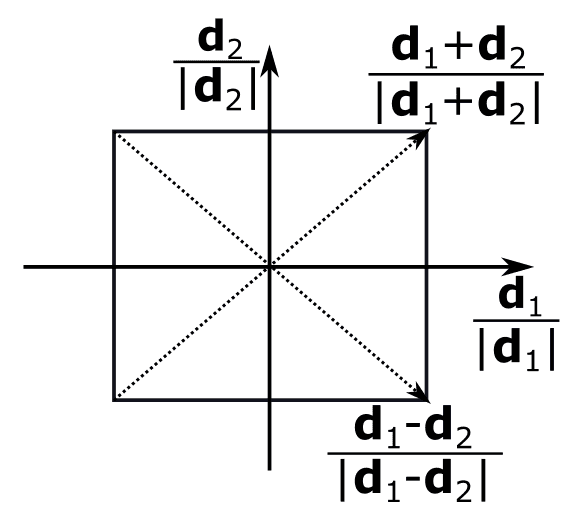}		
	\caption{Diagram depicting the investigated 1-D directions.}
	\label{fig_dir_diags}
\end{figure}

The chosen directions are the unit vectors along the uncoupled axes, $\frac{\boldsymbol{d}_1}{|\boldsymbol{d}_1|} $ and $\frac{\boldsymbol{d}_2}{|\boldsymbol{d}_2|} $, as well as the diagonals of the chosen axis directions, $ \frac{\boldsymbol{d}_1 + \boldsymbol{d}_2}{|\boldsymbol{d}_1 + \boldsymbol{d}_2|}$, $\frac{\boldsymbol{d}_1 - \boldsymbol{d}_2}{|\boldsymbol{d}_1 - \boldsymbol{d}_2|}$, shown in Figure~\ref{fig_dir_diags}. These directions are arbitrarily chosen in order to explore examples of univariate loss function characteristics that might be encountered in line searches. Typically, these directions are given by an optimization algorithm. If the domain range of the 2-D plots is $[r_{min},r_{max}]$ in both $\frac{\boldsymbol{d}_1}{|\boldsymbol{d}_1|} $ and $\frac{\boldsymbol{d}_2}{|\boldsymbol{d}_2|} $, we sample $\tilde{L}(\boldsymbol{x})$ and $\tilde{D}_d(\boldsymbol{x})$ in $n=\{1,2,\dots 100\}$ discrete steps, $\alpha = \frac{r_{max}-r_{min}}{n_{max}}$, along $\frac{\boldsymbol{d}_1}{|\boldsymbol{d}_1|}$ and $\frac{\boldsymbol{d}_2}{|\boldsymbol{d}_2|}$ and correspondingly, $ \alpha = \sqrt{2} \cdot \frac{r_{max}-r_{min}}{n_{max}} $ along $ \frac{\boldsymbol{d}_1 + \boldsymbol{d}_2}{|\boldsymbol{d}_1 + \boldsymbol{d}_2|}$ and $\frac{\boldsymbol{d}_1 - \boldsymbol{d}_2}{|\boldsymbol{d}_1 - \boldsymbol{d}_2|}$, while exhaustively counting the number of local minimizers and SNN-GPPs along each direction. Minima are counted using Definition \ref{def_locmin}, by comparing the function values of 3 adjacent discrete points, $\tilde{L}(\boldsymbol{x}_{n})$, $\tilde{L}(\boldsymbol{x}_{n+1})$ and $\tilde{L}(\boldsymbol{x}_{n+2})$ respectively. If $\tilde{L}(\boldsymbol{x}_{n}) > \tilde{L}(\boldsymbol{x}_{n+1}) < \tilde{L}(\boldsymbol{x}_{n+2})$, a local minimum is identified and counted. SNN-GPPs are counted when a sign change in directional derivative $\tilde{D}_d(\boldsymbol{x})$ from negative, i.e. $\tilde{D}_d(\boldsymbol{x}_n)<0$, to positive, i.e. $\tilde{D}_d(\boldsymbol{x}_{n+1})>0$, is observed, along the search direction. The SNN-GPP is indicated at the positive sample, $\tilde{D}_d(\boldsymbol{x}_{n+1})$, for consistency sake. The total number of candidate optima along all four directions is counted and given in the legend of the plots. We also show histograms of the spatial distribution of the minimizers and SNN-GPPs along the respective search directions.

\begin{figure}[h!]
	\centering
	\begin{subfigure}{.5\textwidth}
		\centering 
		\includegraphics[width=1\linewidth]{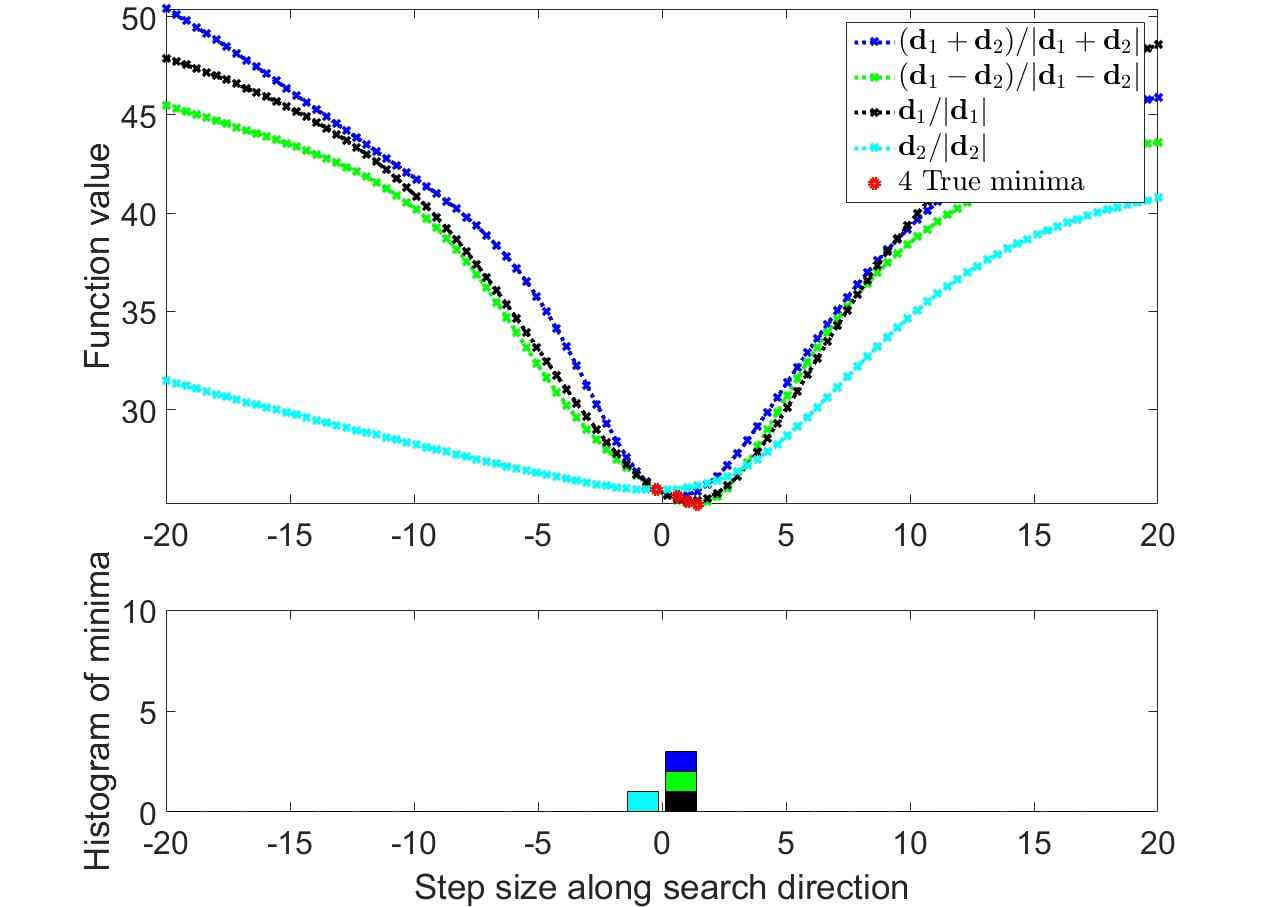}
		\caption{}
	\end{subfigure}%
	\begin{subfigure}{.5\textwidth}
		\centering
		\includegraphics[width=1\linewidth]{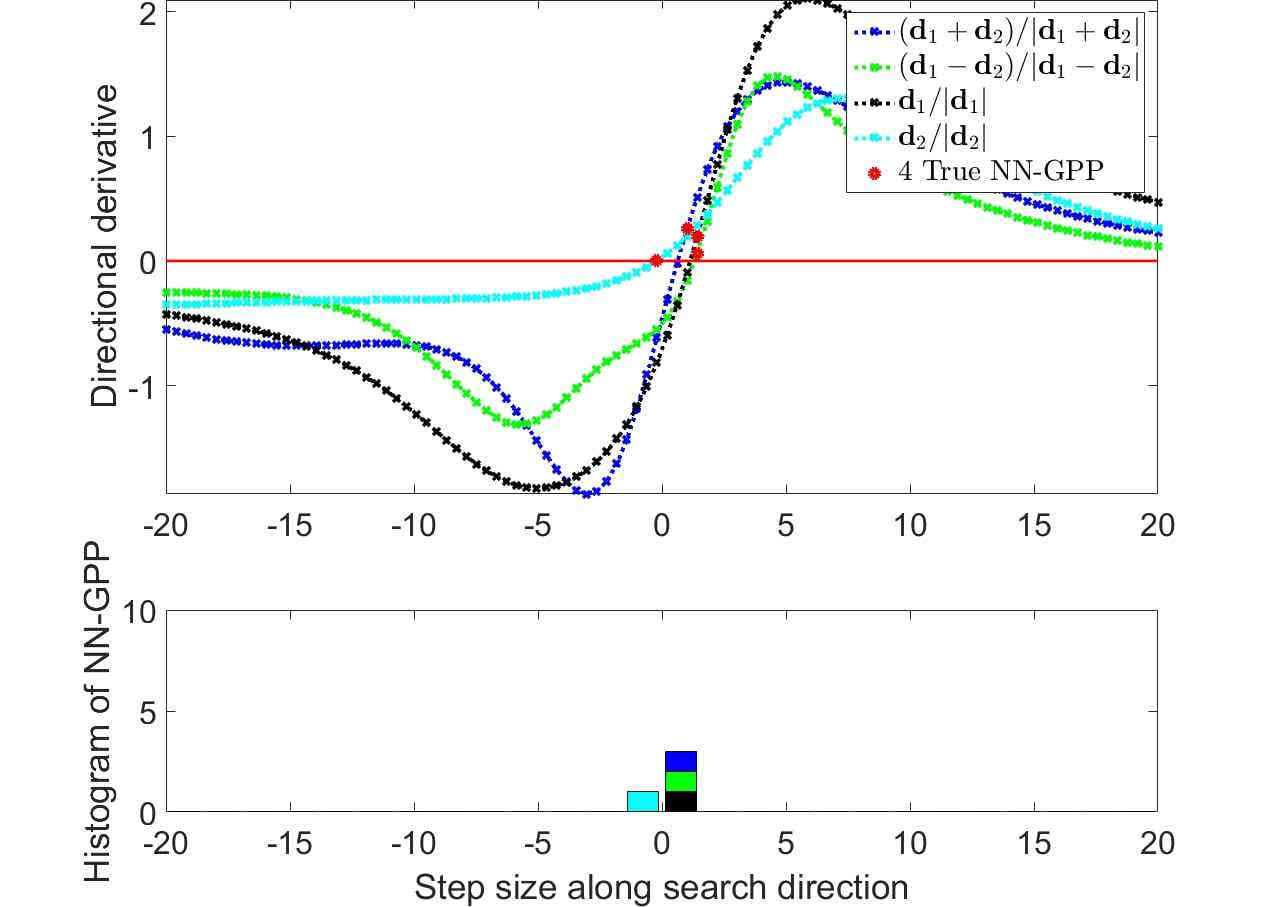}
		\caption{}
	\end{subfigure}%
	
	\begin{subfigure}{.5\textwidth}
		\centering 
		\includegraphics[width=1\linewidth]{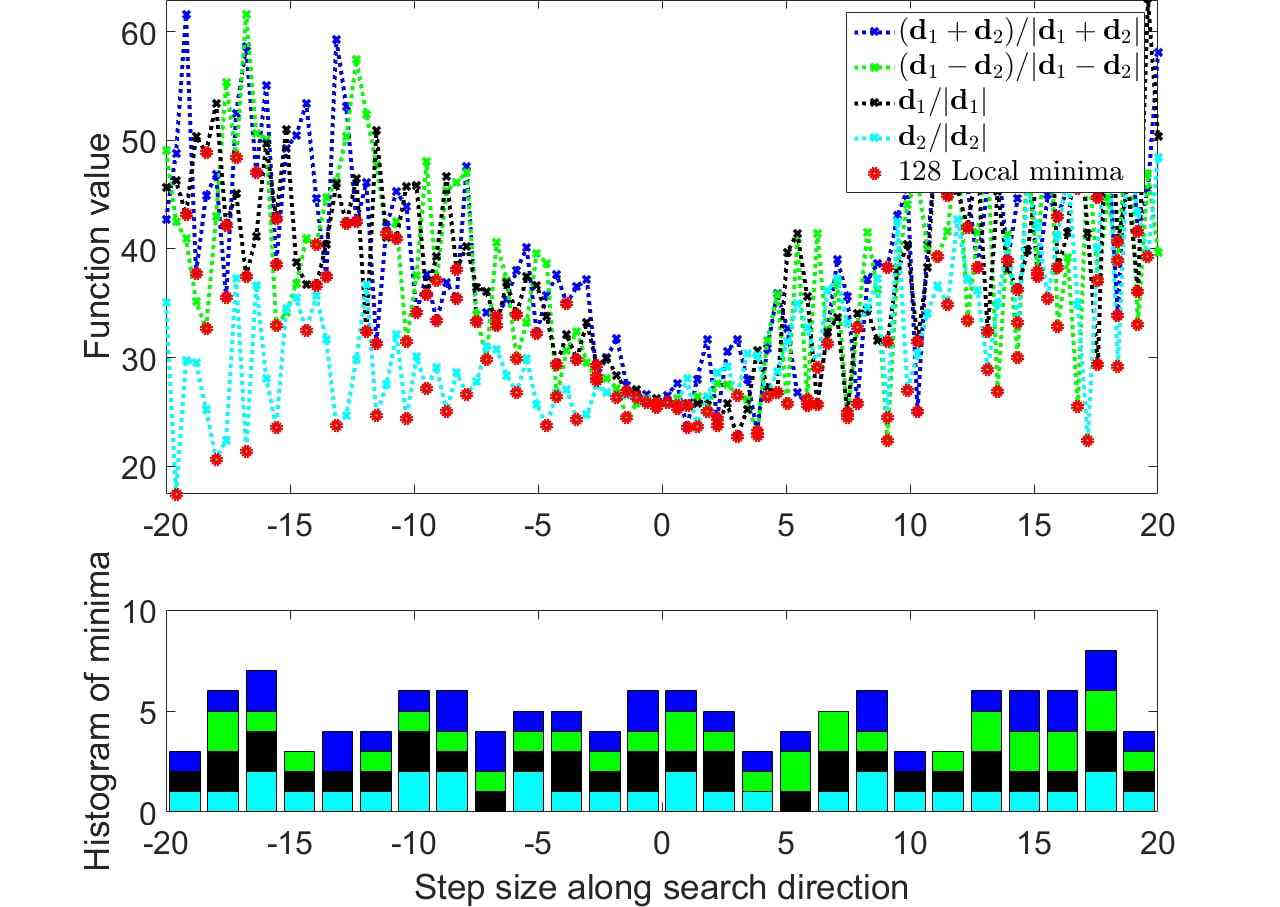}
		\caption{}
	\end{subfigure}%
	\begin{subfigure}{.5\textwidth}
		\centering
		\includegraphics[width=1\linewidth]{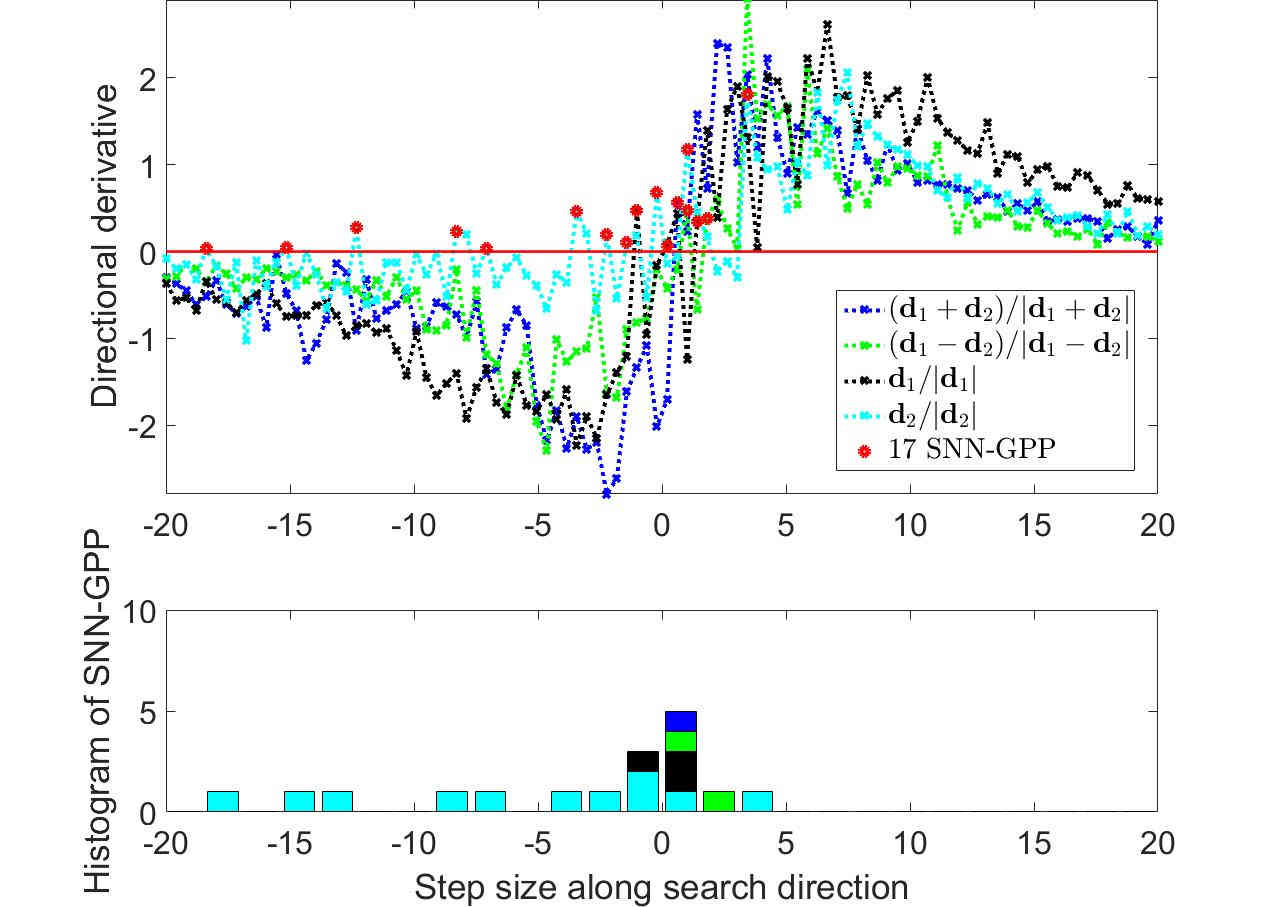}
		\caption{}
	\end{subfigure}%
	
	\caption{1-D plots of the (a) function value and (b) directional derivative along fixed directions of the 2-D domains. We compare two states, using the full-batch loss with $M=150$, and dynamic MBSS loss with, $ |\mathcal{B}_{n,i}| = 10$. We note the number local minima and SNN-GPP in either case according to the respective optimality criteria. Again, the locations of the local minima are uniformly distributed along the sampled domains for all directions. In the case of the directional derivative, directions that contain high curvature, such as $\boldsymbol{d}_1$, have highly localized sign changes. Direction $\boldsymbol{d}_2$ progresses along the area of uncertainty in Figure~\ref{fig_sig_vars} for some of its domain, which makes the location of SNN-GPP more widespread.}
	\label{fig_sig_lines}
\end{figure}

The true optima of the 1-D curves are indicated by the full batches in Figure~\ref{fig_sig_lines}(a) and (b), where we see equivalence between true minima and true NN-GPPs for each respective search direction. As was the case in the 2-D plots of Figure~\ref{fig_sig_vars}, when dynamic MBSS is implemented, the spatial distribution of local minima is spread uniformly along the search directions, as shown in Figure~\ref{fig_sig_lines}(c). The total number of local minima counted over the 4 search directions is 128. This means that a minimization line search {is exposed to} 124 potential false {candidate solutions} across the chosen search directions. { Arguably, the sensitivity of a minimization line search to the number of candidate solutions depends on its particular algorithmic formulation \cite{Arora2011}. A significant probability exists, that a discontinuity occurs at a given loss function evaluation in dynamic MBSS. Therefore, every increment (step) within a line search poses the risk of exposing a spurious minimum. However, not all local minima might be accepted by a line search, especially when inexact line searches are conducted. The well known Armijo and Wolfe conditions \cite{Arora2011}, as often used in inexact line searches, provide rules which can reduce the range of candidate minima to be accepted. Nevertheless, the local minimum definition provides a large number of candidate optima within a given formulation's acceptable range, which the particular formulation is required to contend with, reducing its effectiveness \cite{Kafka2019jogo}.} In turn, the spatial distribution of SNN-GPPs much more localized around the true optima as shown in Figure~\ref{fig_sig_lines}(d). The direction with the highest number of SNN-GPPs is $\boldsymbol{d}_2$, since the first half of $\boldsymbol{d}_2$ runs along an area of uncertainty for SNN-GPP, seen in Figure~\ref{fig_sig_vars}(d). Due to the low magnitude in directional derivative and the correspondingly high variance in the sampled gradient values, it is possible to encounter spurious sign changes along this ridge. The remaining three directions show SNN-GPPs to be are tightly clustered around the true optimum, indicating small spatial domains of uncertainty. This means that a line search strategy that looks for SNN-GPPs would be significantly more effective in approximating the location of the true optimum.

Subsequently, we use the tools developed to investigate the Sigmoid AF, to extend our study to a number of popular activation functions, investigating their effect on finding minimizers and SNN-GPPs as well as observing their variance around the true optimum along a search direction. 

\section{Alternative Activation Functions}
\label{sec_AltActs}

\begin{figure}[h]
	\centering
	\begin{subfigure}{.5\textwidth}
		\centering 
		\includegraphics[width=0.9\linewidth]{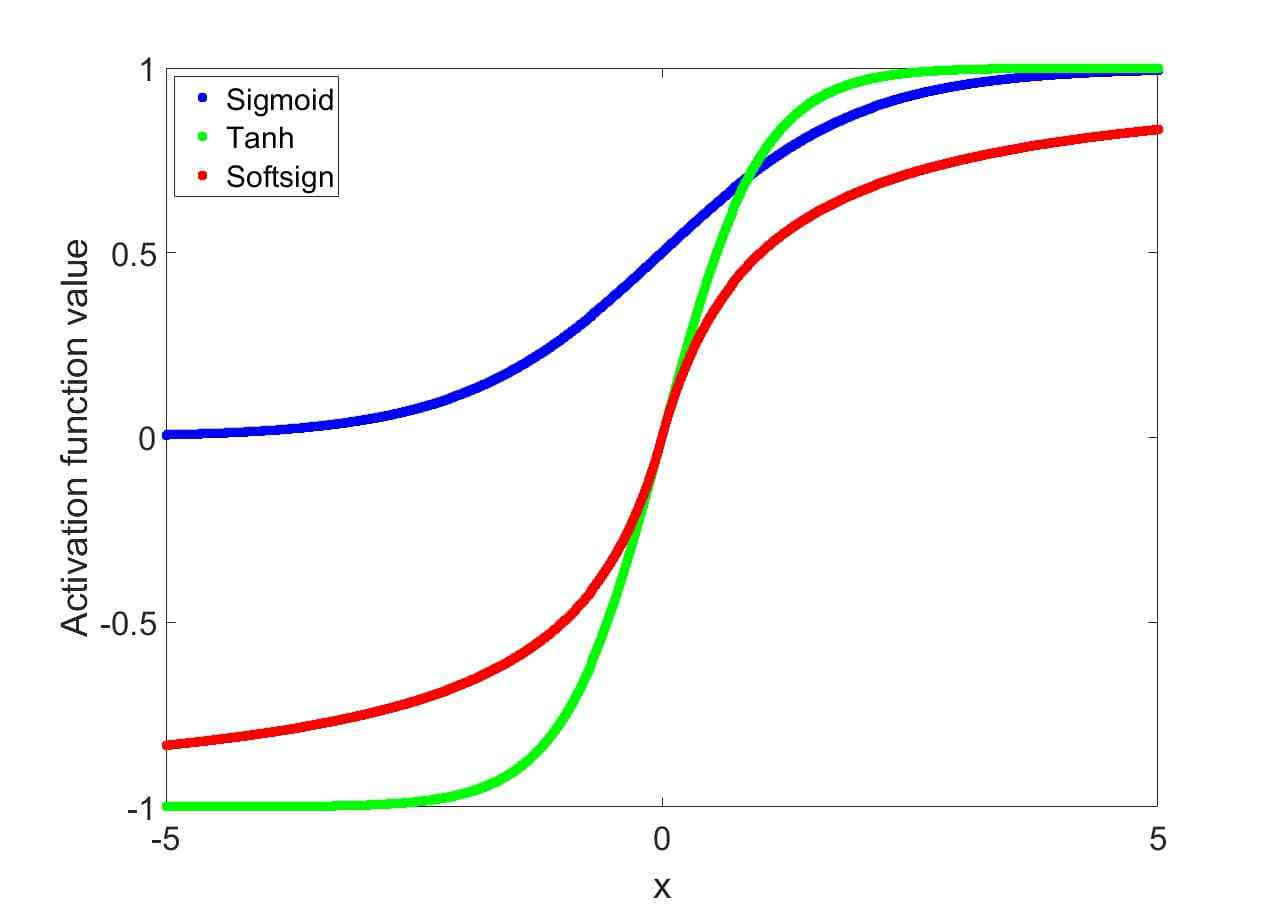}
		\caption{Saturation class function values}
		\label{fig_acts_g1_f}
	\end{subfigure}%
	\begin{subfigure}{.5\textwidth}
		\centering
		\includegraphics[width=0.9\linewidth]{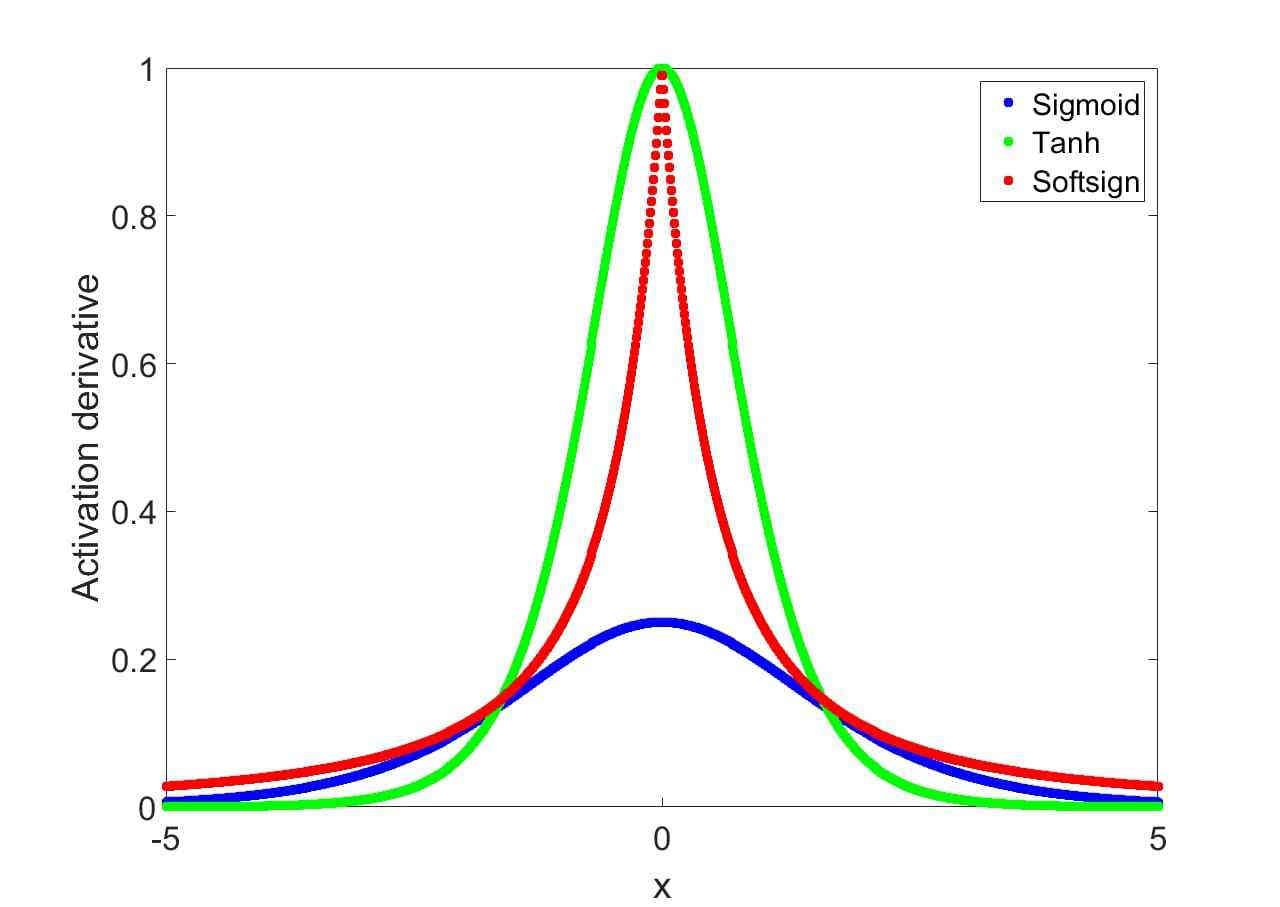}
		\caption{Saturation class derivatives}
		\label{fig_acts_g1_g}
	\end{subfigure}%
	
	\begin{subfigure}{.5\textwidth}
		\centering 
		\includegraphics[width=0.9\linewidth]{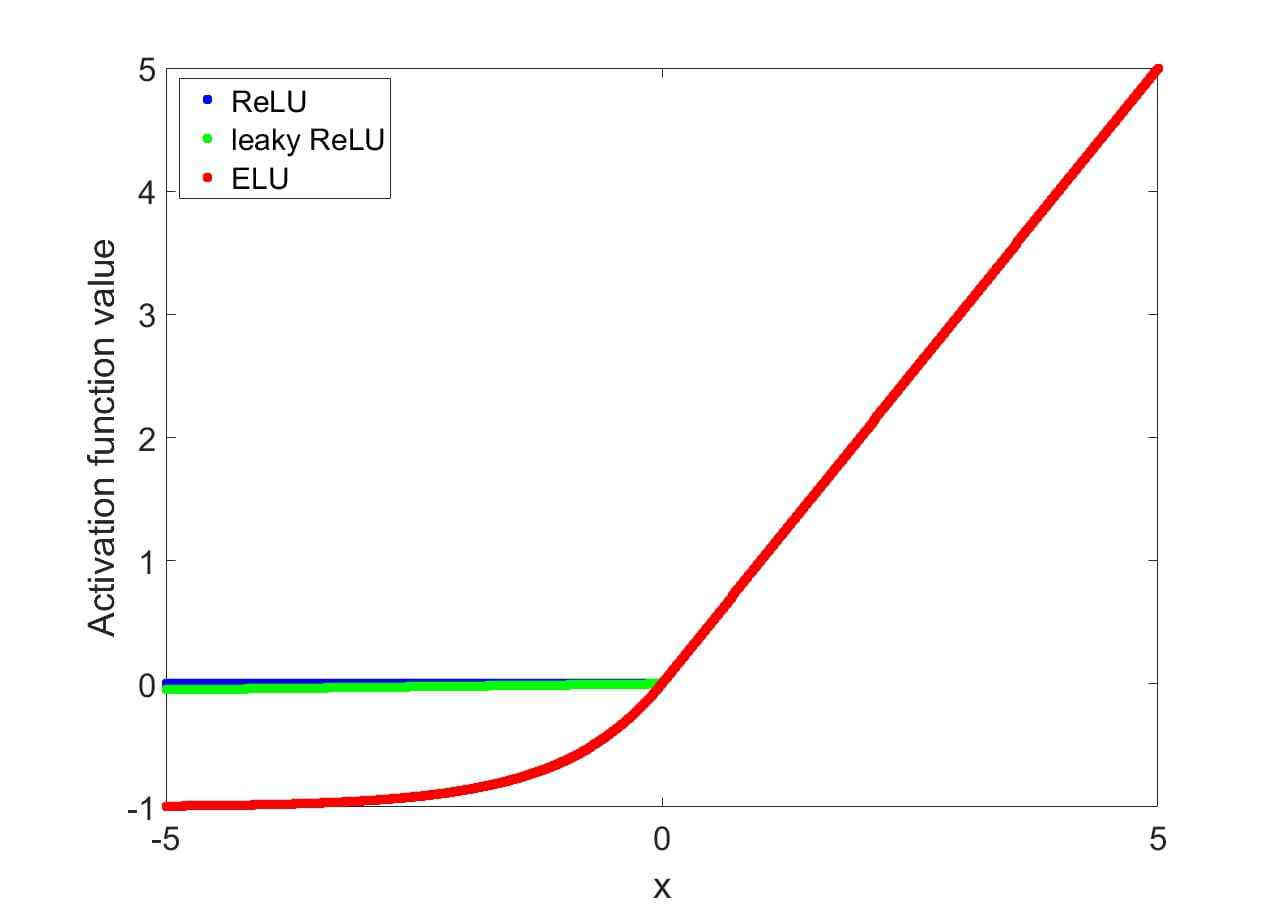}
		\caption{Sparsity class function values}
		\label{fig_acts_g2_f}
	\end{subfigure}%
	\begin{subfigure}{.5\textwidth}
		\centering
		\includegraphics[width=0.9\linewidth]{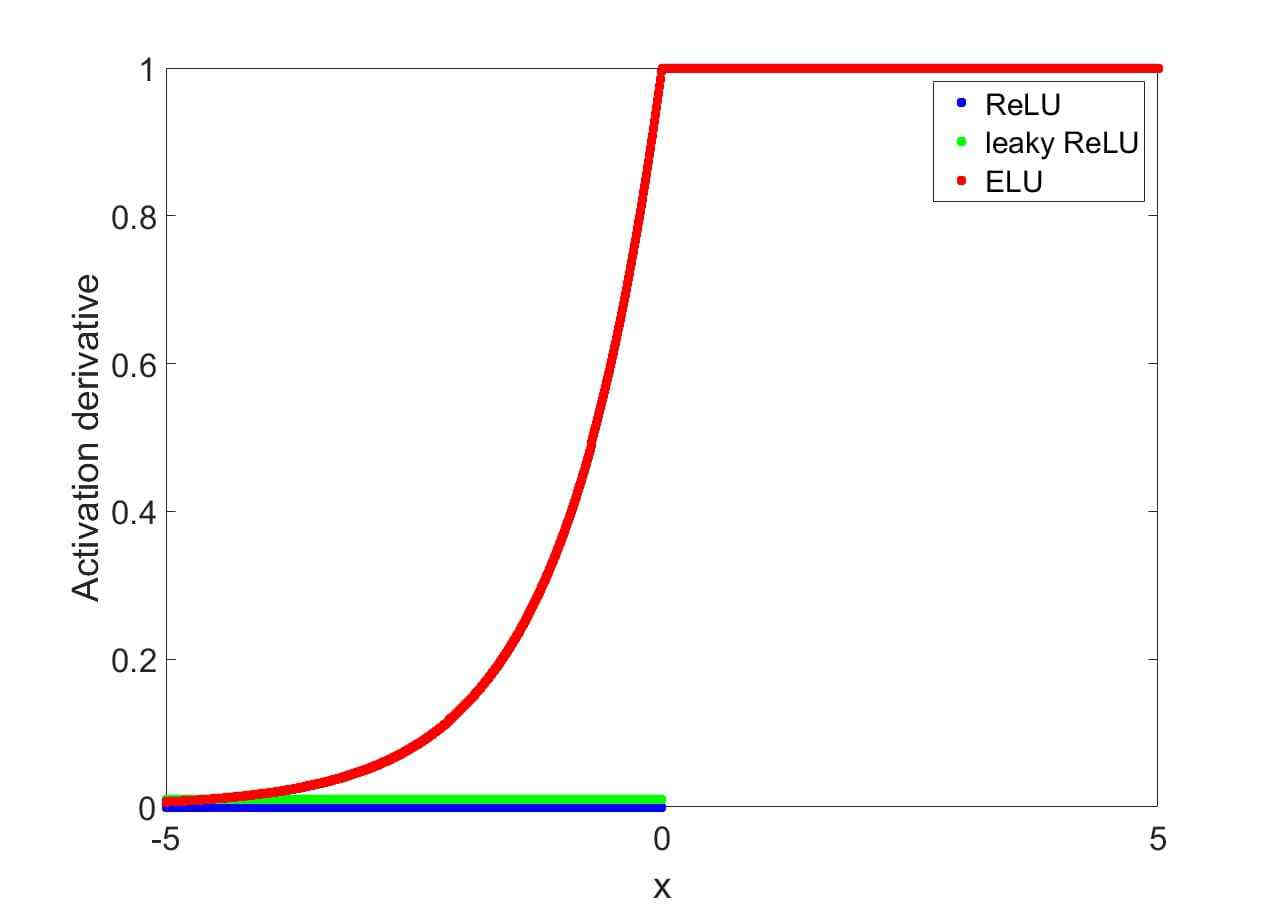}
		\caption{Sparsity class derivatives}
		\label{fig_acts_g2_g}
	\end{subfigure}%

	\caption{(a,c) Function value and (b,d) derivatives of activation functions considered in our investigations. These are grouped together into (a,b) saturation and (c,d) sparsity classes respectively. The saturation class makes use of the non-linearities of the activation functions to construct mappings. The sparsity class makes use of the architecture in the network to introduce the non-linearities, where specific "channels" in the network are selectively activated, depending on whether the activation function transmits information or not.}
	\label{fig_acts}
\end{figure}

Though Sigmoid activations were popular in the past, other smooth activations, such as Tanh \cite{Karlik2015} and Softsign \cite{Bergstra2009}, are often preferred. Collectively, this type of AF has been called the saturation class of AFs \cite{Xu2016}. Today, the state of the art networks use AFs that promote sparsity, namely the ReLU \cite{Glorot2011} family of activations. We include leaky ReLU \cite{Maas2013} and ELU \cite{Clevert2016} into this category and refer to this family as the sparsity class of AFs. Both these classes are plotted in Figure~\ref{fig_acts} for comparison.

As their name suggests, the dominant characteristic of the saturation class functions is the stagnating, plateau-like behaviour at in both negative and positive extremes of the x-domains. The differences within the saturation class concentrate primarily on the range over which the transition from saturation to the active domain (where the gradient is high) and back to saturation occurs. Activation functions with small "active" ranges are noted as having high saturation characteristics, while those that have higher gradients over a larger domain are said to have low saturation characteristics. The active domains are centred around $x=0$, but not all run through the origin. The Sigmoid AF is only positive, while Tanh and Softsign pass through the origin and allow both negative and positive outputs. As we will demonstrate, the differences between the activation functions, though slight in their elemental form, can lead to significant changes in loss function characteristics.

The sparsity class have fundamentally different characteristics to the saturation class in that they allow a linear relationship in their "activated" positive domain, while enforcing or approximating an "off" characteristic in the negative domain. The ReLU activation function enforces a "hard off", setting the function value and derivative to 0 for the negative x-domain. The leaky ReLU allows a small constant positive derivative in the negative domain of $x$. This allows a small amount of information to permeate the network when the leaky ReLU is in its "off" state. The transition to the positive domain $x$ is discontinuous in the derivative, which like ReLU enforces a drastic change in characteristics within the loss function. The ELU formulation constructs the derivative to be smooth and continuous, in the form of an exponential, when transitioning from negative to positive along $x$.

\subsection{Global activation function characteristics in random directions over [-20,20] domains in weight space}
\label{sec_global}

We now investigate the effect of the above discussed AFs on the distribution of minimizers and SNN-GPPs. To allow for a direct comparison between AF-related loss function characteristics, we use the same problem and architecture as explored with Sigmoids in Section \ref{sec_SigsConcepts}, with the only change being the AFs used in the neural network. The results are summarized in Table~\ref{tbl_AFSum}. From Table~\ref{tbl_AFSum} it is evident that the number of SNN-GPPs is much closer to that of the true local minima, in comparison to the number of local minimizers for all activation functions. In general, there are 12 to 32 times more identified local minimizers than true optima, whereas there are only 1.5 to 6 times more SSN-GPPs than true optima. The number SSN-GPPs is the closest to that of the true optima for the Softmax and ELU AFs, while Sigmoid and Tanh AFs resulted in the largest estimation of SSN-GPPs compared to true optima.

The spatial distribution of the minimizers and SNN-GPPs are given in Appendix A as histogram plots. In the histogram plots, since SNN-GPPs are indicated as being after the sign change in our analyses, cases can occur where the location of the true optimum is close to edge of a bin in the histogram. In such instances the location of the SNN-GPP might be noted in the bin that is adjacent of that indicated for the minima. This is not an error, but simply an artefact of how local minima and SNN-GPP are identified from discrete samples of the loss function and directional derivatives respectively.

\begin{table}[h!]
	\centering
	\begin{tabularx}{\textwidth}{|p{14mm}|c|p{13mm}|p{14mm}|p{10mm}|X|}
		\hline 
		\textbf{AF} & \textbf{Figure} & \textbf{\# True} \newline \textbf{Optima} & \textbf{\#} \newline \textbf{Local} \newline
		\textbf{Minima} & \textbf{\#} \newline \textbf{SNN-GPPs}  & \textbf{Comments}  \\ 
		\hline 
		\textbf{Sigmoid} & \ref{fig_sig}, \ref{fig_sig_lines} & 4 & 128 & 17 & Smooth features, spurious SNN-GPPs in low curvature directions. \\ 
		\hline 
		\textbf{Tanh} & \ref{fig_tan} & 4 & 126 & 23 & Higher curvature; SNN-GPPs more localized, but high saturation, leading to spurious SNN-GPPs at edges. \\ 
		\hline 
		\textbf{Softsign} & \ref{fig_soft} & 4 & 117 & 6 & More curvature than Sigmoid, less than Tanh; more curvature at origin gives better localization; less saturation at ends means less spurious SNN-GPPs at edges. \\ 
		\hline 
		\textbf{ReLU} & \ref{fig_relu} & 8 & 104 & 13 & Exponential behaviour in MSE loss; Globally convex shape, but locally intricate; more true optima; no saturation, no spurious SNN-GPPs at edges. \\ 
		\hline 
		\textbf{Leaky ReLU} & \ref{fig_lrelu} & 8 & 101 & 17 & No different to ReLU at global scale; low curvature directions have more spurious SNN-GPP due to leaky gradient (as opposed to hard-zero). \\ 
		\hline 
		\textbf{ELU} & \ref{fig_elu} & 4 & 99 & 6 & Less true optima than other ReLUs; smooth loss function features; higher gradient curvature, resulting in more localized SNN-GPPs. \\ 
		\hline 
	\end{tabularx} 
	
	\caption{Summary of observations from analyses conducted by traversing random directions over a scalar range of [-20,20] to give an impression of "global" loss function characteristics. The corresponding Figures~\ref{fig_tan}-\ref{fig_elu} are included in Appendix A.}
	\label{tbl_AFSum}
\end{table}

One of the prominent differences between the saturation and sparsity activation function classes is the behaviour on the extremities of the sampled domains. The saturation functions have "mountainous" features, in the sense that the landscape alternates between high and low curvature regions. This occurs as different nodes in the network saturate, or activate. When most but not all nodes have saturated, flat planes with low directional derivative values can be observed. This can have a negative effect on the spatial distribution of SNN-GPPs, as seen in Figures~\ref{fig_sig_lines}(d) and \ref{fig_tan}(h). In cases where saturation is high (the AF derivative tends quickly towards $0$, as $x \rightarrow \pm \infty$, Figure~\ref{fig_acts}(b)), the directional derivative magnitude is low. Hence, the variance of the directional derivative is lower than other areas of the loss function, but due to the small directional derivative magnitude, it may still create spurious SNN-GPPs. This is particularly prevalent for the Tanh activation function, see Figure~\ref{fig_tan}(h). Conversely, SNN-GPPs are highly localized for the Softsign analysis, since saturation occurs much further from $x=0$, with higher derivatives at the edges of the domain, see Figures~\ref{fig_acts}(b) and \ref{fig_soft}(h). This demonstrates that the derivative shape of the activation function matters with respect to finding SNN-GPPs.	

The sparsity class exhibits very steep behaviour at the outer limits of the domain. This is a feature related to the "on" nature of the activation functions. When most/all activations of the network nodes are in the positive domain, the input data is passed through the network and potentially amplified by larger weights. Since we implement the MSE loss for these examples, any classification error above 1 gets amplified, resulting in the aggressive increase in error. The consequence of this is a relatively convex looking loss function on a "global" scale. However, in the centre of the domain, where the error is small, a significant amount of detail is present, that is lost at this scale. It is also notable, that the number of true optima obtained when using ReLU and leaky ReLU AFs are different to the rest for the same problem. This is a clear indication, that there are unique features closer to the origin, that distinguish these activations from the rest. In search for these features, we conduct a second analysis over a more local domain in the weight space in Section \ref{sec_local}.

\subsection{Local activation function characteristics in descent directions over [-2,2] domains in weight space}
\label{sec_local}

In this section we modify the definitions of $\boldsymbol{d}_1$ and $\boldsymbol{d}_2$, to be the steepest descent direction of the full batch, $\boldsymbol{d}_1 = - \nabla\mathcal{L}(\boldsymbol{x})$ at initial starting point $\boldsymbol{x}_0$ located at (0,0), and $\boldsymbol{d}_2$ is chosen to be a random direction perpendicular to $\boldsymbol{d}_1$. We also reduce the grid range to [-2,2], to give a more detailed perspective of characteristics around local minima for different AFs. A summary of the observations made is given in Table~\ref{tbl_AF_z_Sum}. 

In this case, the number of local minima counted is 21 to 33 times higher than the number of true optima present in the loss function. In comparison, the equivalent occurrence of SNN-GPPs compared to true optima is a factor of 3 to 9. Due to the smaller relative domains, the step sizes along respective search directions are smaller, creating a higher chance of SNN-GPPs occurring close to a true optimum. This accounts for the slight increase in the ratio between counted SNN-GPPs and true optima, compared to the larger [-20,20] domains. The AF resulting in the least identified SNN-GPPs for the saturation and sparsity classes are Softsign and ELU respectively, while the corresponding worst performers are Tanh and leaky ReLU.

\begin{table}[h!]
	\centering
	\begin{tabularx}{\textwidth}{|p{14mm}|c|p{13mm}|p{14mm}|p{10mm}|X|}
		\hline 
		\textbf{AF} & \textbf{Figure} & \textbf{\# True} \newline \textbf{Optima} & \textbf{\#} \newline \textbf{Local} \newline
		\textbf{Minima} & \textbf{\#} \newline \textbf{SNN-GPPs}  & \textbf{Comments}  \\ 
		\hline 
		\textbf{Sigmoid} & \ref{fig_z_sig} & 4 & 117 & 36 & Smooth features; spurious SNN-GPP again in low curvature directions; mainly the contour direction. \\ 
		\hline 
		\textbf{Tanh} & \ref{fig_z_tan} & 4 & 130 & 15 & Higher curvature than Sigmoid; SNN-GPP fewer, more localized; contour direction generates spurious SNN-GPP over whole domain. \\ 
		\hline 
		\textbf{Softsign} & \ref{fig_z_soft} & 4 & 123 & 13 & More local curvature than Tanh; SNN-GPP highly localized. \\ 
		\hline 
		\textbf{ReLU} & \ref{fig_z_relu}, \ref{fig_z_relu_0s} & 5 & 105 & 17 & Piece-wise, multi-modal with shallow basins; directional derivatives are discontinuous; existence of flat planes when all nodes are "off" (insensitive to sub-sampling); quadratic when all nodes are "on" \\ 
		\hline 
		\textbf{Leaky ReLU} & \ref{fig_z_lrelu}, \ref{fig_z_relu_0s} & 5 & 124 & 17 & Similar features to ReLU, but "flat planes" have slight curvature\\ 
		\hline 
		\textbf{ELU} & \ref{fig_z_elu} & 4 & 121 & 11 & Smooth loss function features; uni-modal in diagonal directions; continuous directional derivative; higher gradient curvature; most localized SNN-GPP for ReLUs. \\ 
		\hline 
	\end{tabularx} 
	
	\caption{Summary of observations from analyses conducted by traversing a scalar range of [-2,2] to give an impression of "local" loss function characteristics. Here the search directions are $\boldsymbol{d}_1 = - \nabla\mathcal{L}(\boldsymbol{x})$ and $\boldsymbol{d}_2$ is a random perpendicular direction to $\boldsymbol{d}_1$. The corresponding Figures~\ref{fig_z_sig}-\ref{fig_z_elu} are shown in Appendix B.}
	\label{tbl_AF_z_Sum}
\end{table}

Considering specific AFs, the most prominent feature of the ReLU and leaky ReLU loss functions is that they do not have uniform characteristics. As is shown in Figure~\ref{fig_z_relu}(a), there are flat planes, low curvature optima, high curvature optima, and quadratic characteristics almost seemingly "stitched" together. There are up to two optima in a given search direction. Some are in very shallow basins (in the negative domains) and others are in basins of higher curvature (around 0.5 units). This "stitched" behaviour in the loss function are a result of the discontinuous nature of the AF derivative, seen in Figures~\ref{fig_z_relu}(b) and \ref{fig_z_lrelu}(b). Here, the steps indicate different nodes switching "on" and "off". The less nodes are active, the less curvature is present in the loss function. Conversely, if many nodes are simultaneously "on" with high weight values, the loss function increases quadratically. This explains the multi-modality in a given search direction, which is in stark contrast to the other activation functions, that are smooth and continuous.

Importantly, there are also domains in the ReLU loss function where the directional derivatives are exactly 0. These are domains where all nodes are "off" and no information is able to pass through the network. These areas are unaffected by mini-batch sub-sampling, exhibiting directional derivatives that are consistently 0, see Figure~\ref{fig_z_relu_0s}(a). This occurs when the network weights are such that all the incoming information from the data is pushed into the negative domain of the ReLU activation function. These are problematic areas for neural network training, since most training algorithms make use of gradient information to update $\boldsymbol{x}$. If there is no gradient information, there will be no updates to the state of the network and training comes to a premature halt. These flat planes are also present for the leaky ReLU. However, these do not have derivatives that are 0 (see Figure~\ref{fig_z_relu_0s}), thus preventing training algorithms from stagnating during optimization. The additional curvature and smooth transition of gradients given by the ELU activation function further aid in constructing smooth features in the loss function. In this analysis these characteristics have led to a lower number of local optima, and have further aided the location of SNN-GPP in dynamic MBSS loss functions (see Figure~\ref{fig_z_elu}).

For the saturation class of AFs, we see similar trends  in this analysis compared to that of the "global" domains: High curvature directions contain fewer and more localized SNN-GPPs. Low curvature directions result in a wider range of possible SNN-GPP locations and in some cases can even cause the likely location of SNN-GPP to shift. Specifically, the close-up Softsign analysis shows the disappearance of a SNN-GPP out of the edge of the sampled domain. The true NN-GPP at step size around $\pm -1.9$ moves beyond [-2,2] to fall between [-3.5,-3.1] (confirmed by multiple mini-batch analyses in this extended domain). This SNN-GPP is far from the true NN-GPP. If an optimization method were to find the solution in the [-3.5,-3.1] range, it would be a poor representation of the true optimum, though the consequence in terms of error from the model might be low, since the change in error is not high around this area. However, this underlines the argument of preferring directions of high curvature for optimization purposes.

Overall, the same general trend holds for investigations on both the global and local domains conducted in this paper: The characteristics of the optimality formulations explored in Section \ref{sec_apply2sig} hold across all considered activation functions. The number of local minima counted is often around an order of magnitude higher than the equivalent number of SNN-GPPs in dynamic MBSS loss functions. Local minima are spatially spread across the whole domain, whereas only low curvature directions exhibit a larger number of spurious SNN-GPPs. Due to their favourable derivative characteristics, the activation functions that produced the best results for finding SNN-GPPs were Softsign and ELU for their respective classes. Smooth derivatives and high curvature directions are concepts that align well with attempts to incorporate mathematical programming \cite{Arora2011} methods into machine learning \cite{Schraudolph2006,Martens2010}.

\section{Sensitivity of optimum formulations to mini-batch sample size}

\begin{figure}[h!]
	\centering	
	\begin{subfigure}{.45\textwidth}
		\centering 
		\includegraphics[width=1\linewidth]{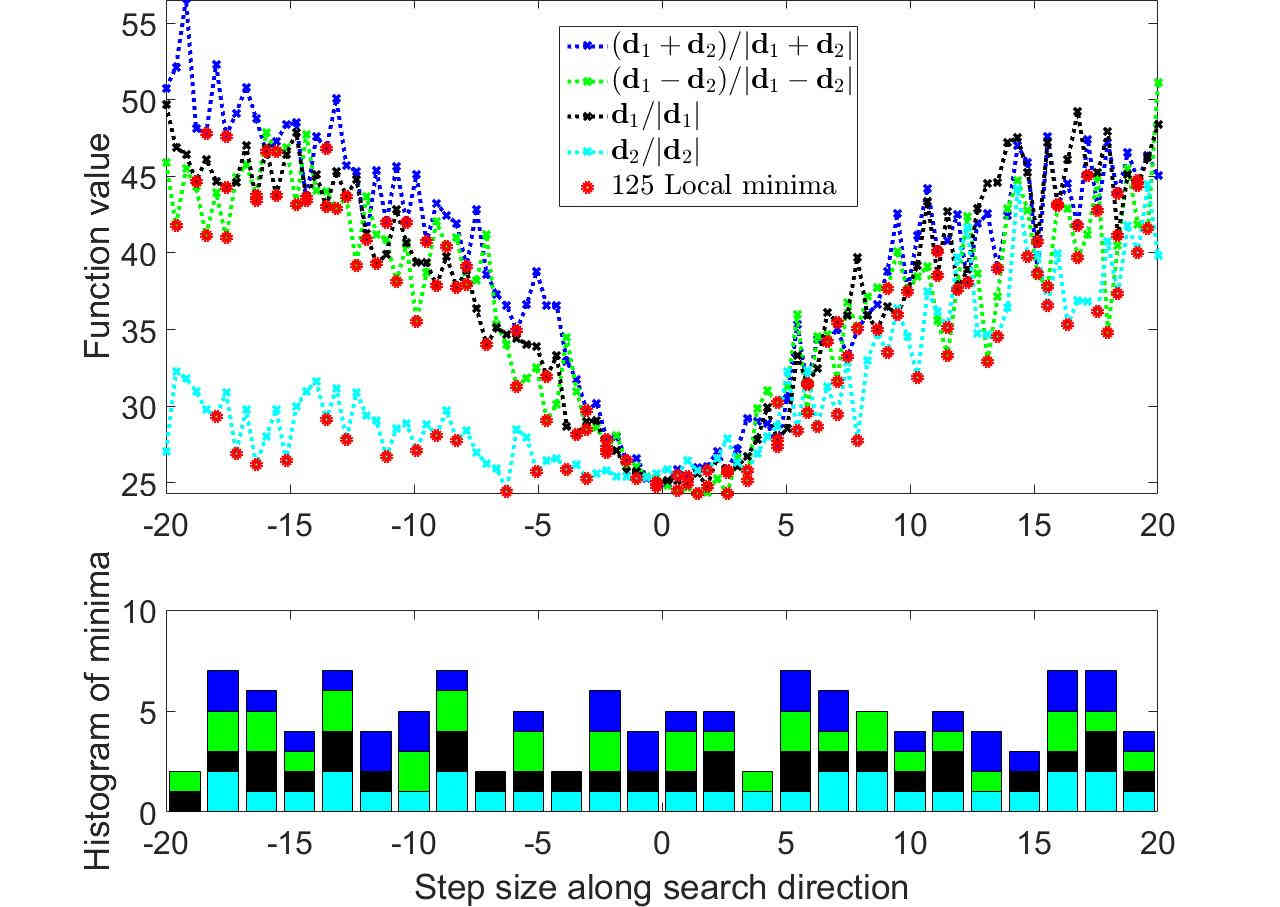}
		\caption{Function value over [-20,20]}
	\end{subfigure}%
	\begin{subfigure}{.45\textwidth}
		\centering
		\includegraphics[width=1\linewidth]{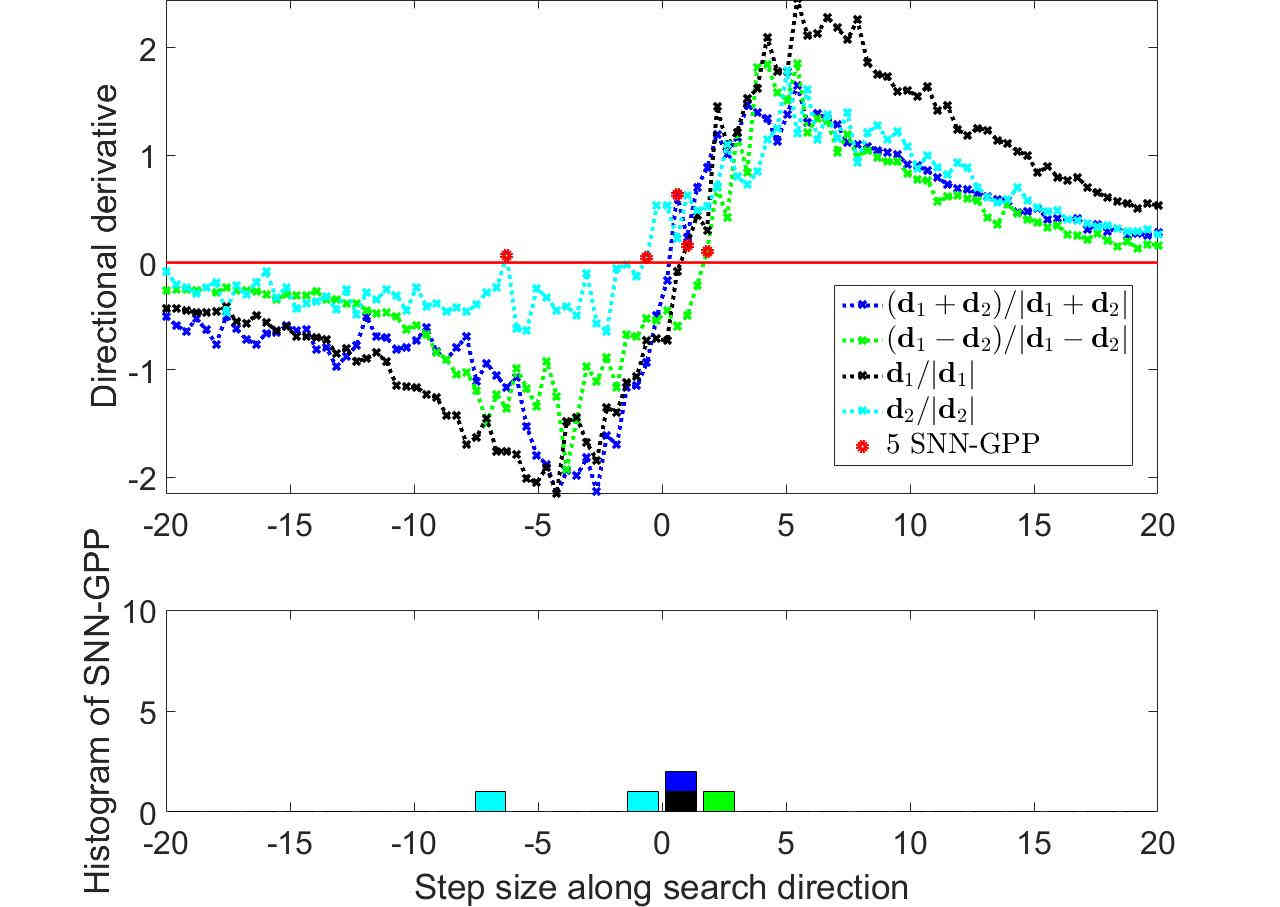}
		\caption{Directional derivative over [-20,20]}
	\end{subfigure}%
	
	\begin{subfigure}{.45\textwidth}
		\centering 
		\includegraphics[width=1\linewidth]{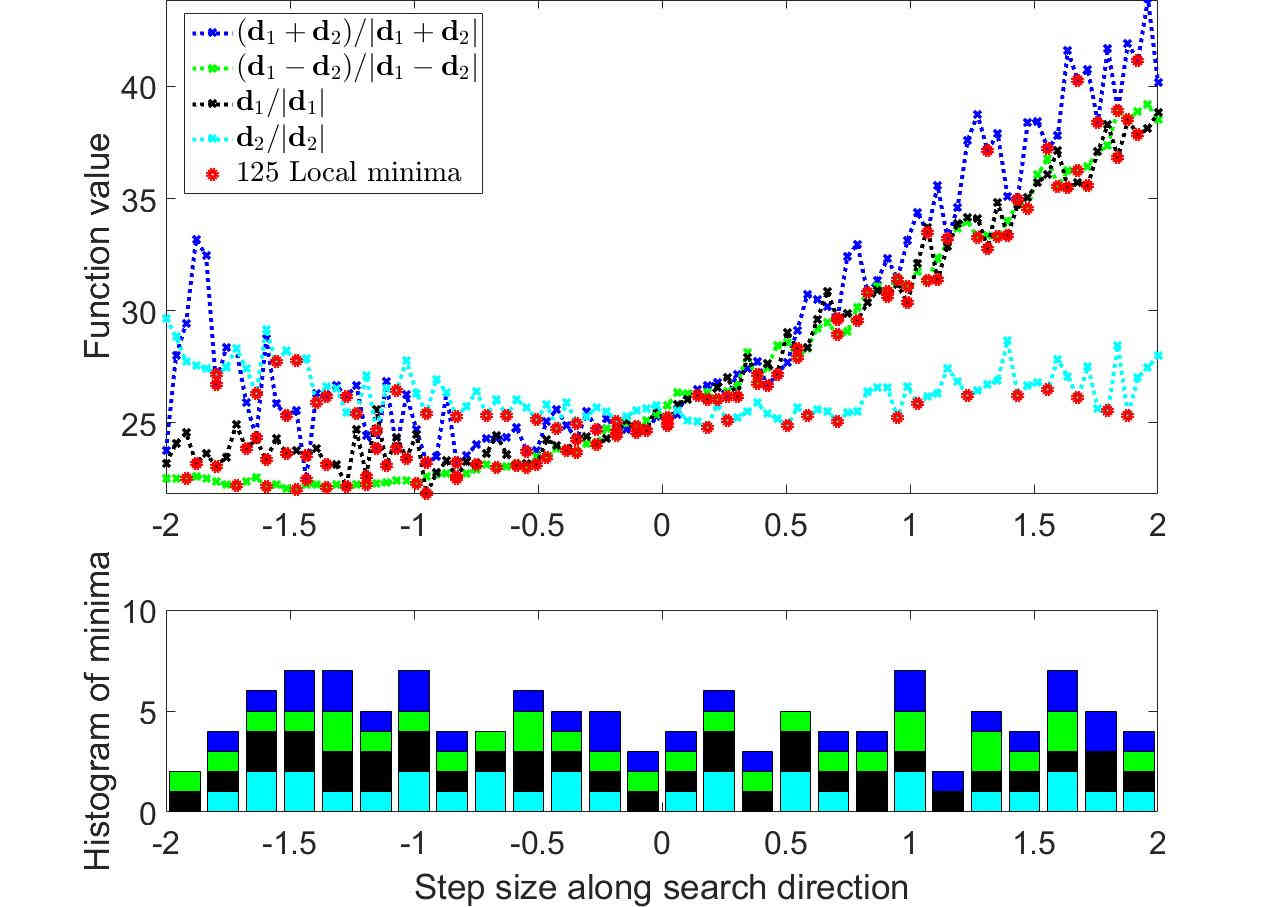}
		\caption{Function value over [-2,2]}
	\end{subfigure}%
	\begin{subfigure}{.45\textwidth}
		\centering
		\includegraphics[width=1\linewidth]{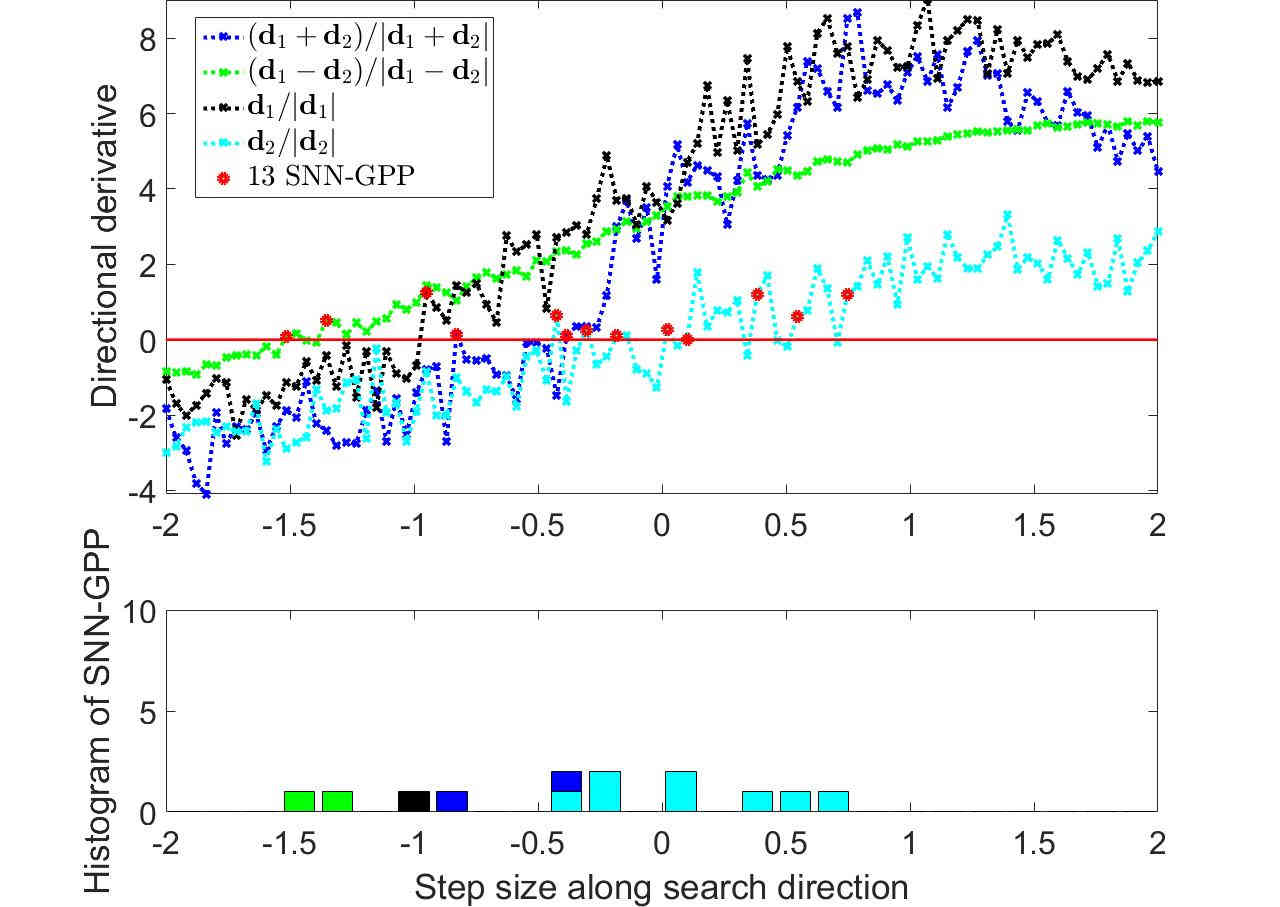}
		\caption{Directional derivative over [-2,2]}
	\end{subfigure}%
	
	\caption{Function values and directional derivatives along fixed search directions using mini-batch size $ |\mathcal{B}_{n,i}|=50 $. The increase in batch size reduces the severity of sampling error. This is particularly effective in low curvature directions, causing the locations of SNN-GPPs to be more consistent. However, the localization of local minima is not positively affected by this increase in  $ |\mathcal{B}_{n,i}| $.}
	\label{fig_sig_lines_n50}
\end{figure}

Through our analyses of different activation functions with a constant batch size of $|\mathcal{B}_{n,i}|= 10$, we have demonstrated the sensitivity of optimality criteria to curvature in the loss function. To round off the discussion around different optimum formulations, we show a representative example of sensitivity to batch size. We again use the Sigmoid AF (the worst performer in the saturation class) and conduct the same analyses with domains and directions as shown Sections \ref{sec_global} and \ref{sec_local}, only changing the mini-batch sample size to $|\mathcal{B}_{n,i}|= 50$.

The results shown in Figure~\ref{fig_sig_lines_n50}(b) and (d) immediately exhibit a lower occurrence and smaller spatial localization of SNN-GPP, especially in the low-curvature direction, $\boldsymbol{d}_2$, compared to the results in Figures~\ref{fig_sig_lines}(d) and \ref{fig_z_sig}(h). The increase in sample size reduces the variance in the discontinuities due to changing mini-batches, which affects both function values and directional derivatives. This is particularly effective for increasing the robustness of locating SNN-GPPs in low curvature domains.

We compare the quantitative data of analyses with $|\mathcal{B}_{n,i}|= 10$ and $|\mathcal{B}_{n,i}|= 50$ in Table~\ref{tbl_n50}. In both the global and local domains, the number of SNN-GPPs counted dropped by roughly a factor of 3. With regards to function values, the number of local minima remained similar to those counted for $|\mathcal{B}_{n,i}|=10$. This indicates that while local minima are largely insensitive to a decrease in loss function variance, SNN-GPPs react positively to a reduction in gradient variance. As the sample size increases, the number and location of SNN-GPPs more closely approximates that of the true optima. However, as demonstrated by Figure~\ref{fig_sig}, this does not occur as rapidly with the increase of mini-batch size for the local minima formulation. We can only say with certainty that finding local minima is only reliable in the limit case of using all data, or keeping the mini-batch static. The use of SNN-GPPs offers an improved alternative to estimate the solution, in particular, when loss functions are computed using dynamic MBSS with smaller batch sizes.

\begin{table}[h!]
	\centering
	\begin{tabularx}{\textwidth}{|p{13mm}|p{11mm}|p{17mm}|p{17mm}|p{17mm}|X|}
		\hline 
		\textbf{domain} & \textbf{\# True} \newline \textbf{Optima} & \textbf{\#} \textbf{Local Minima} \textbf{for} \newline  $|\mathcal{B}_{n,i}|=10$ & \textbf{\#}  \textbf{Local Minima}  \textbf{for} \newline  $|\mathcal{B}_{n,i}|=50$ & \textbf{\#} \textbf{SNN-GPPs} \textbf{for} \newline  $|\mathcal{B}_{n,i}|=10$ &  \textbf{\#} \textbf{SNN-GPPs} \textbf{for} \newline  $|\mathcal{B}_{n,i}|=50$  \\ 
		\hline 
		\textbf{[-20,20]} & 4 & 128 & 125 & 17 & 5  \\ 
		\hline 
		\textbf{[-2,2]} & 4 & 117 & 125 & 36 & 13 \\ 
		\hline
	\end{tabularx} 
	
	\caption{Number of local minima and SNN-GPPs counted in different domain sizes as a function of mini-batch size, $|\mathcal{B}_{n,i}|$.}
	\label{tbl_n50}
\end{table}

\section{Conclusion}

This paper visually explores the local minimum and stochastic non-negative associated gradient projection point (SNN-GPP) definitions in the context of dynamic mini-batch sub-sampled (MBSS) loss functions of neural networks with different activation functions. We have highlighted the differences between static and dynamic MBSS strategies and have linked their properties to the bias-variance trade-off respectively. Static sub-samples result in a smooth loss function, which is beneficial to minimization optimization methods, but results in a bias towards the selected mini-batch. Conversely, dynamic MBSS results in an unbiased but high variance loss function with discontinuities, which obstructs minimization based optimizers. However, the fact that new information is presented at every cycle in dynamic MBSS, seems to have performance benefits to an optimization method that is able to operate in discontinuous loss functions.

We investigate the ability of local function minimizers and SNN-GPPs to localize true (full-batch) optima when dynamic MBSS is implemented. Function minimizers formulated as the standard mathematical programming problem give rise to a large number of spurious local minima, which are high in number and uniformly distributed throughout the domain. This may significantly hamper the effectiveness of minimization line searches in neural network training using both exact and inexact line searches. Although inexact line searches may overcome some minima, it is not guaranteed, especially when monotonic descent is enforced to achieve convergent line search strategies.

SNN-GPPs, formulated as the solution to the gradient-only optimization problem, have been shown to be spatially localized around the true optimum. Our results have shown that SNN-GPPs have on average an order of magnitude less chance of occurrence than local minima over the same domain. With increasing mini-batch sub-sample sizes, the chance of spurious SNN-GPPs decreases even further, while local minima remain spatially spread. This holds for both "global" and "local" investigations performed. The spatial variance of SNN-GPPs depends on the curvature in the loss function along the search direction. High curvature tends to result in spatially concentrated SNN-GPPs, while low curvature search directions result in larger areas of uncertainty. The curvature of the loss function is sensitive to the activation function chosen in the neural network. In the problem considered, we showed that it is of interest to choose activation functions that result in high curvature in the directional derivative, while avoiding characteristics that might lead to the construction of numerous optima, or flat planes with zero derivatives. 

We divide the investigated activation functions into two classes, namely: Saturation and sparsity. These classes have different characteristics and therefore can be selected based on requirements of the given problem. According to our investigations, the activation functions for each class with the most favourable characteristics for locating high quality SNN-GPPs in dynamically sub-sampled loss functions were Softsign and ELU respectively. The ability to find optima more reliably in dynamic MBSS loss functions allows for exploration in constructing better optimizers for memory-restricted machine learning applications. Neural network models in particular can benefit from considering training as finding SNN-GPP as opposed to minimizing the loss function. This perspective of optimization may improve the construction of efficient and effective line searches in dynamic MBSS losses, which is still an open problem.

\begin{acknowledgements}
This work was supported by the Centre for Asset and Integrity Management (C-AIM), Department of Mechanical and Aeronautical Engineering, University of Pretoria, Pretoria, South Africa. 
\end{acknowledgements}

%
%

\bibliographystyle{spmpsci}      
\bibliography{BibC2}   

%
%

\newpage

\appendix

\section*{Appendix A.}
\label{AppA}

\setcounter{figure}{10}

\begin{figure}[h!]
	\centering
	\begin{subfigure}{.45\textwidth}
		\centering 
		\includegraphics[width=0.9\linewidth]{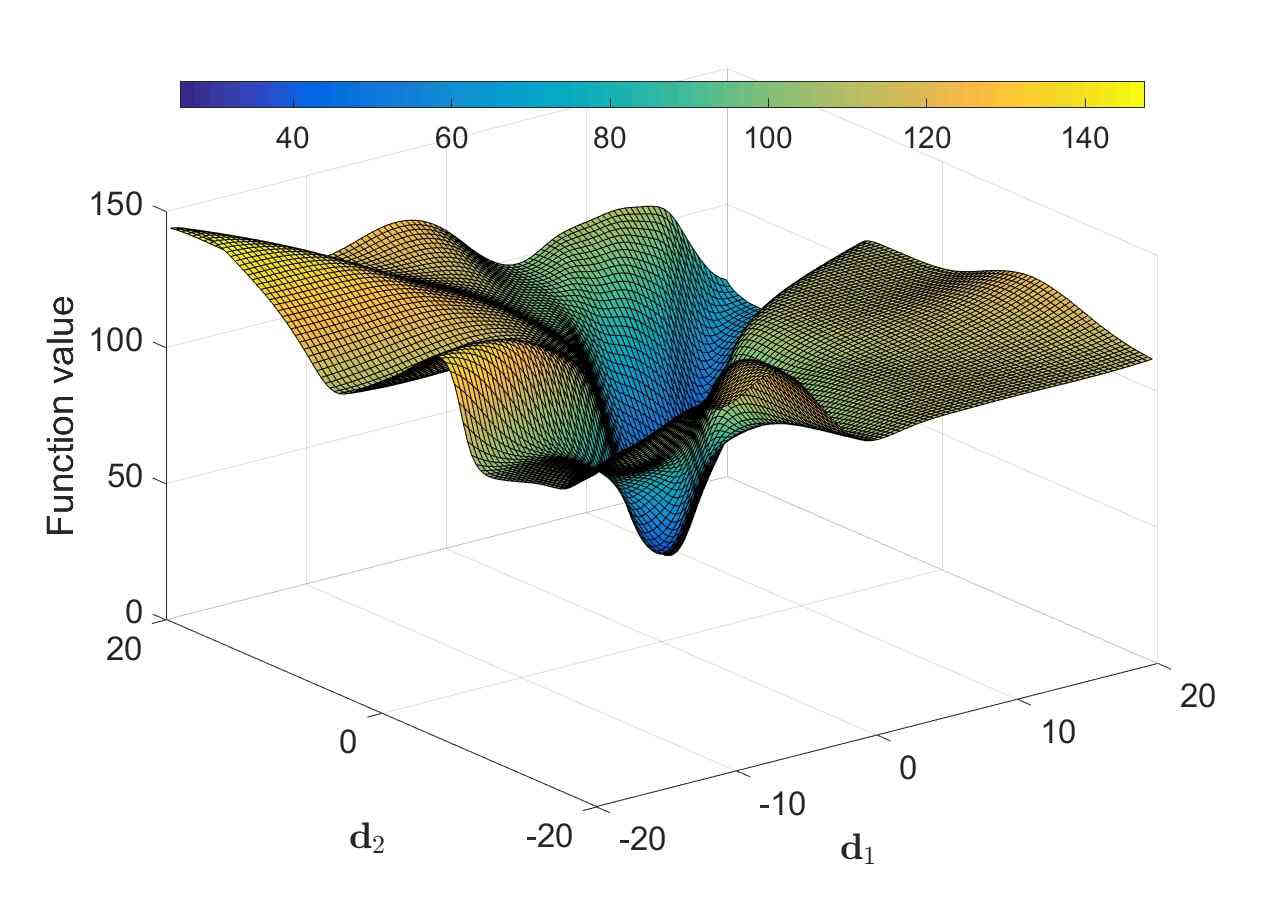}
		\caption{Function value, $M = 150$}
		\label{fig_tan_func_B}
	\end{subfigure}%
	\begin{subfigure}{.45\textwidth}
		\centering
		\includegraphics[width=0.9\linewidth]{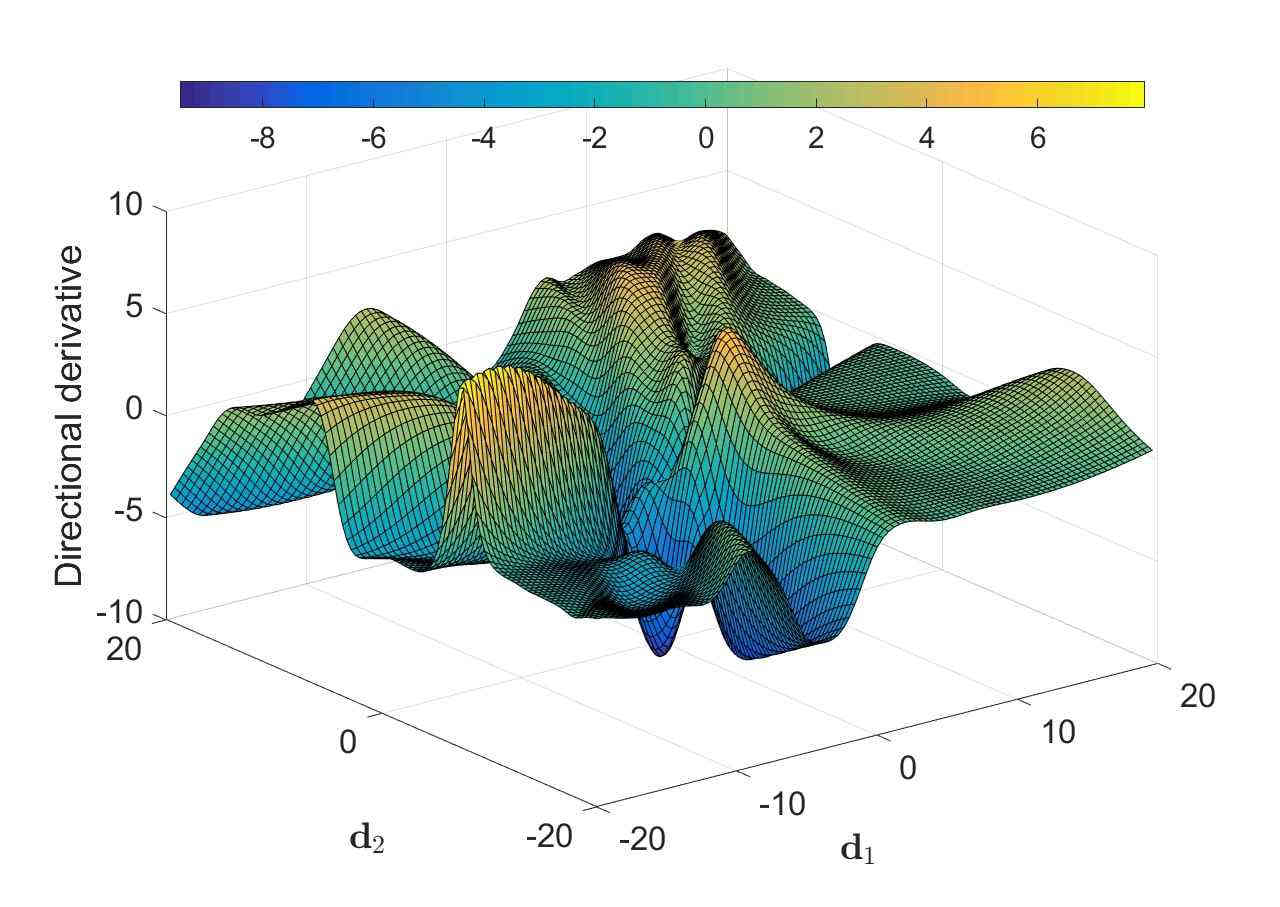}
		\caption{Directional derivative, $M = 150$}
		\label{fig_tan_dd_B}
	\end{subfigure}%
	
	\begin{subfigure}{.45\textwidth}
		\centering 
		\includegraphics[width=0.9\linewidth]{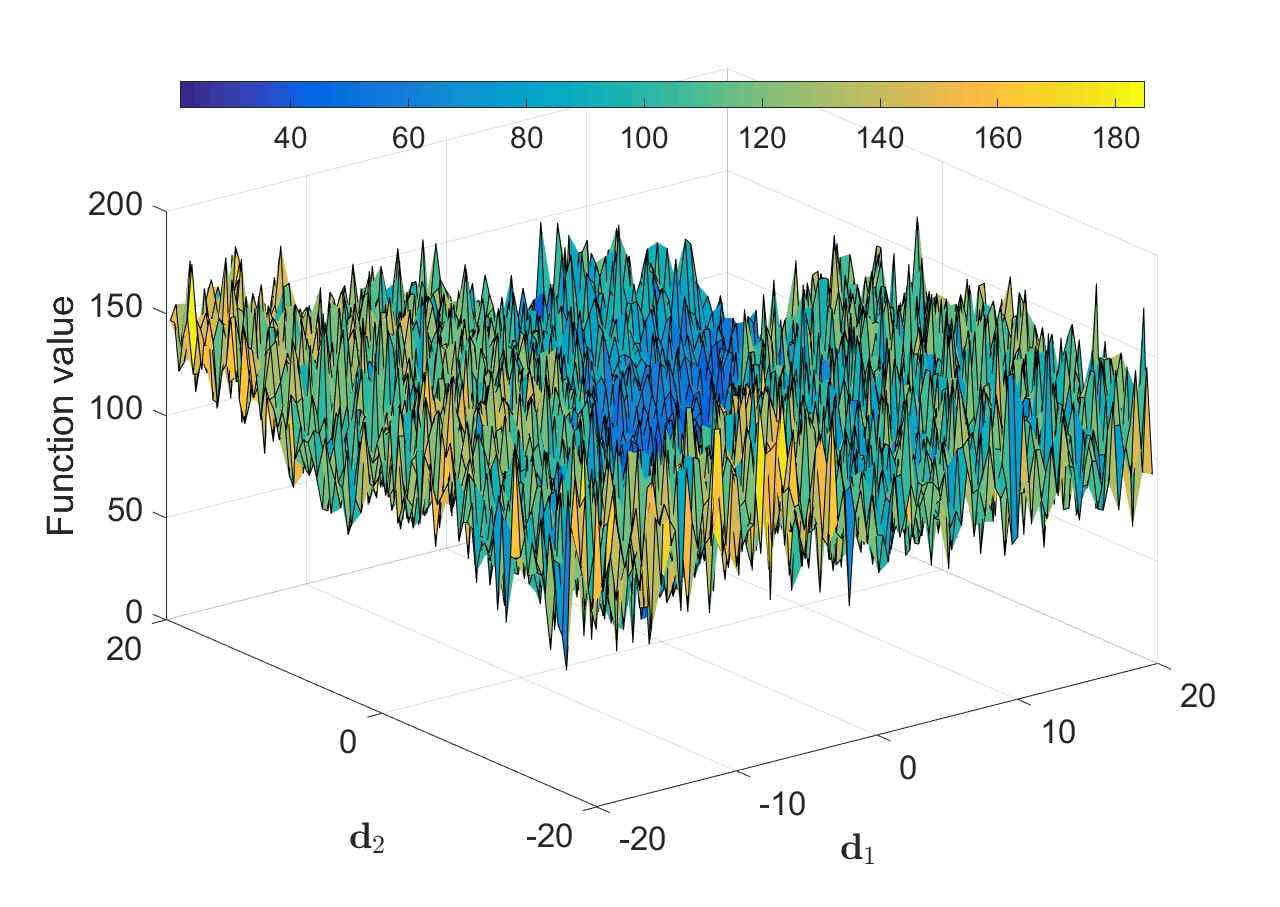}
		\caption{Function value, $|\mathcal{B}_{n,i}| = 10$}
		\label{fig_tan_func_M}
	\end{subfigure}%
	\begin{subfigure}{.45\textwidth}
		\centering
		\includegraphics[width=0.9\linewidth]{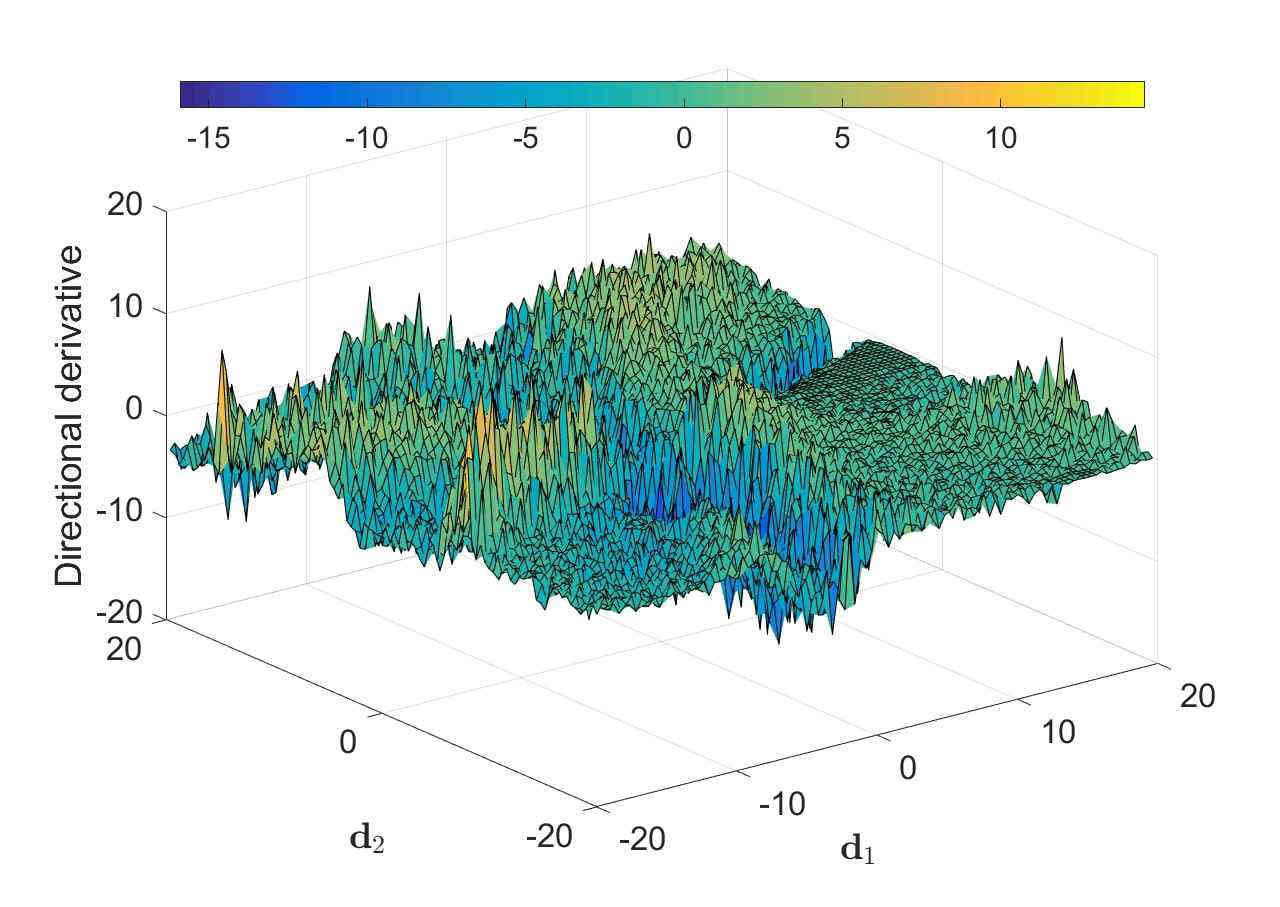}
		\caption{Directional derivative, $|\mathcal{B}_{n,i}| = 10$}
		\label{fig_tan_dd_M}
	\end{subfigure}%
	
	\begin{subfigure}{.45\textwidth}
		\centering 
		\includegraphics[width=0.9\linewidth]{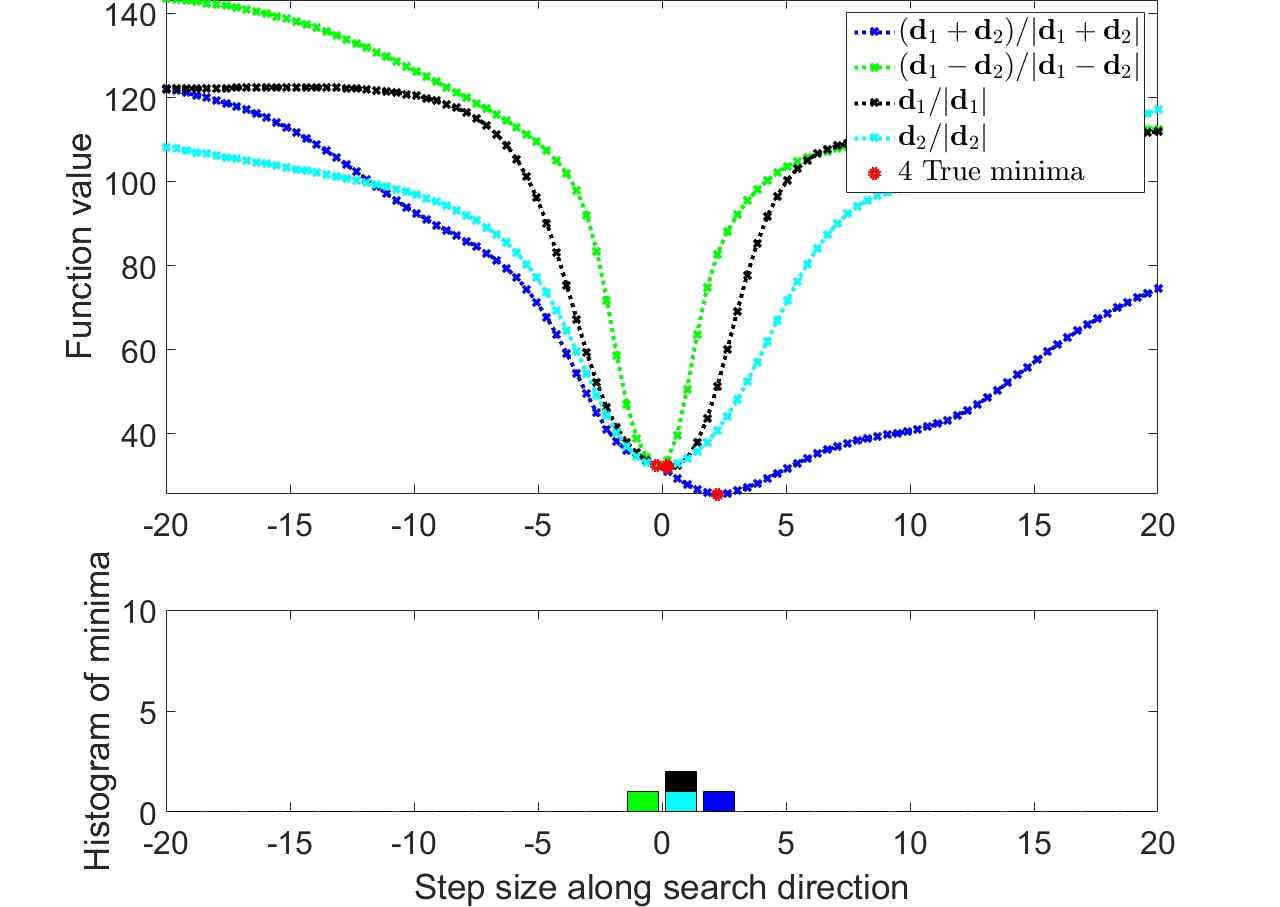}
		\caption{True minima along search directions}
		\label{fig_tan_fline_B}
	\end{subfigure}%
	\begin{subfigure}{.45\textwidth}
		\centering
		\includegraphics[width=0.9\linewidth]{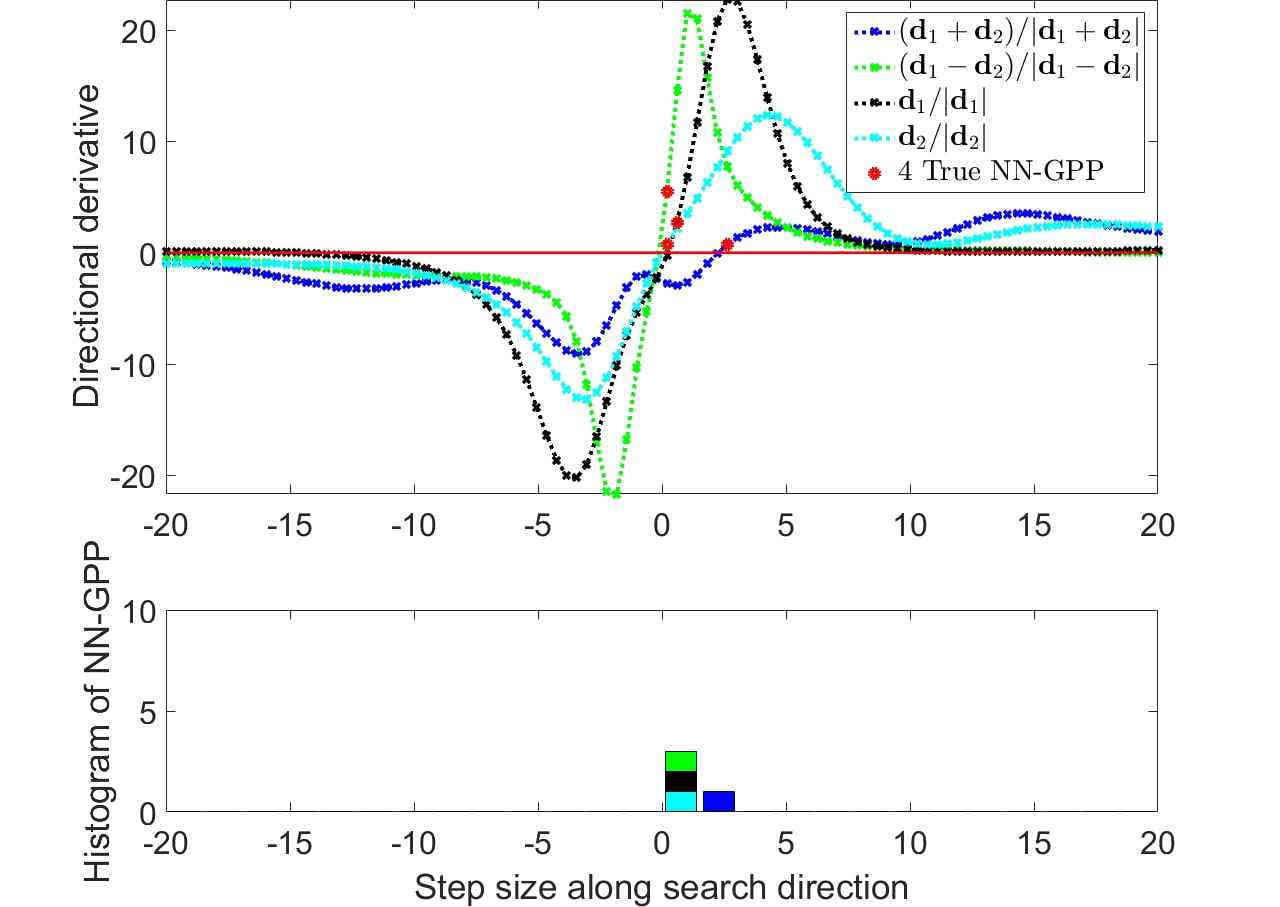}
		\caption{True NN-GPPs along search directions}
		\label{fig_tan_dline_B}
	\end{subfigure}%
	
	\begin{subfigure}{.45\textwidth}
		\centering 
		\includegraphics[width=0.9\linewidth]{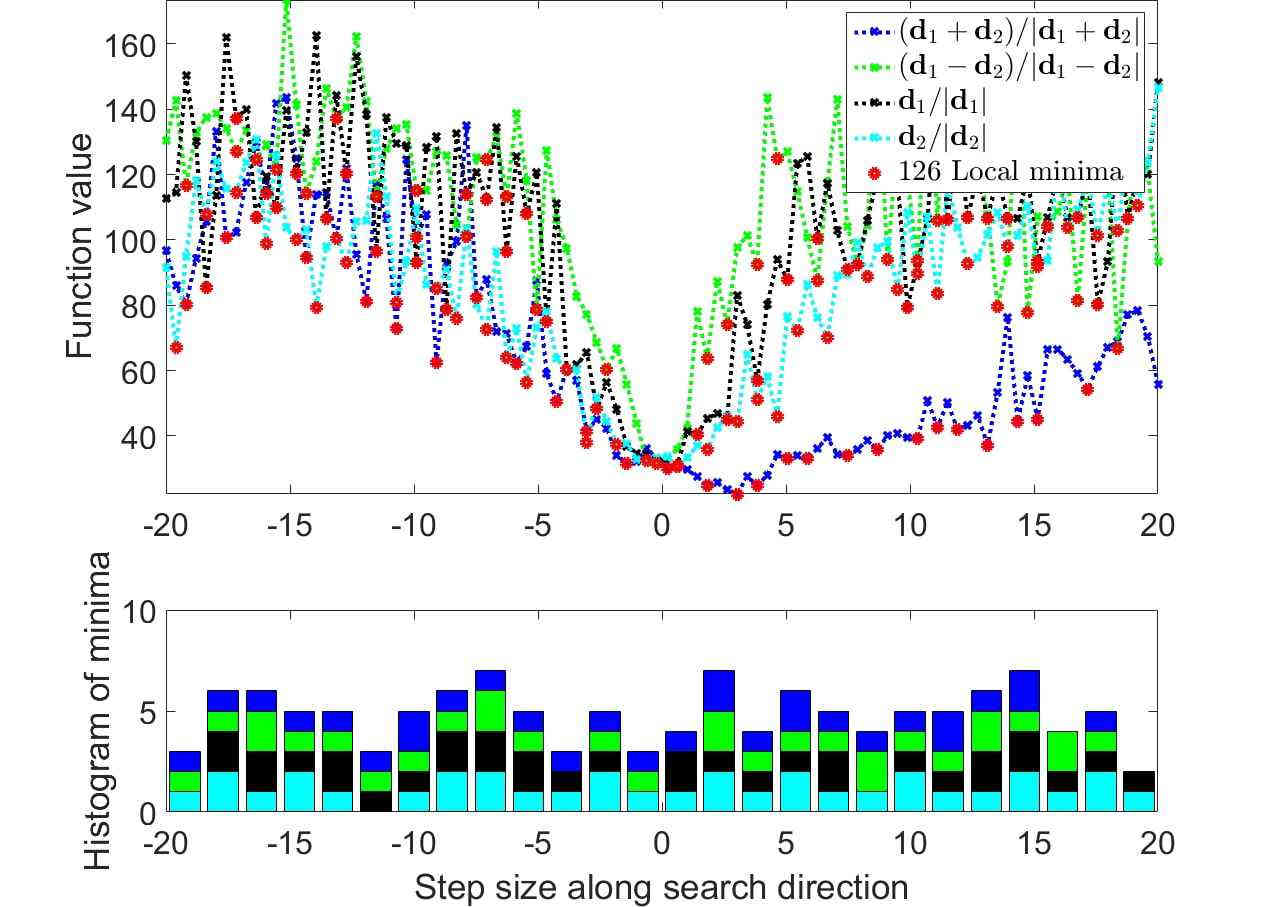}
		\caption{Local minima along search directions}
		\label{fig_tan_fline_M}
	\end{subfigure}%
	\begin{subfigure}{.45\textwidth}
		\centering
		\includegraphics[width=0.9\linewidth]{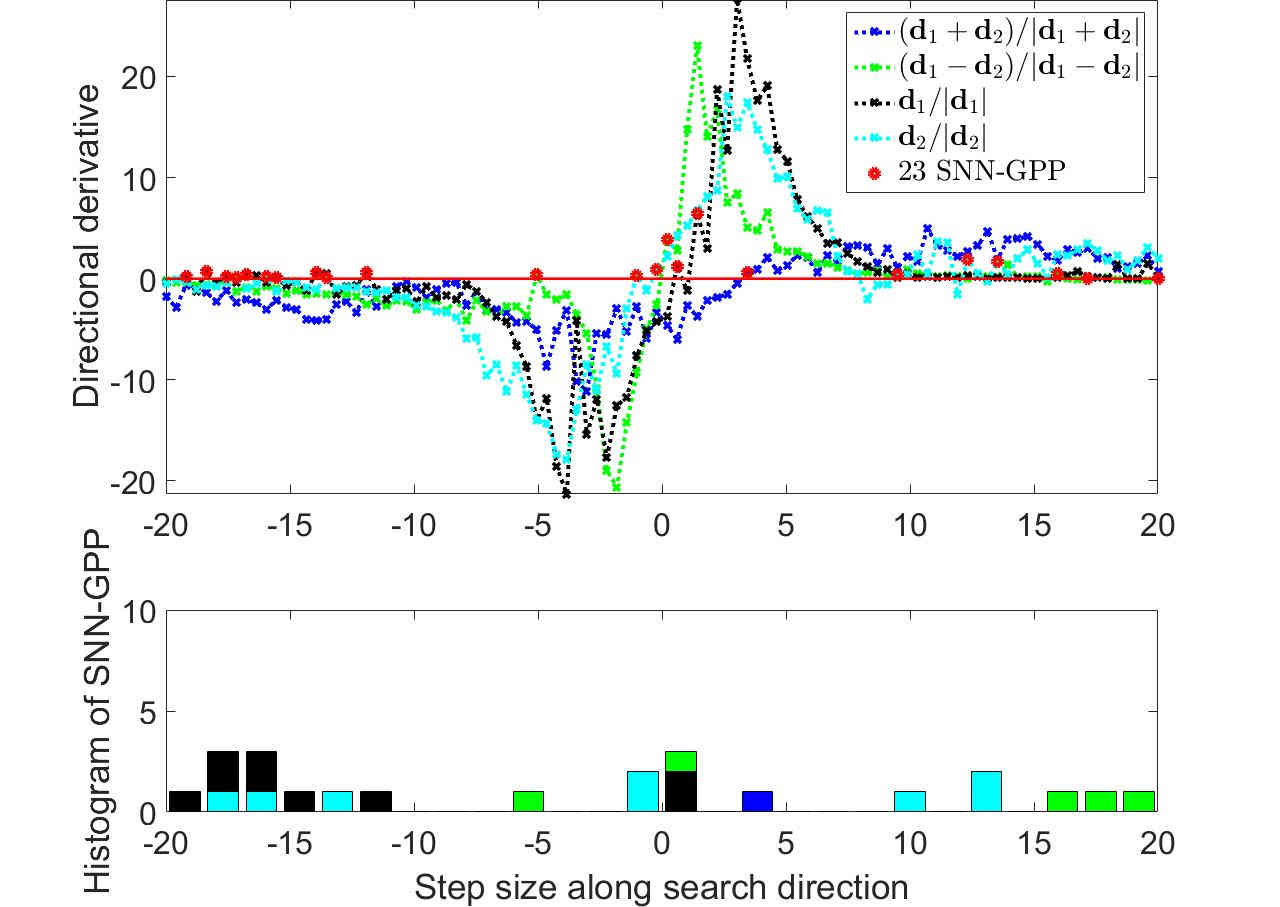}
		\caption{SNN-GPPs along search directions}
		\label{fig_tan_dline_M}
	\end{subfigure}%
	\caption{The Tanh AF: A steeper AF derivative around 0 means higher curvature in the loss gradients, manifesting as steeper directional derivatives. (h) SNN-GPPs are correctly localized around 0, but spuriously at saturation.}
	\label{fig_tan}
\end{figure}

\begin{figure}[h!]
	\centering
	\begin{subfigure}{.45\textwidth}
		\centering 
		\includegraphics[width=0.9\linewidth]{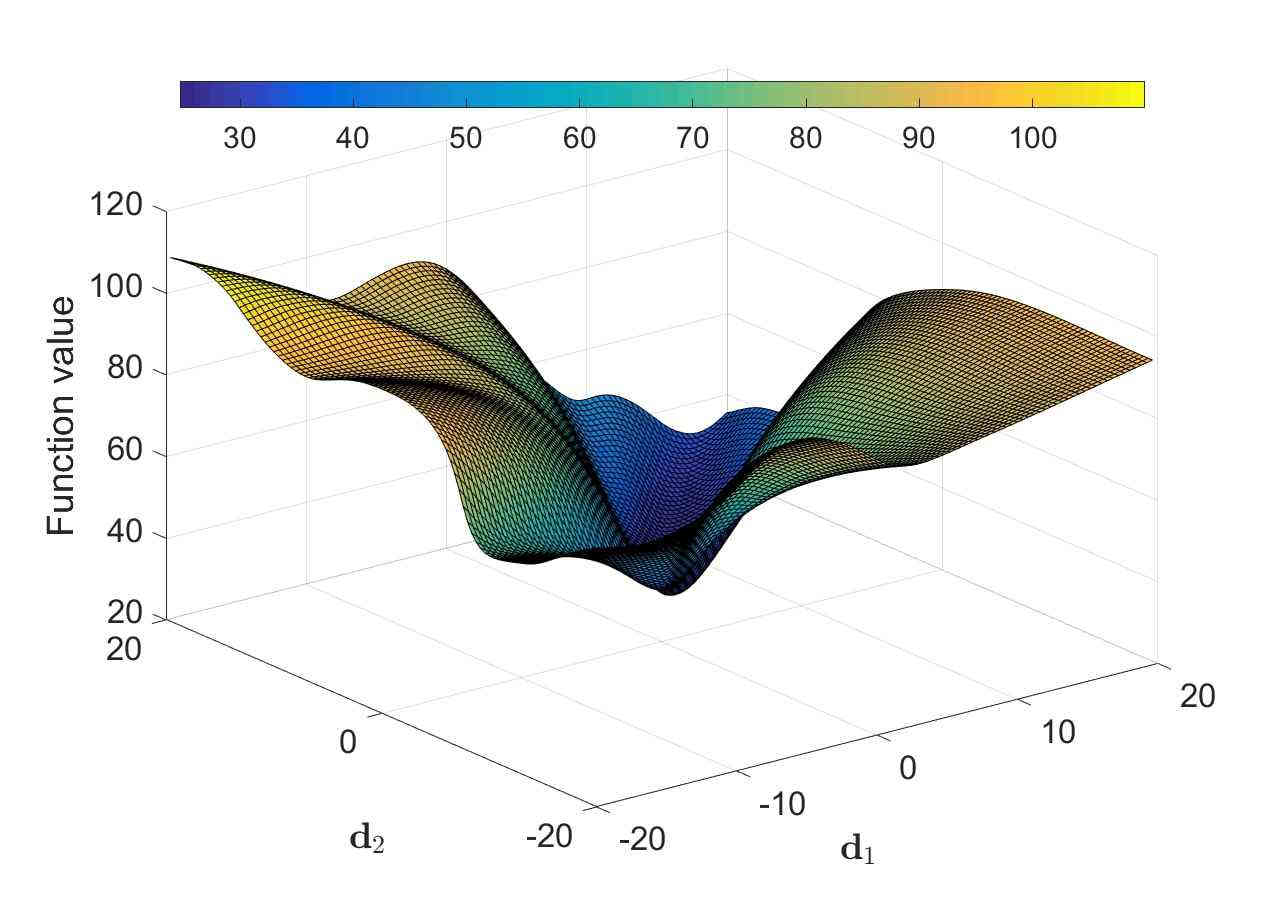}
		\caption{Function value, $M = 150$}
		\label{fig_soft_func_B}
	\end{subfigure}%
	\begin{subfigure}{.45\textwidth}
		\centering
		\includegraphics[width=0.9\linewidth]{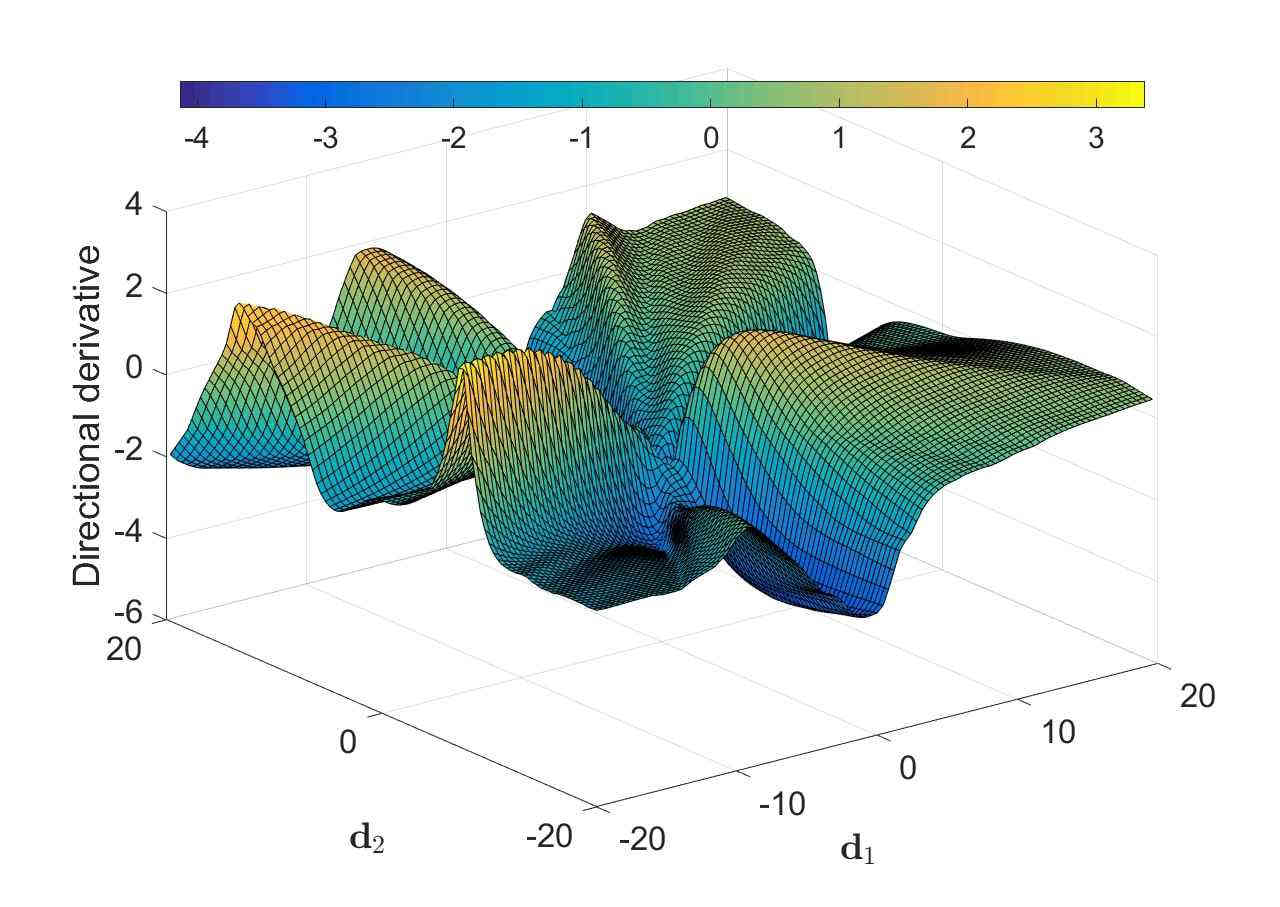}
		\caption{Directional derivative, $M = 150$}
		\label{fig_soft_dd_B}
	\end{subfigure}%
	
	\begin{subfigure}{.45\textwidth}
		\centering 
		\includegraphics[width=0.9\linewidth]{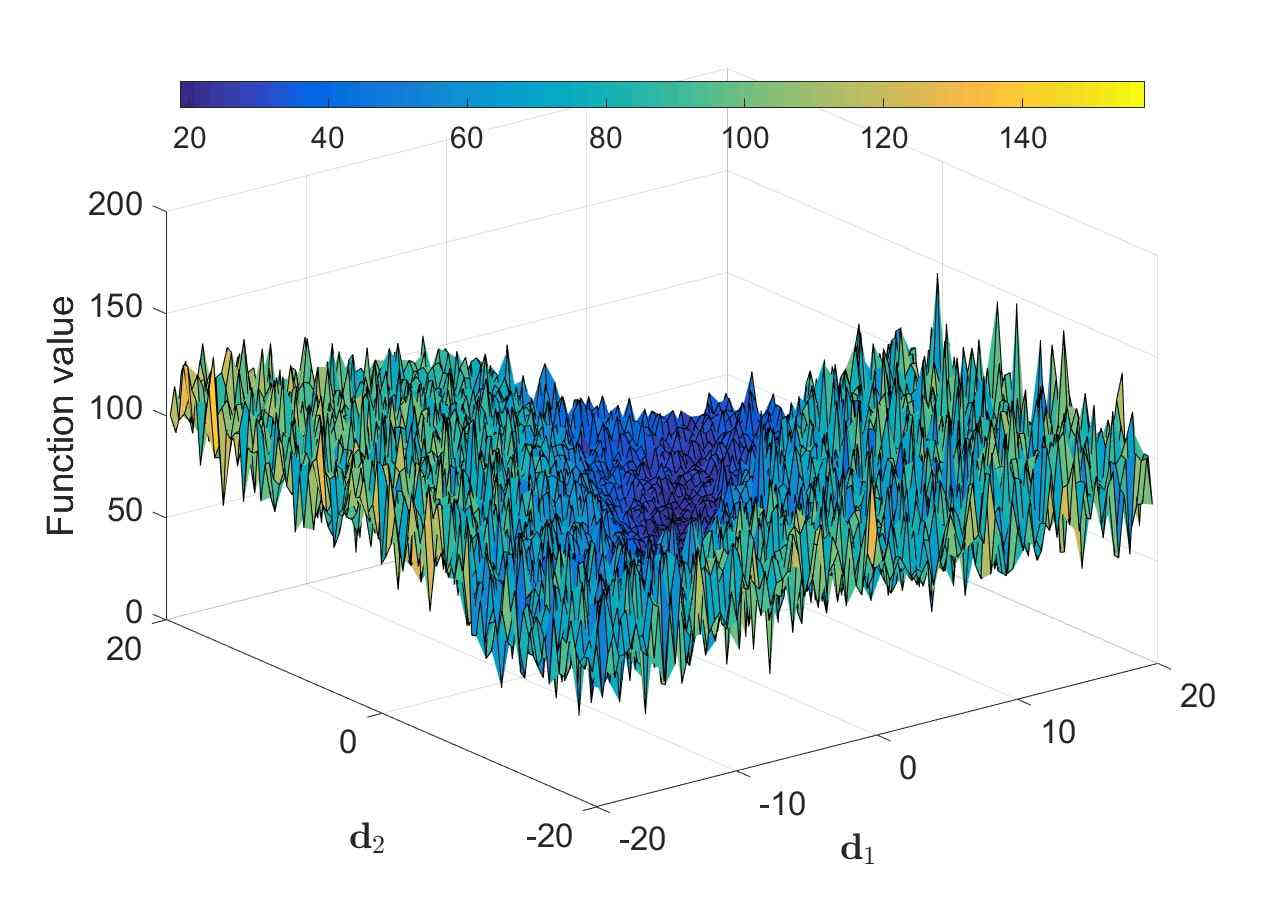}
		\caption{Function value, $|\mathcal{B}_{n,i}| = 10$}
		\label{fig_soft_func_M}
	\end{subfigure}%
	\begin{subfigure}{.45\textwidth}
		\centering
		\includegraphics[width=0.9\linewidth]{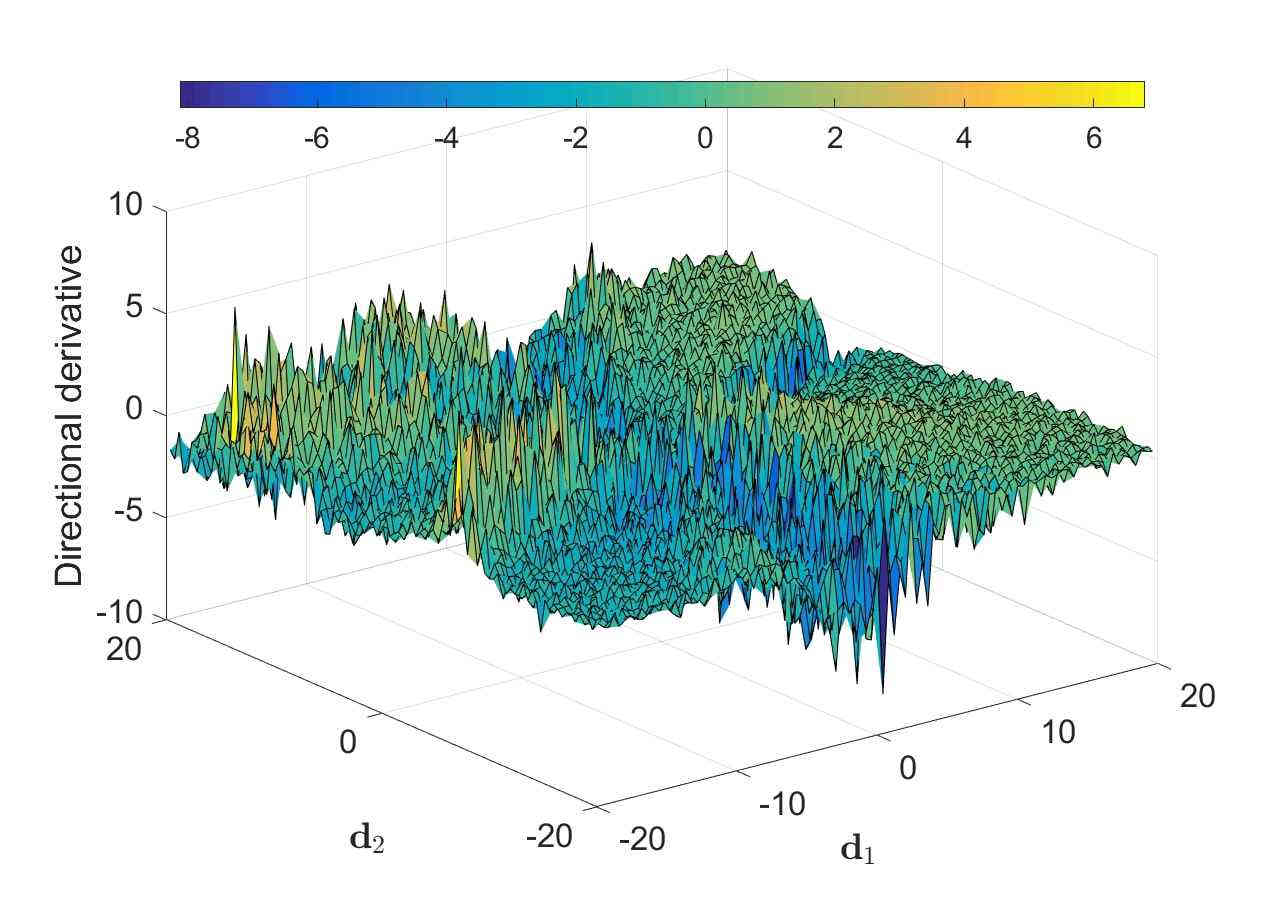}
		\caption{Directional derivative, $|\mathcal{B}_{n,i}| = 10$}
		\label{fig_soft_dd_M}
	\end{subfigure}%
	
	\begin{subfigure}{.45\textwidth}
		\centering 
		\includegraphics[width=0.9\linewidth]{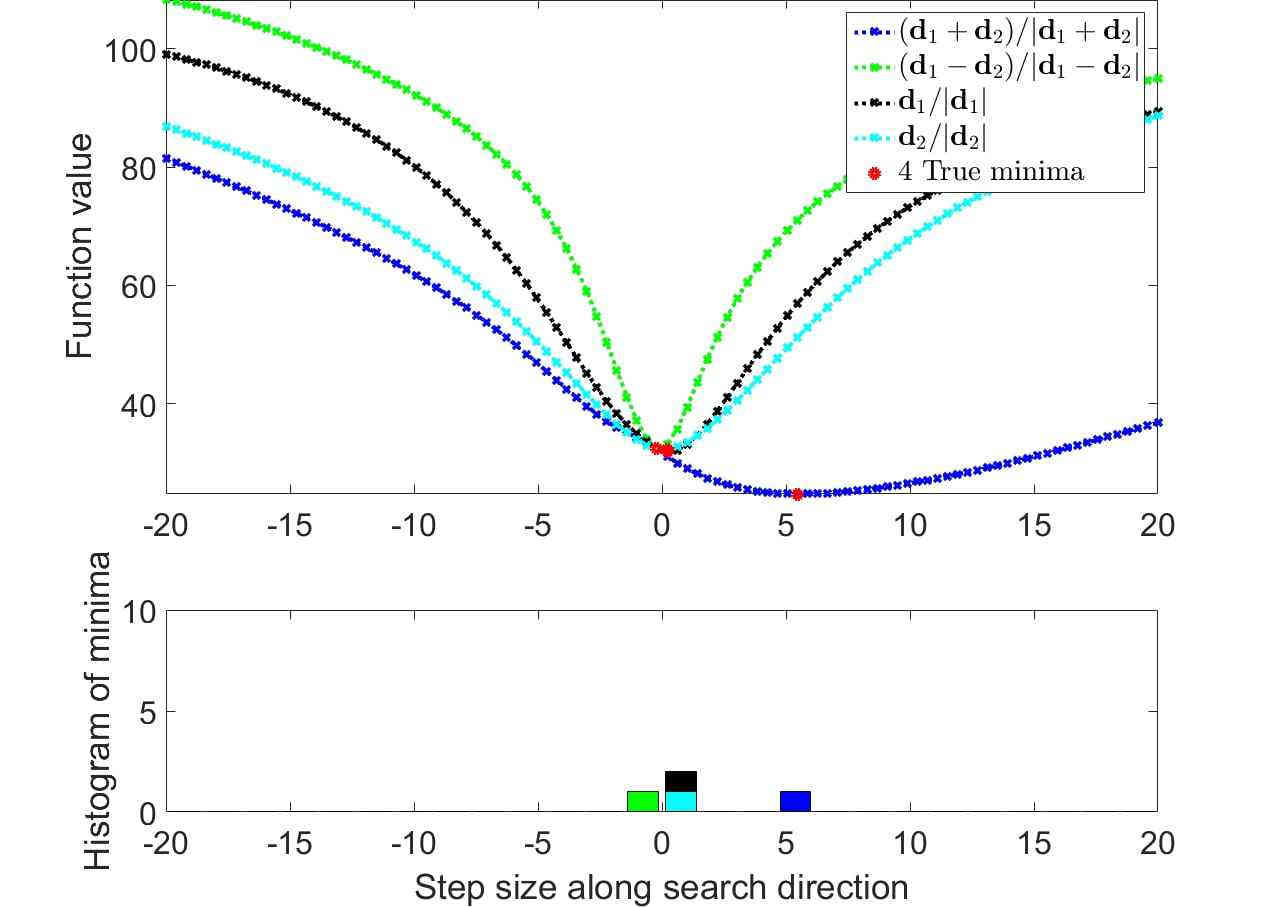}
		\caption{True minima along search directions}
		\label{fig_soft_fline_B}
	\end{subfigure}%
	\begin{subfigure}{.45\textwidth}
		\centering
		\includegraphics[width=0.9\linewidth]{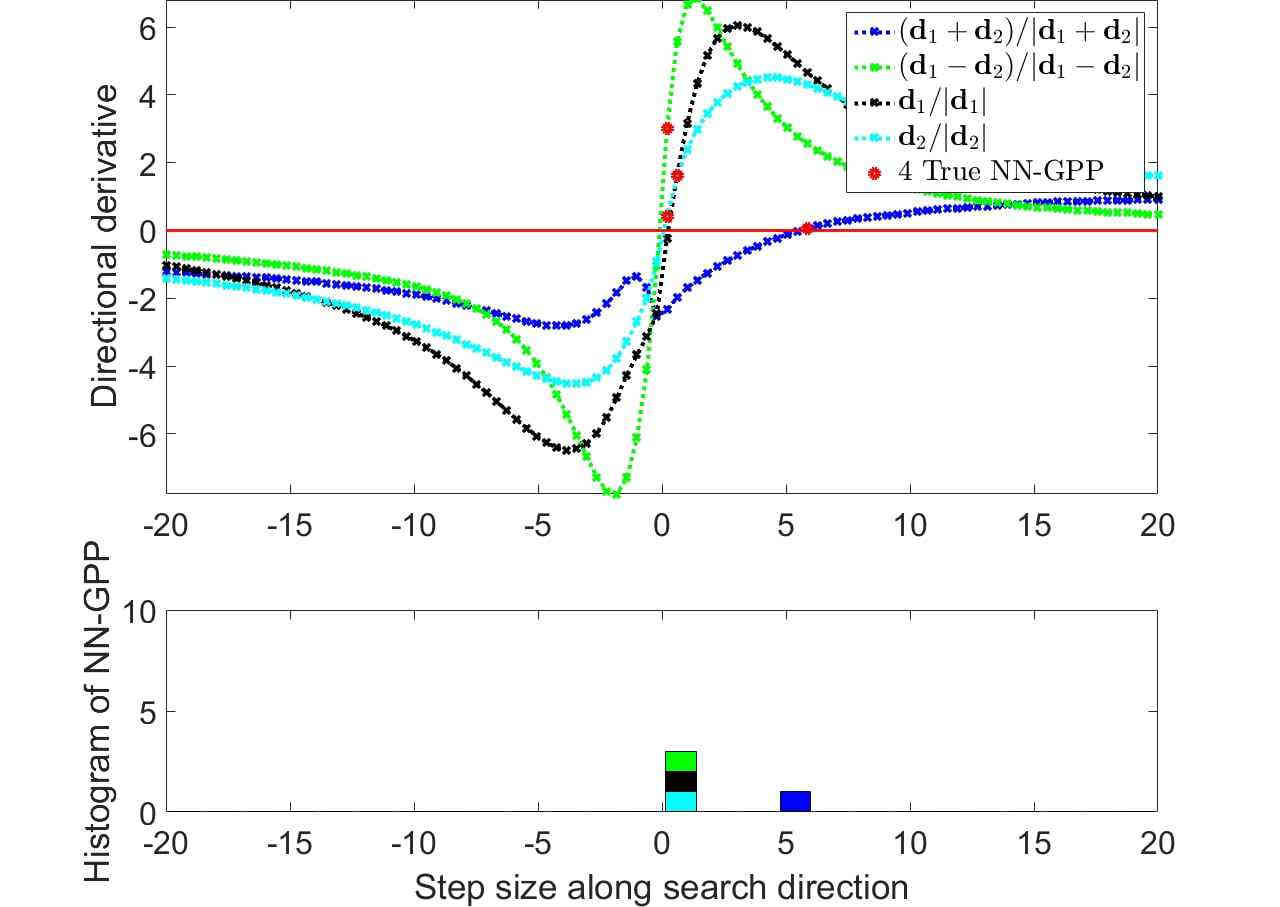}
		\caption{True NN-GPPs along search directions}
		\label{fig_soft_dline_B}
	\end{subfigure}%
	
	\begin{subfigure}{.45\textwidth}
		\centering 
		\includegraphics[width=0.9\linewidth]{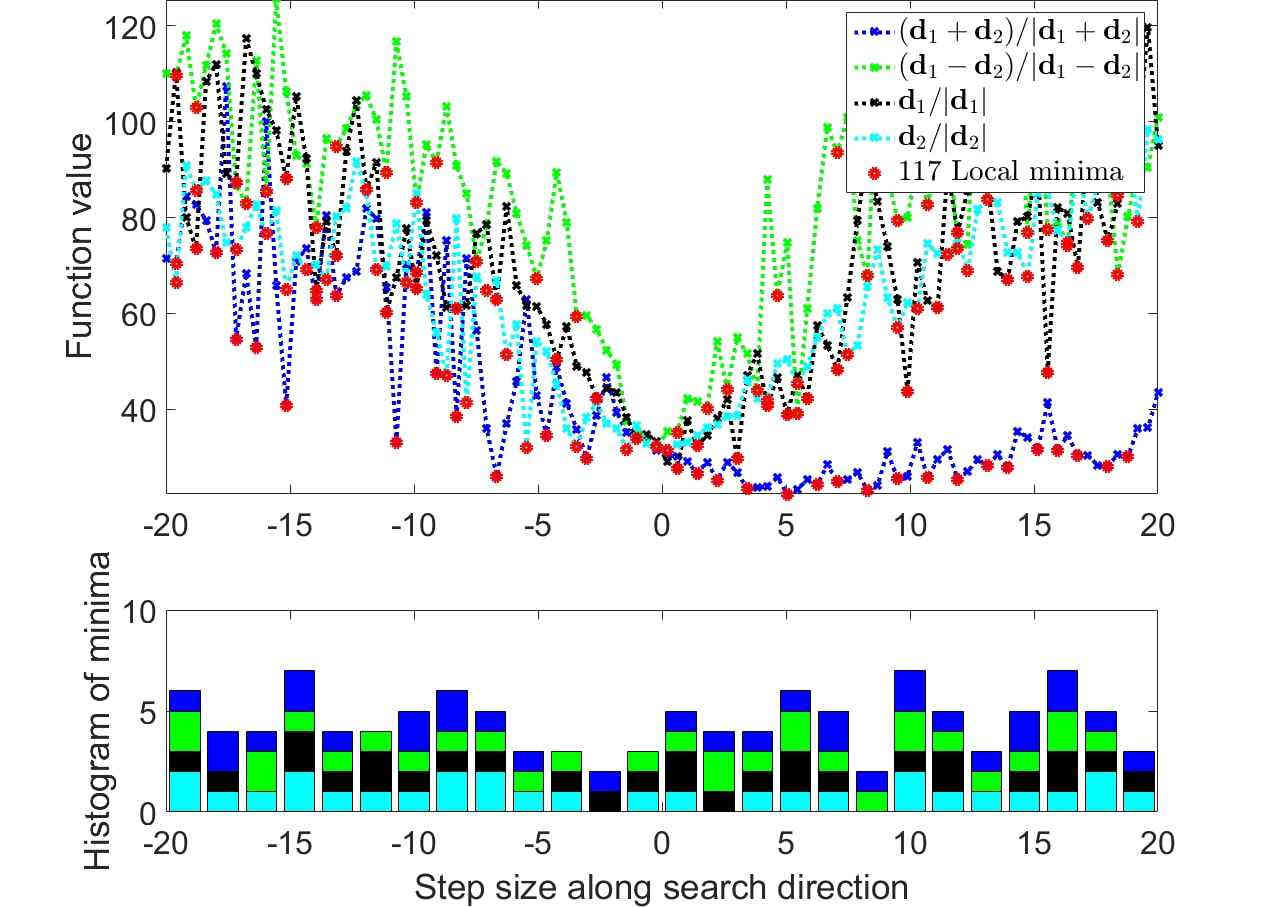}
		\caption{Local minima along search directions}
		\label{fig_soft_fline_M}
	\end{subfigure}%
	\begin{subfigure}{.45\textwidth}
		\centering
		\includegraphics[width=0.9\linewidth]{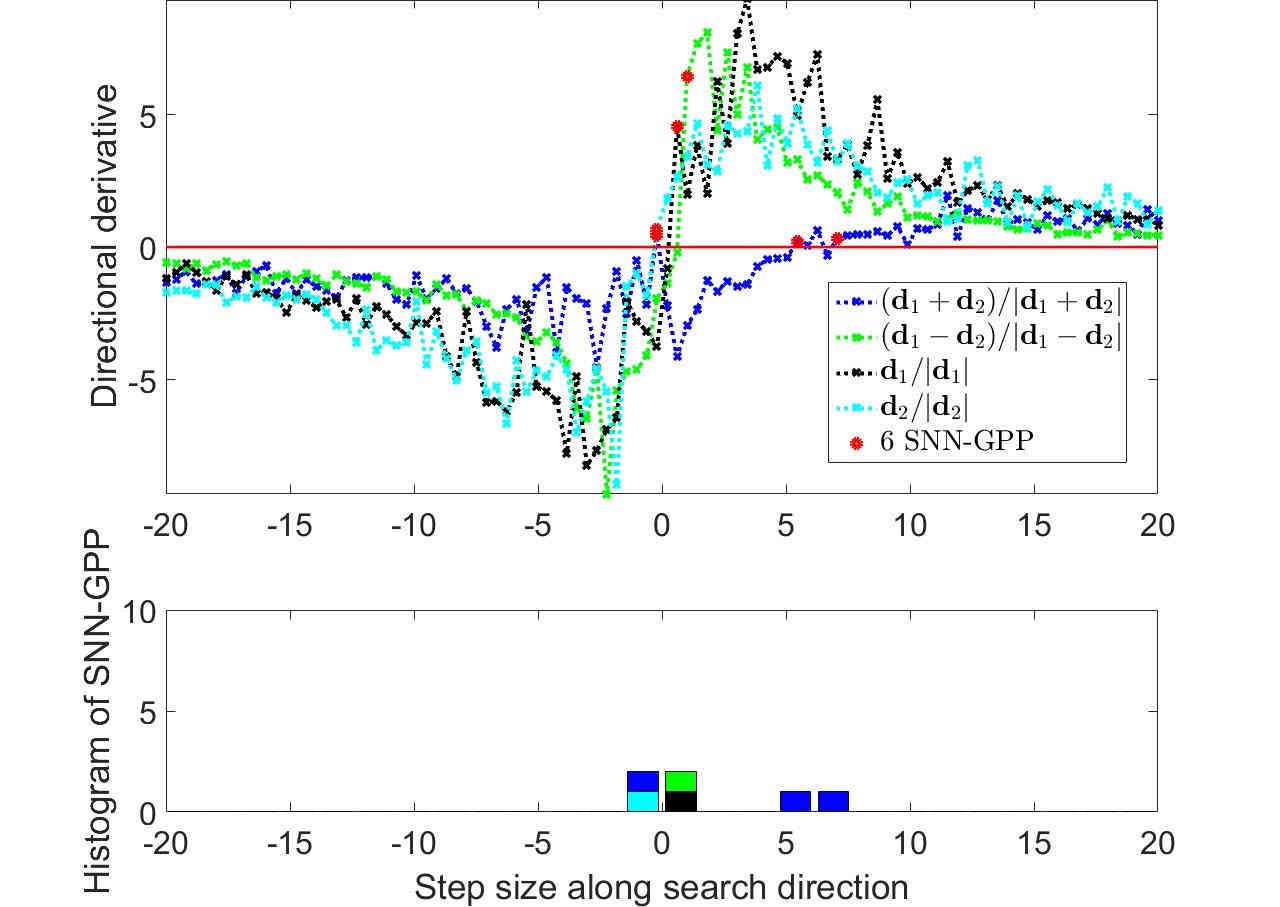}
		\caption{SNN-GPPs along search directions}
		\label{fig_soft_dline_M}
	\end{subfigure}%
	\caption{The Softsign AF: The less aggressive taper-off in the AF derivative reduces the chance for spurious SNN-GPPs in the saturation regions. In the centre domain, SNN-GPPs are highly localized around the full-batch, true optima.}
	\label{fig_soft}
\end{figure}

\begin{figure}[h!]
	\centering
	\begin{subfigure}{.45\textwidth}
		\centering 
		\includegraphics[width=0.9\linewidth]{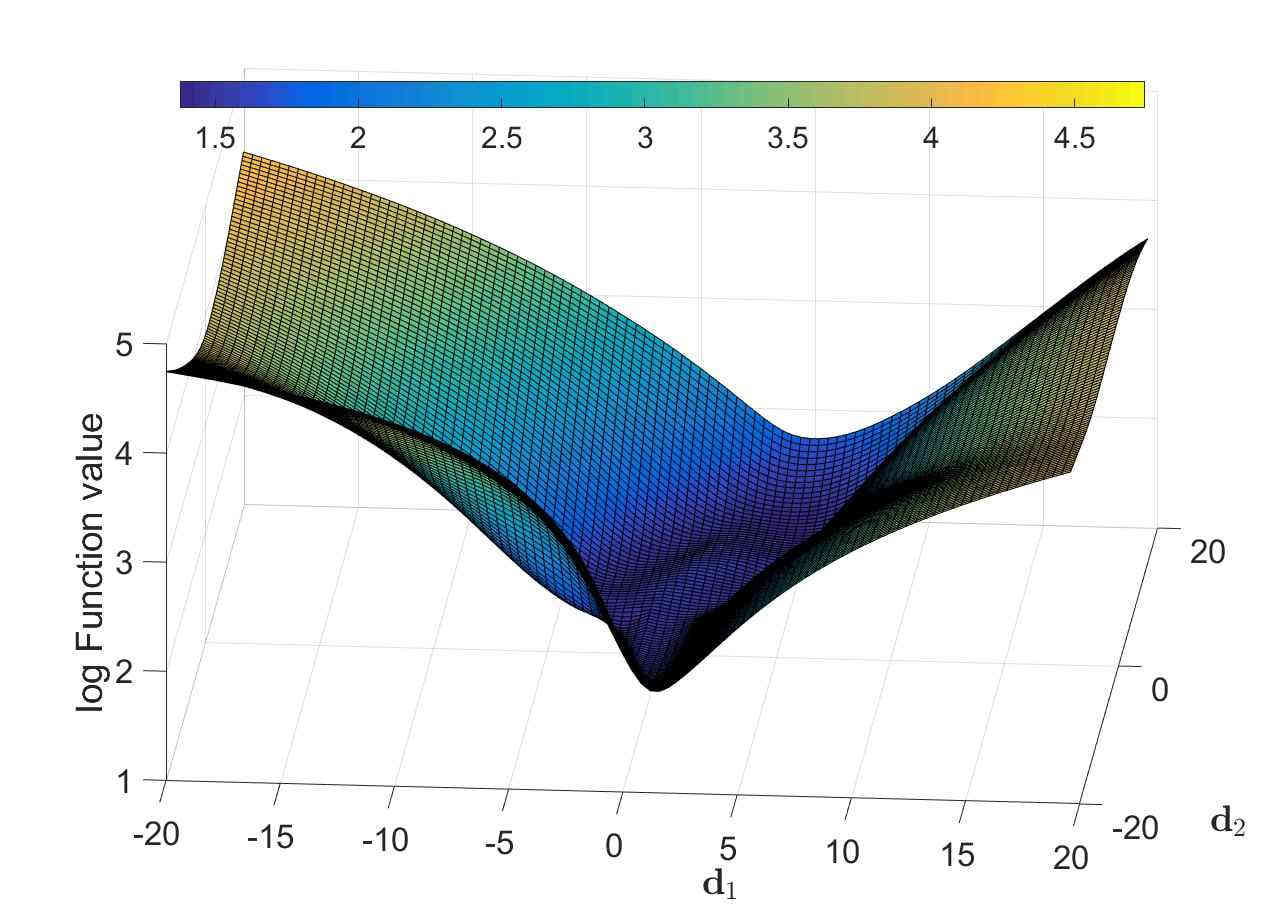}
		\caption{Function value, $M = 150$}
		\label{fig_relu_func_B}
	\end{subfigure}%
	\begin{subfigure}{.45\textwidth}
		\centering
		\includegraphics[width=0.9\linewidth]{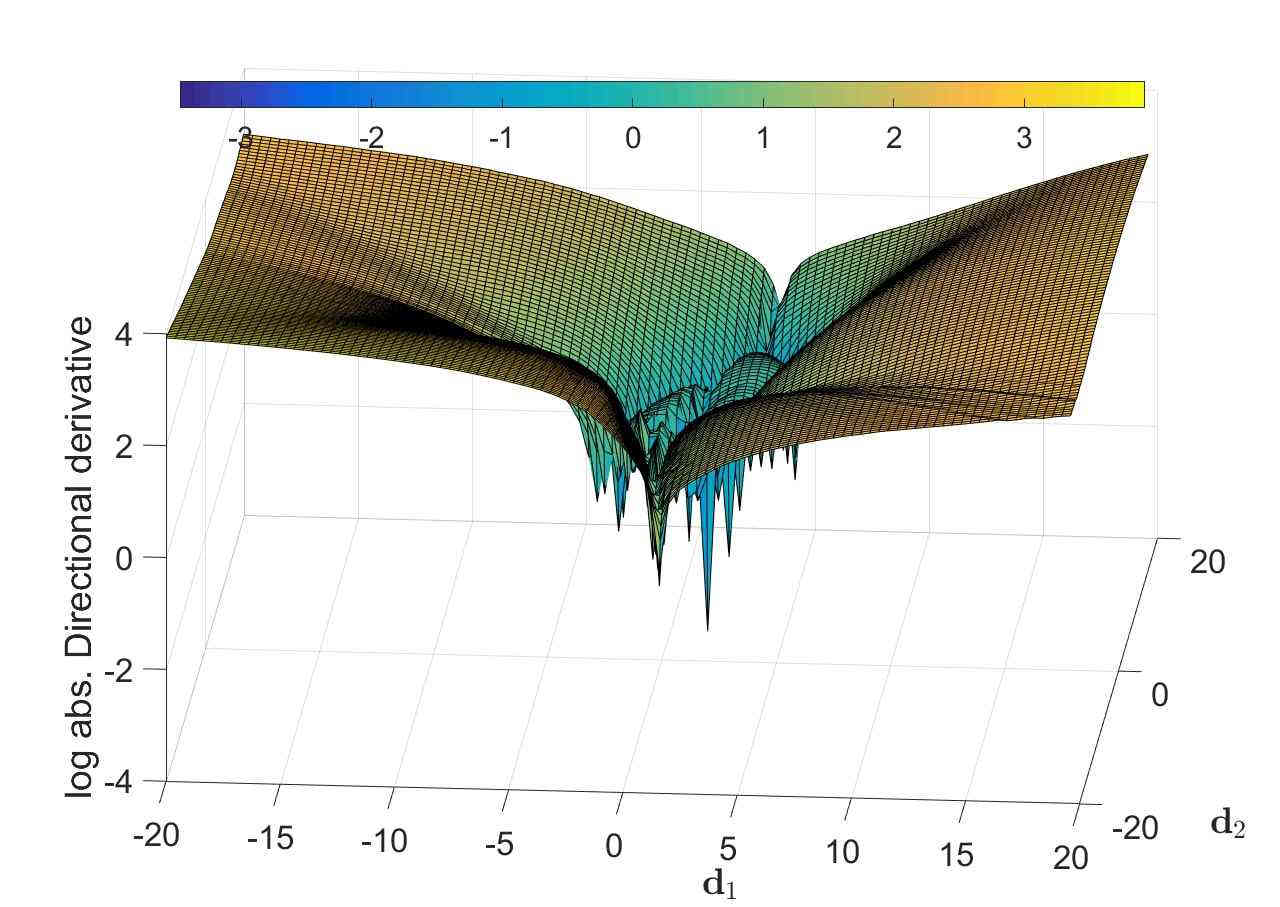}
		\caption{Directional derivative, $M = 150$}
		\label{fig_relu_dd_B}
	\end{subfigure}%
	
	\begin{subfigure}{.45\textwidth}
		\centering 
		\includegraphics[width=0.9\linewidth]{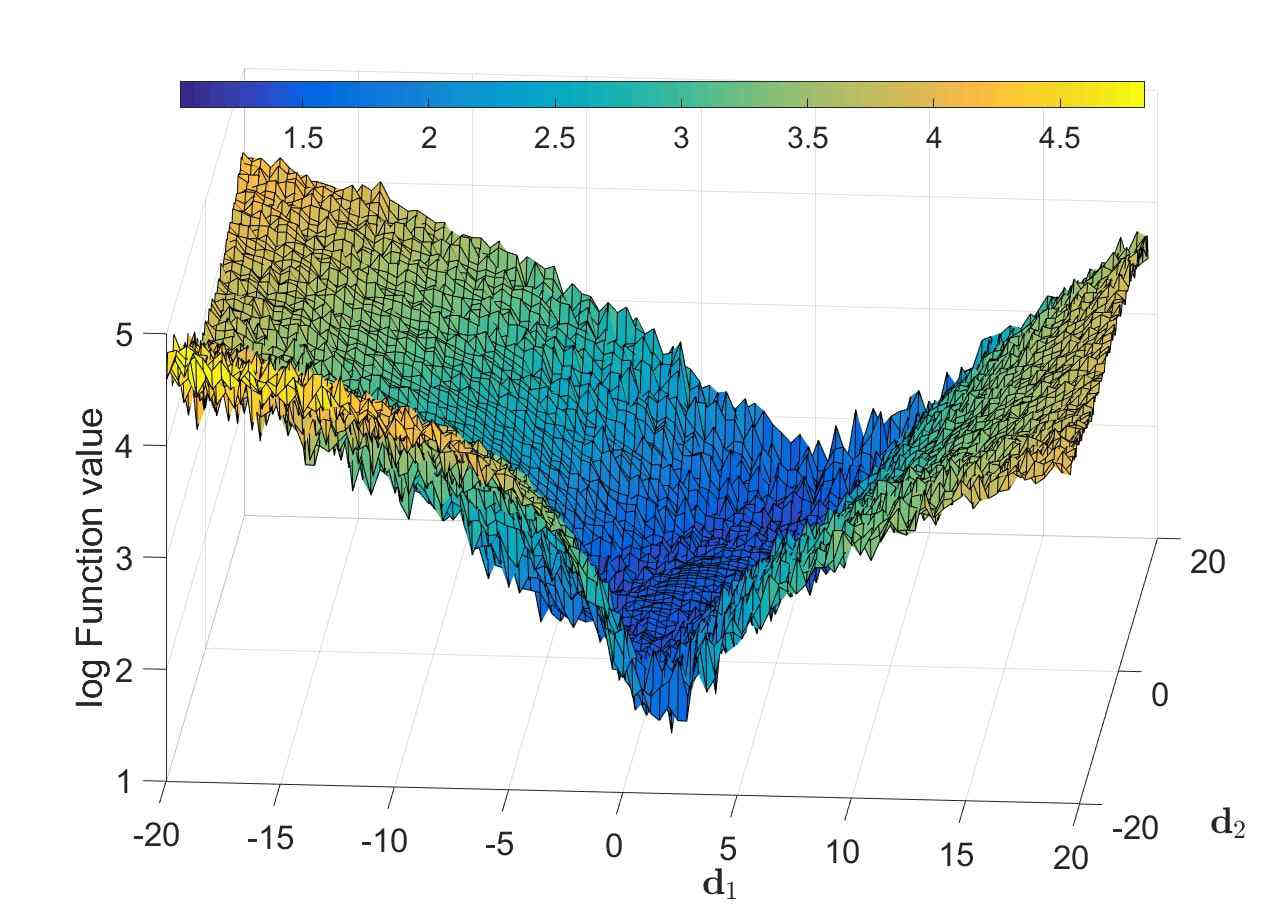}
		\caption{Directional derivative, $|\mathcal{B}_{n,i}| = 10$}
		\label{fig_relu_func_M}
	\end{subfigure}%
	\begin{subfigure}{.45\textwidth}
		\centering
		\includegraphics[width=0.9\linewidth]{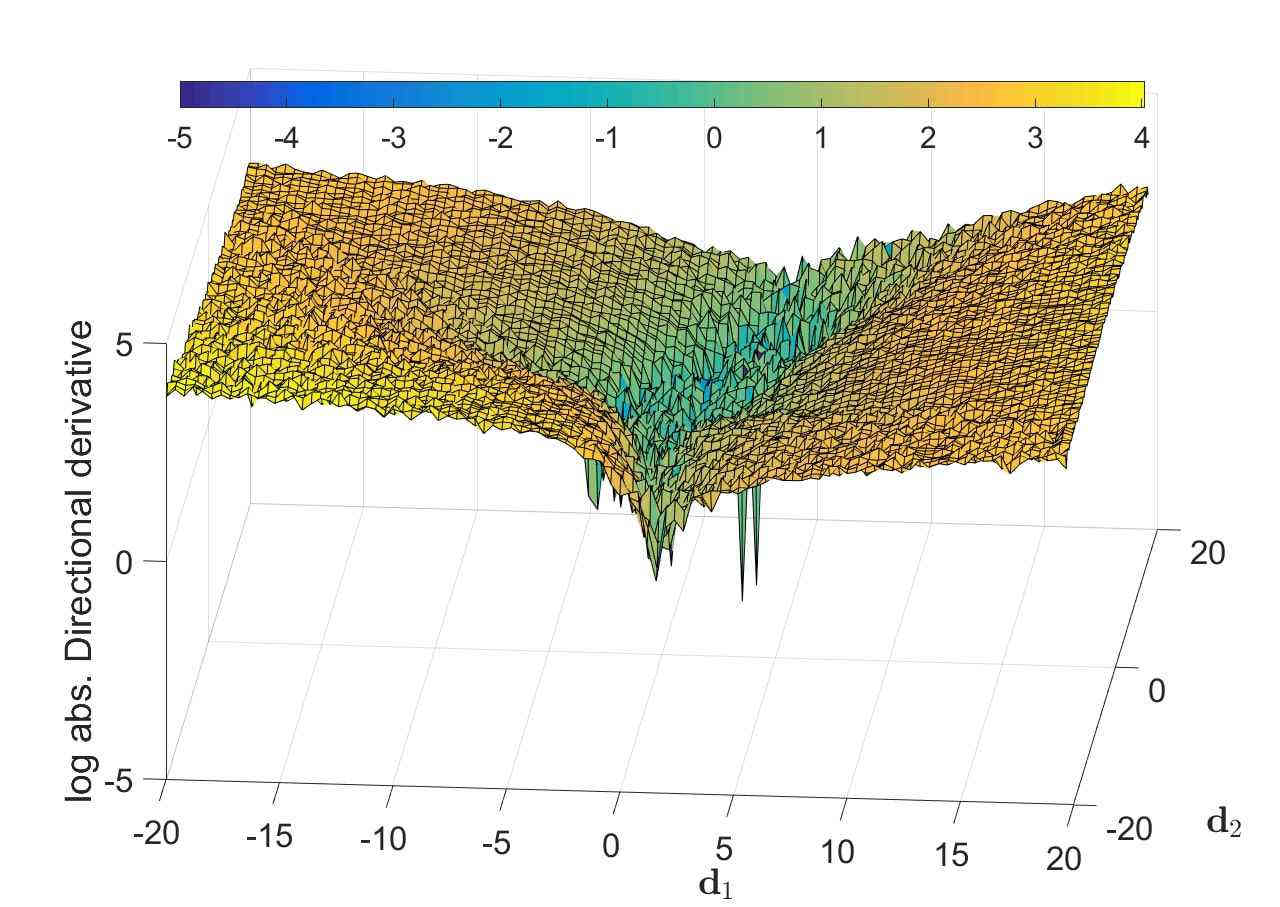}
		\caption{Function value, $|\mathcal{B}_{n,i}| = 10$}
		\label{fig_relu_dd_M}
	\end{subfigure}%
	
	\begin{subfigure}{.45\textwidth}
		\centering 
		\includegraphics[width=0.9\linewidth]{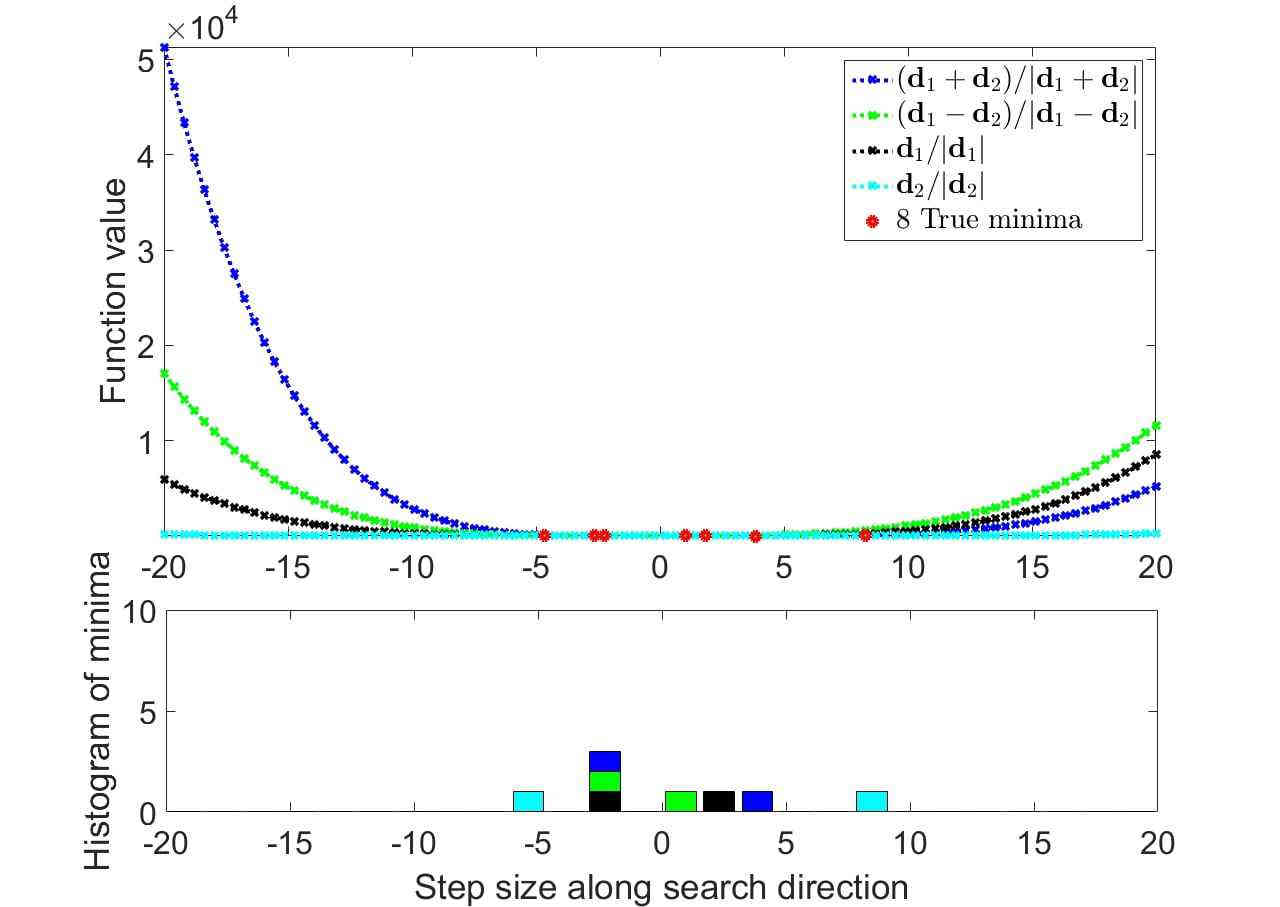}
		\caption{True minima along search directions}
		\label{fig_relu_fline_B}
	\end{subfigure}%
	\begin{subfigure}{.45\textwidth}
		\centering
		\includegraphics[width=0.9\linewidth]{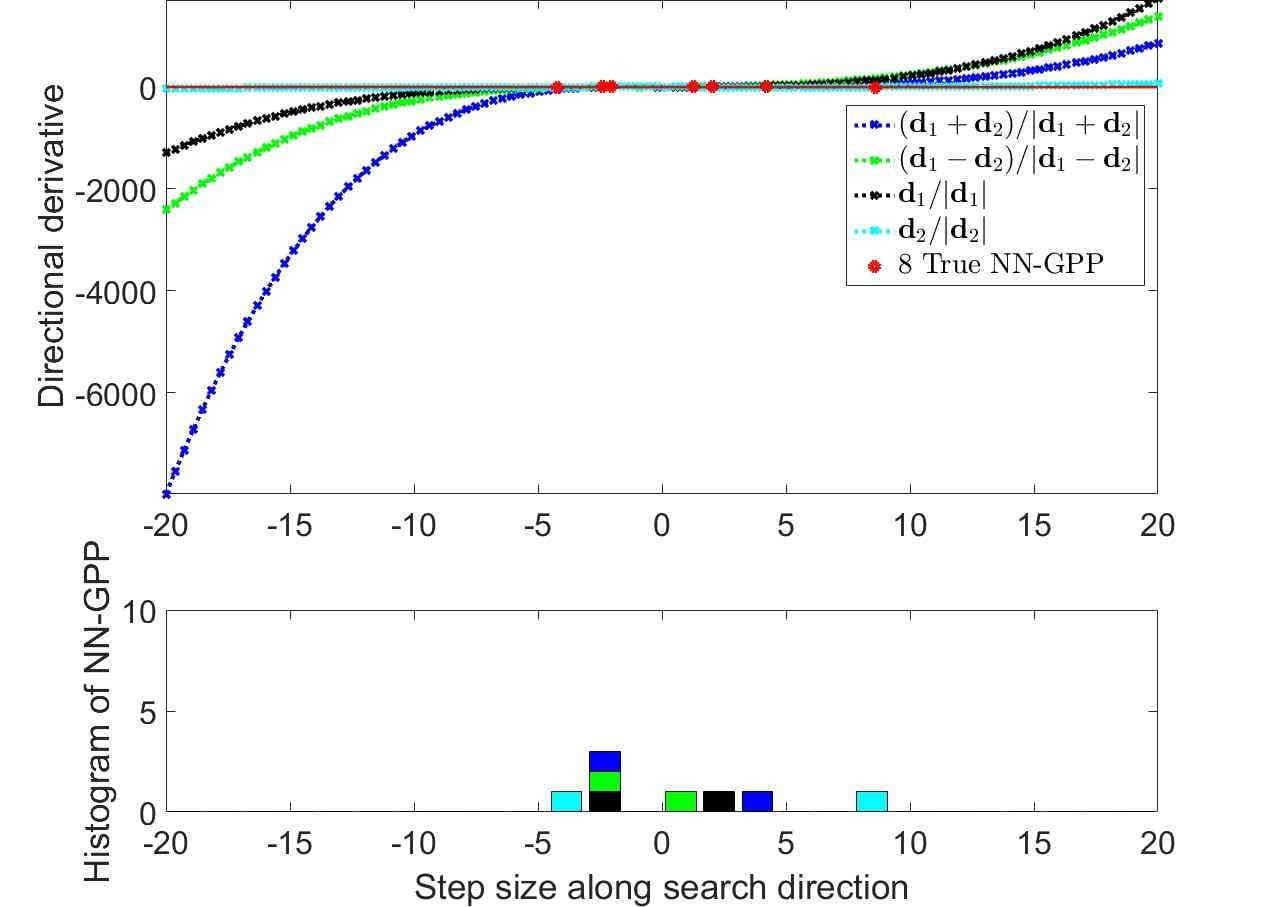}
		\caption{True NN-GPPs along search directions}
		\label{fig_relu_dline_B}
	\end{subfigure}%
	
	\begin{subfigure}{.45\textwidth}
		\centering 
		\includegraphics[width=0.9\linewidth]{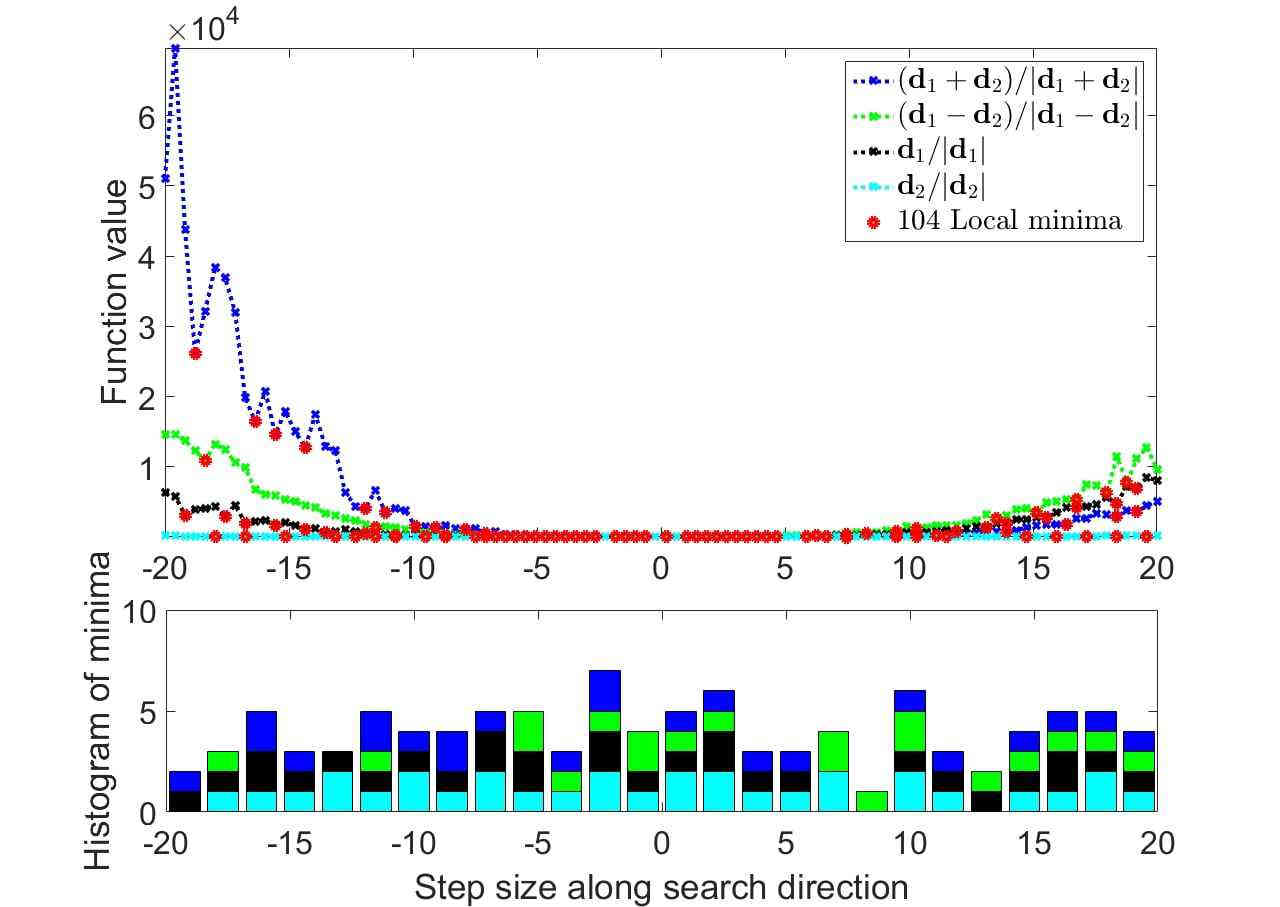}
		\caption{Local minima along search directions}
		\label{fig_relu_fline_M}
	\end{subfigure}%
	\begin{subfigure}{.45\textwidth}
		\centering
		\includegraphics[width=0.9\linewidth]{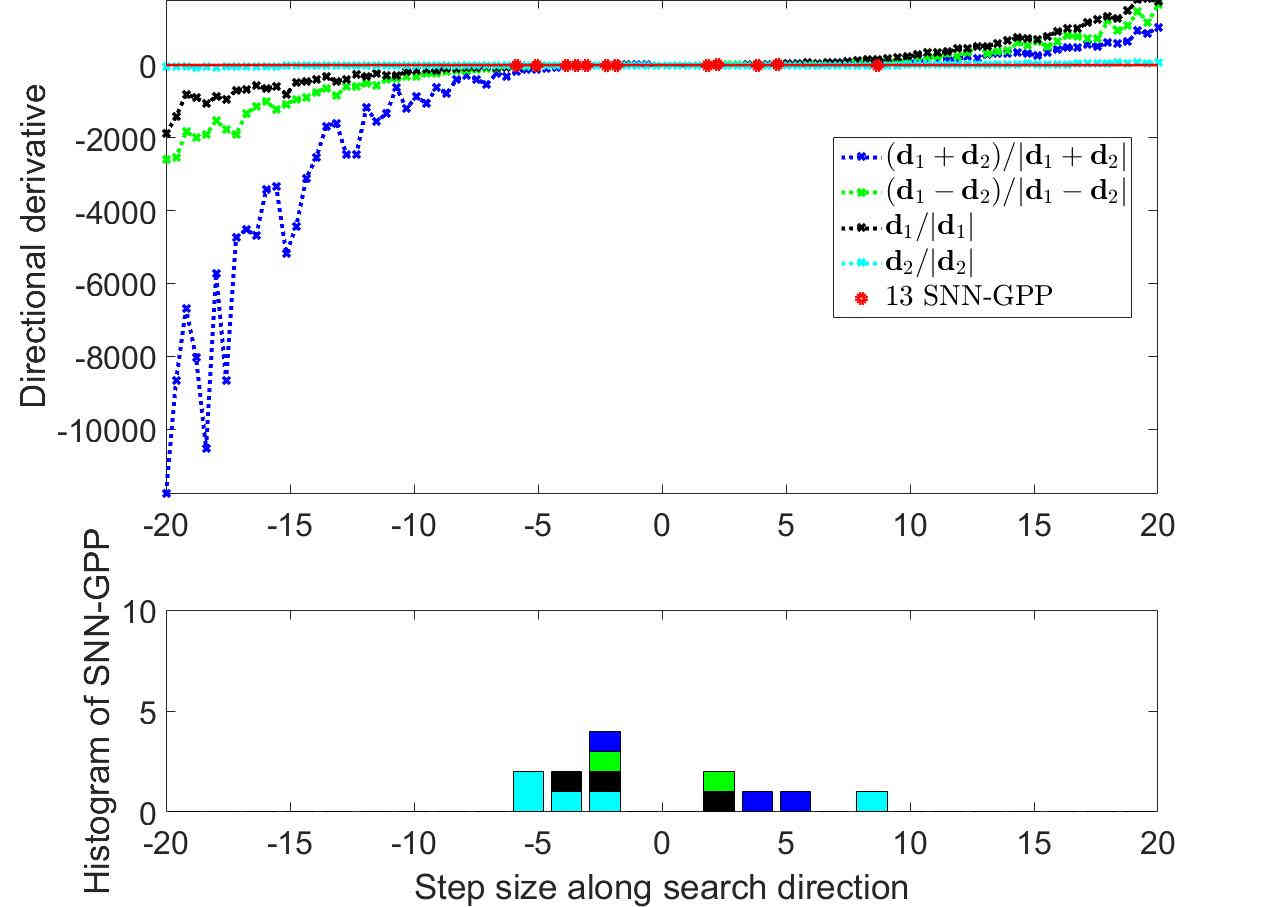}
		\caption{SNN-GPPs along search directions}
		\label{fig_relu_dline_M}
	\end{subfigure}%
	
	\caption{The ReLU AF: Pushing ReLU activations far into their active domain results in convex behaviour of the MSE loss function on a large scale. However, directional derivative and true optima plots indicate that more detail is contained in the basin.}
	\label{fig_relu}
\end{figure}

\begin{figure}[h!]
	\centering
	\begin{subfigure}{.45\textwidth}
		\centering 
		\includegraphics[width=0.9\linewidth]{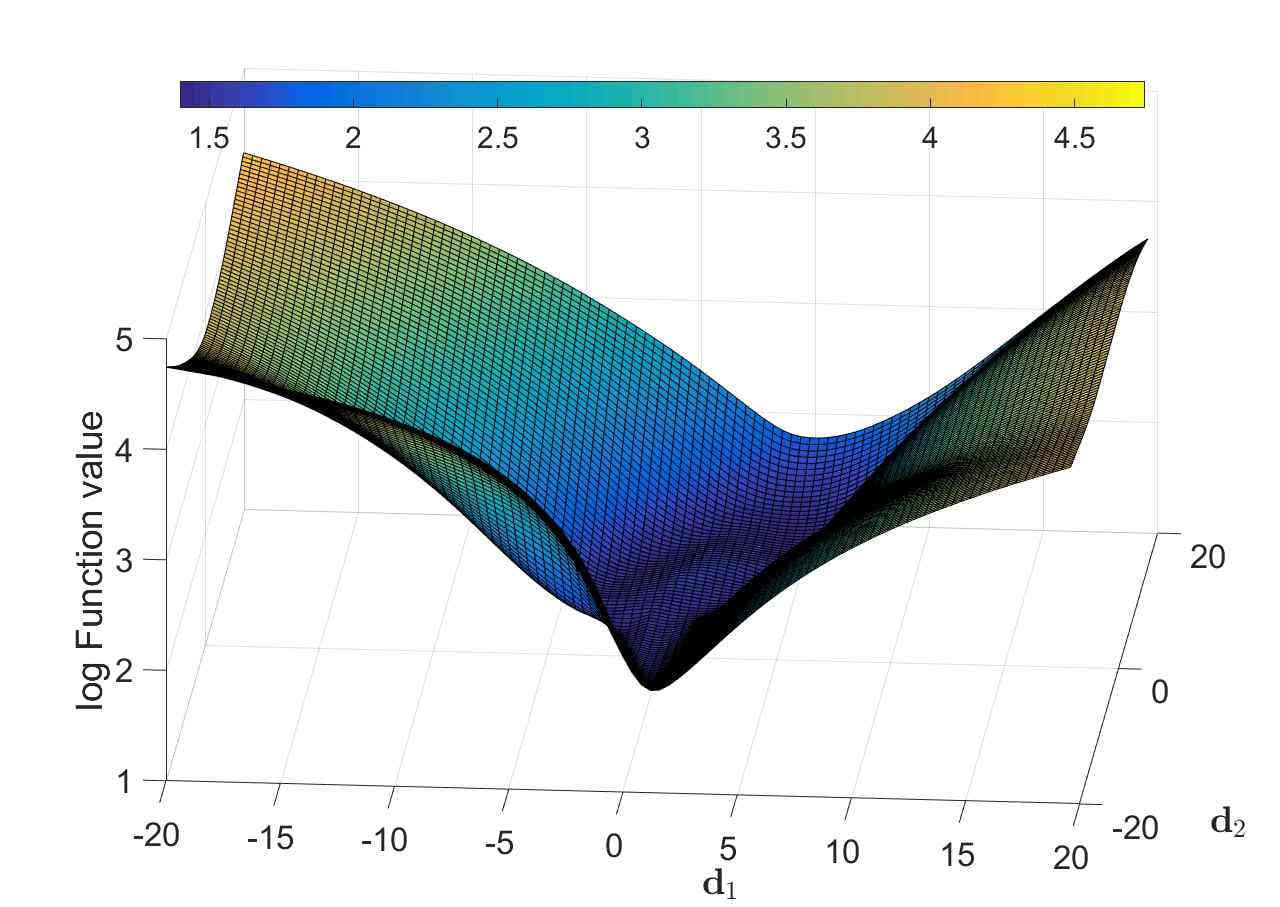}
		\caption{Function value, $M = 150$}
		\label{fig_lrelu_func_B}
	\end{subfigure}%
	\begin{subfigure}{.45\textwidth}
		\centering
		\includegraphics[width=0.9\linewidth]{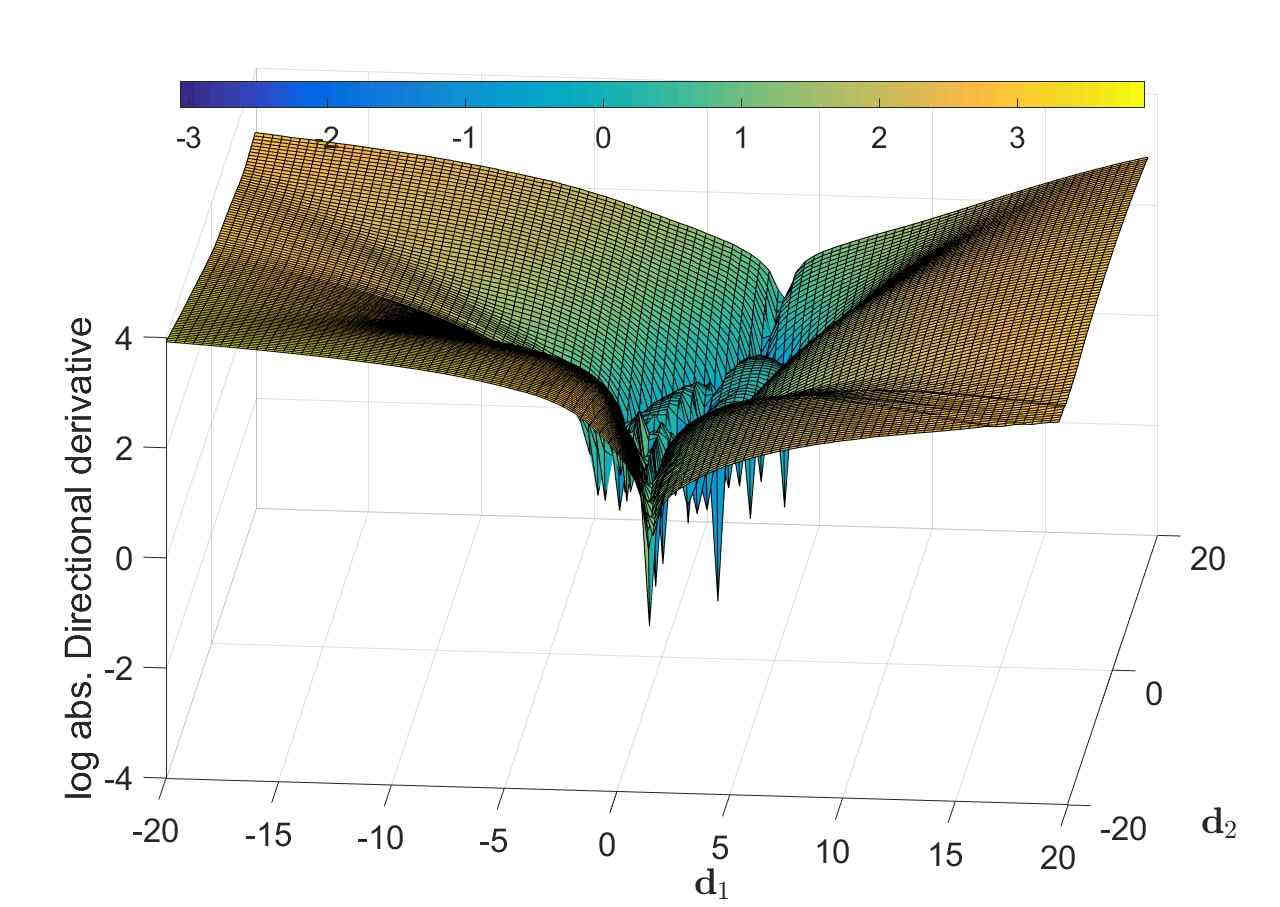}
		\caption{Directional derivative, $M = 150$}
		\label{fig_lrelu_dd_B}
	\end{subfigure}%
	
	\begin{subfigure}{.45\textwidth}
		\centering 
		\includegraphics[width=0.9\linewidth]{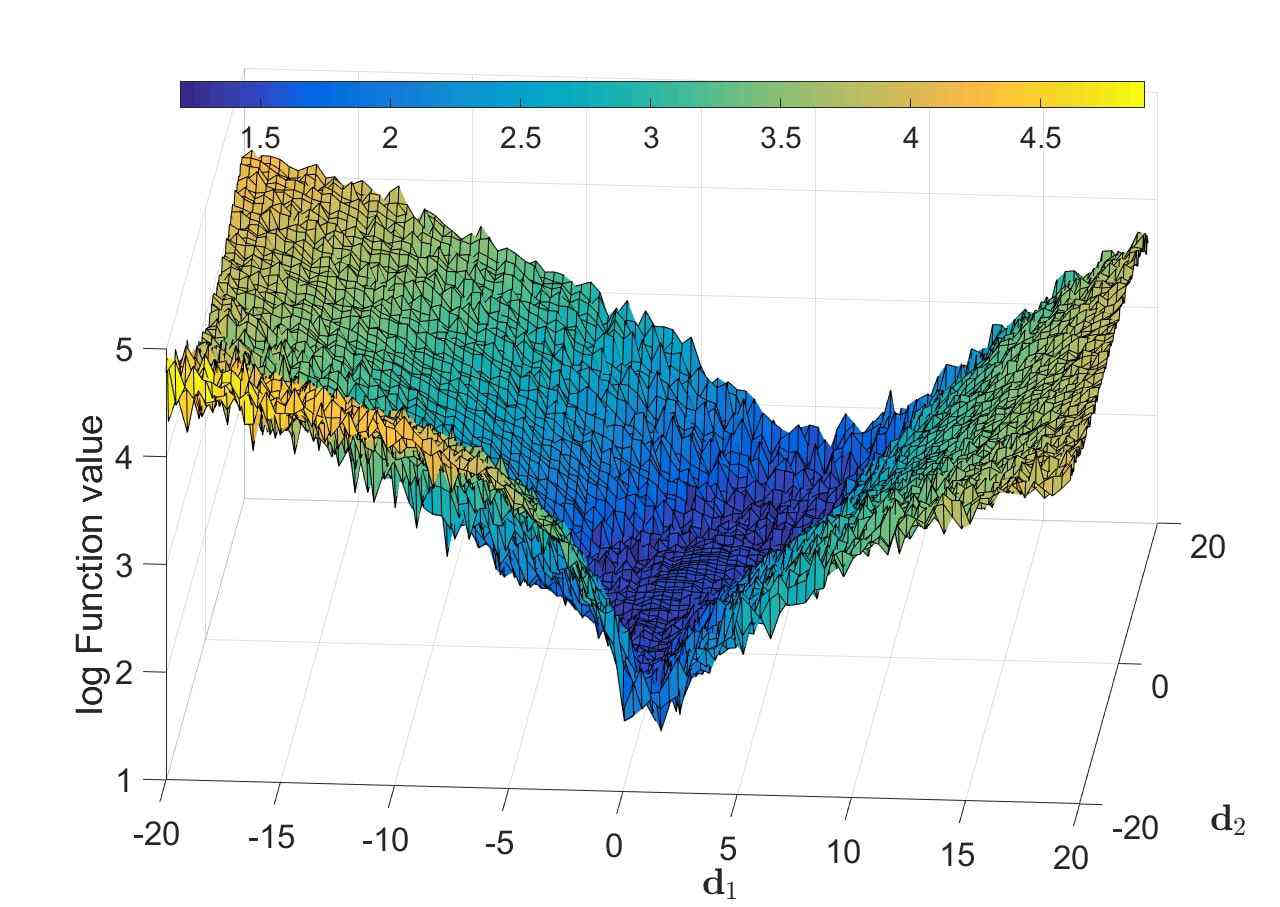}
		\caption{Function value, $|\mathcal{B}_{n,i}| = 10$}
		\label{fig_lrelu_func_M}
	\end{subfigure}%
	\begin{subfigure}{.45\textwidth}
		\centering
		\includegraphics[width=0.9\linewidth]{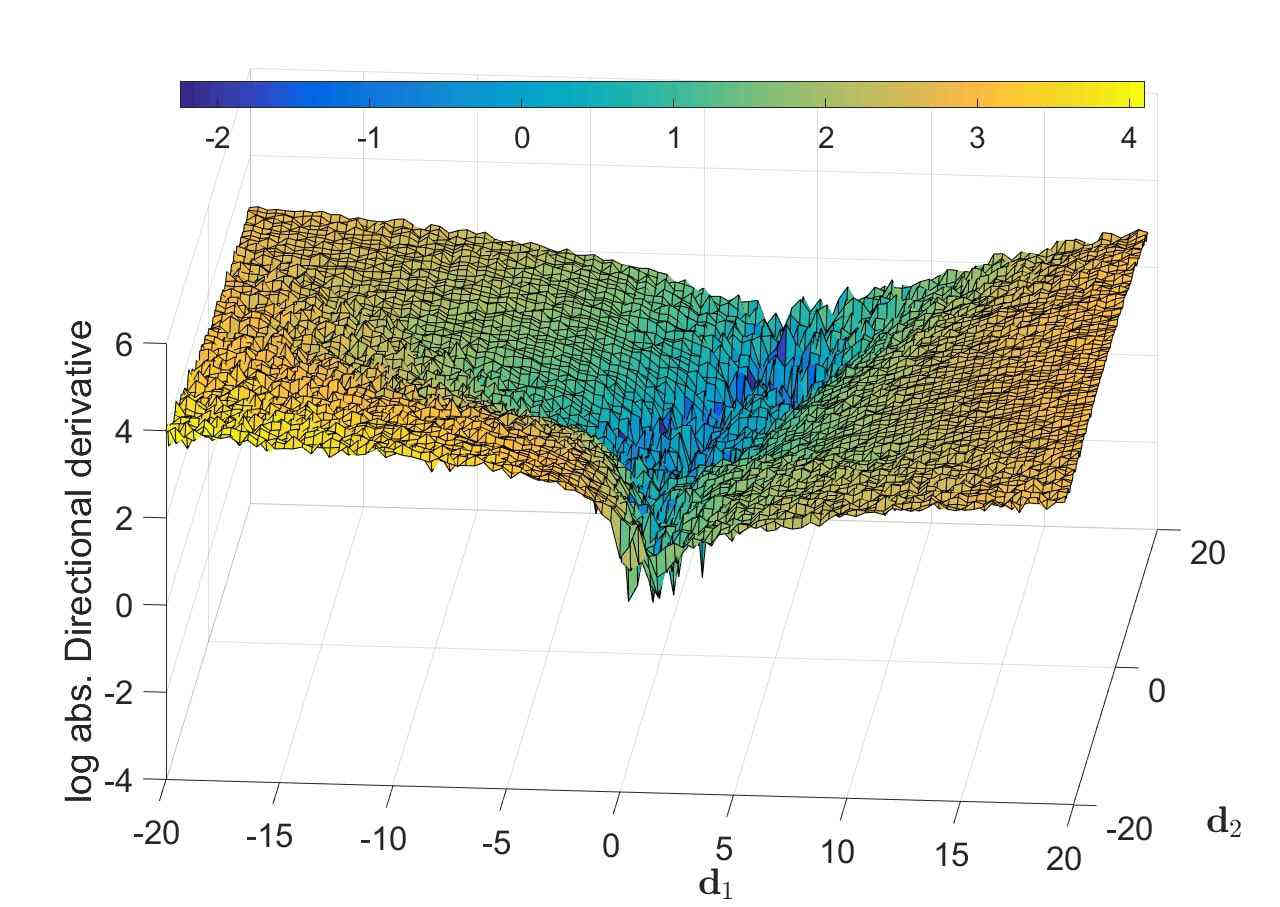}
		\caption{Directional derivative, $|\mathcal{B}_{n,i}| = 10$}
		\label{fig_lrelu_dd_M}
	\end{subfigure}%
	
	\begin{subfigure}{.45\textwidth}
		\centering 
		\includegraphics[width=0.9\linewidth]{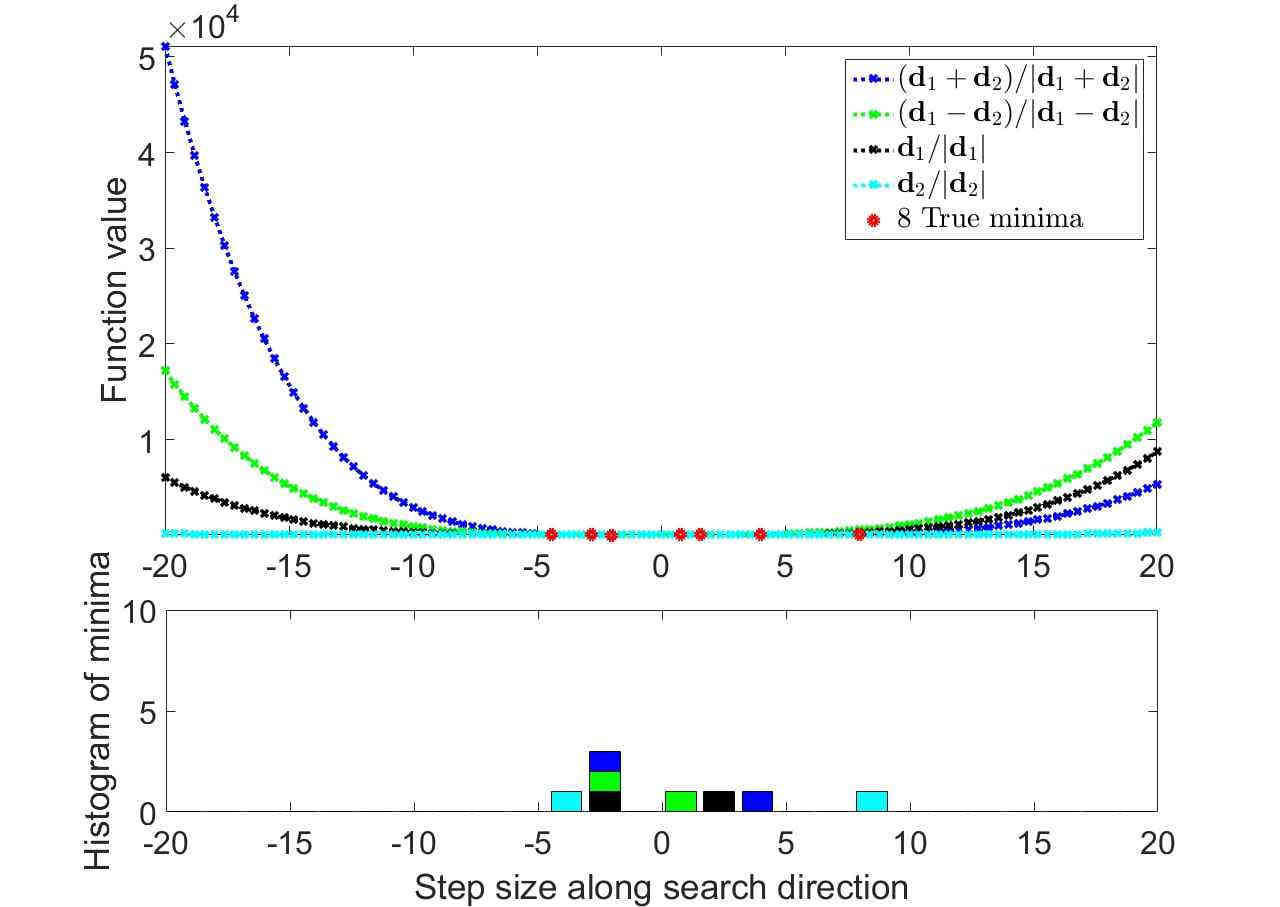}
		\caption{True minima along search directions}
		\label{fig_lrelu_fline_B}
	\end{subfigure}%
	\begin{subfigure}{.45\textwidth}
		\centering
		\includegraphics[width=0.9\linewidth]{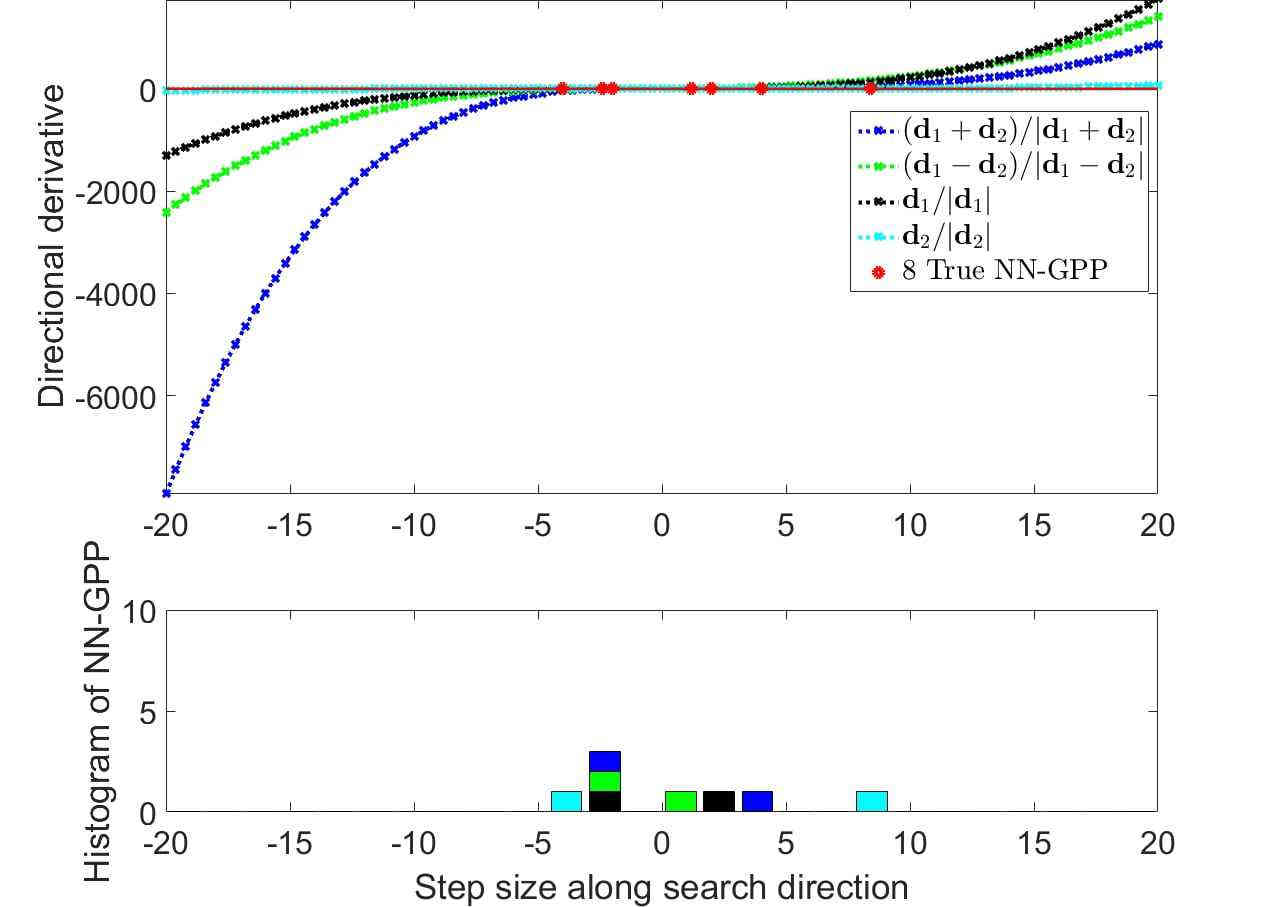}
		\caption{True NN-GPPs along search directions}
		\label{fig_lrelu_dline_B}
	\end{subfigure}%
	
	\begin{subfigure}{.45\textwidth}
		\centering 
		\includegraphics[width=0.9\linewidth]{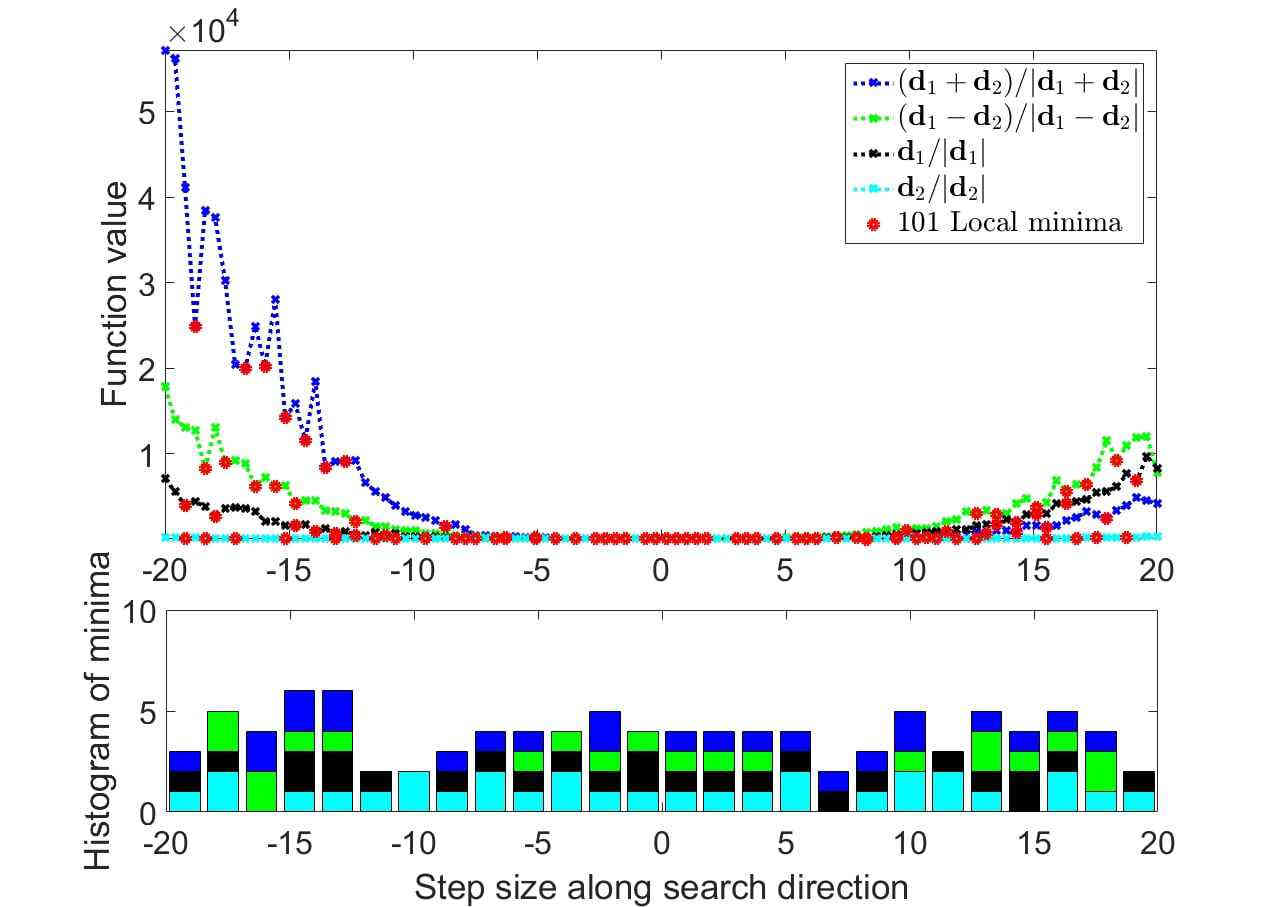}
		\caption{Local minima along search directions}
		\label{fig_lrelu_fline_M}
	\end{subfigure}%
	\begin{subfigure}{.45\textwidth}
		\centering
		\includegraphics[width=0.9\linewidth]{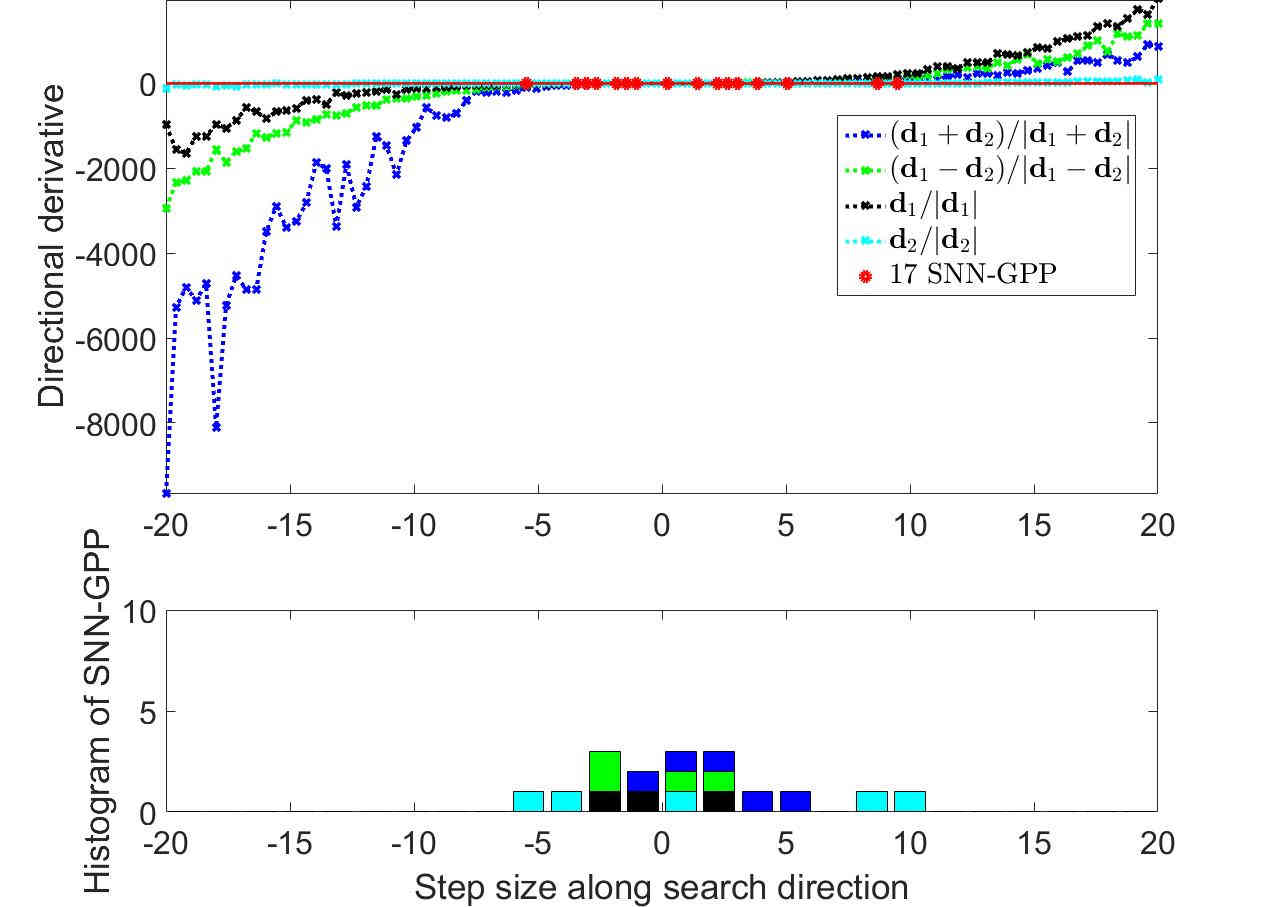}
		\caption{SNN-GPPs along search directions}
		\label{fig_lrelu_dline_M}
	\end{subfigure}%
	
	\caption{The leaky ReLU AF: Since the magnitude of the "leaky" gradient is relatively small, its contribution is not apparent at this length scale. Therefore, the plots look very similar to those of ReLU. SNN-GPPs are concentrated around the centre, where the magnitude of the directional derivatives is small.}
	\label{fig_lrelu}
\end{figure}

\begin{figure}[h!]
	\centering
	\begin{subfigure}{.45\textwidth}
		\centering 
		\includegraphics[width=0.9\linewidth]{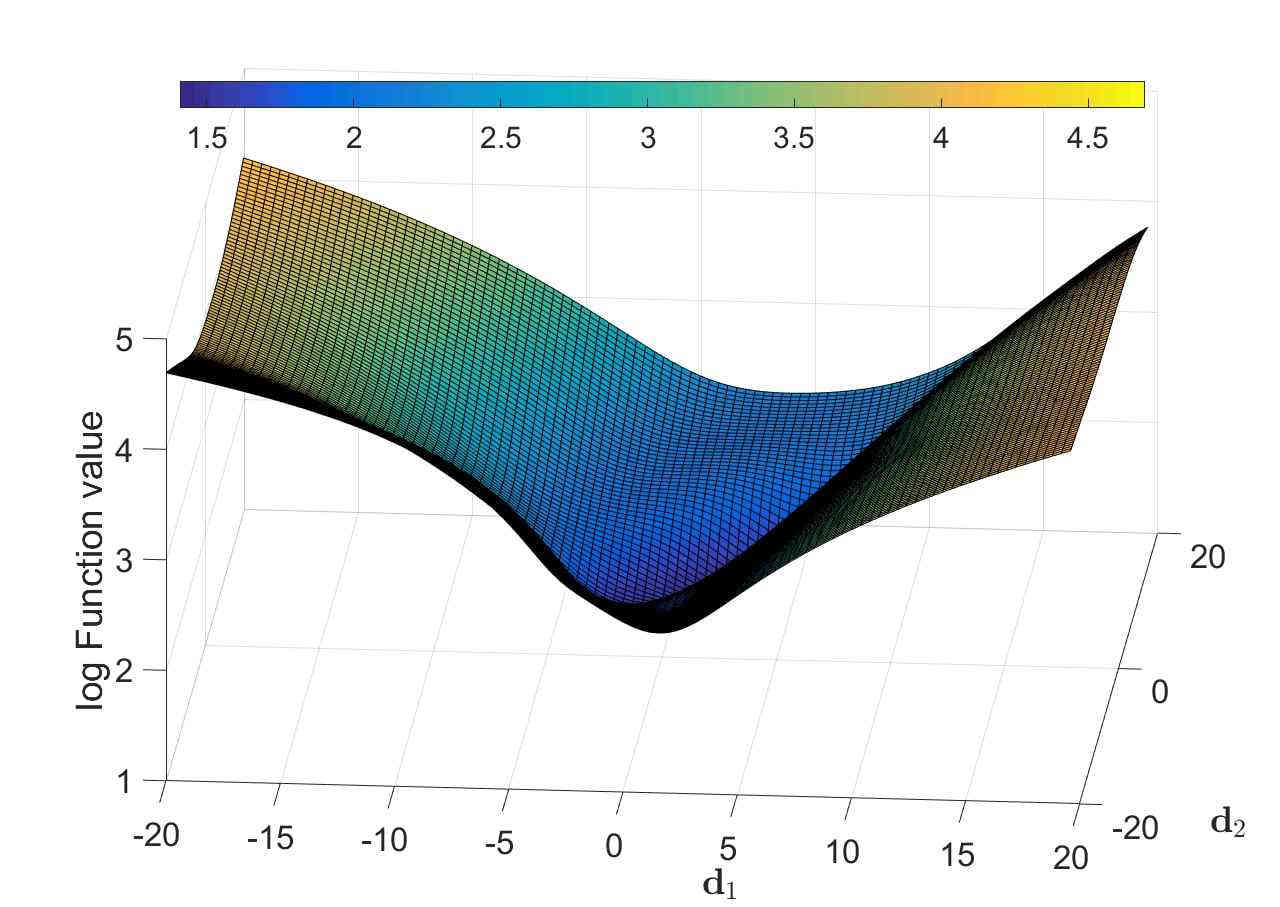}
		\caption{Function value, $M = 150$}
		\label{fig_elu_func_B}
	\end{subfigure}%
	\begin{subfigure}{.45\textwidth}
		\centering
		\includegraphics[width=0.9\linewidth]{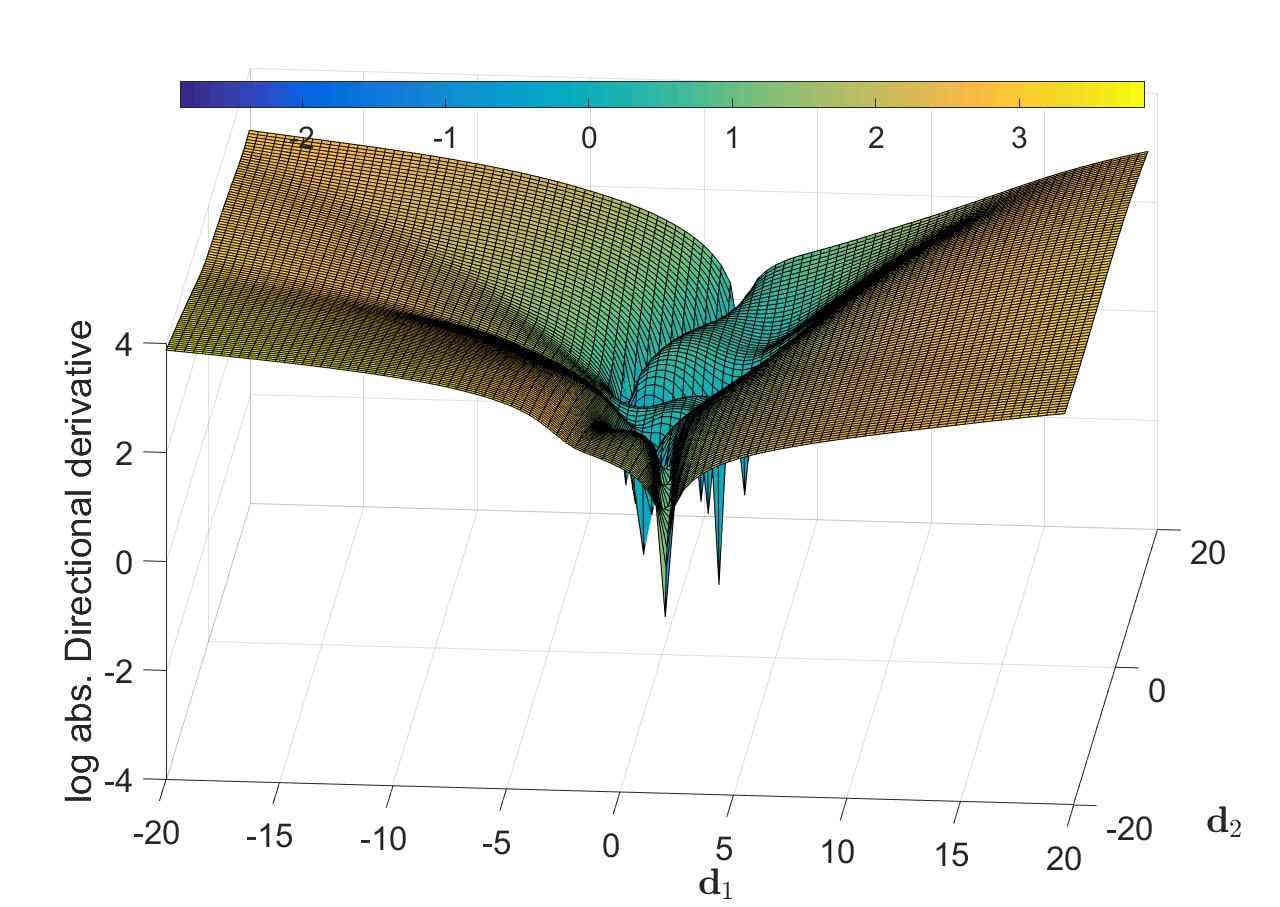}
		\caption{Directional derivative, $M = 150$}
		\label{fig_elu_dd_B}
	\end{subfigure}%
	
	\begin{subfigure}{.45\textwidth}
		\centering 
		\includegraphics[width=0.9\linewidth]{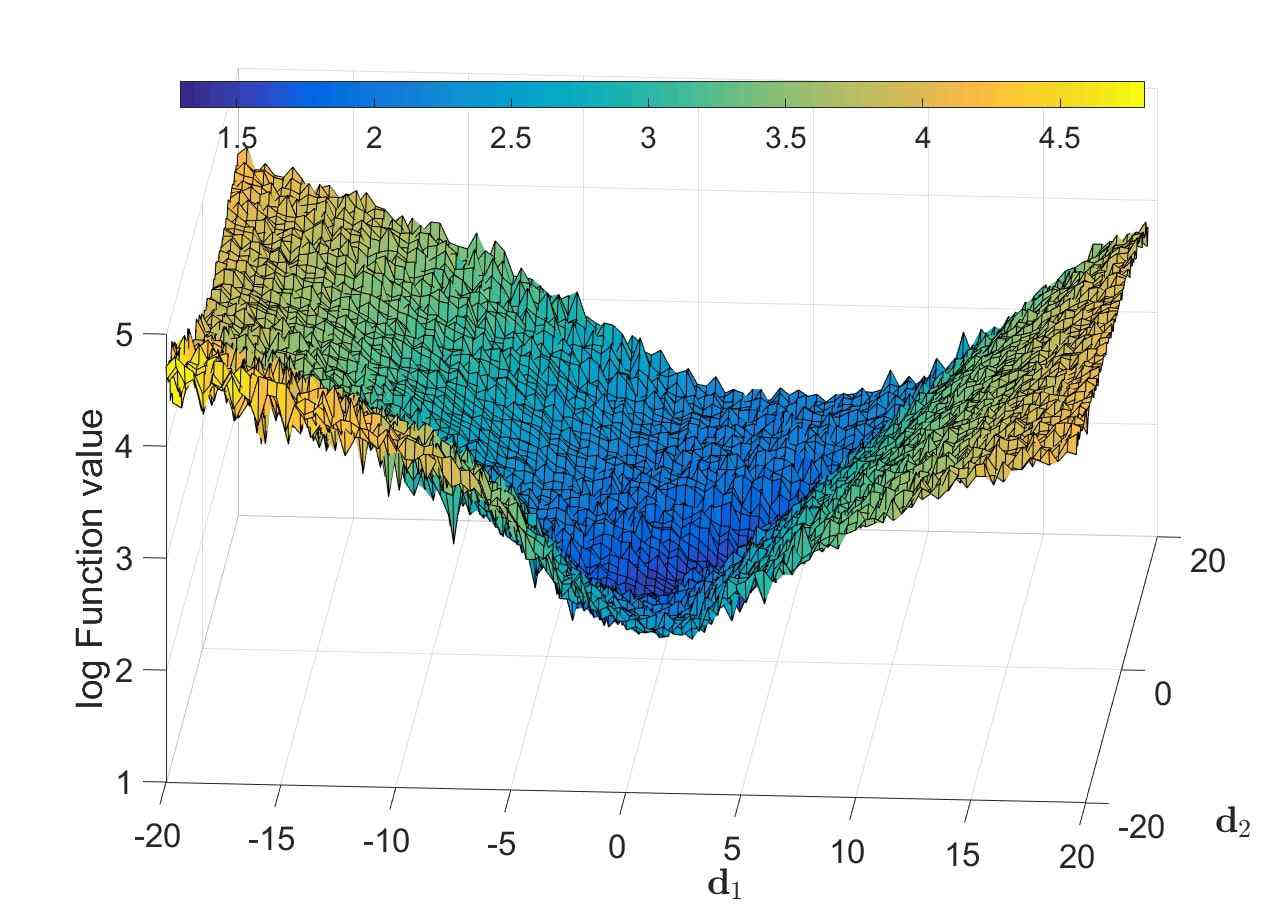}
		\caption{Function value, $|\mathcal{B}_{n,i}| = 10$}
		\label{fig_elu_func_M}
	\end{subfigure}%
	\begin{subfigure}{.45\textwidth}
		\centering
		\includegraphics[width=0.9\linewidth]{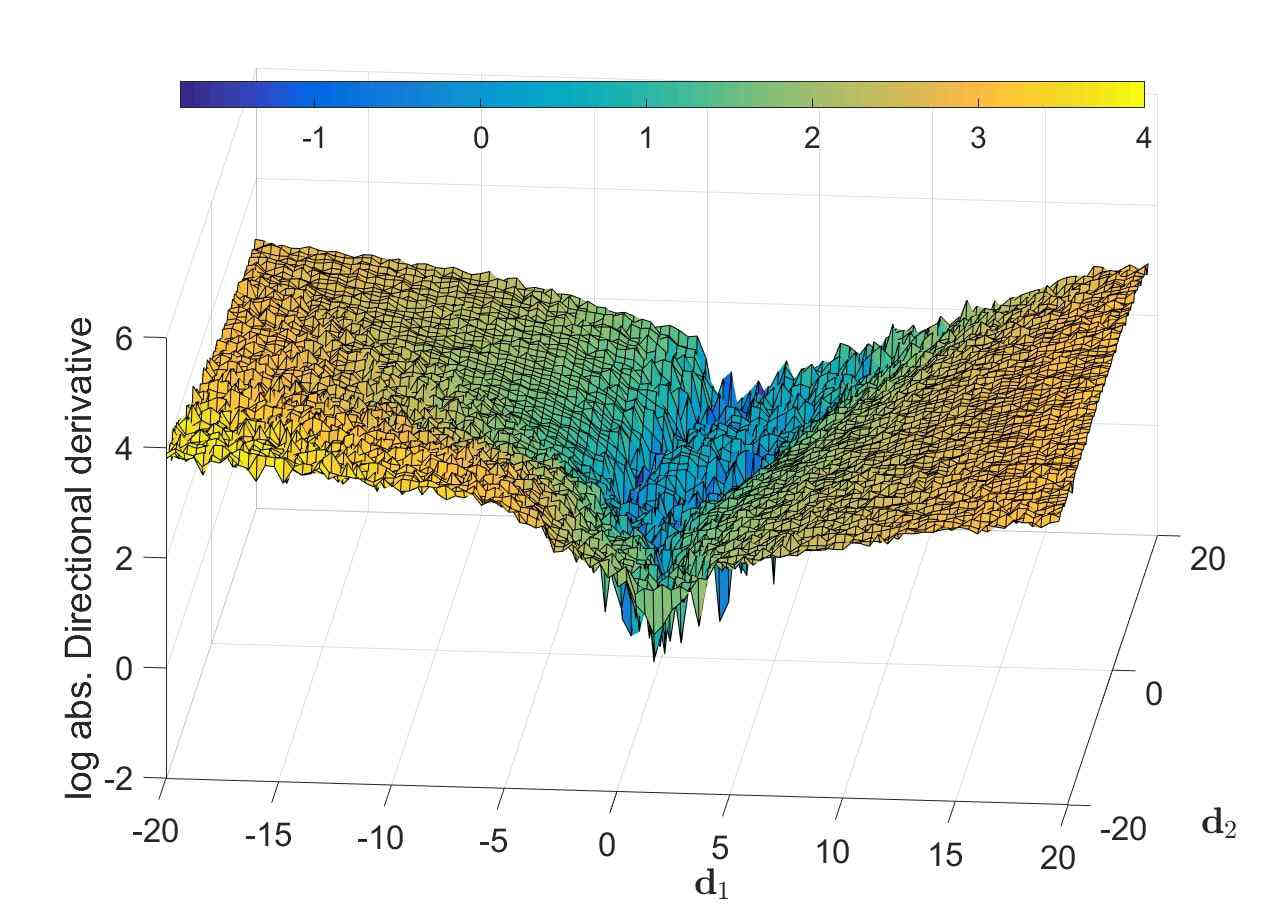}
		\caption{Directional derivative, $|\mathcal{B}_{n,i}| = 10$}
		\label{fig_elu_dd_M}
	\end{subfigure}%
	
	\begin{subfigure}{.45\textwidth}
		\centering 
		\includegraphics[width=0.9\linewidth]{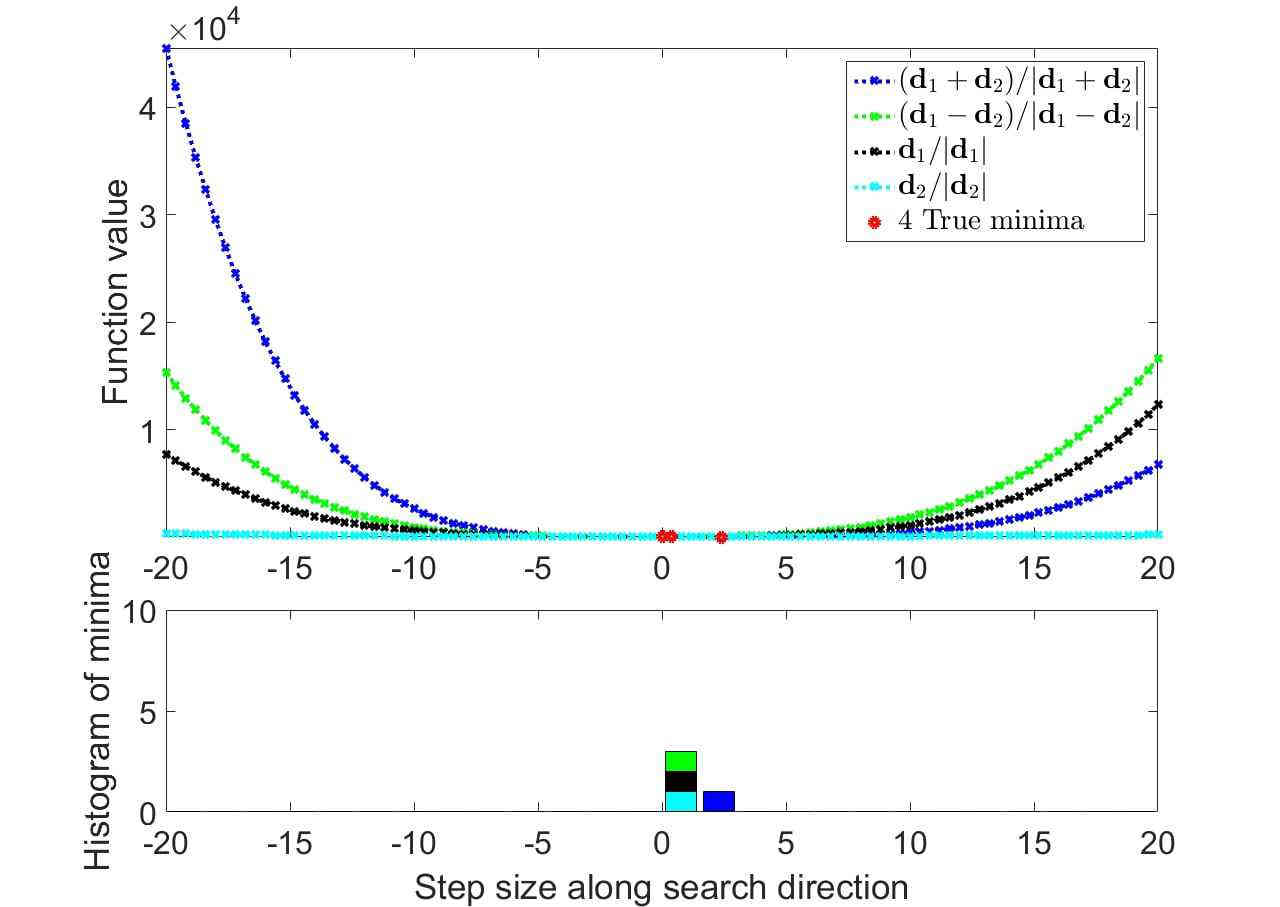}
		\caption{True minima along search directions}
		\label{fig_elu_fline_B}
	\end{subfigure}%
	\begin{subfigure}{.45\textwidth}
		\centering
		\includegraphics[width=0.9\linewidth]{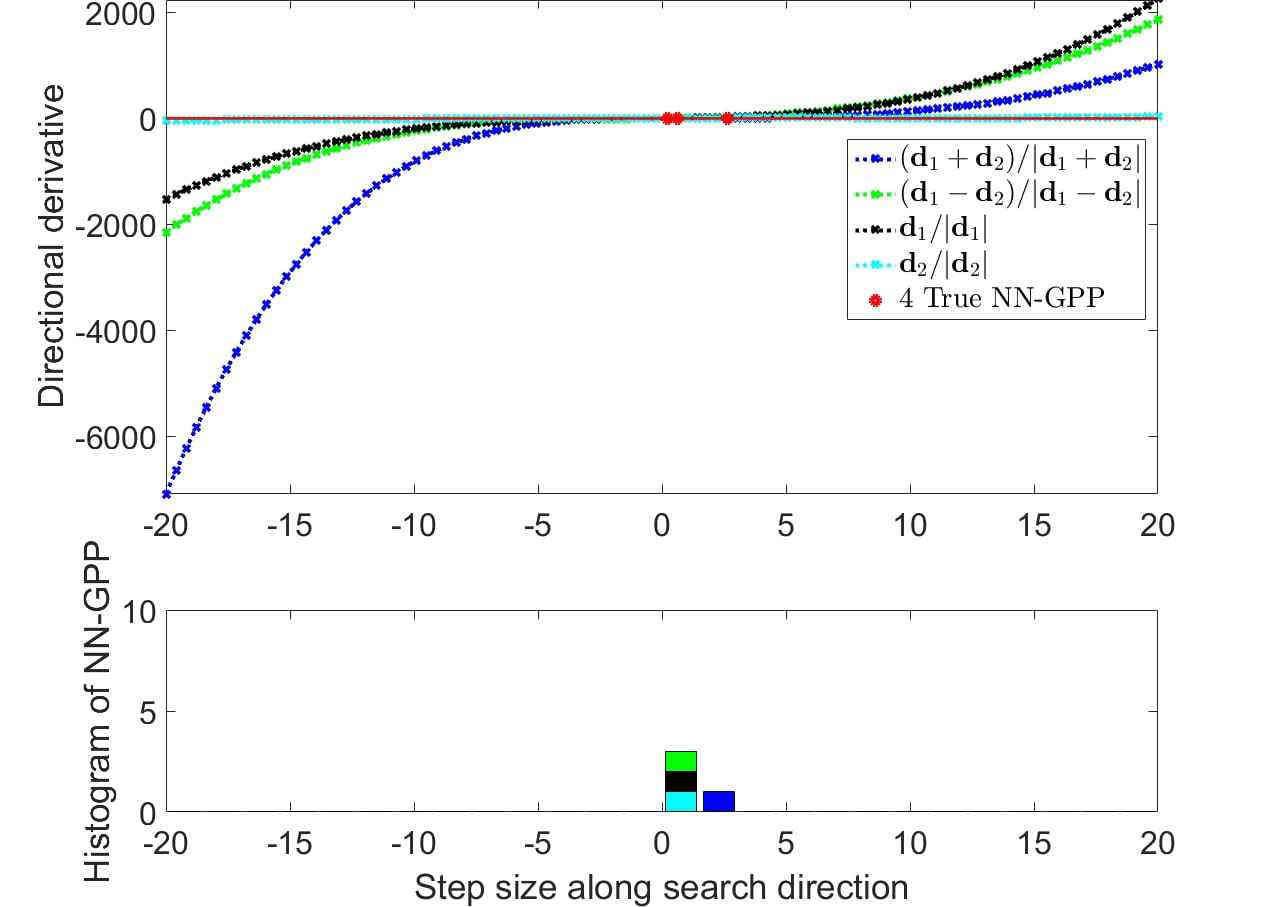}
		\caption{True NN-GPPs along search directions}
		\label{fig_elu_dline_B}
	\end{subfigure}%
	
	\begin{subfigure}{.45\textwidth}
		\centering 
		\includegraphics[width=0.9\linewidth]{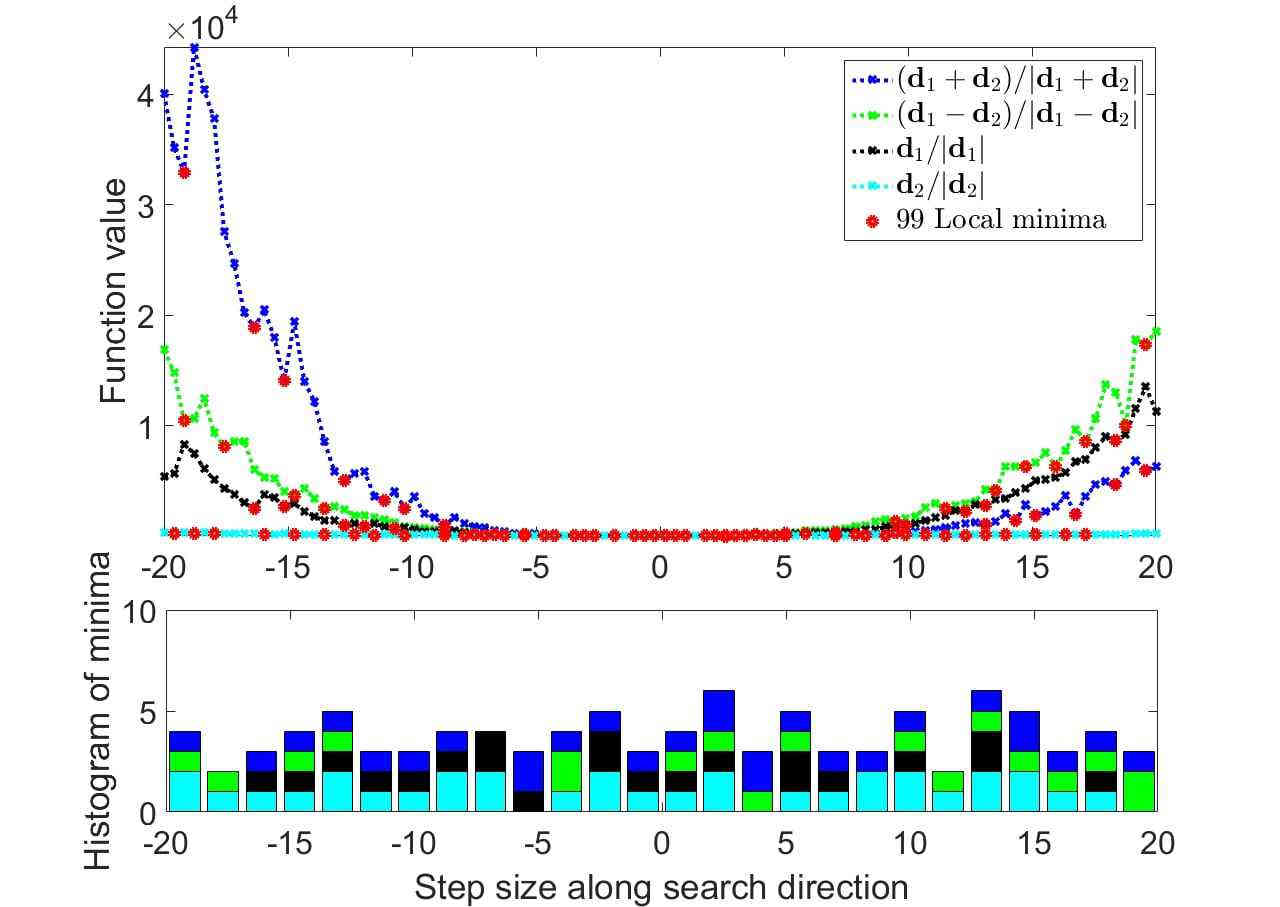}
		\caption{Local minima along search directions}
		\label{fig_elu_fline_M}
	\end{subfigure}%
	\begin{subfigure}{.45\textwidth}
		\centering
		\includegraphics[width=0.9\linewidth]{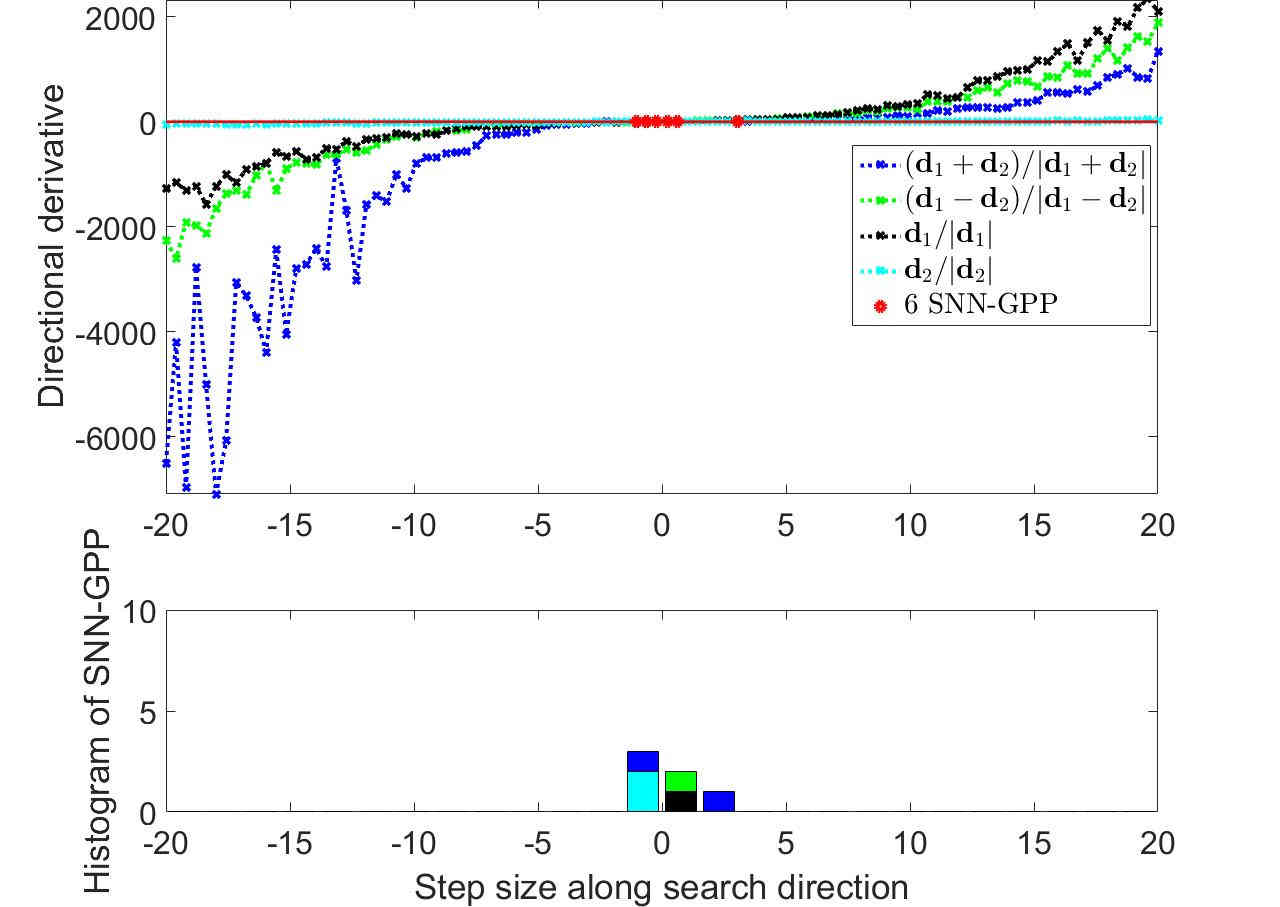}
		\caption{SNN-GPPs along search directions}
		\label{fig_elu_dline_M}
	\end{subfigure}%
	\caption{The ELU AF: The shapes of the convex element of the loss function are similar to those of the other ReLUs. However, the structure in the basin seems to be different. This is confirmed by the number of true optima. The narrow spatial grouping of SNN-GPPs is the contribution of the continuous AF derivative.}
	\label{fig_elu}
\end{figure}

\section*{Appendix B.}
\label{AppB}

\begin{figure}[h!]
	\centering
	\begin{subfigure}{.45\textwidth}
		\centering 
		\includegraphics[width=0.9\linewidth]{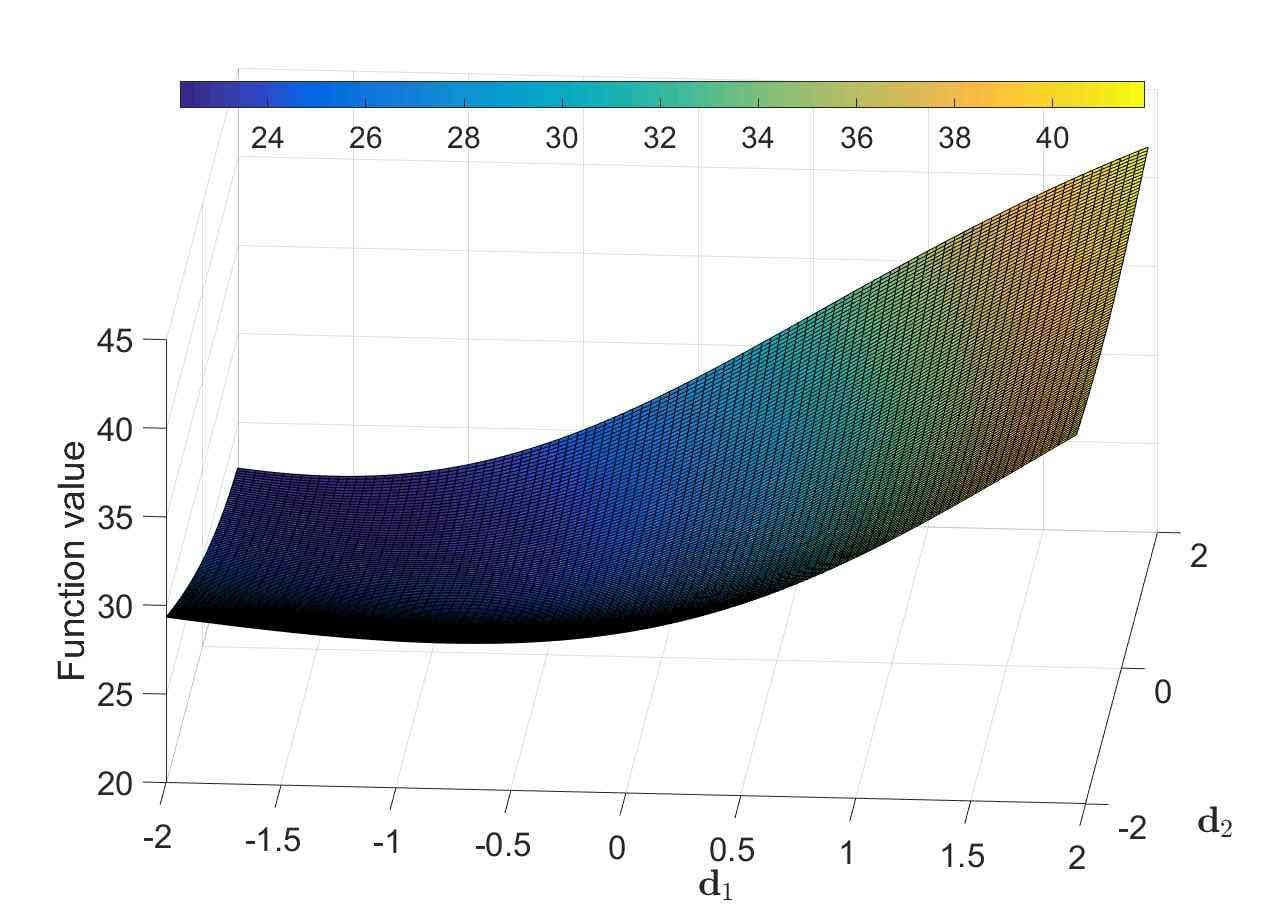}
		\caption{Function value, $M = 150$}
	\end{subfigure}%
	\begin{subfigure}{.45\textwidth}
		\centering
		\includegraphics[width=0.9\linewidth]{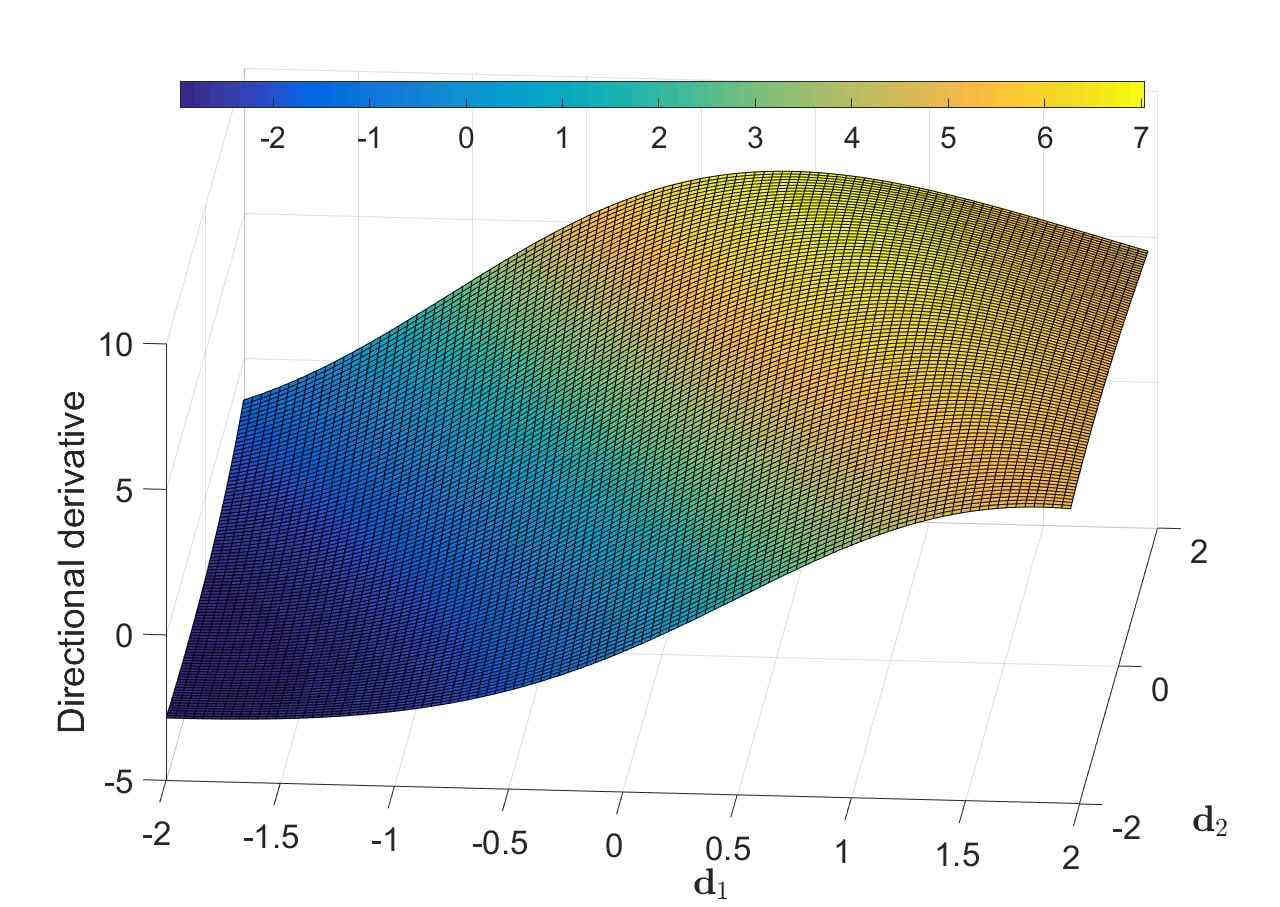}
		\caption{Directional derivative, $M = 150$}
	\end{subfigure}%
	
	\begin{subfigure}{.45\textwidth}
		\centering 
		\includegraphics[width=0.9\linewidth]{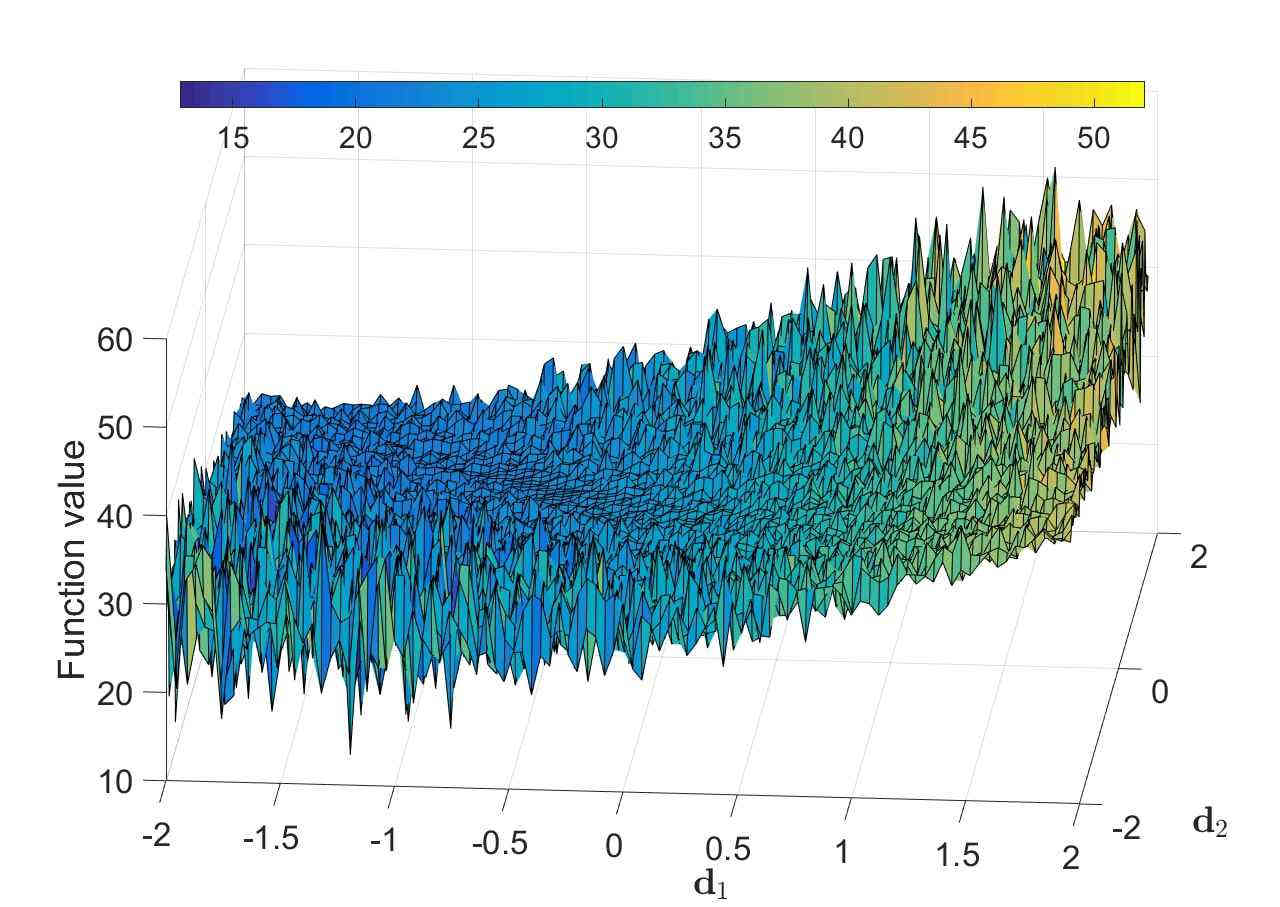}
		\caption{Function value, $|\mathcal{B}_{n,i}| = 10$}
	\end{subfigure}%
	\begin{subfigure}{.45\textwidth}
		\centering
		\includegraphics[width=0.9\linewidth]{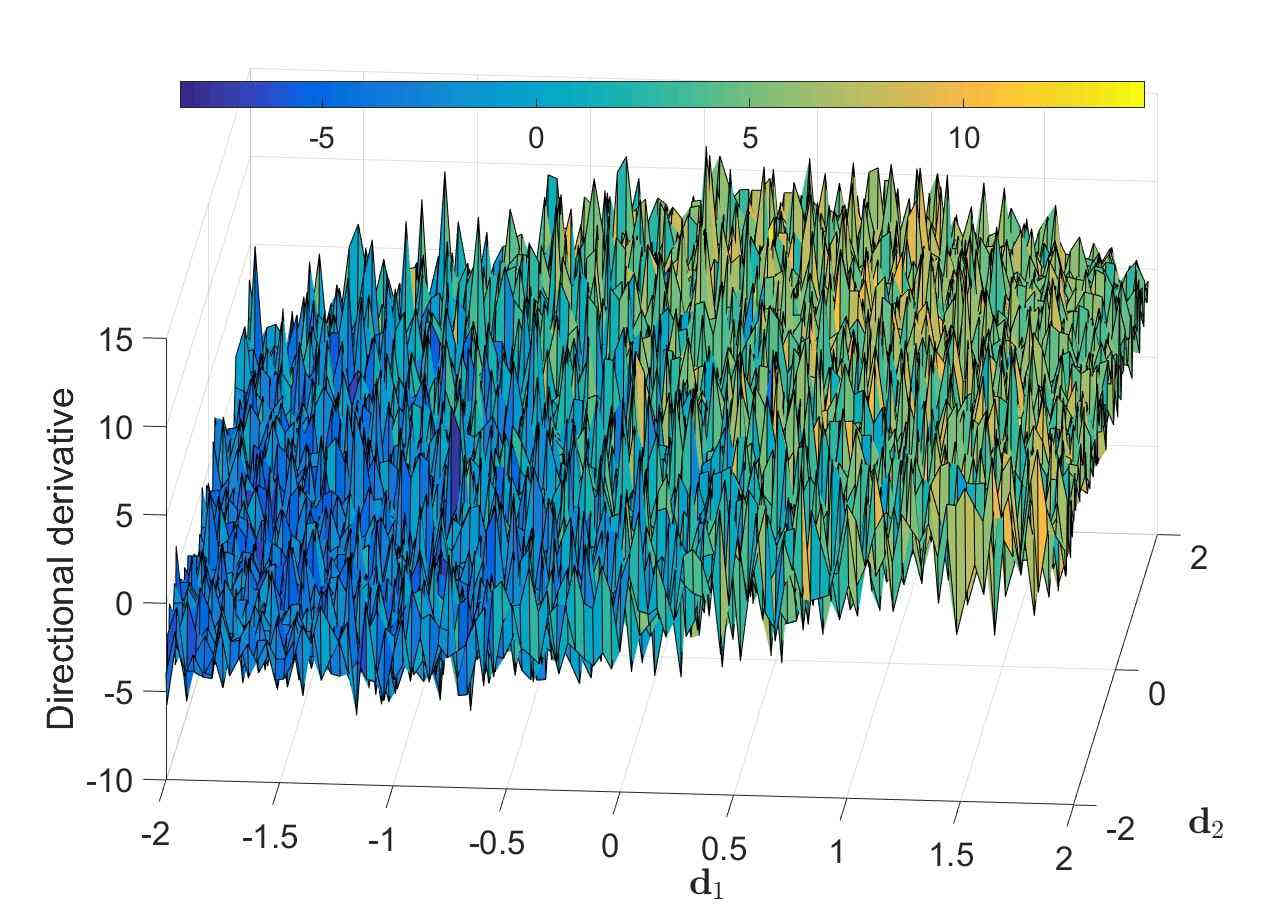}
		\caption{Directional derivative, $|\mathcal{B}_{n,i}| = 10$}
	\end{subfigure}%
	
	\begin{subfigure}{.45\textwidth}
		\centering 
		\includegraphics[width=0.9\linewidth]{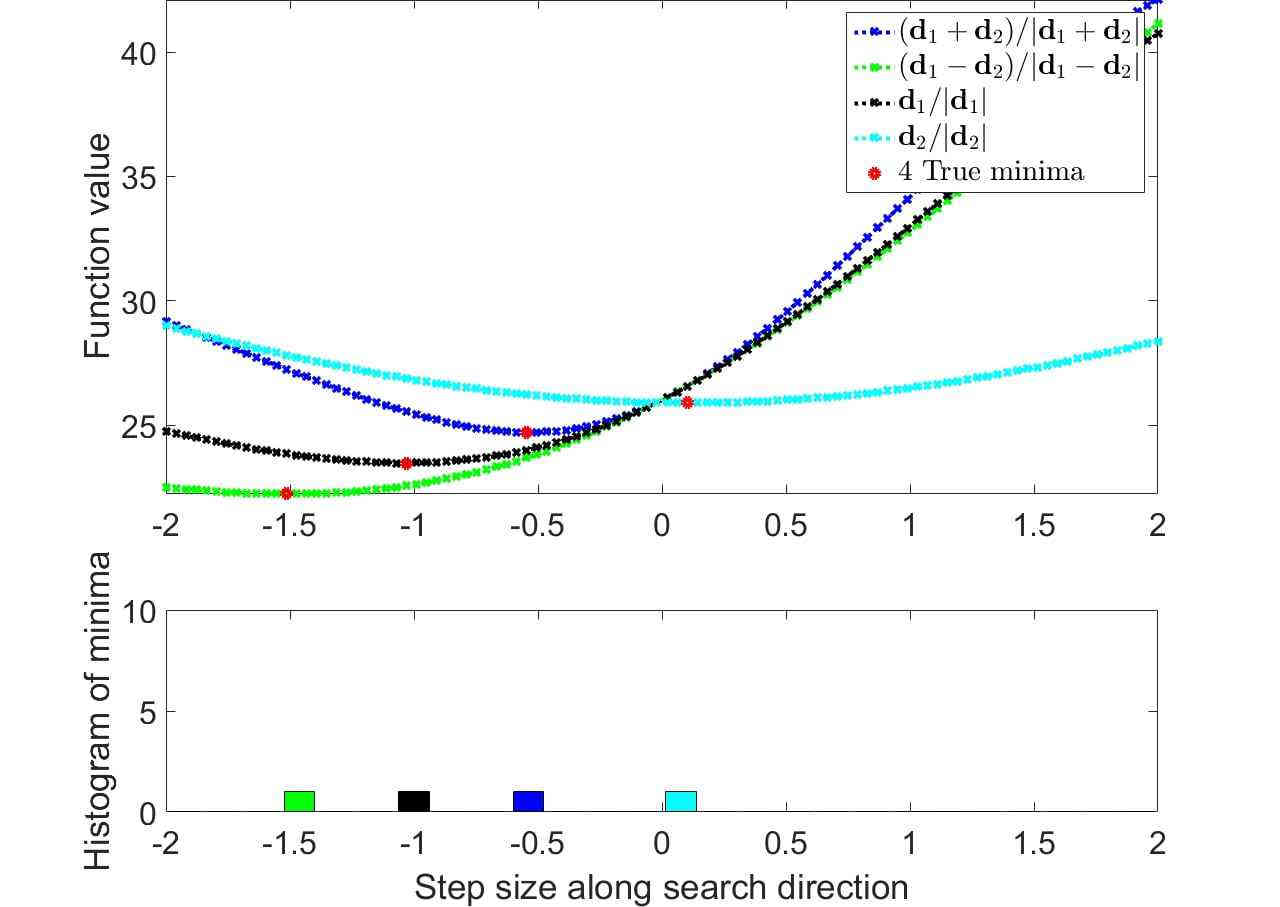}
		\caption{True minima along search directions}
	\end{subfigure}%
	\begin{subfigure}{.45\textwidth}
		\centering
		\includegraphics[width=0.9\linewidth]{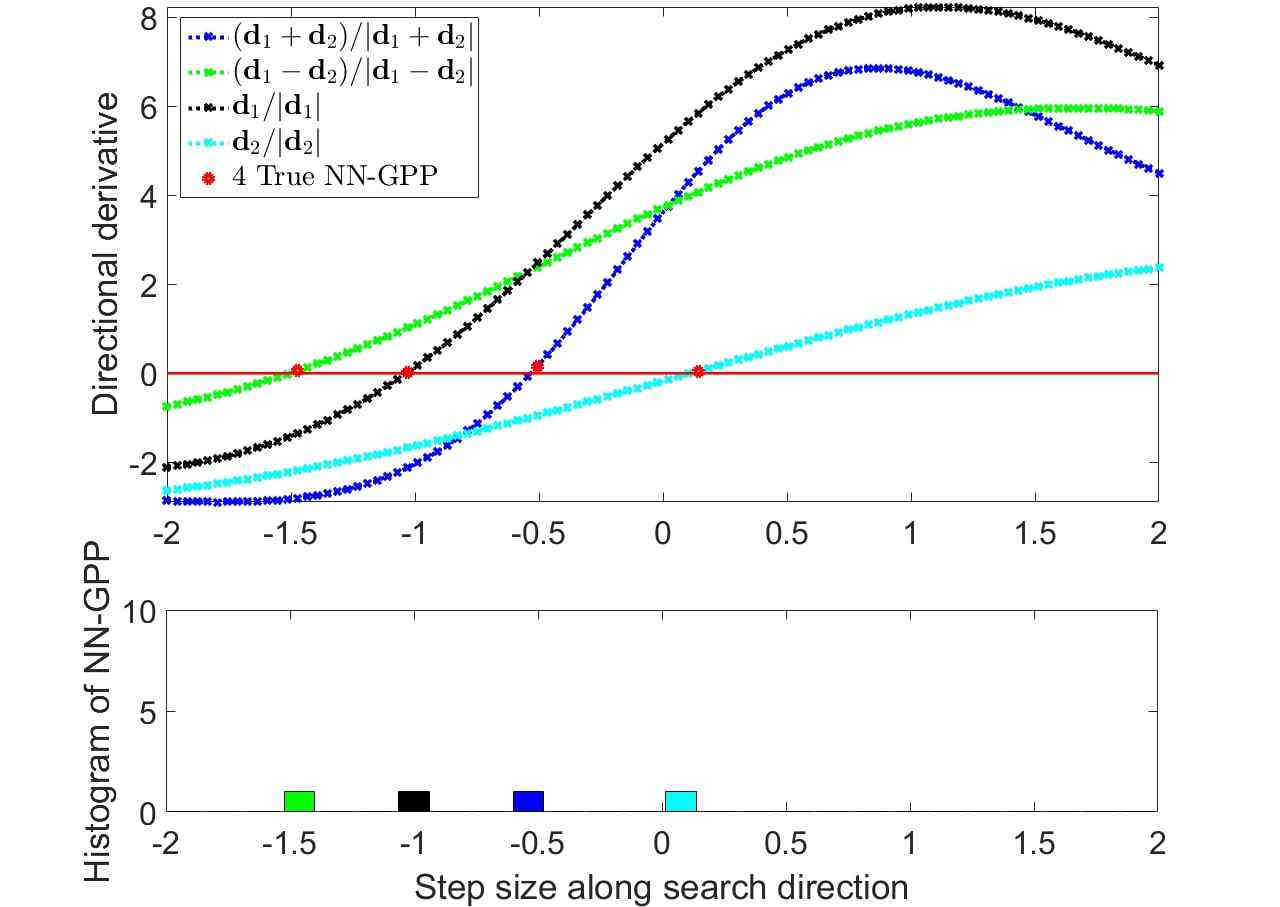}
		\caption{True NN-GPPs along search directions}
	\end{subfigure}%
	
	\begin{subfigure}{.45\textwidth}
		\centering 
		\includegraphics[width=0.9\linewidth]{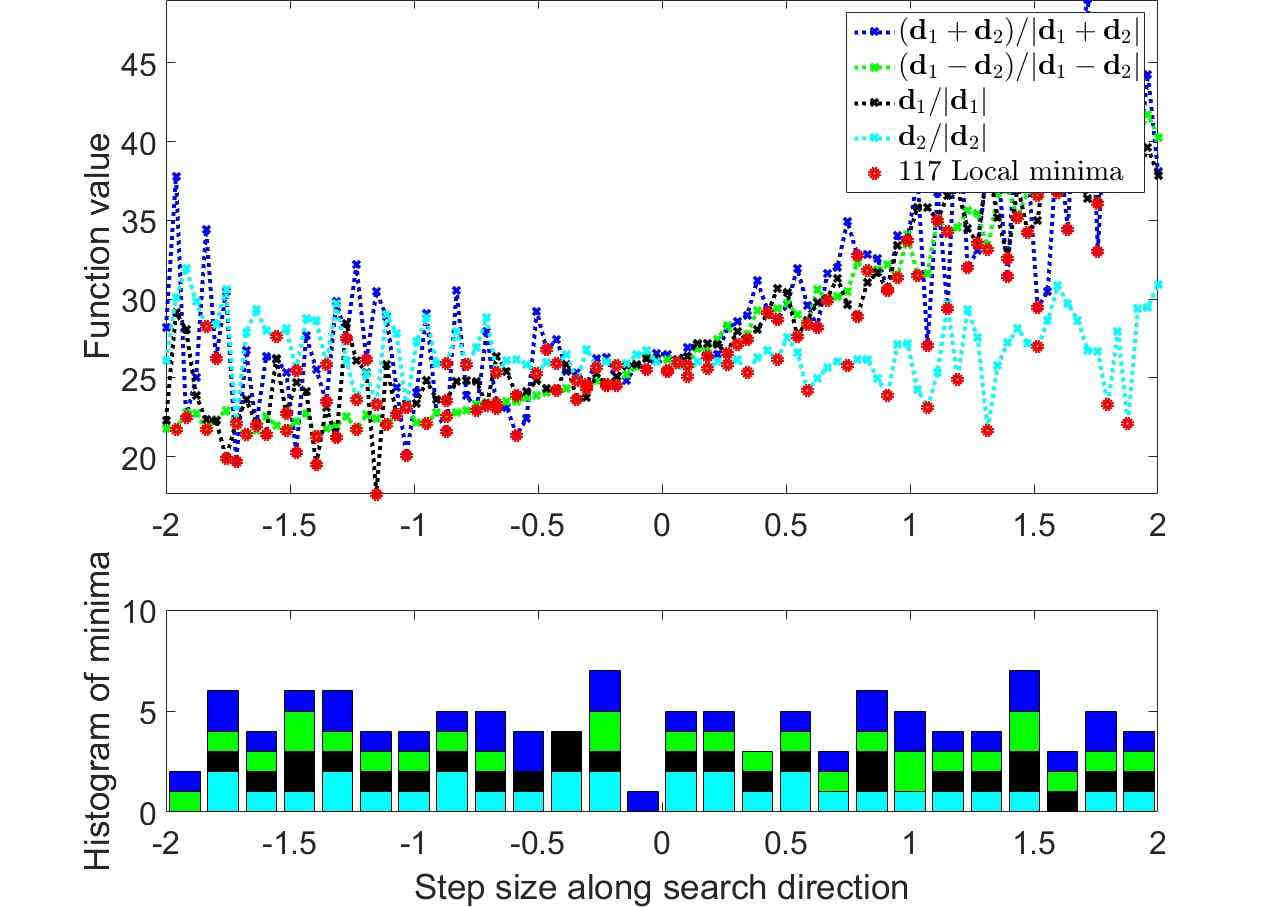}
		\caption{Local minima along search directions}
	\end{subfigure}%
	\begin{subfigure}{.45\textwidth}
		\centering
		\includegraphics[width=0.9\linewidth]{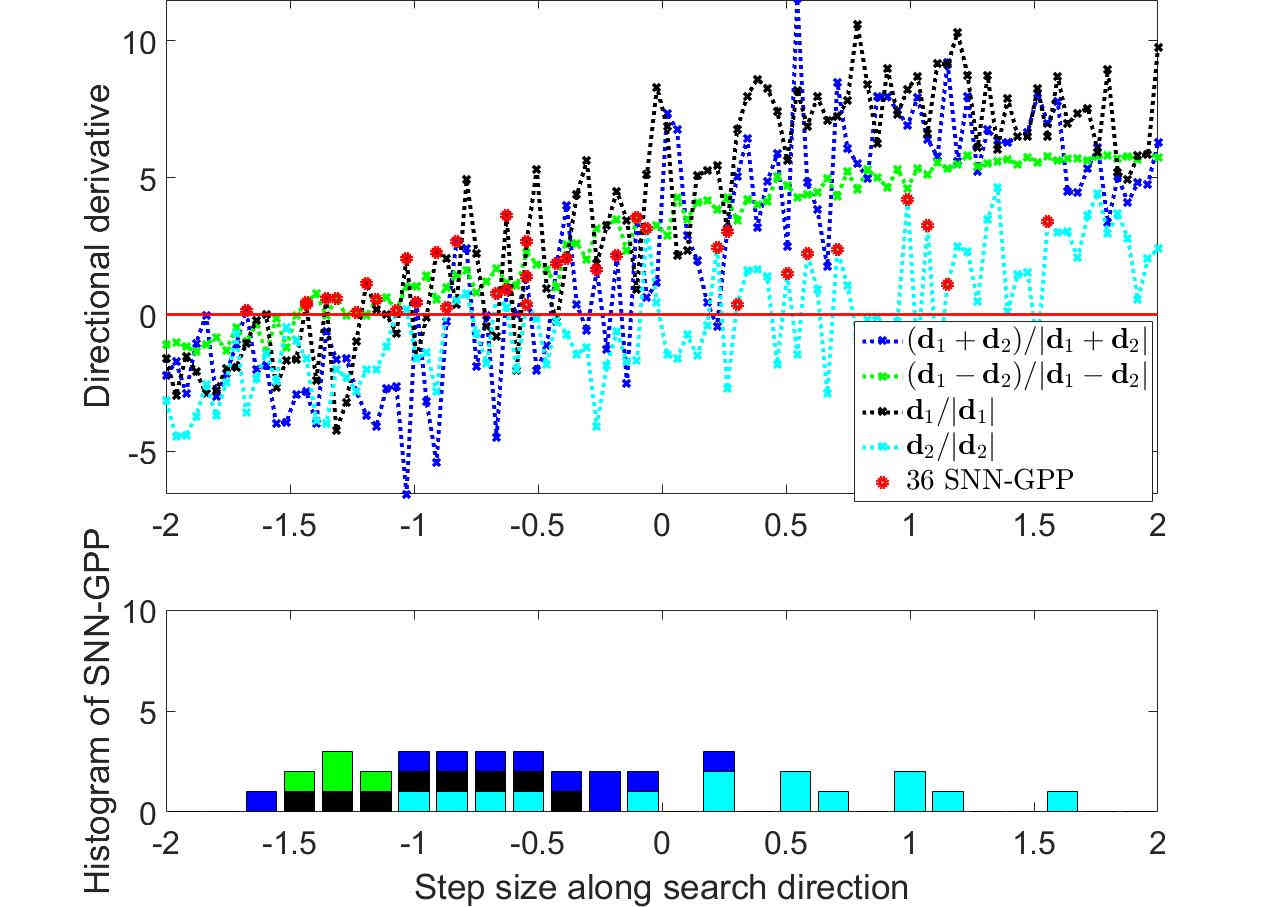}
		\caption{SNN-GPPs along search directions}
	\end{subfigure}%
	
	\caption{The Sigmoid AF close-up: Shapes are smooth and have little curvature. SNN-GPPs have a smaller spatial range in directions where the curvature is larger, while being more spread out in low curvature directions.}
	\label{fig_z_sig}
\end{figure}

\begin{figure}[h!]
	\centering
	\begin{subfigure}{.45\textwidth}
		\centering 
		\includegraphics[width=0.9\linewidth]{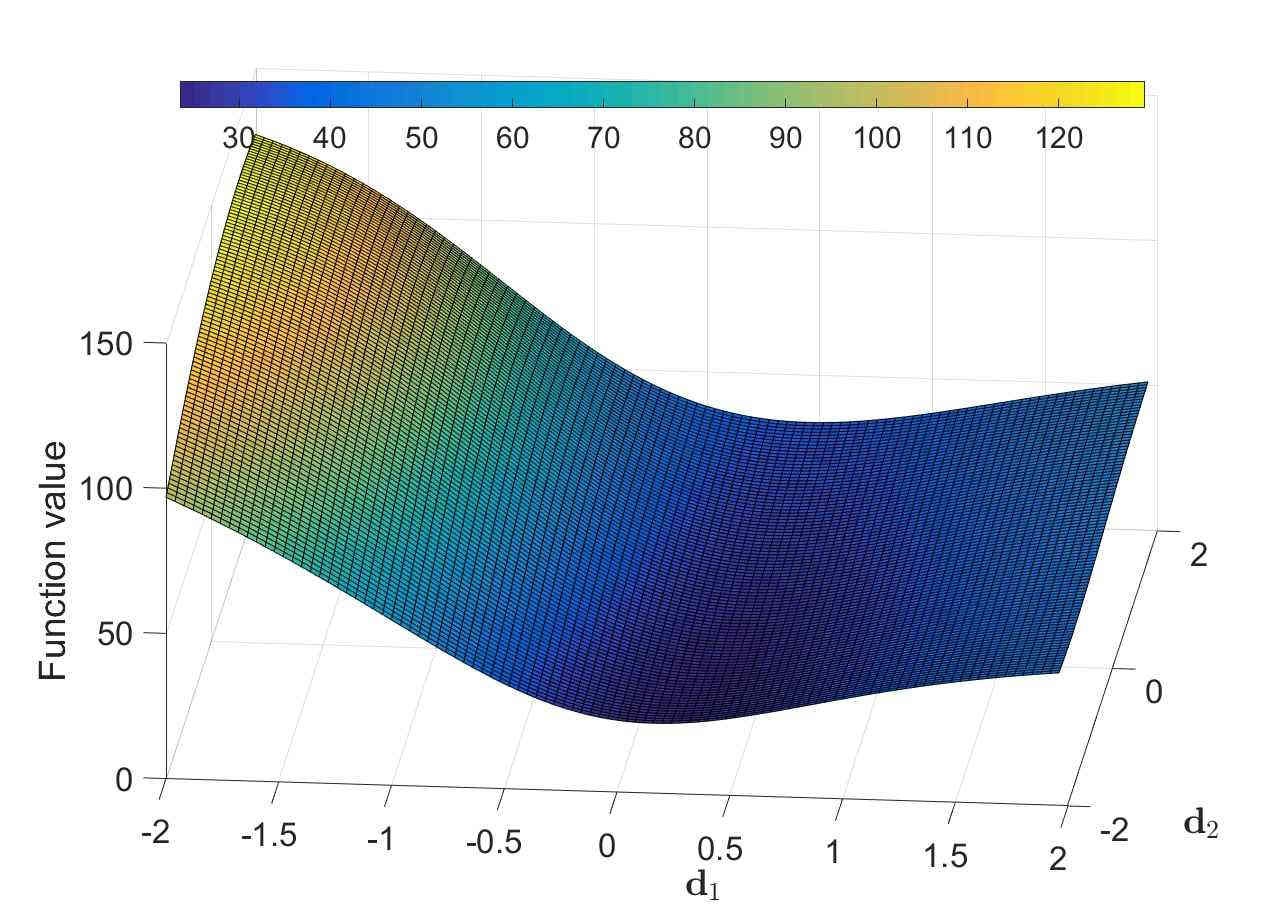}
		\caption{Function value, $M = 150$}
		\label{fig_tan_z_func_B}
	\end{subfigure}%
	\begin{subfigure}{.45\textwidth}
		\centering
		\includegraphics[width=0.9\linewidth]{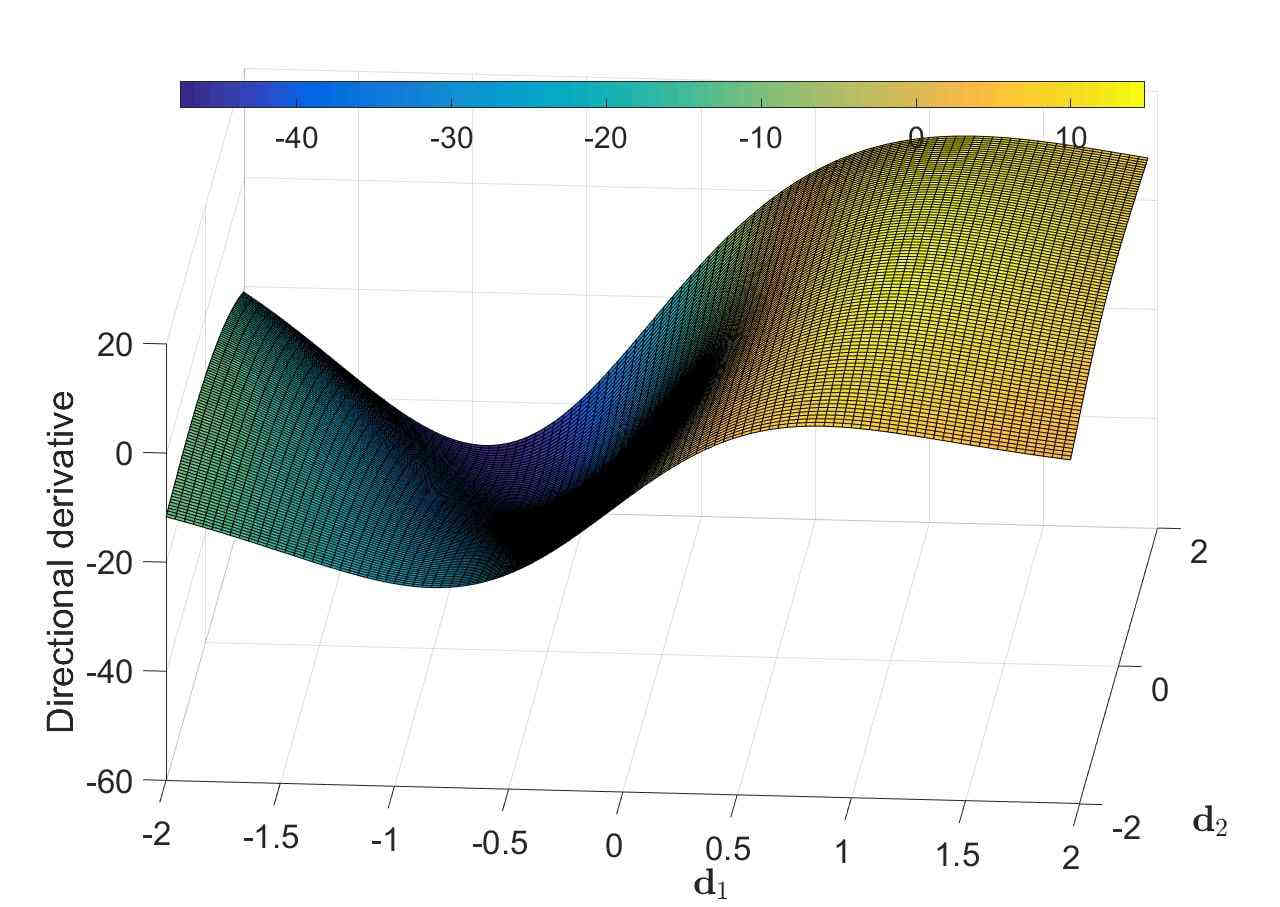}
		\caption{Directional derivative, $M = 150$}
		\label{fig_tan_z_dd_B}
	\end{subfigure}%
	
	\begin{subfigure}{.45\textwidth}
		\centering 
		\includegraphics[width=0.9\linewidth]{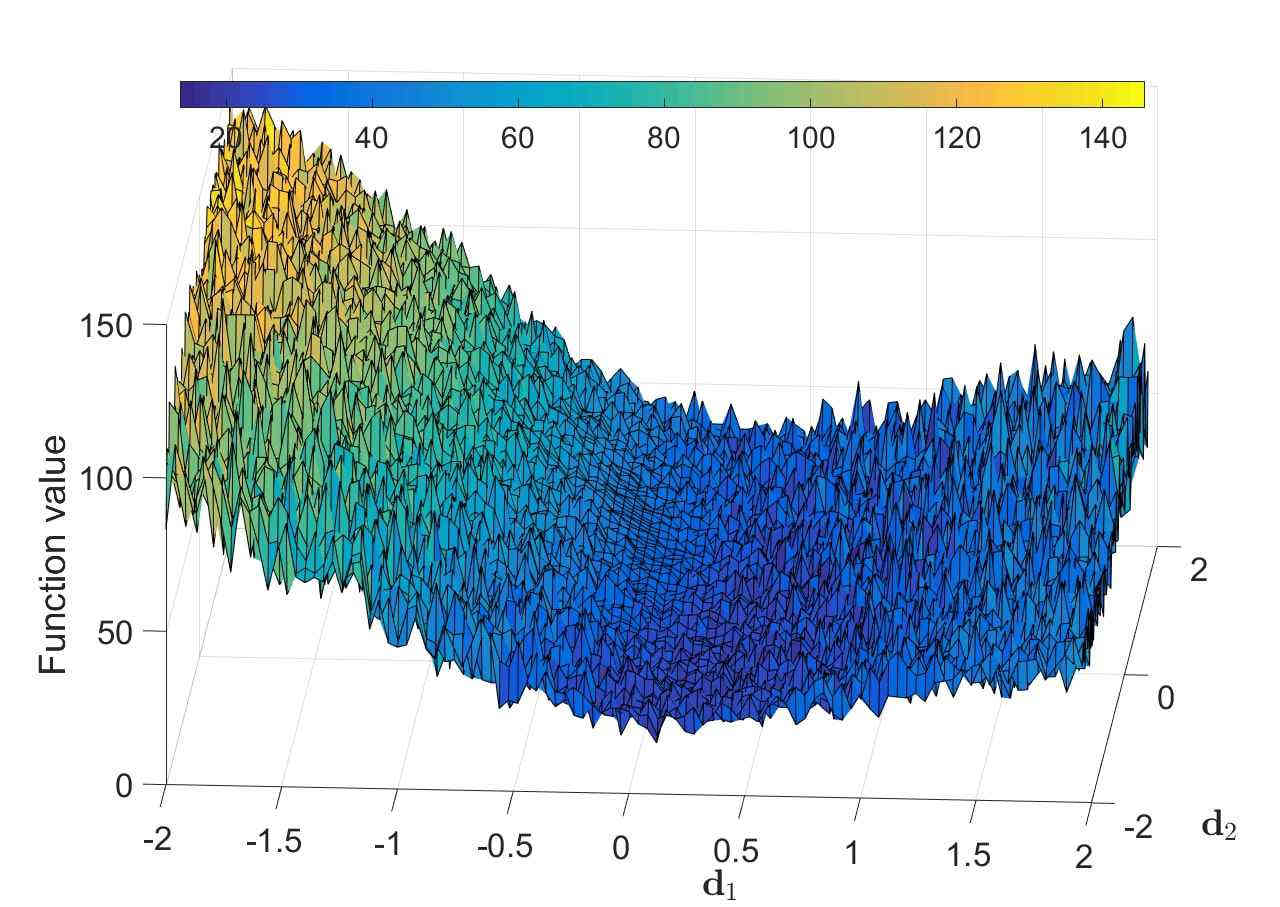}
		\caption{Function value, $|\mathcal{B}_{n,i}| = 10$}
		\label{fig_tan_z_func_M}
	\end{subfigure}%
	\begin{subfigure}{.45\textwidth}
		\centering
		\includegraphics[width=0.9\linewidth]{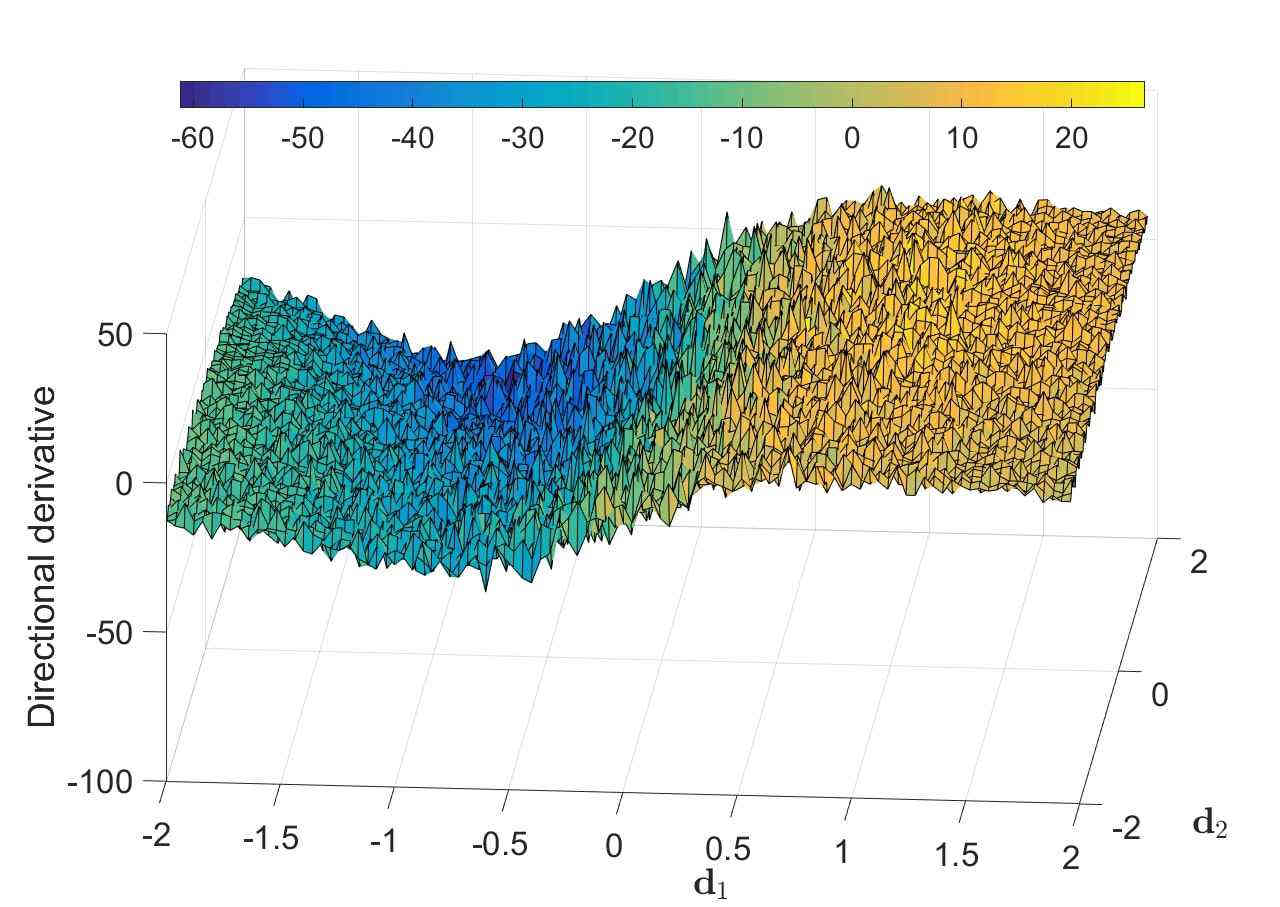}
		\caption{Directional derivative, $|\mathcal{B}_{n,i}| = 10$}
		\label{fig_tan_z_dd_M}
	\end{subfigure}%
	
	\begin{subfigure}{.45\textwidth}
		\centering 
		\includegraphics[width=0.9\linewidth]{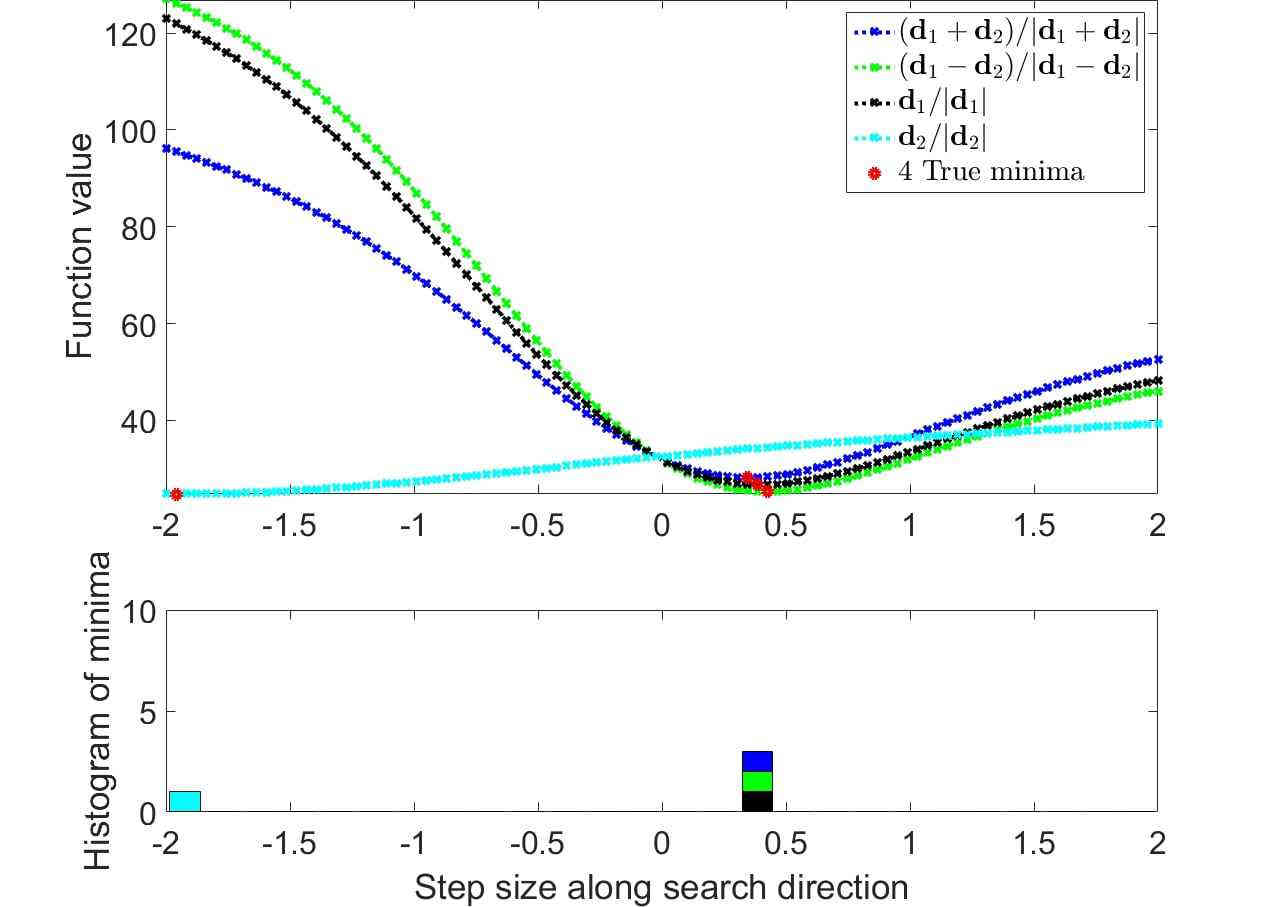}
		\caption{True minima along search directions}
		\label{fig_tan_z_fline_B}
	\end{subfigure}%
	\begin{subfigure}{.45\textwidth}
		\centering
		\includegraphics[width=0.9\linewidth]{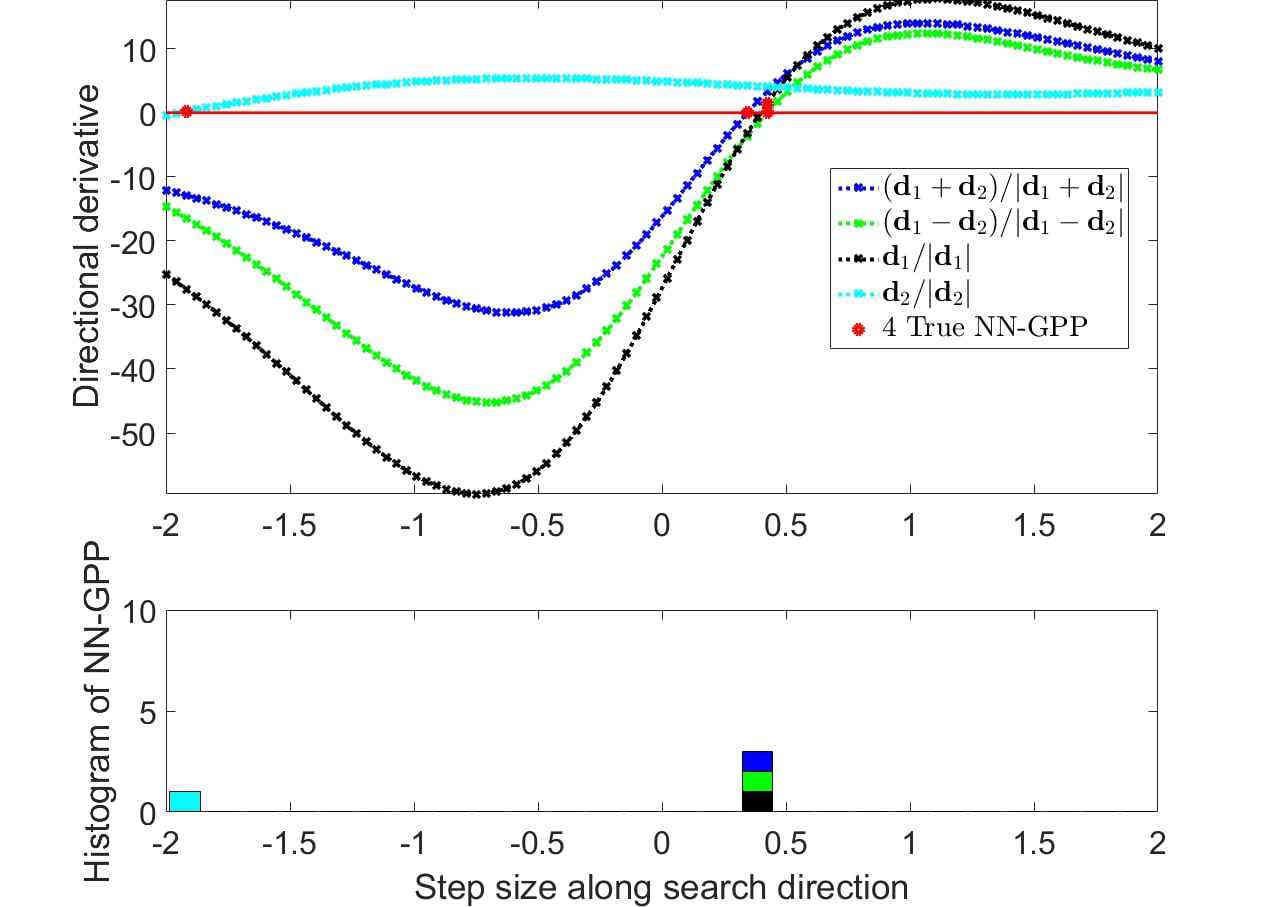}
		\caption{True NN-GPPs along search directions}
		\label{fig_tan_z_dline_B}
	\end{subfigure}%
	
	\begin{subfigure}{.45\textwidth}
		\centering 
		\includegraphics[width=0.9\linewidth]{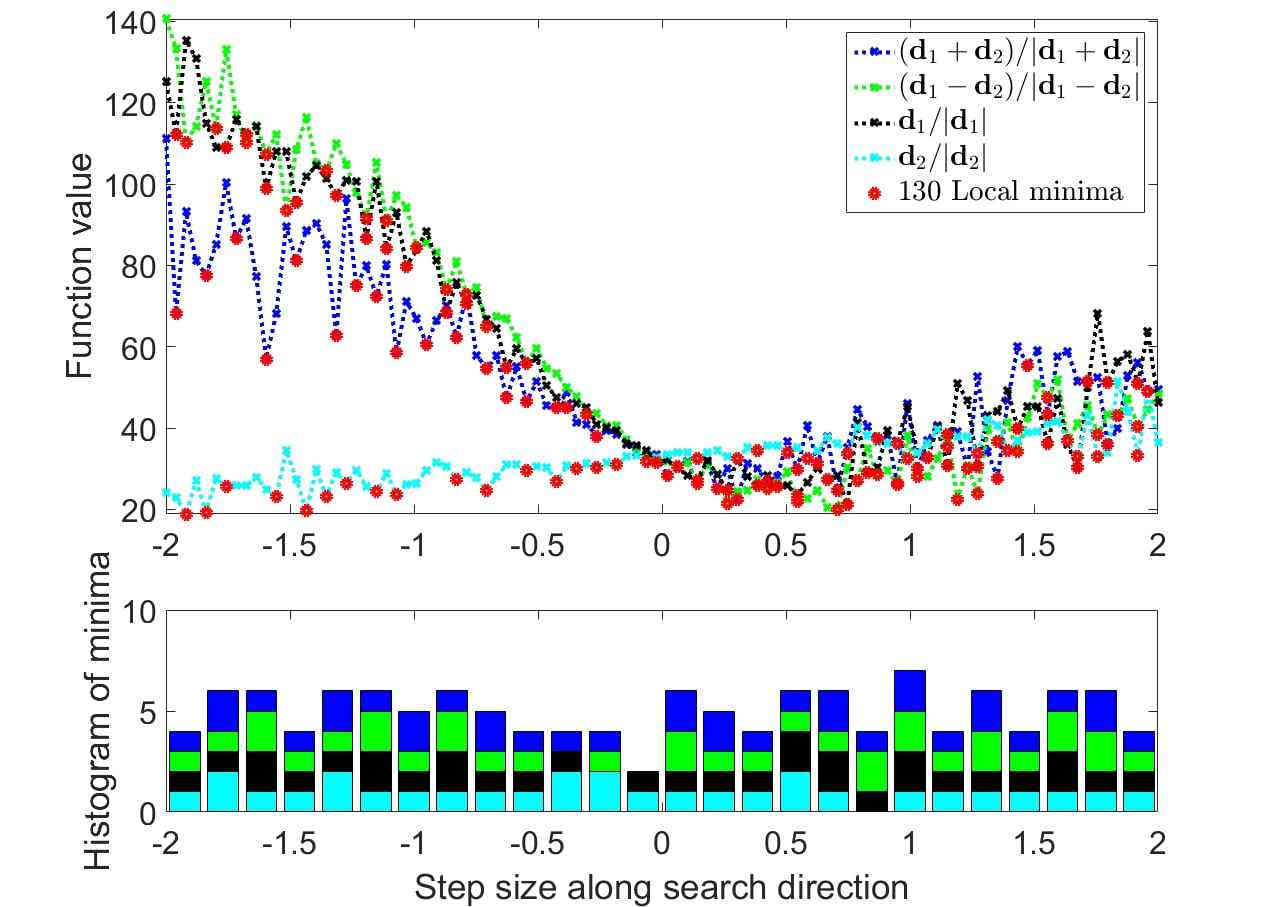}
		\caption{Local minima along search directions}
		\label{fig_tan_z_fline_M}
	\end{subfigure}%
	\begin{subfigure}{.45\textwidth}
		\centering
		\includegraphics[width=0.9\linewidth]{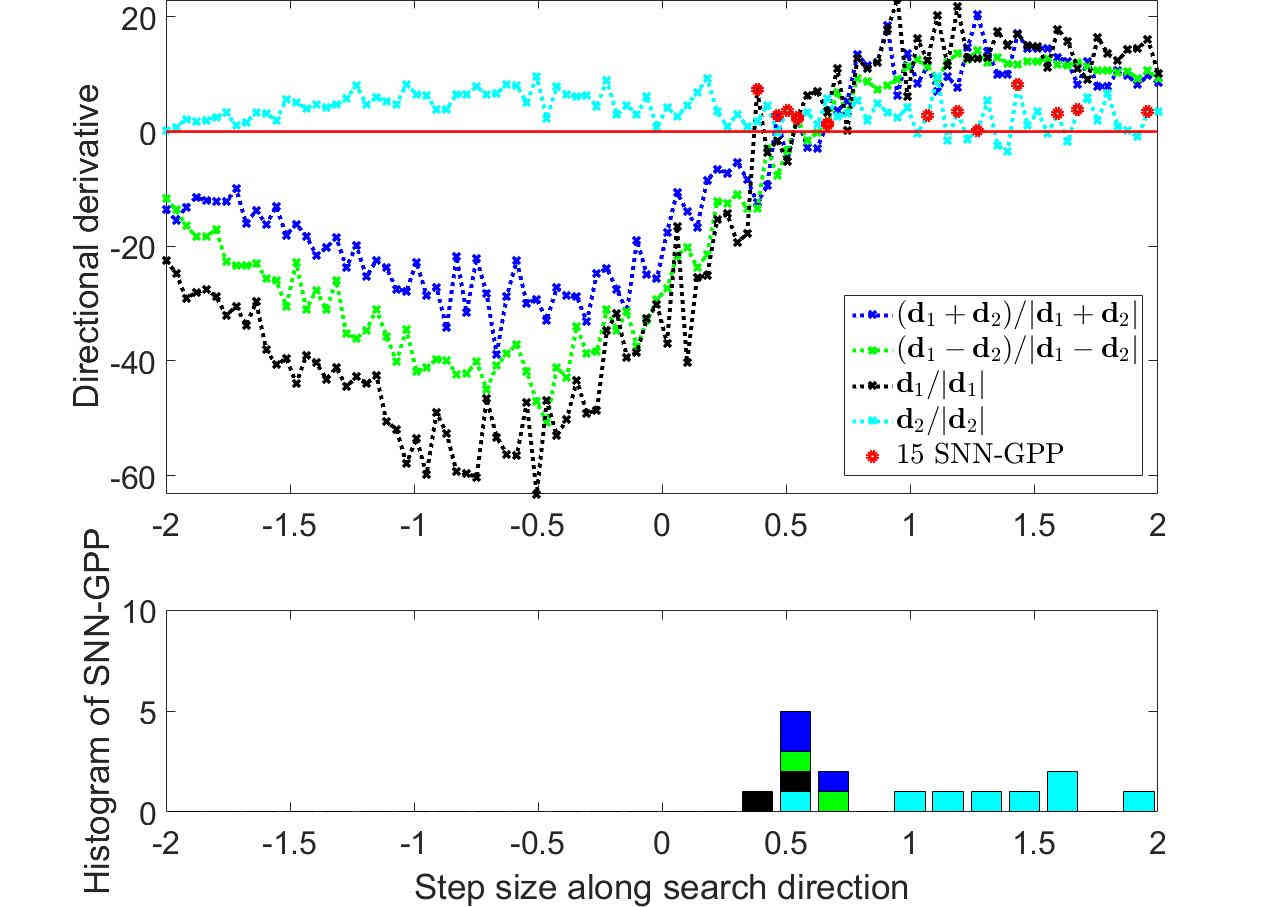}
		\caption{SNN-GPPs along search directions}
		\label{fig_tan_z_dline_M}
	\end{subfigure}%
	
	\caption{The Tanh AF close-up: The higher curvature of Tanh helps localize SNN-GPPs. Though this example captures lower variance directions in the function value (see (g), $\boldsymbol{d}_2$), this does not alleviate the problem of uniformly spread spurious local minima.}
	\label{fig_z_tan}
\end{figure}

\begin{figure}[h!]
	\centering
	\begin{subfigure}{.45\textwidth}
		\centering 
		\includegraphics[width=0.9\linewidth]{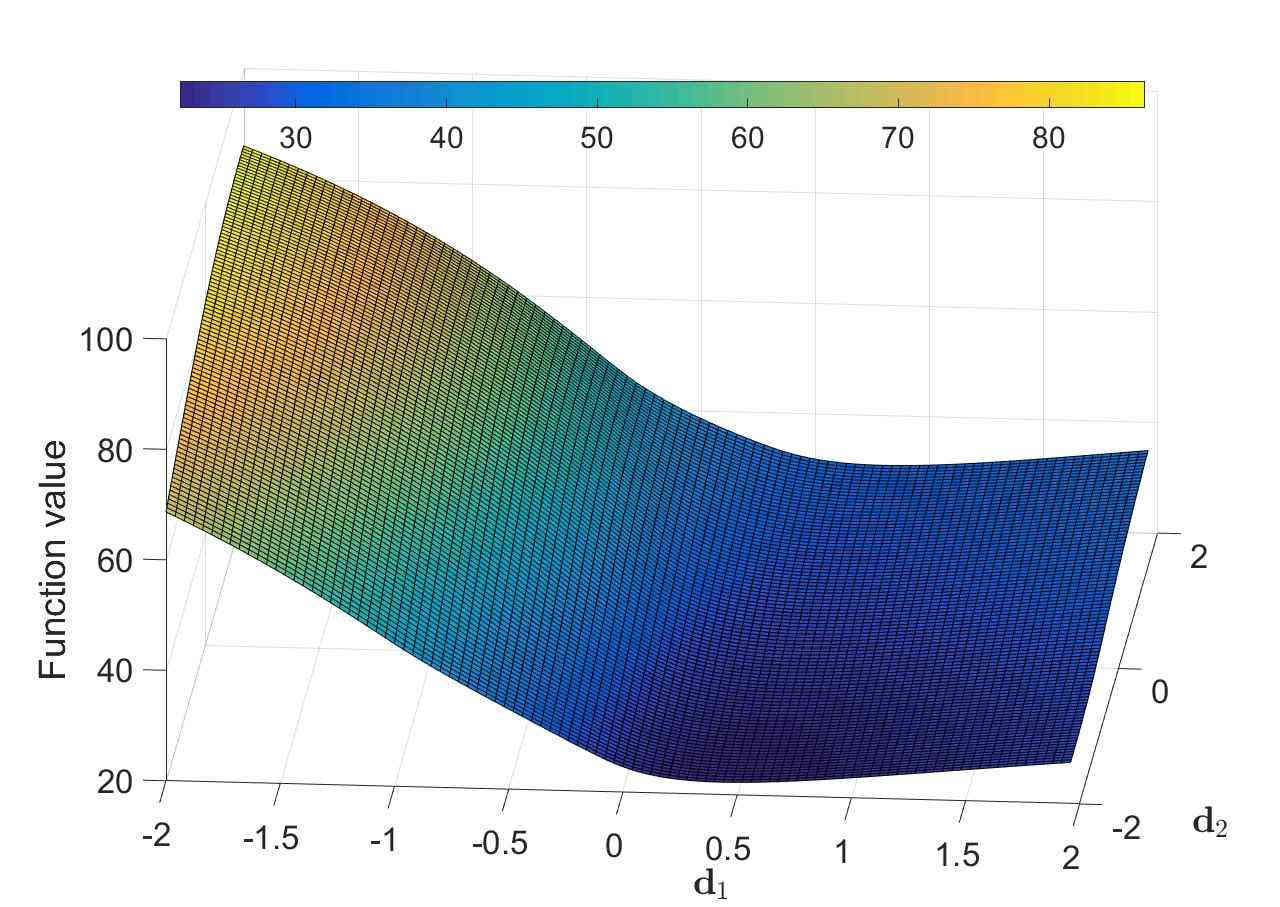}
		\caption{Function value, $M = 150$}
		\label{fig_soft_z_func_B}
	\end{subfigure}%
	\begin{subfigure}{.45\textwidth}
		\centering
		\includegraphics[width=0.9\linewidth]{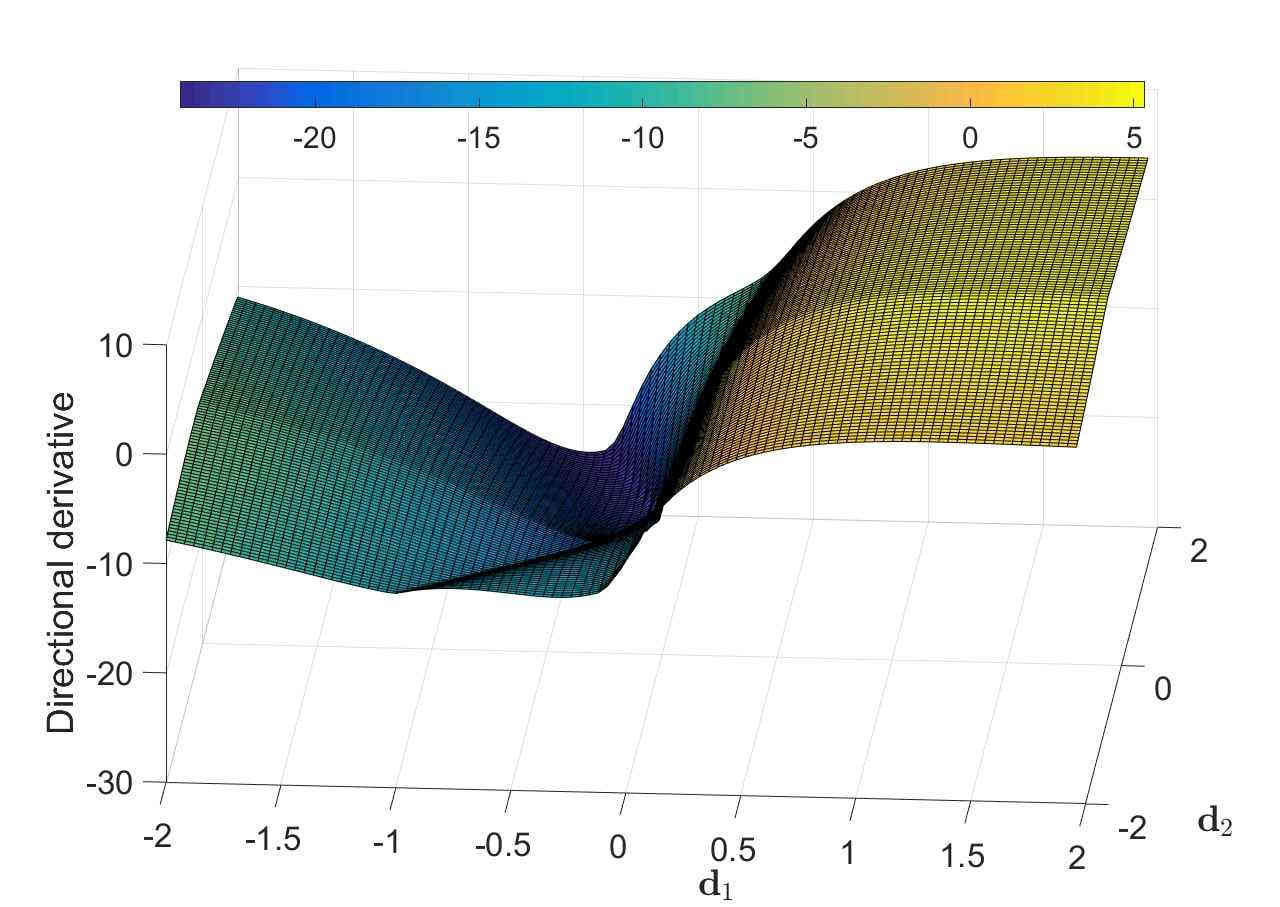}
		\caption{Directional derivative, $M = 150$}
		\label{fig_soft_z_dd_B}
	\end{subfigure}%
	
	\begin{subfigure}{.45\textwidth}
		\centering 
		\includegraphics[width=0.9\linewidth]{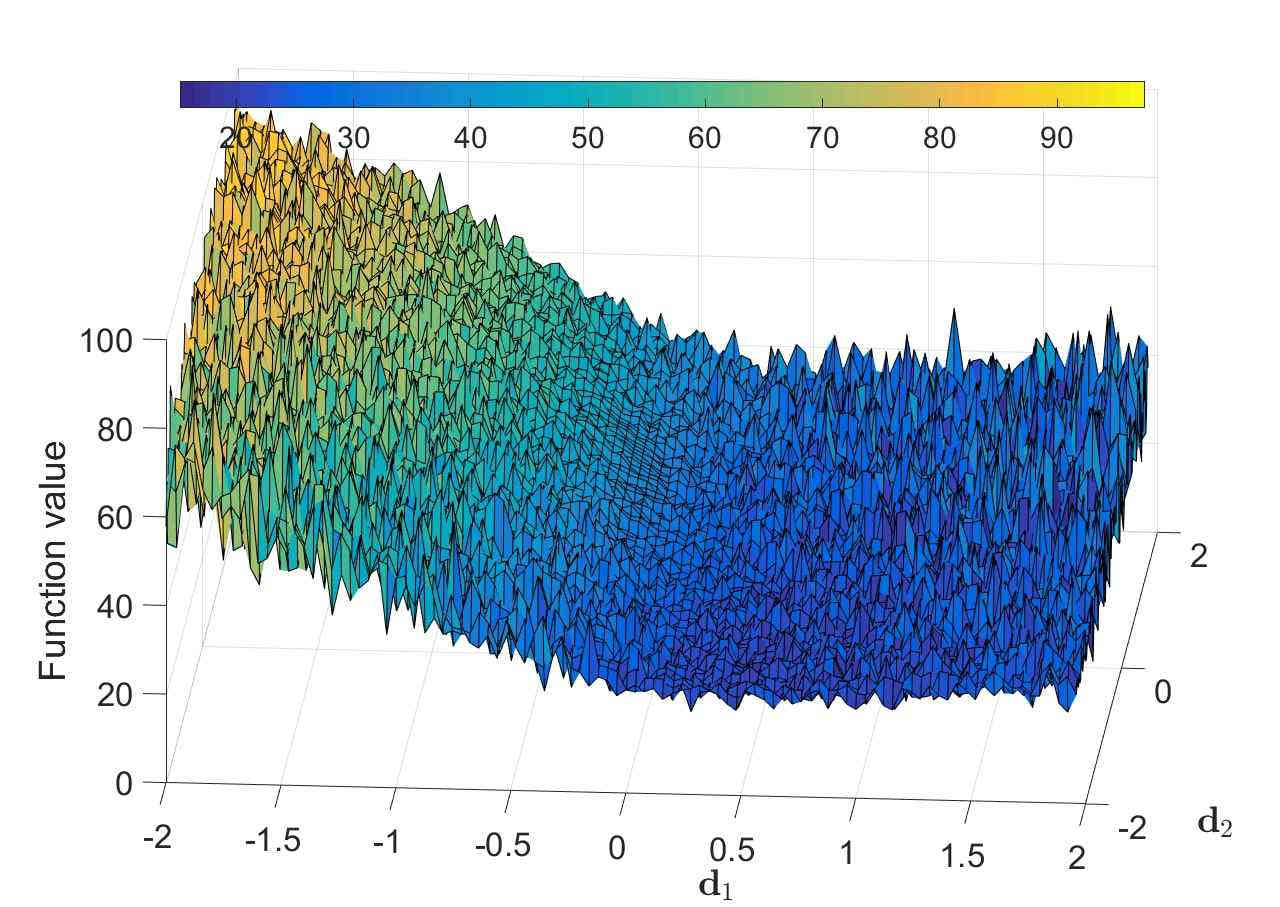}
		\caption{Function value, $|\mathcal{B}_{n,i}| = 10$}
		\label{fig_soft_z_func_M}
	\end{subfigure}%
	\begin{subfigure}{.45\textwidth}
		\centering
		\includegraphics[width=0.9\linewidth]{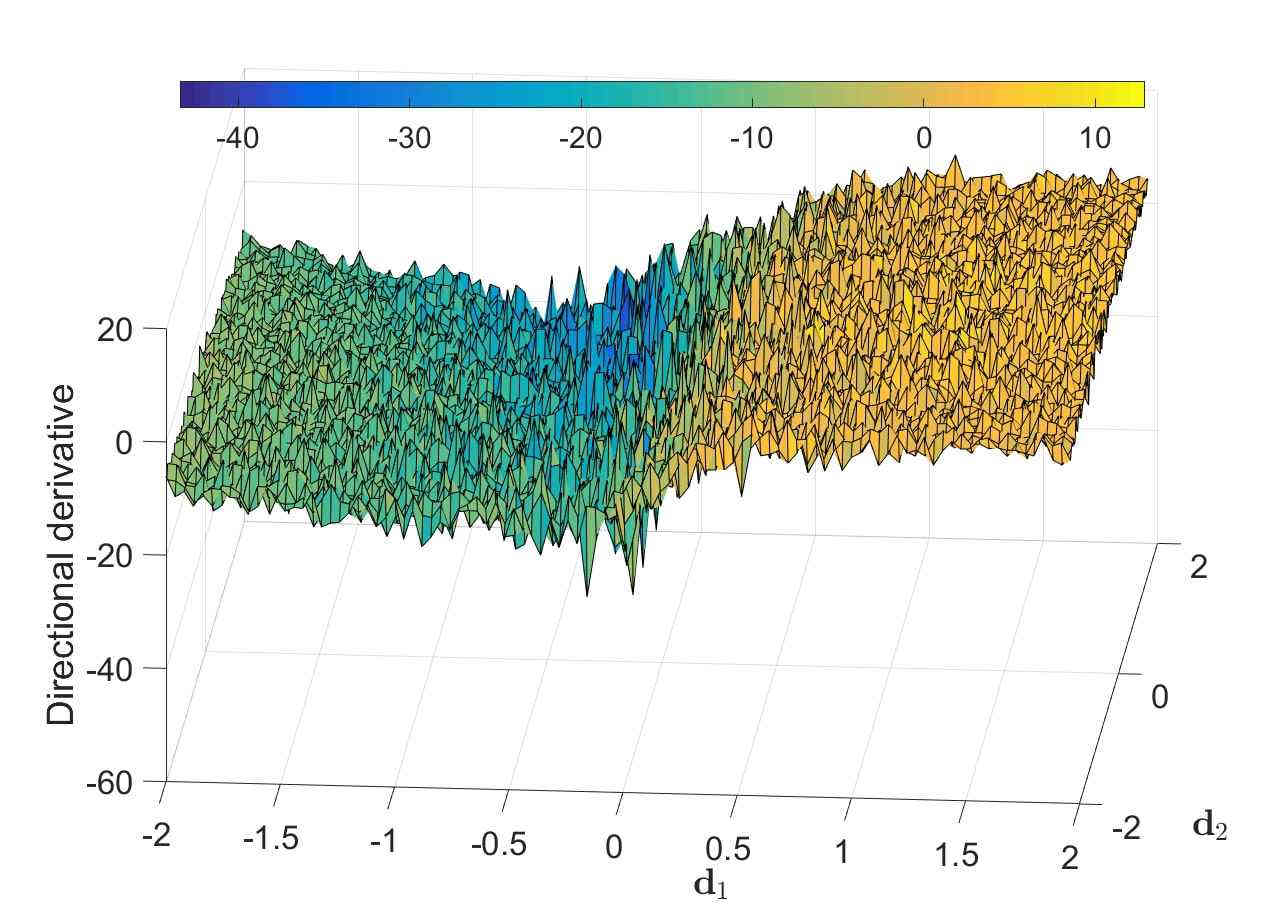}
		\caption{Directional derivative, $|\mathcal{B}_{n,i}| = 10$}
		\label{fig_soft_z_dd_M}
	\end{subfigure}%
	
	\begin{subfigure}{.45\textwidth}
		\centering 
		\includegraphics[width=0.9\linewidth]{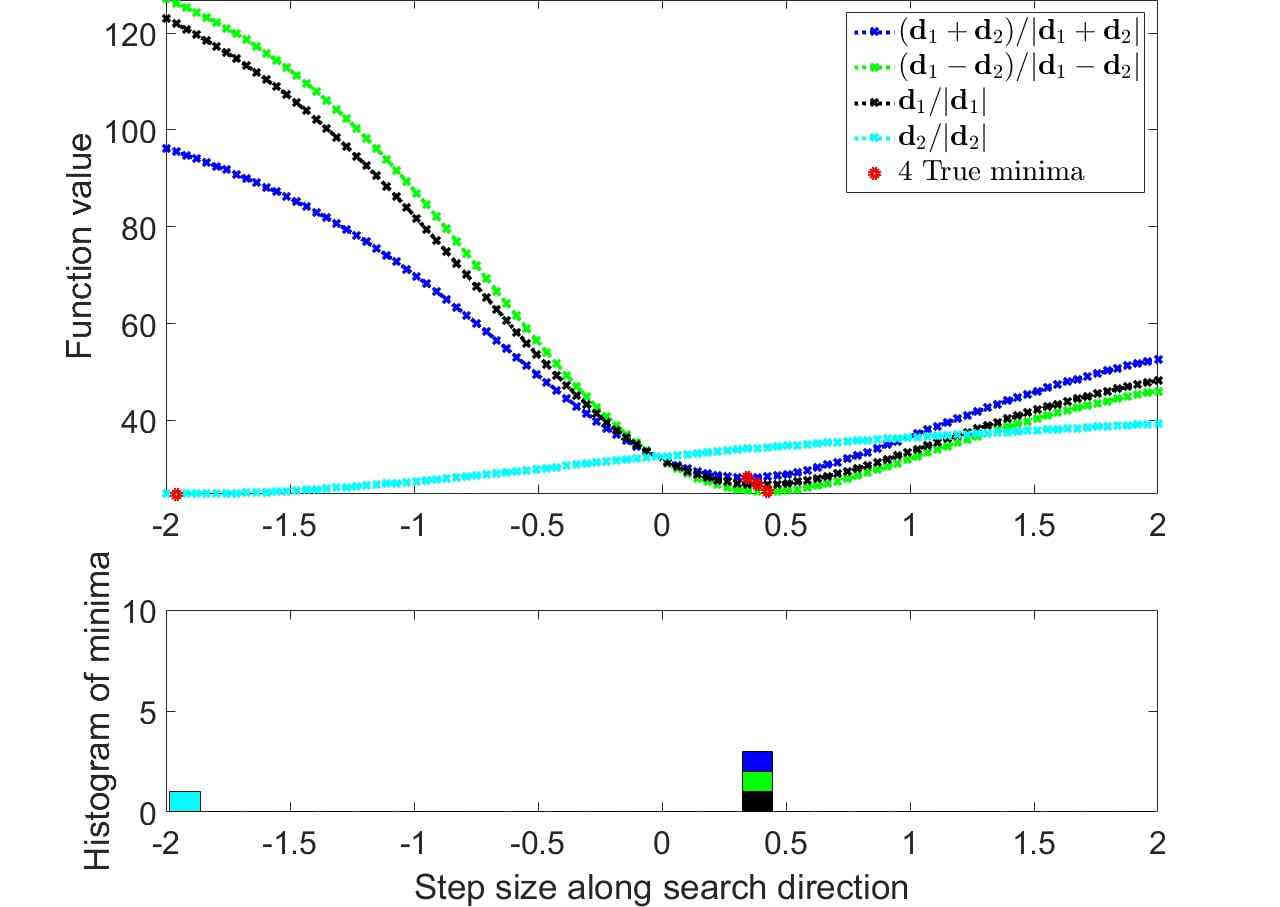}
		\caption{True minima along search directions}
		\label{fig_soft_z_fline_B}
	\end{subfigure}%
	\begin{subfigure}{.45\textwidth}
		\centering
		\includegraphics[width=0.9\linewidth]{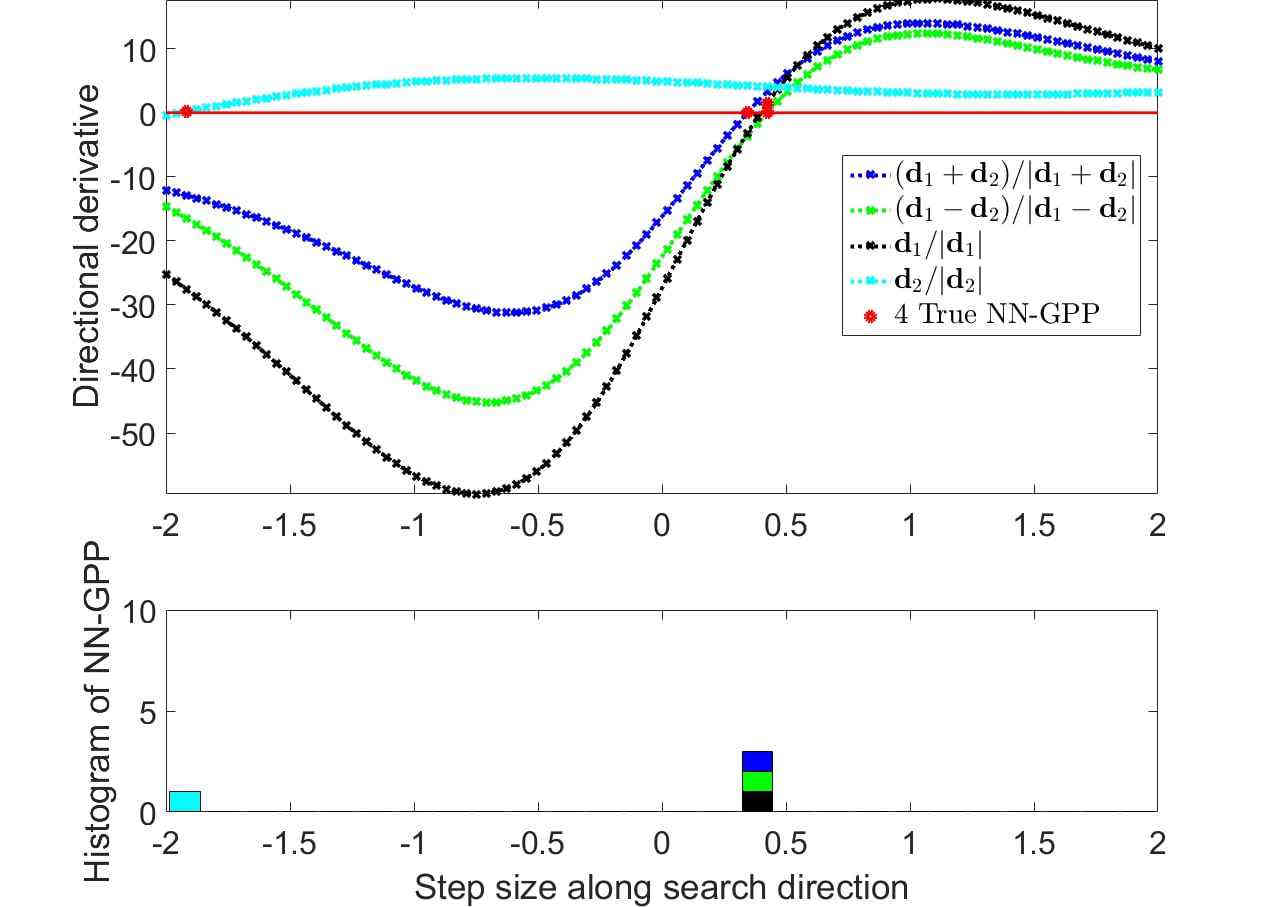}
		\caption{True NN-GPPs along search directions}
		\label{fig_soft_z_dline_B}
	\end{subfigure}%
	
	\begin{subfigure}{.45\textwidth}
		\centering 
		\includegraphics[width=0.9\linewidth]{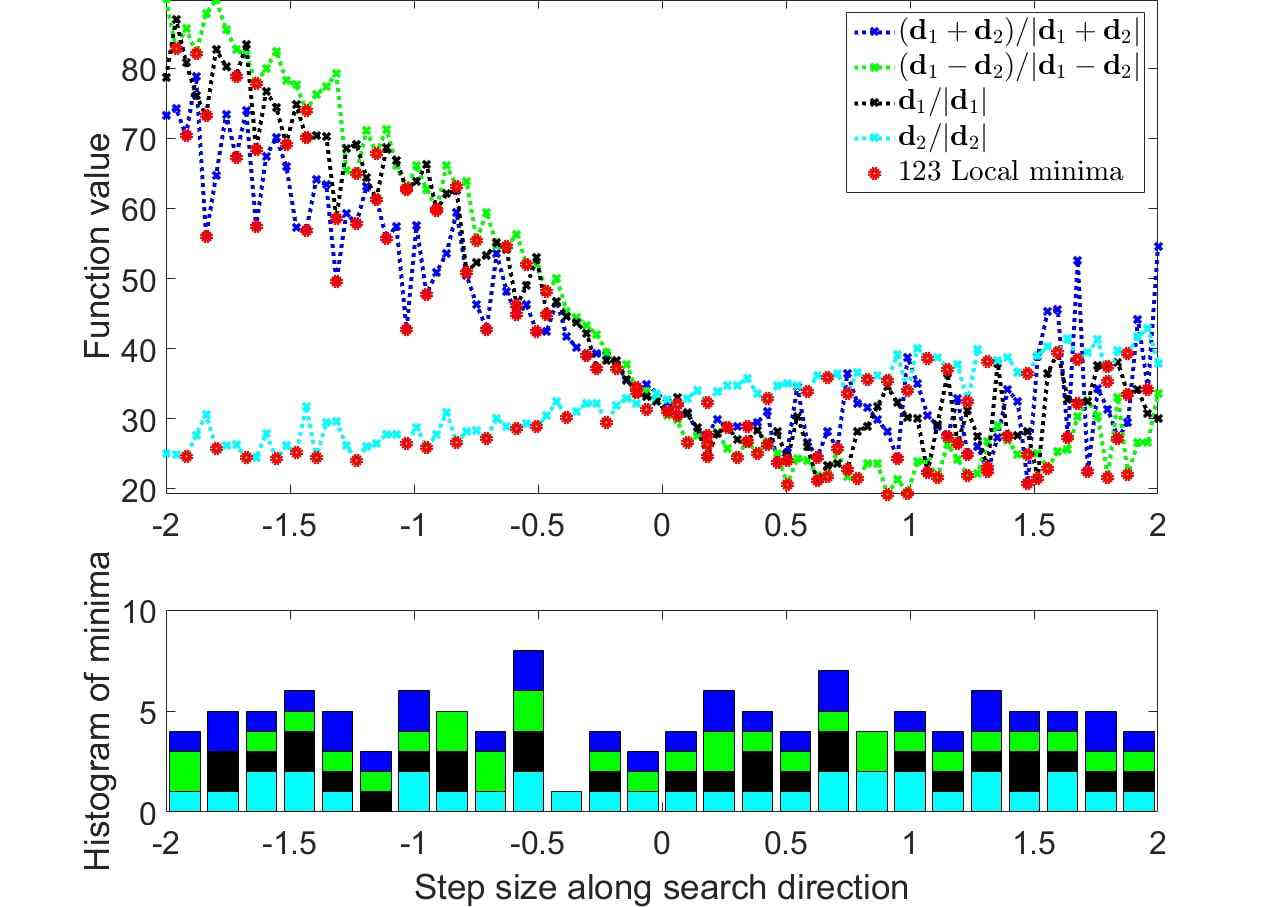}
		\caption{Local minima along search directions}
		\label{fig_soft_z_fline_M}
	\end{subfigure}%
	\begin{subfigure}{.45\textwidth}
		\centering
		\includegraphics[width=0.9\linewidth]{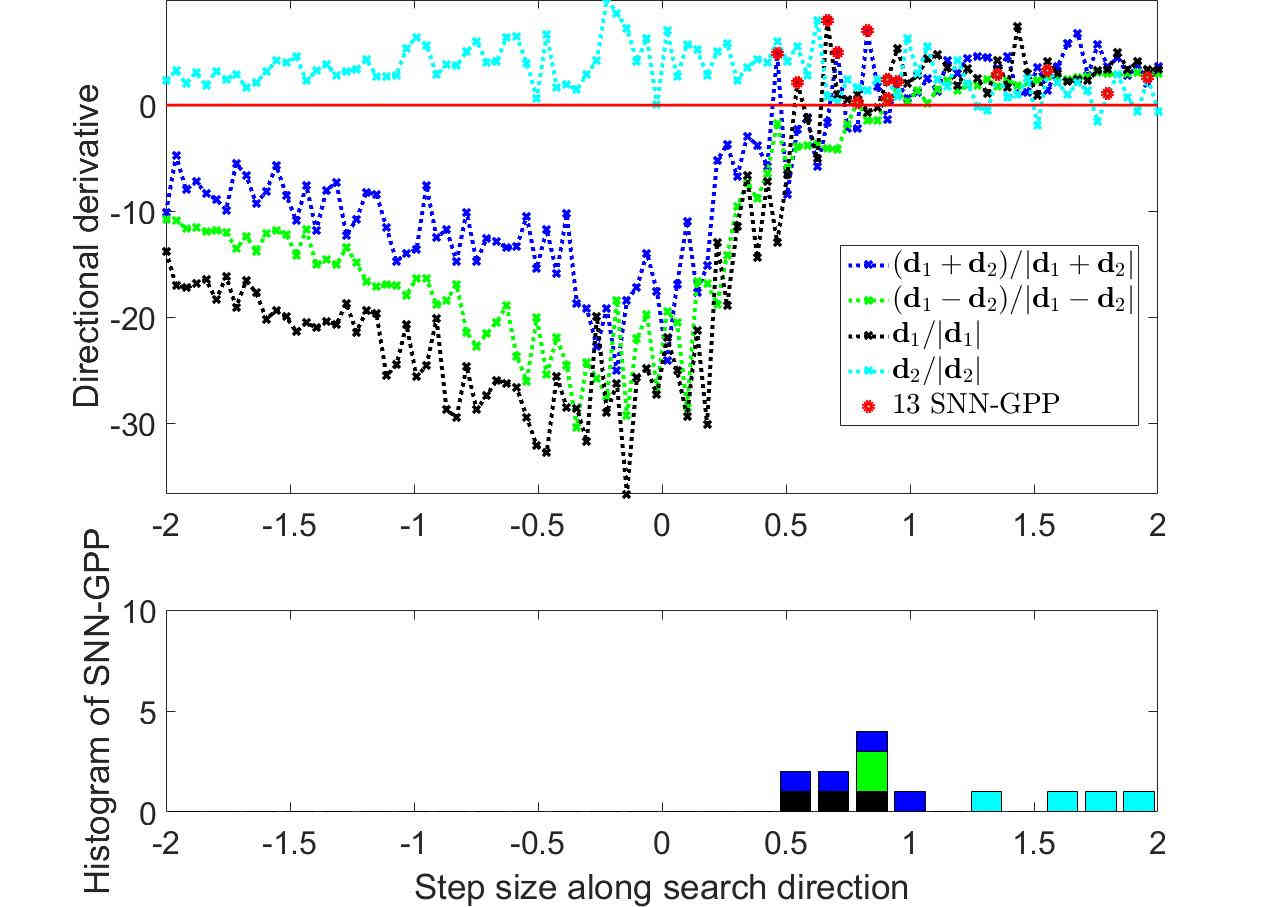}
		\caption{SNN-GPPs along search directions}
		\label{fig_soft_z_dline_M}
	\end{subfigure}%
	
	\caption{The Softsign AF close-up: The small domain investigation confirms that the less aggressive taper-off of the AF derivative contributes to localizing SNN-GPPs around the true optimum, while avoiding spurious instances at saturation. Local minima remain uniformly distributed.}
	\label{fig_z_soft}
\end{figure}

\begin{figure}[h!]
	\centering
	\begin{subfigure}{.45\textwidth}
		\centering 
		\includegraphics[width=0.9\linewidth]{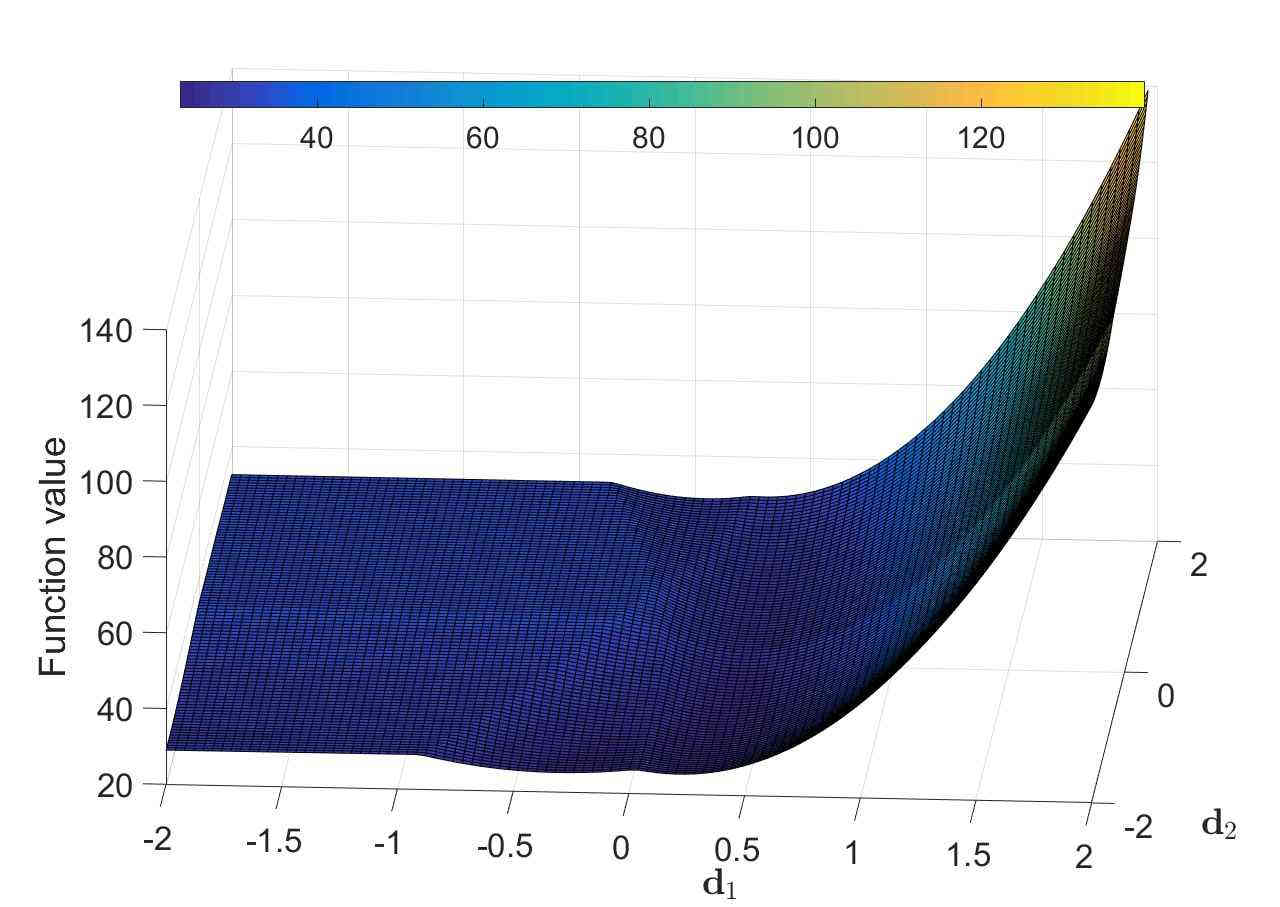}
		\caption{Function value, $M = 150$}
		\label{fig_relu_z_func_B}
	\end{subfigure}%
	\begin{subfigure}{.45\textwidth}
		\centering
		\includegraphics[width=0.9\linewidth]{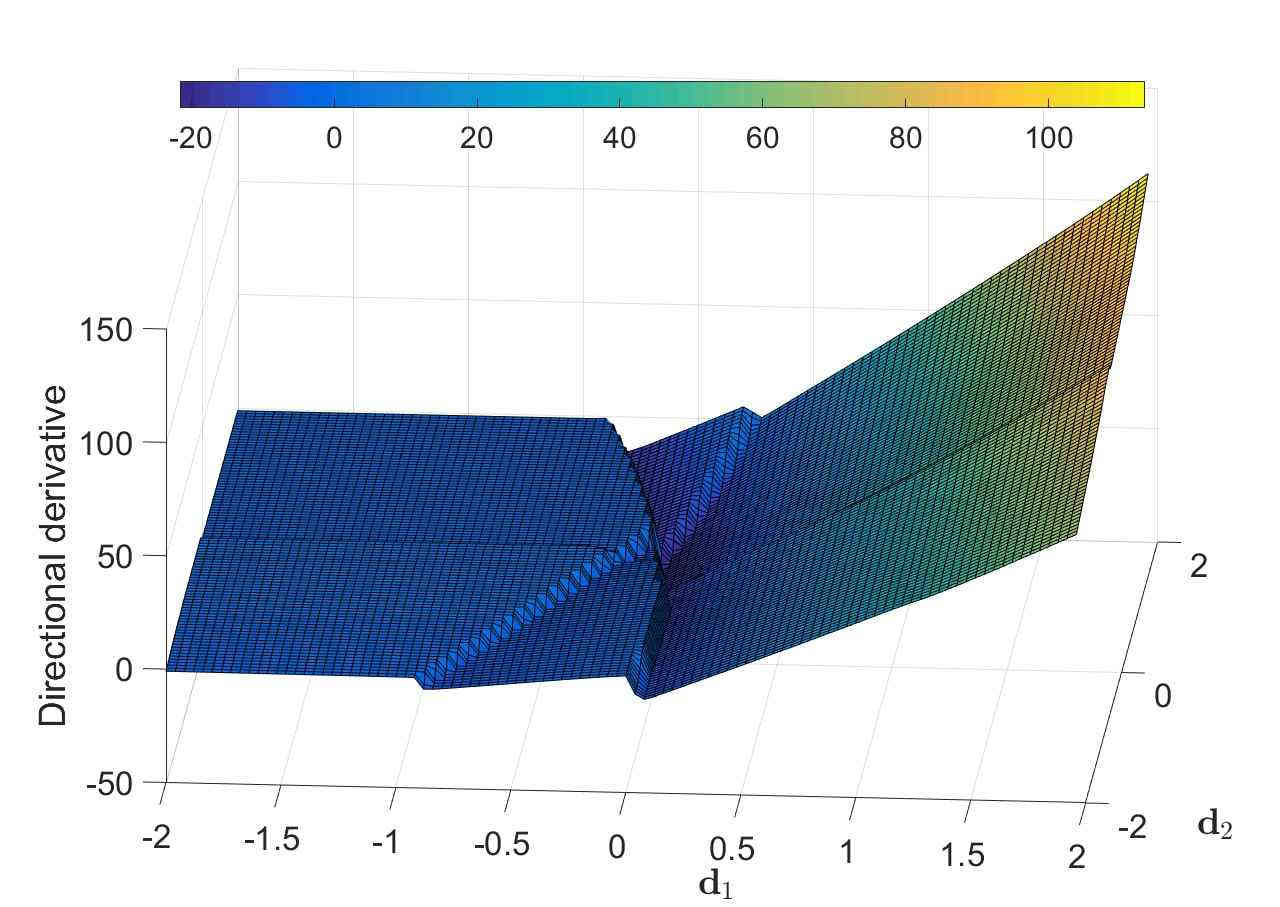}
		\caption{Directional derivative, $M = 150$}
		\label{fig_relu_z_dd_B}
	\end{subfigure}%
	
	\begin{subfigure}{.45\textwidth}
		\centering 
		\includegraphics[width=0.9\linewidth]{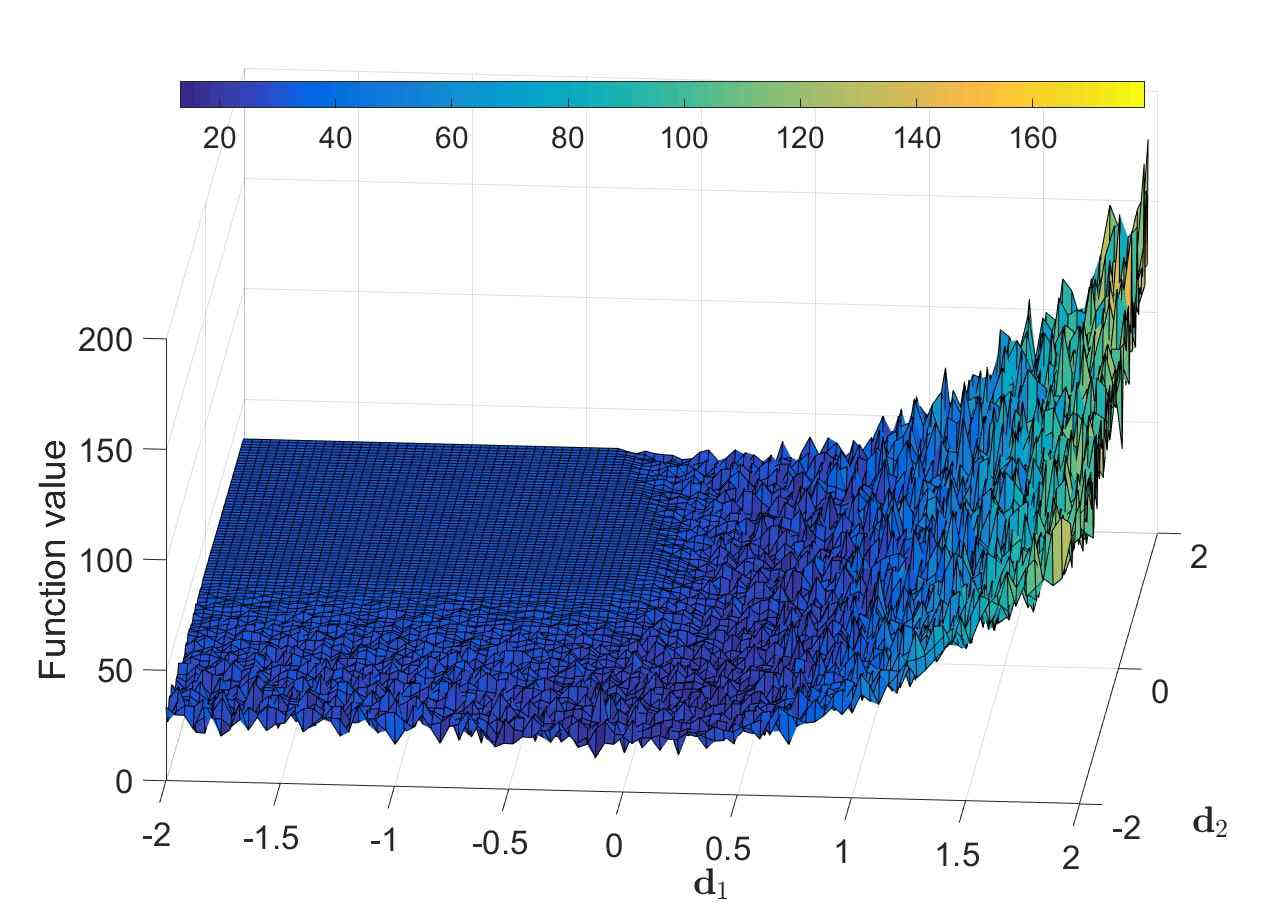}
		\caption{Function value, $|\mathcal{B}_{n,i}| = 10$}
		\label{fig_relu_z_func_M}
	\end{subfigure}%
	\begin{subfigure}{.45\textwidth}
		\centering
		\includegraphics[width=0.9\linewidth]{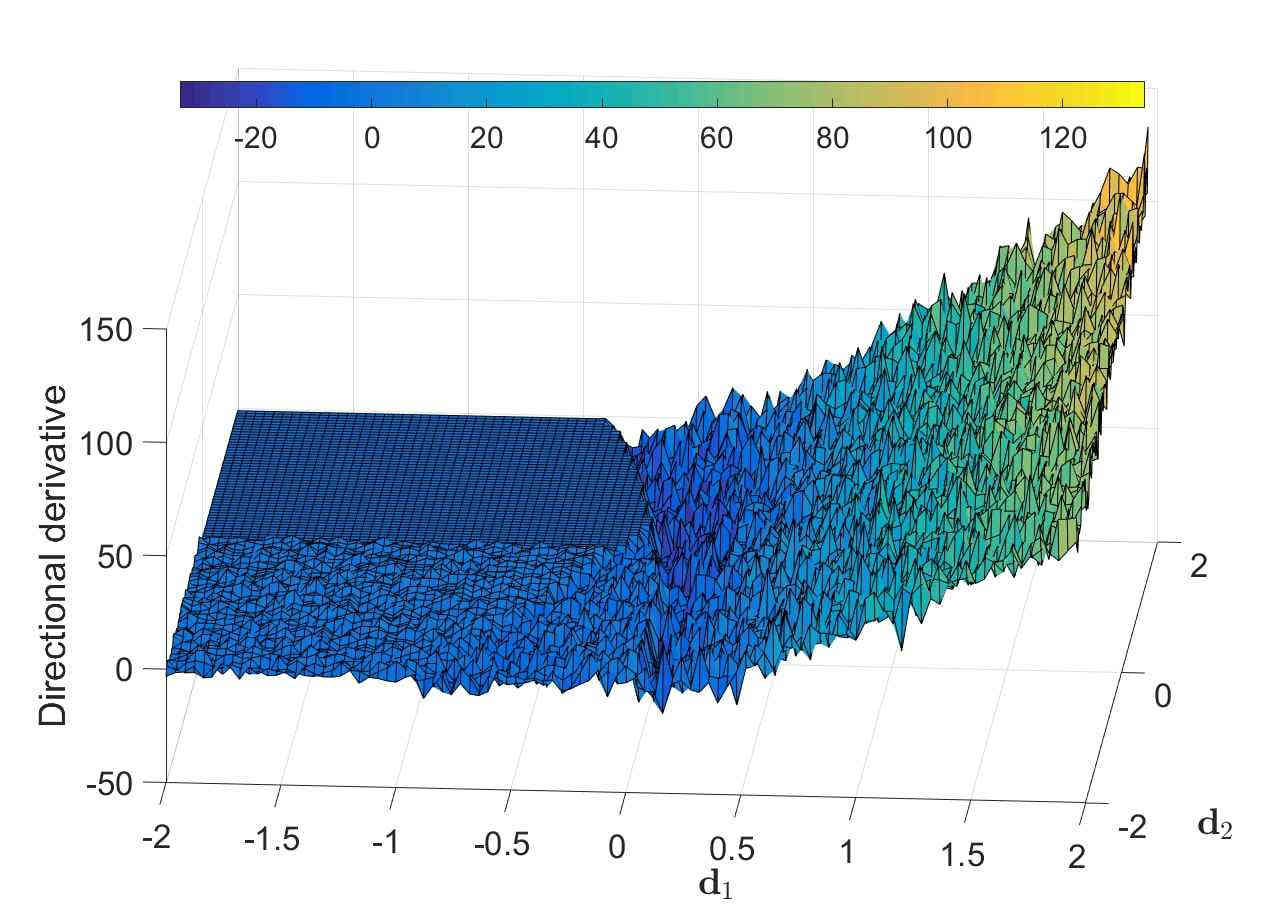}
		\caption{Directional derivative, $|\mathcal{B}_{n,i}| = 10$}
		\label{fig_relu_z_dd_M}
	\end{subfigure}%
	
	\begin{subfigure}{.45\textwidth}
		\centering 
		\includegraphics[width=0.9\linewidth]{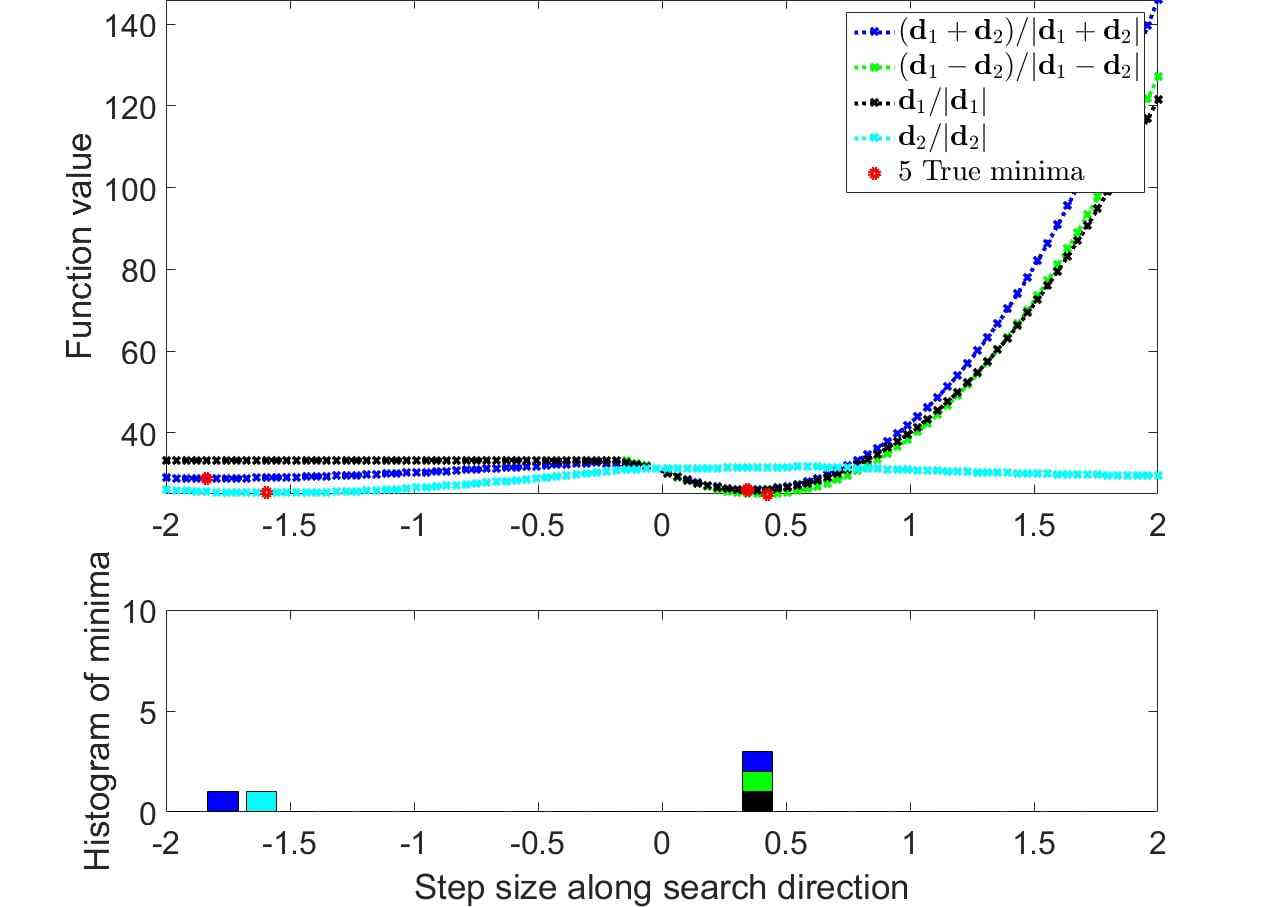}
		\caption{True minima along search directions}
		\label{fig_relu_z_fline_B}
	\end{subfigure}%
	\begin{subfigure}{.45\textwidth}
		\centering
		\includegraphics[width=0.9\linewidth]{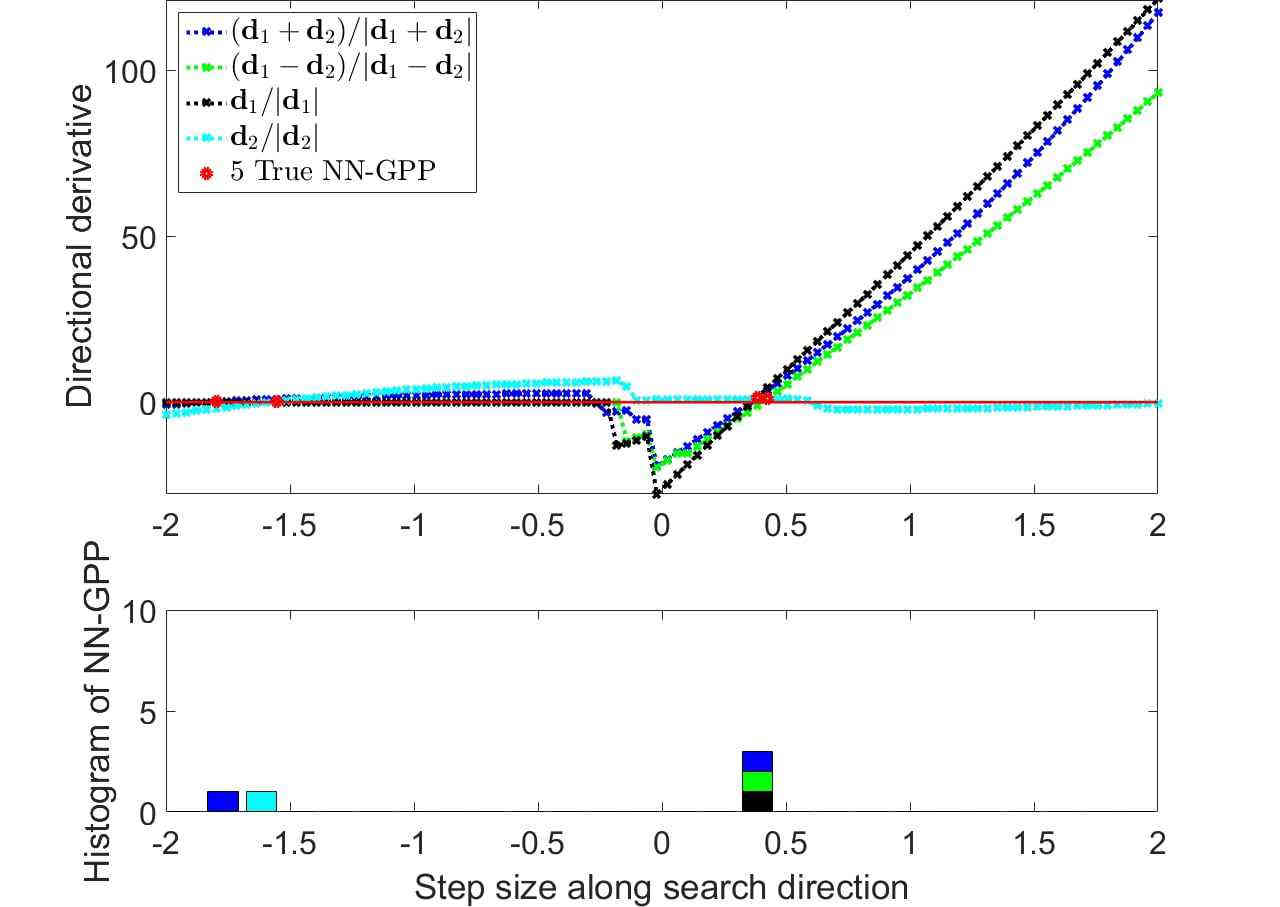}
		\caption{True NN-GPPs along search directions}
		\label{fig_relu_z_dline_B}
	\end{subfigure}%
	
	\begin{subfigure}{.45\textwidth}
		\centering 
		\includegraphics[width=0.9\linewidth]{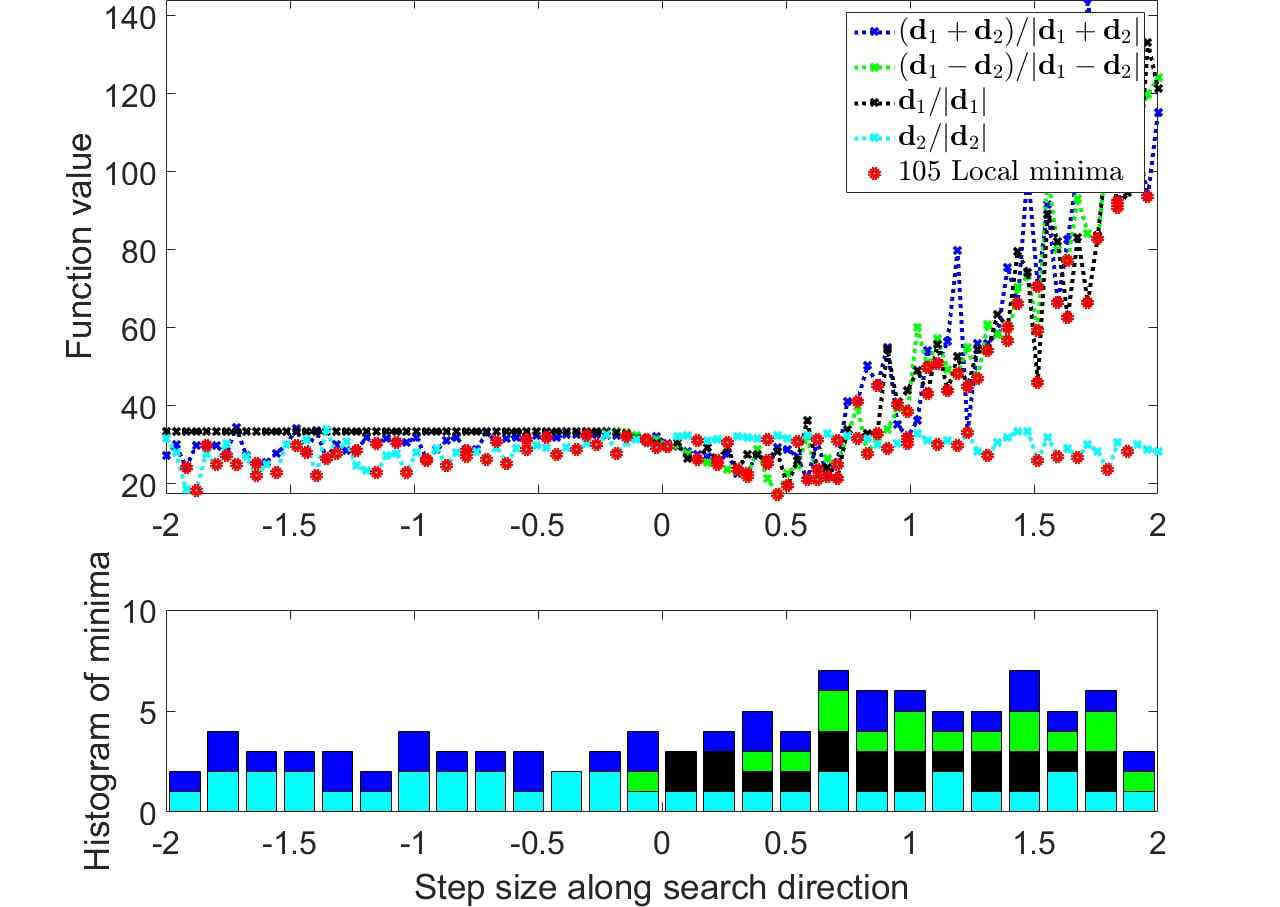}
		\caption{Local minima along search directions}
		\label{fig_relu_z_fline_M}
	\end{subfigure}%
	\begin{subfigure}{.45\textwidth}
		\centering
		\includegraphics[width=0.9\linewidth]{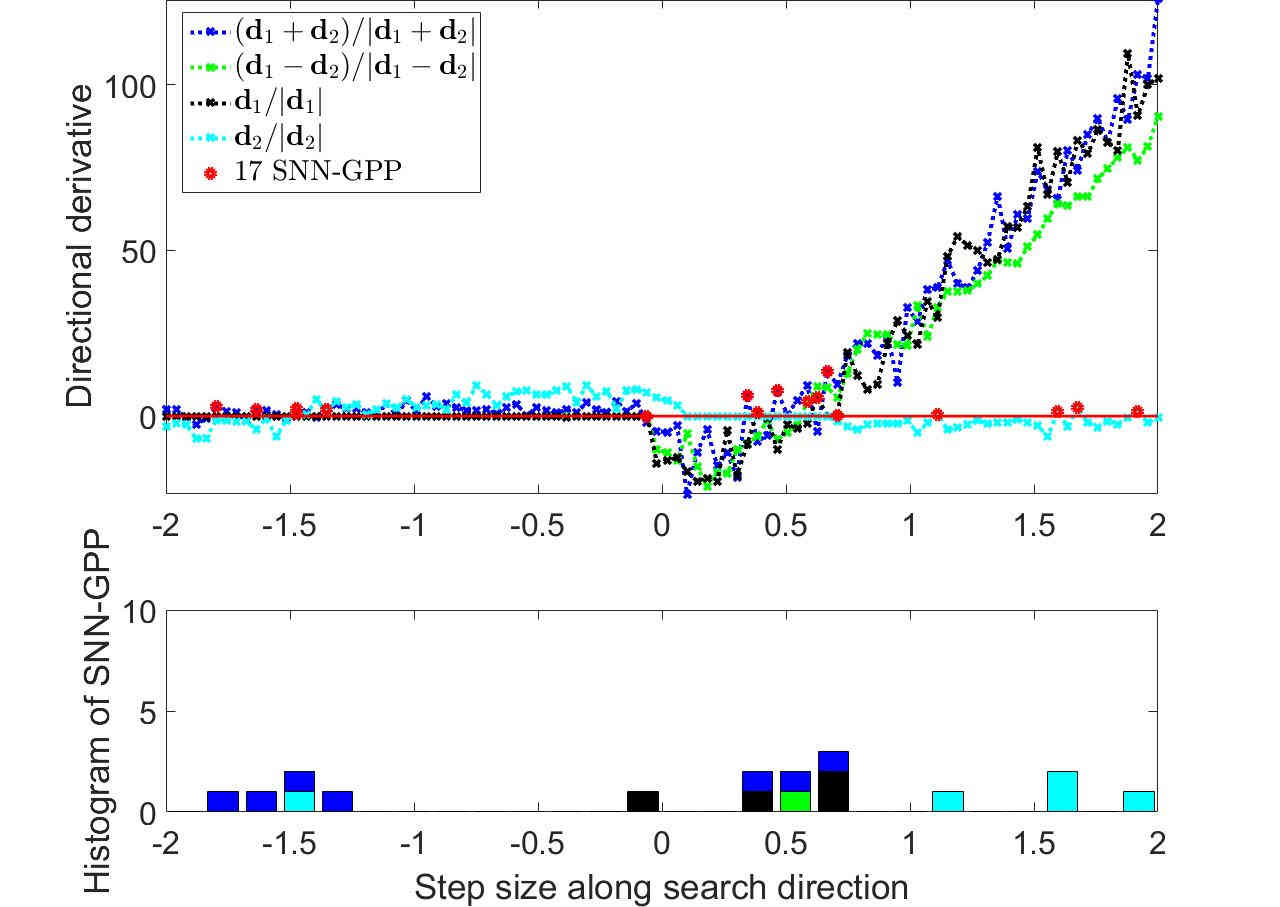}
		\caption{SNN-GPPs along search directions}
		\label{fig_relu_z_dline_M}
	\end{subfigure}%
	
	\caption{The ReLU AF close-up: Notable features are the stark changes in the directional derivatives. These correspond to the "activation" and "deactivation" of various nodes in the network. A unique feature to the ReLU activation is the presence of flat planes where the directional derivative is 0. These areas denote weight spaces where no information passes through the network, for all mini-batches.}
	\label{fig_z_relu}
\end{figure}

\begin{figure}[h!]
	\centering
	\begin{subfigure}{.45\textwidth}
		\centering 
		\includegraphics[width=0.9\linewidth]{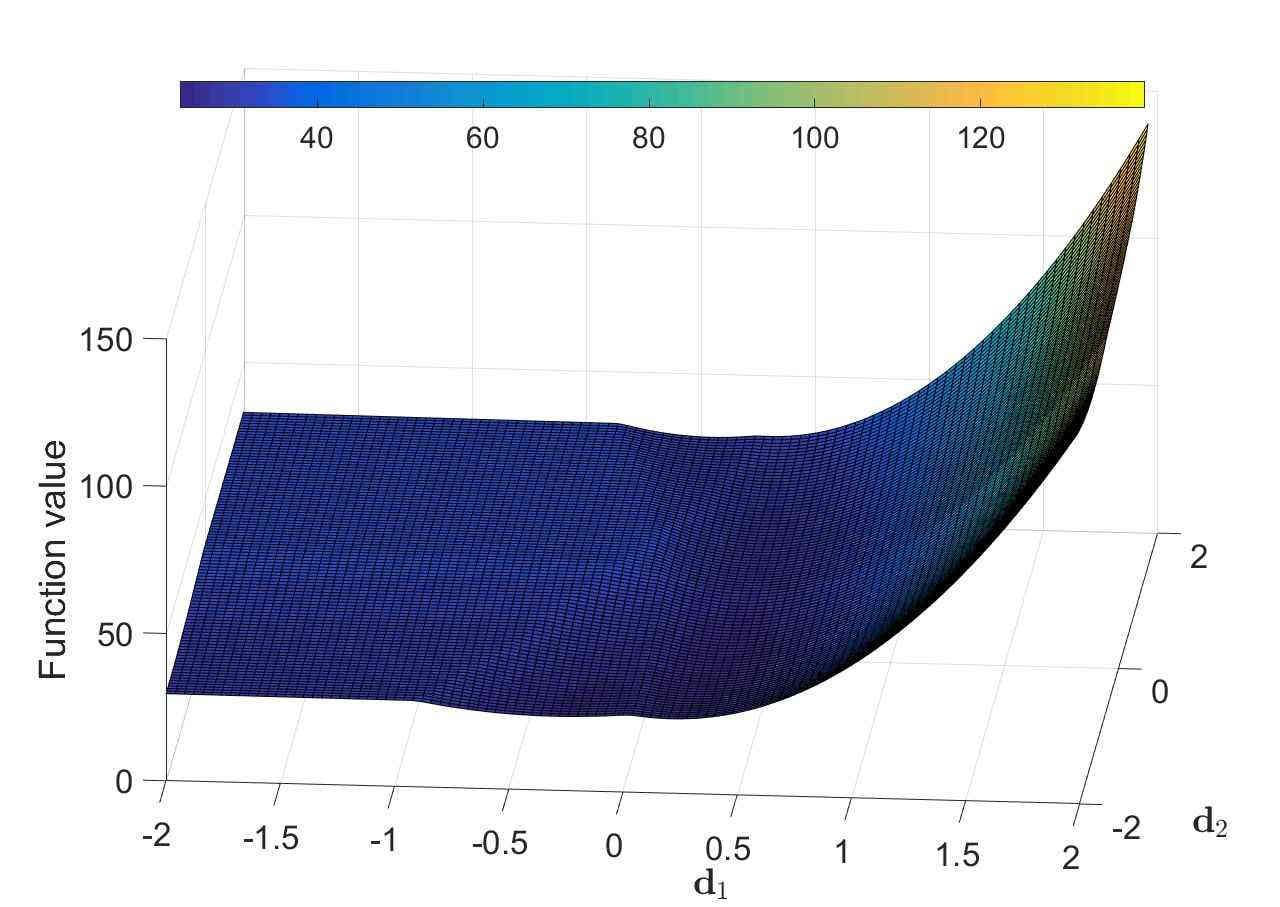}
		\caption{Function value, $M = 150$}
		\label{fig_lrelu_z_func_B}
	\end{subfigure}%
	\begin{subfigure}{.45\textwidth}
		\centering
		\includegraphics[width=0.9\linewidth]{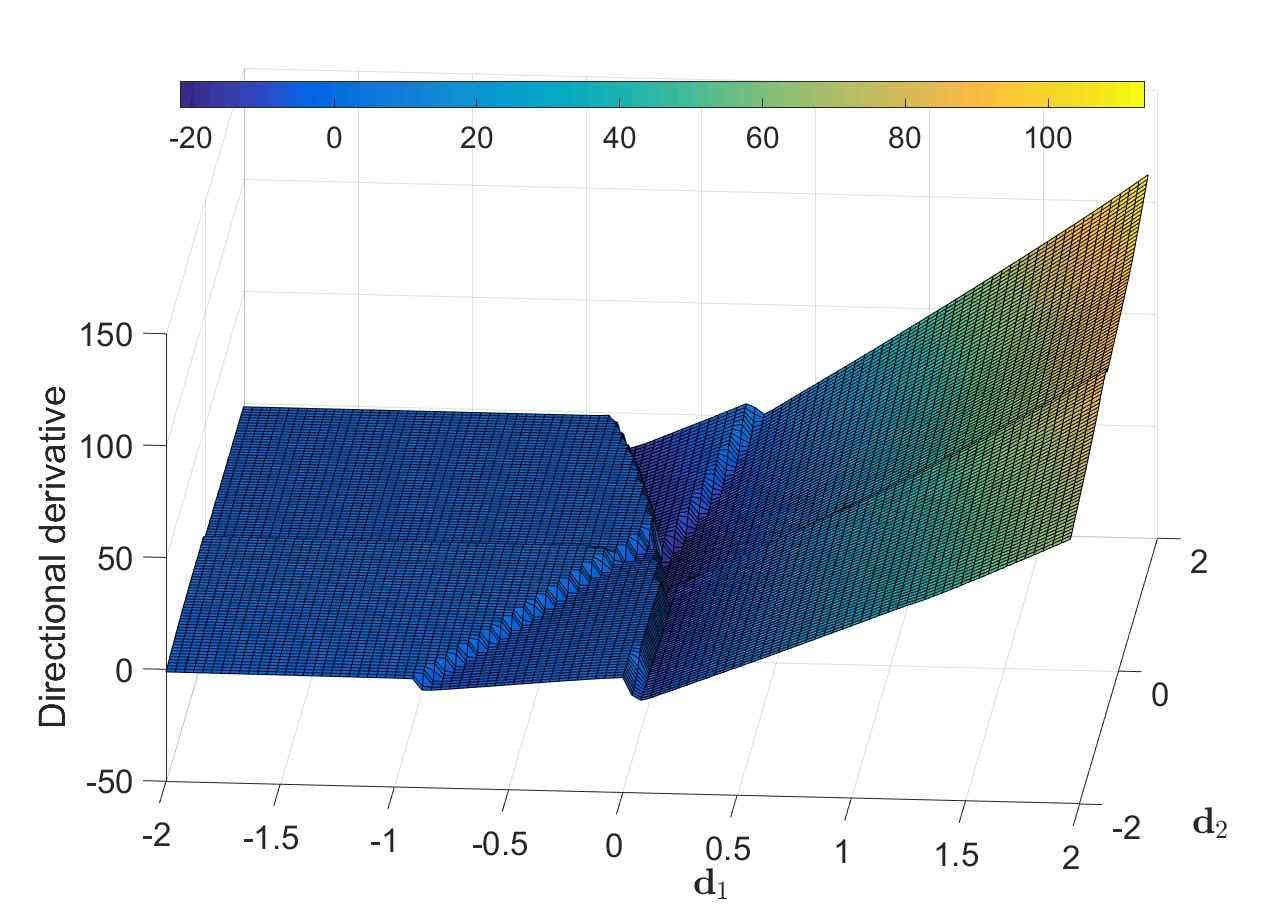}
		\caption{Directional derivative, $M = 150$}
		\label{fig_lrelu_z_dd_B}
	\end{subfigure}%
	
	\begin{subfigure}{.45\textwidth}
		\centering 
		\includegraphics[width=0.9\linewidth]{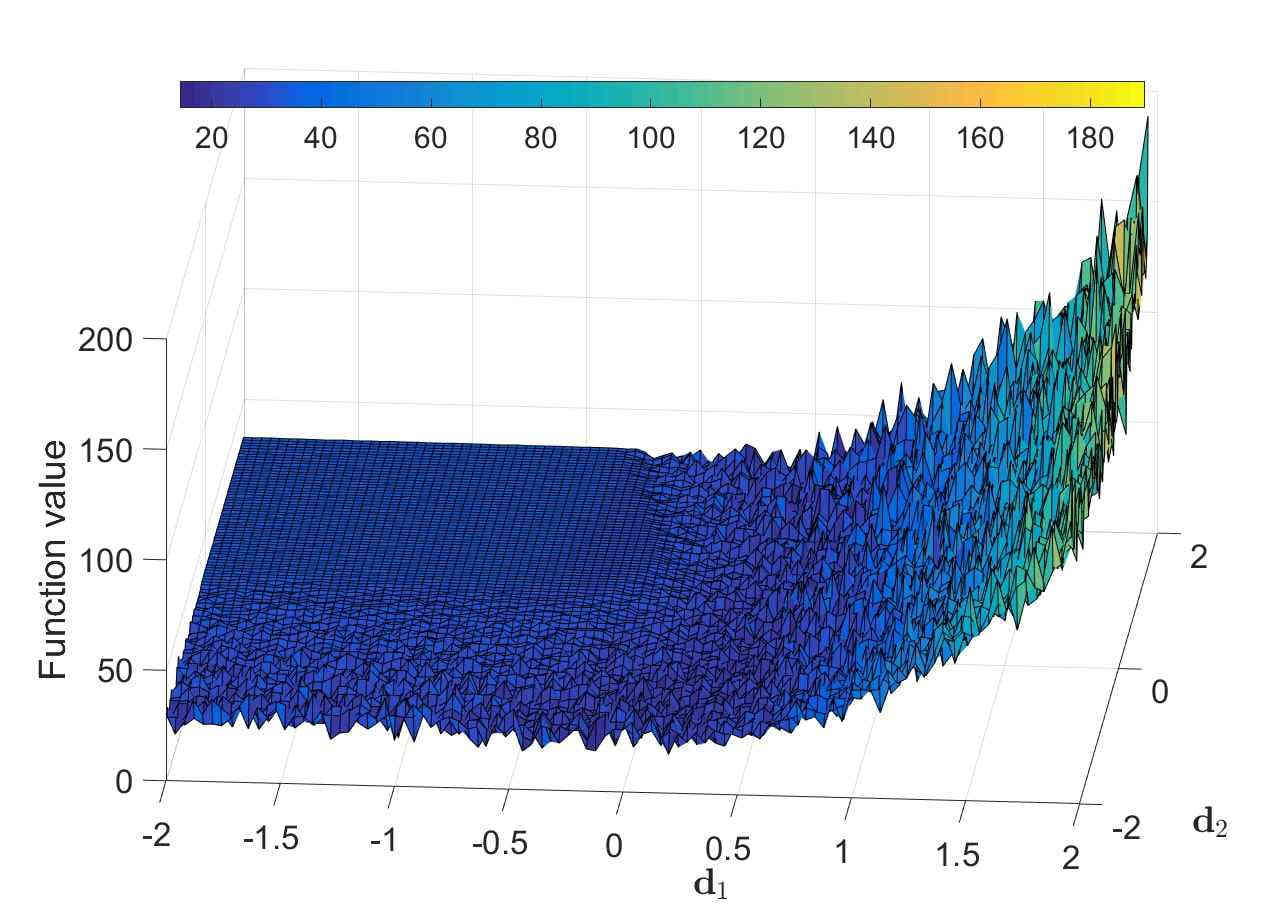}
		\caption{Function value, $|\mathcal{B}_{n,i}| = 10$}
		\label{fig_lrelu_z_func_M}
	\end{subfigure}%
	\begin{subfigure}{.45\textwidth}
		\centering
		\includegraphics[width=0.9\linewidth]{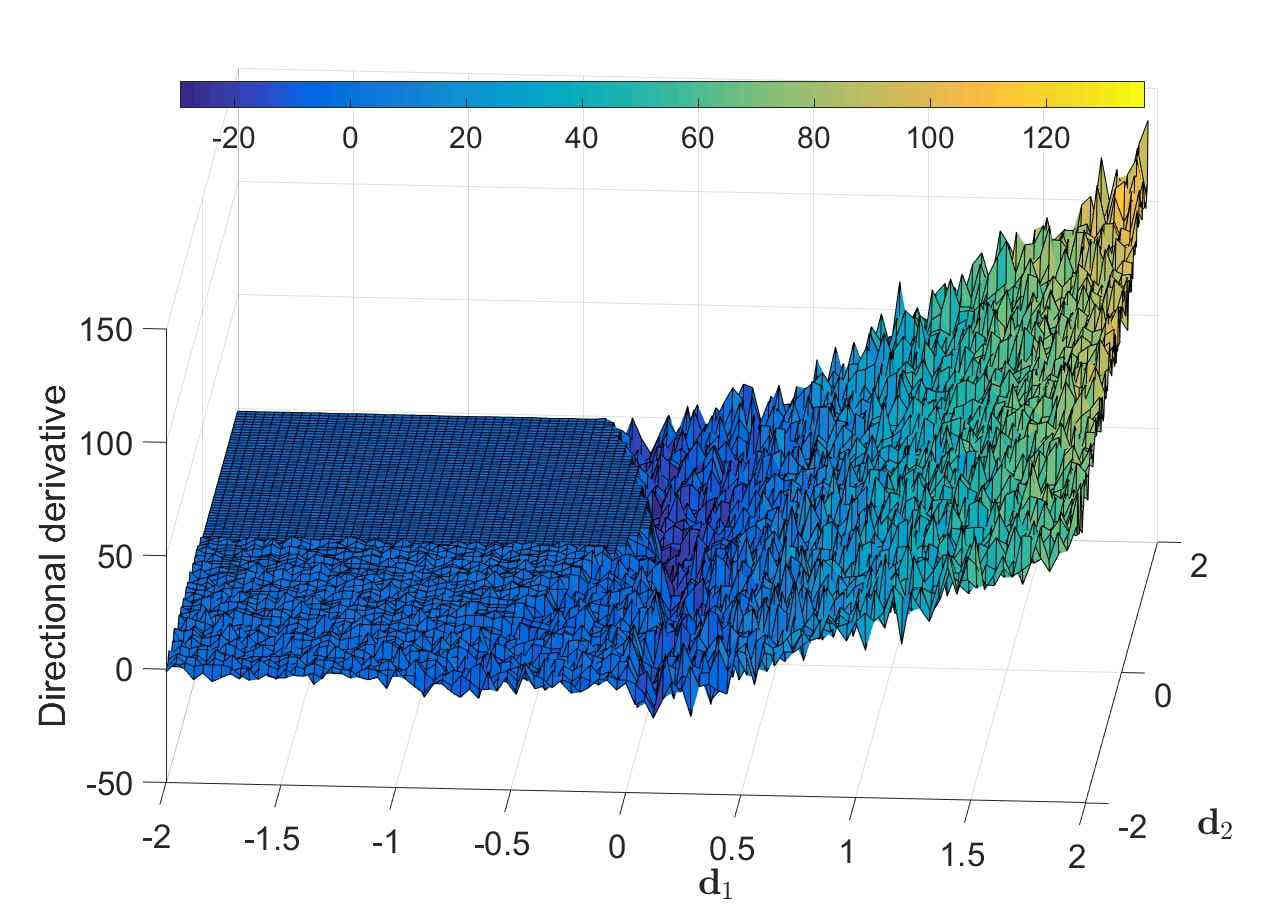}
		\caption{Directional derivative, $|\mathcal{B}_{n,i}| = 10$}
		\label{fig_lrelu_z_dd_M}
	\end{subfigure}%
	
	\begin{subfigure}{.45\textwidth}
		\centering 
		\includegraphics[width=0.9\linewidth]{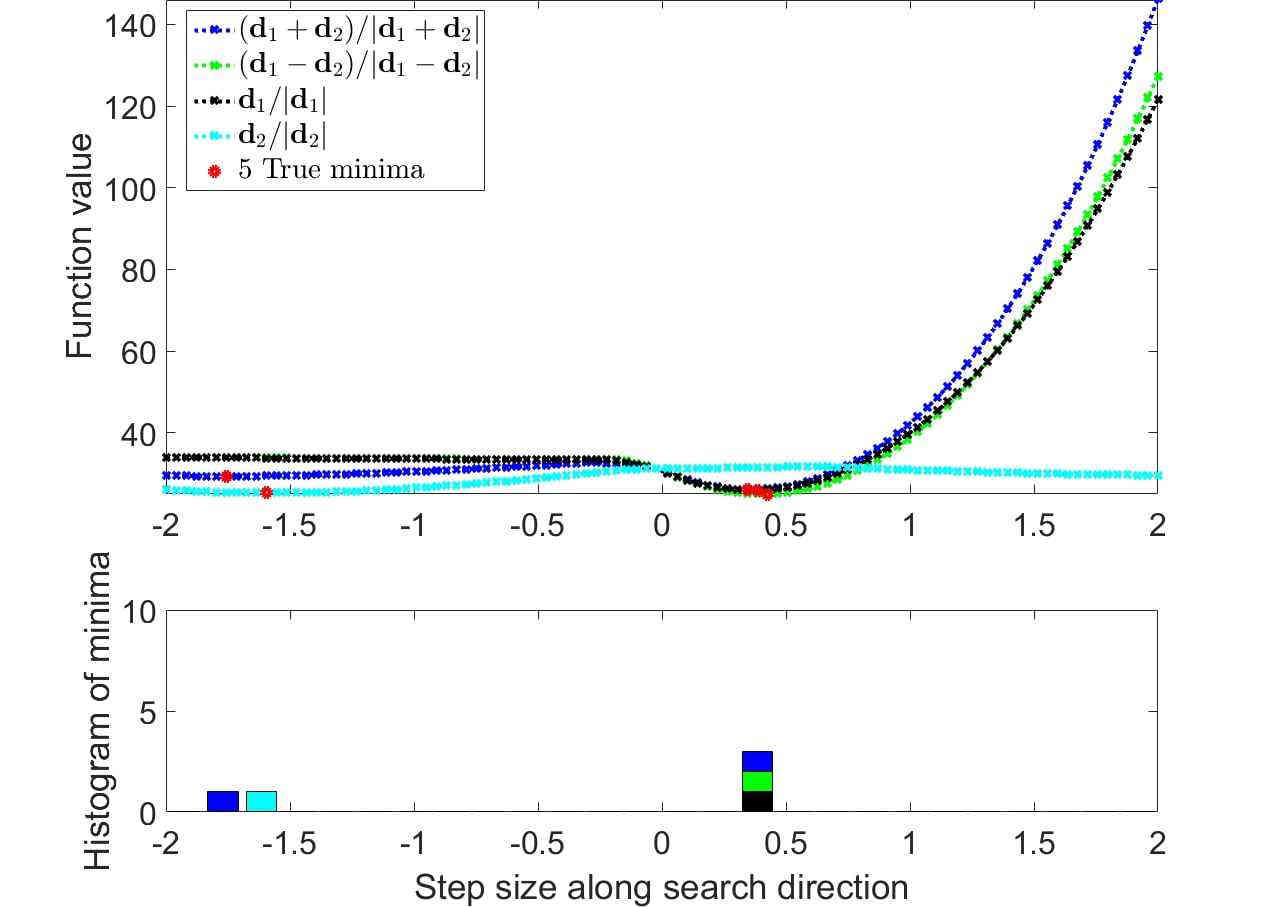}
		\caption{True minima along search directions}
		\label{fig_lrelu_z_fline_B}
	\end{subfigure}%
	\begin{subfigure}{.45\textwidth}
		\centering
		\includegraphics[width=0.9\linewidth]{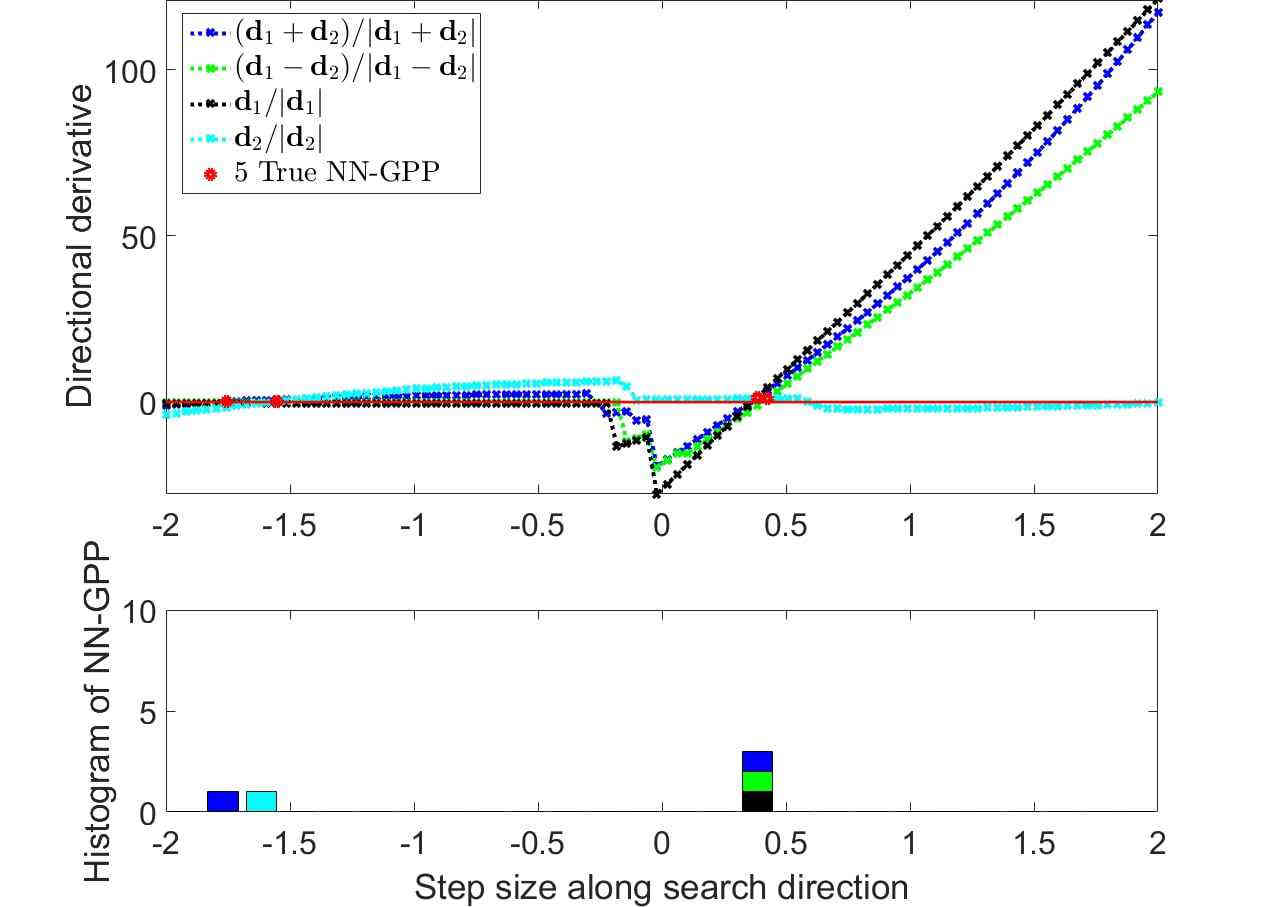}
		\caption{True NN-GPPs along search directions}
		\label{fig_lrelu_z_dline_B}
	\end{subfigure}%
	
	\begin{subfigure}{.45\textwidth}
		\centering 
		\includegraphics[width=0.9\linewidth]{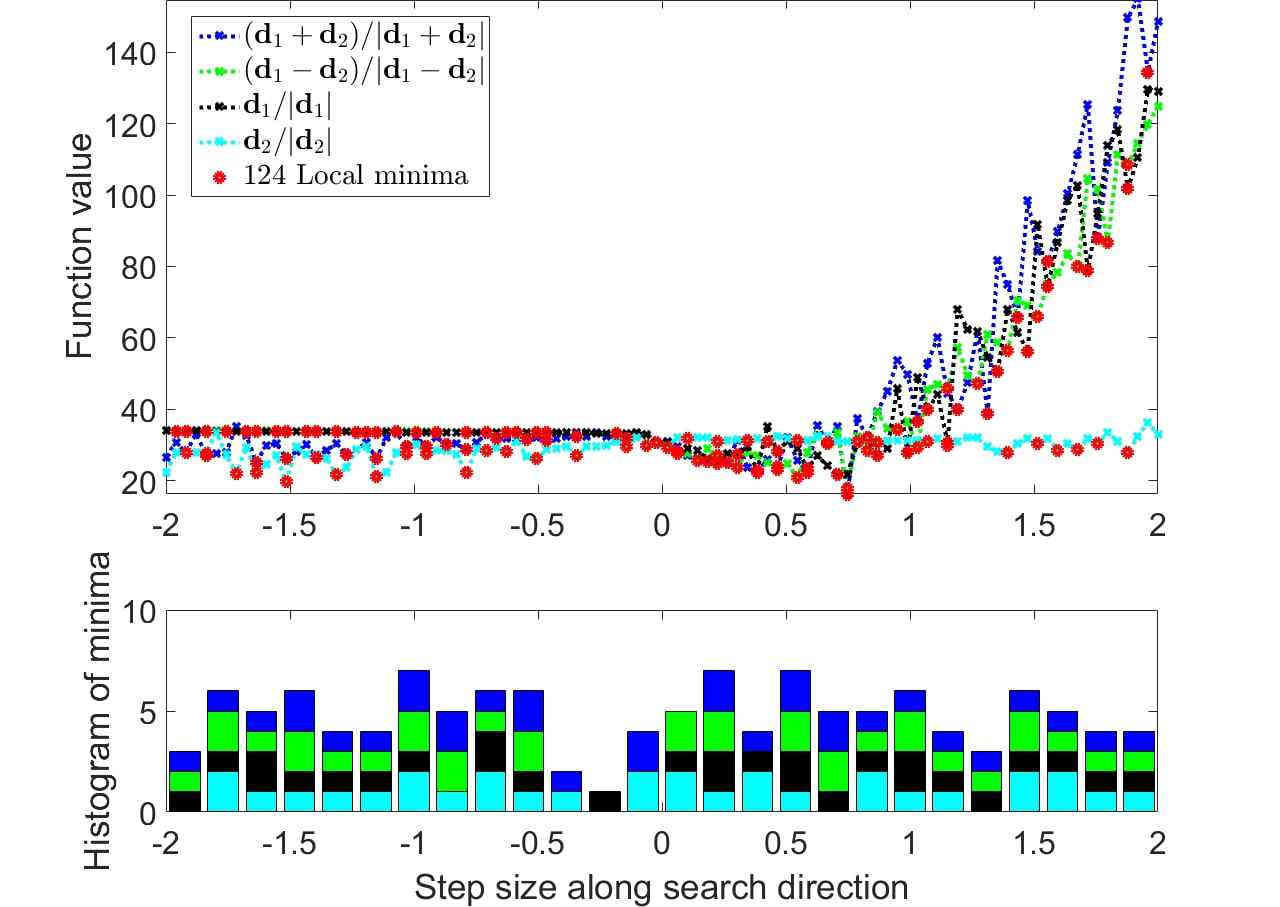}
		\caption{Local minima along search directions}
		\label{fig_lrelu_z_fline_M}
	\end{subfigure}%
	\begin{subfigure}{.45\textwidth}
		\centering
		\includegraphics[width=0.9\linewidth]{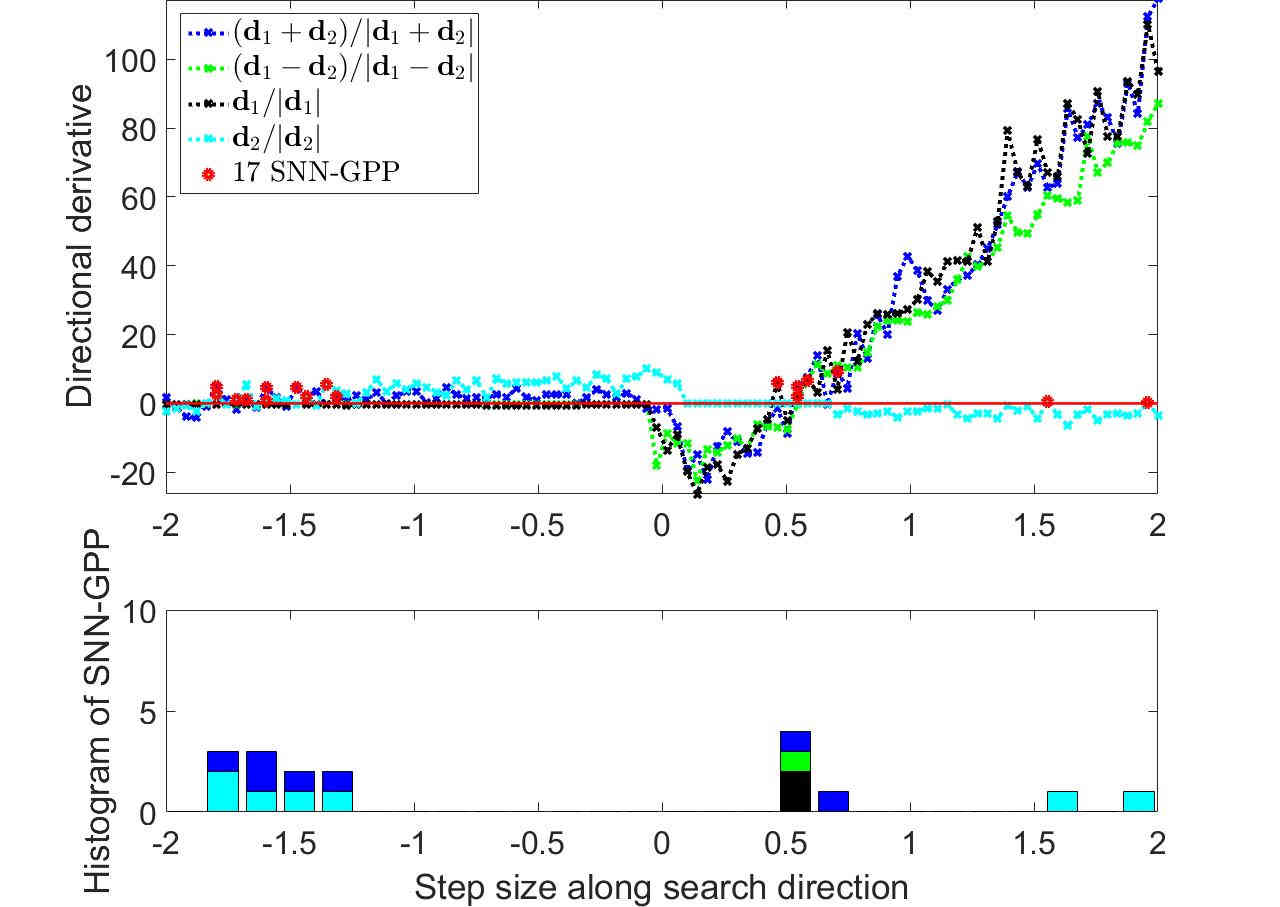}
		\caption{SNN-GPPs along search directions}
		\label{fig_z_lrelu_dline_M}
	\end{subfigure}%
	
	\caption{The leaky ReLU AF close-up: Flat planes with constant directional derivative values are also present here. Although in this case they have a non-zero numerical value, they do not contribute significantly to localizing SNN-GPPs or reducing the number of true optima for the ReLU class.}
	\label{fig_z_lrelu}
\end{figure}

\begin{figure}[h!]
	\centering
	\begin{subfigure}{.45\textwidth}
		\centering 
		\includegraphics[width=0.9\linewidth]{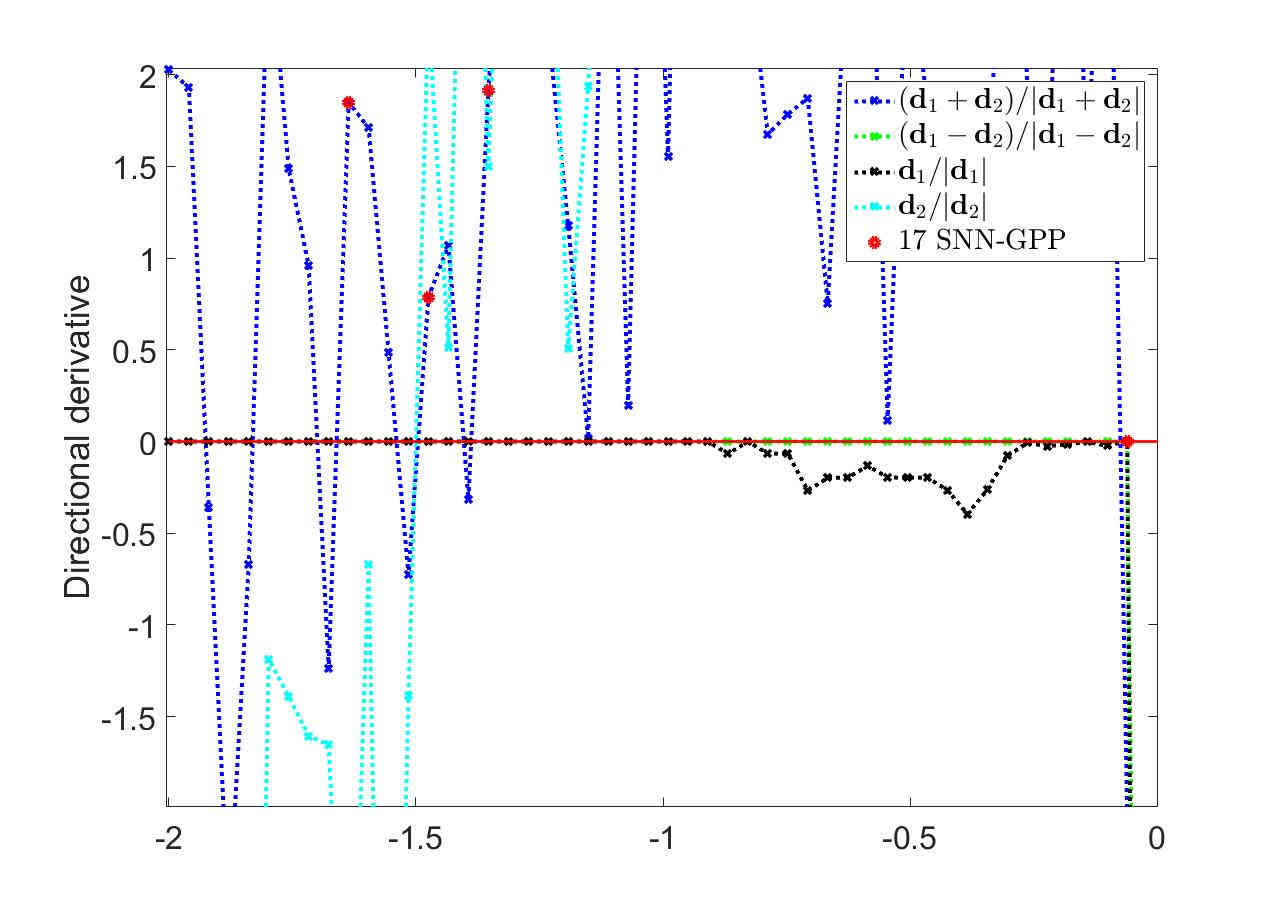}
		\caption{ReLU}
	\end{subfigure}%
	\begin{subfigure}{.45\textwidth}
		\centering
		\includegraphics[width=0.9\linewidth]{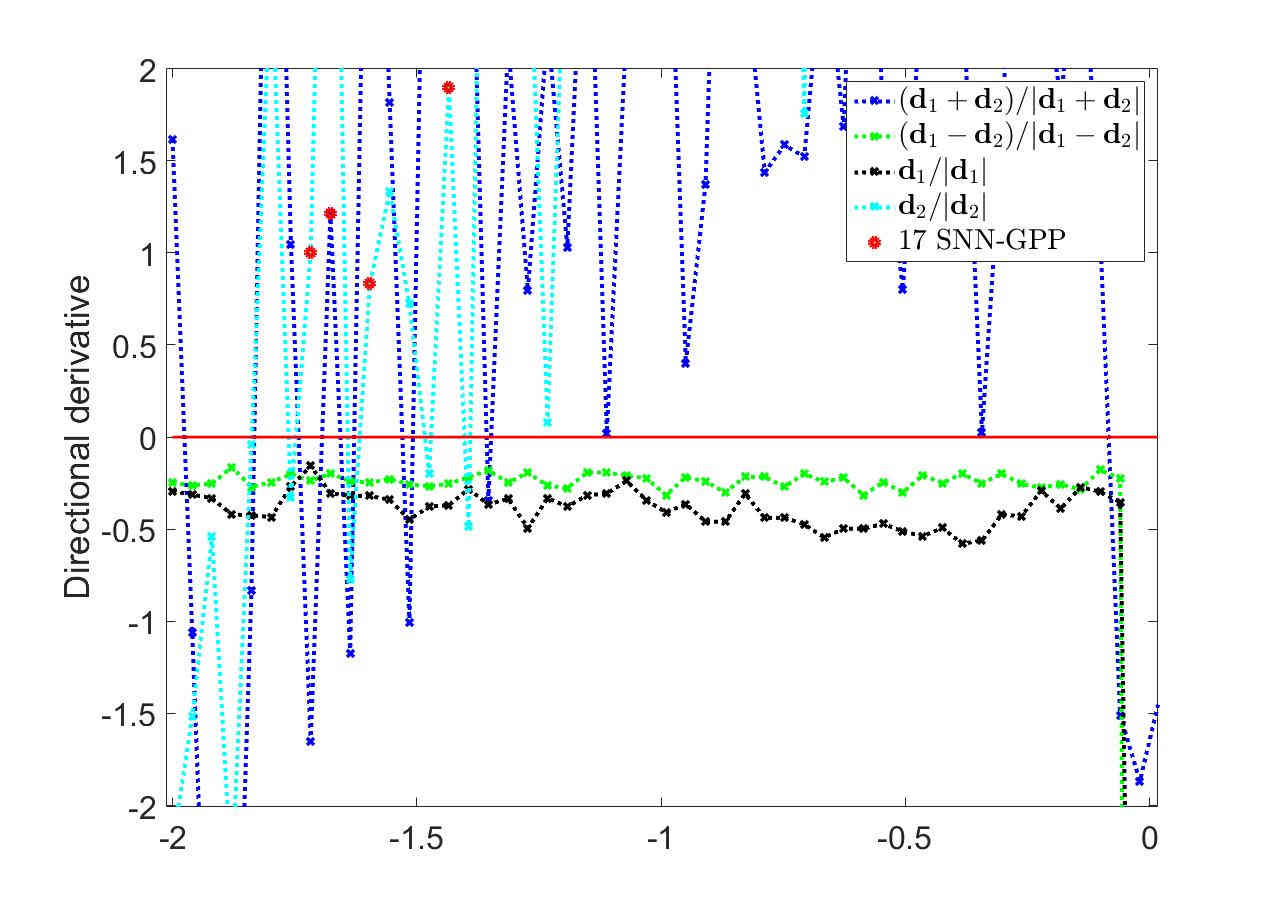}
		\caption{Leaky ReLU}
	\end{subfigure}%
	
	\caption{Detailed comparison of directional derivative plots between ReLU and leaky ReLU activations when hidden units become "inactive". As expected, ReLU units switch "off" entirely, containing no gradient information, while the leaky ReLU results in non-zero directional derivatives.}
	\label{fig_z_relu_0s}
\end{figure}

\begin{figure}[h!]
	\centering
	\begin{subfigure}{.45\textwidth}
		\centering 
		\includegraphics[width=0.9\linewidth]{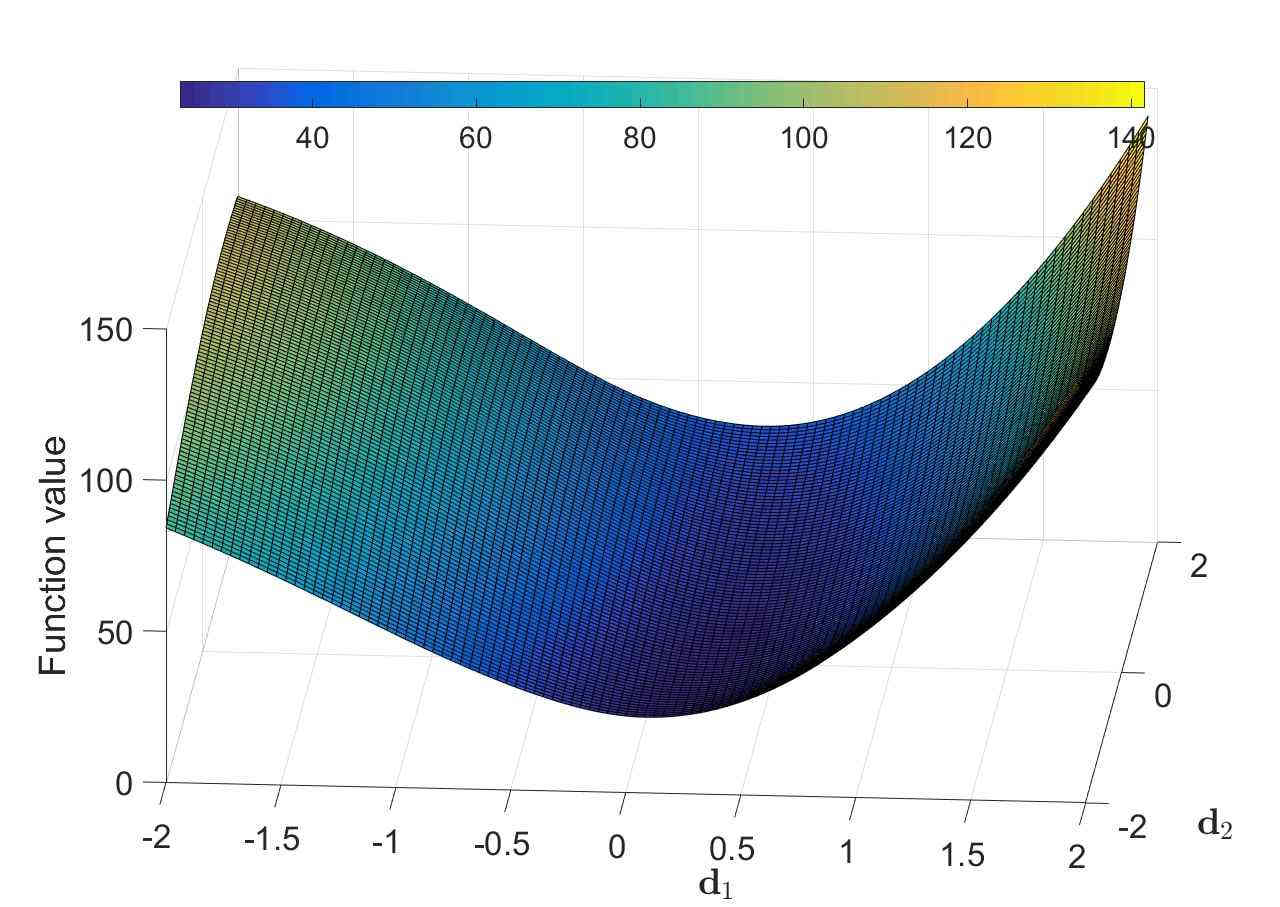}
		\caption{Function value, $M = 150$}
		\label{fig_z_elu_func_B}
	\end{subfigure}%
	\begin{subfigure}{.45\textwidth}
		\centering
		\includegraphics[width=0.9\linewidth]{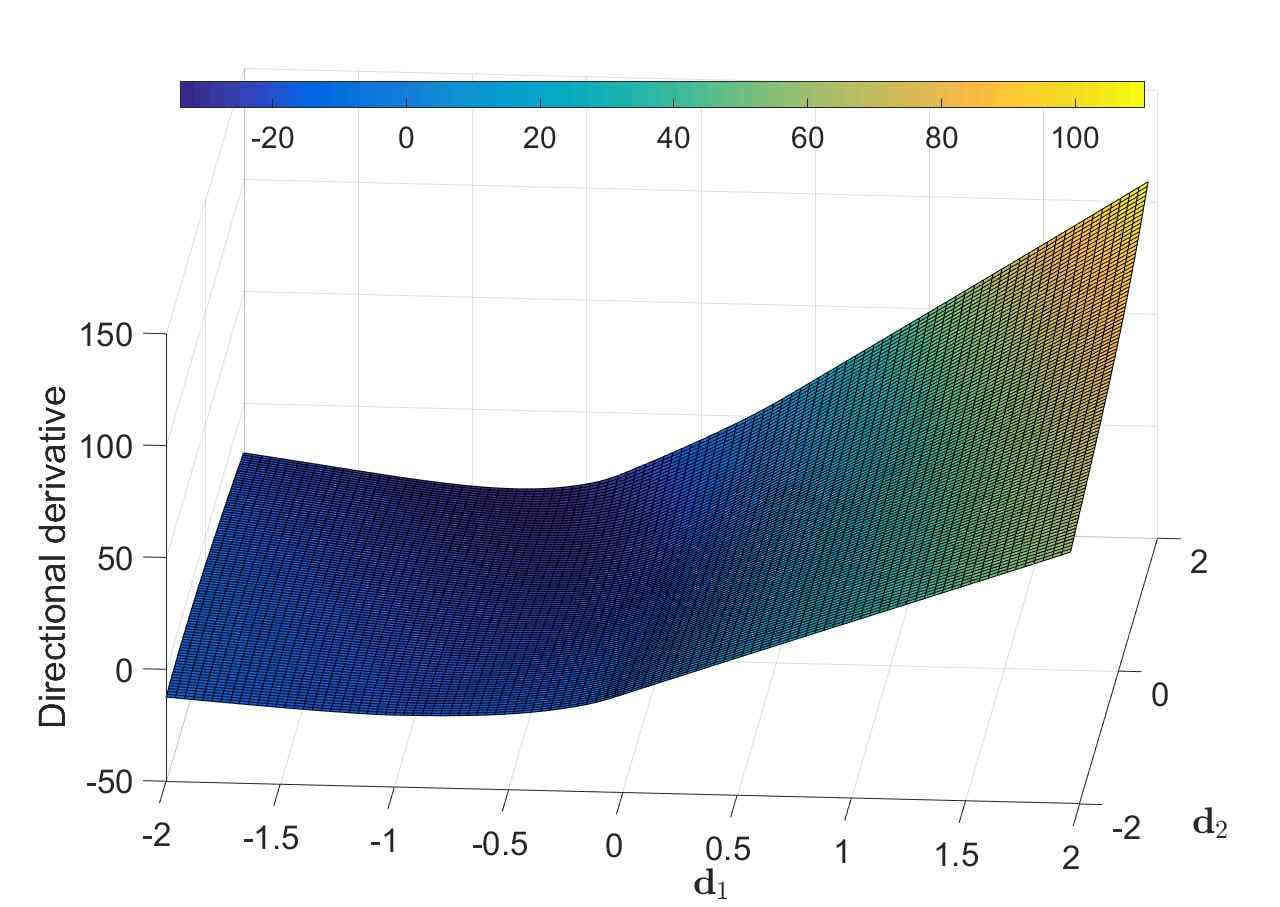}
		\caption{Directional derivative, $M = 150$}
		\label{fig_z_elu_dd_B}
	\end{subfigure}%
	
	\begin{subfigure}{.45\textwidth}
		\centering 
		\includegraphics[width=0.9\linewidth]{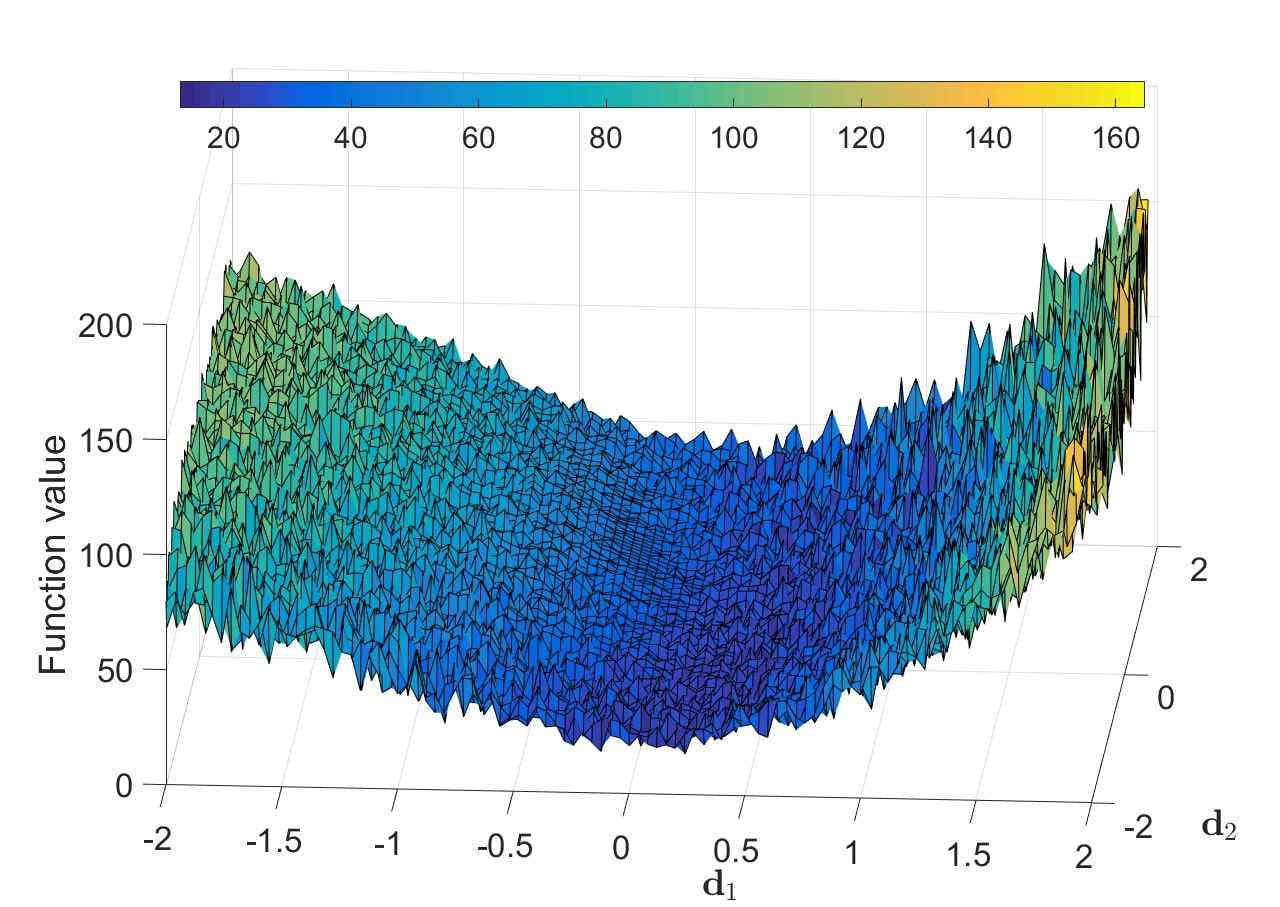}
		\caption{Function value, $|\mathcal{B}_{n,i}| = 10$}
		\label{fig_z_elu_func_M}
	\end{subfigure}%
	\begin{subfigure}{.45\textwidth}
		\centering
		\includegraphics[width=0.9\linewidth]{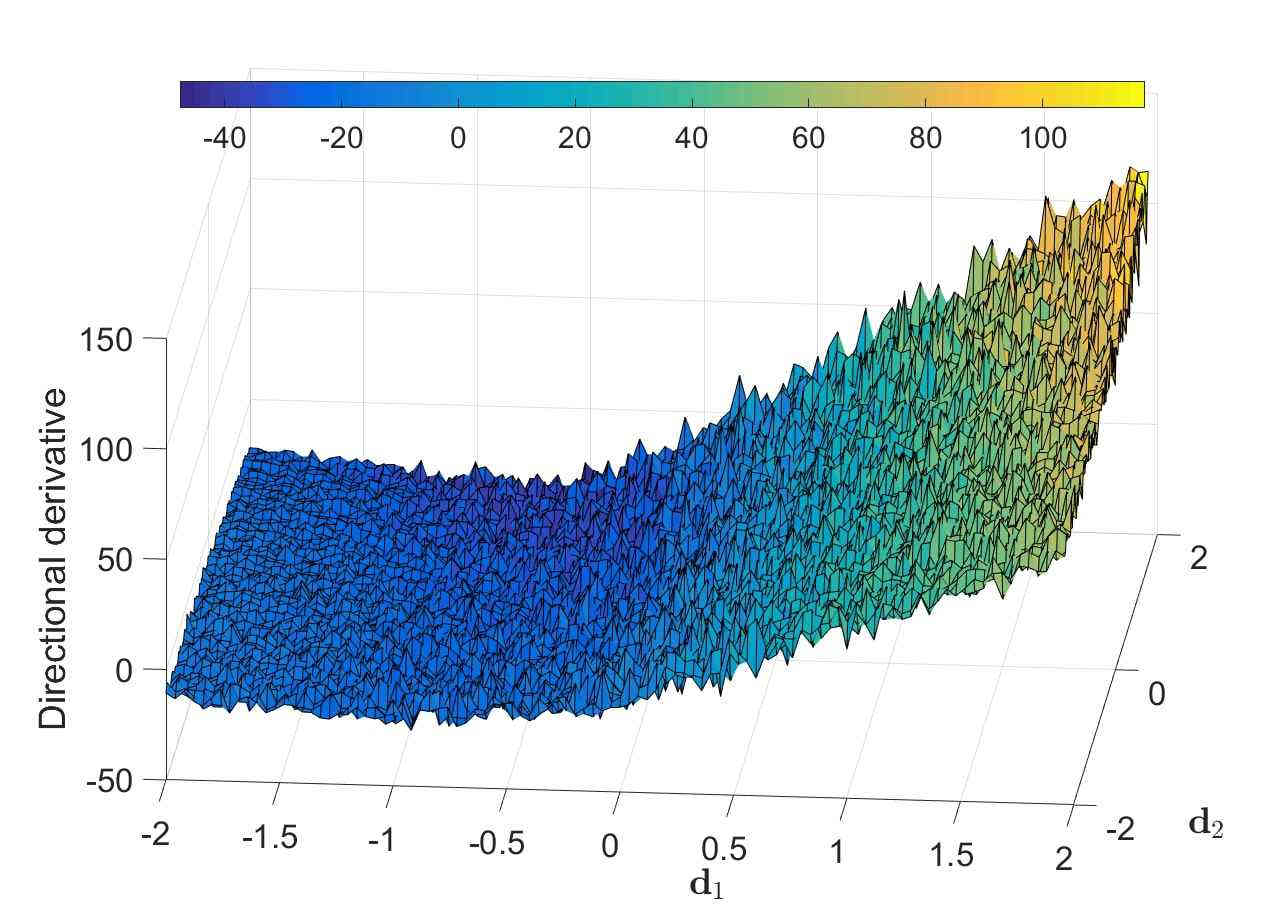}
		\caption{Directional derivative, $|\mathcal{B}_{n,i}| = 10$}
		\label{fig_z_elu_dd_M}
	\end{subfigure}%
	
	\begin{subfigure}{.45\textwidth}
		\centering 
		\includegraphics[width=0.9\linewidth]{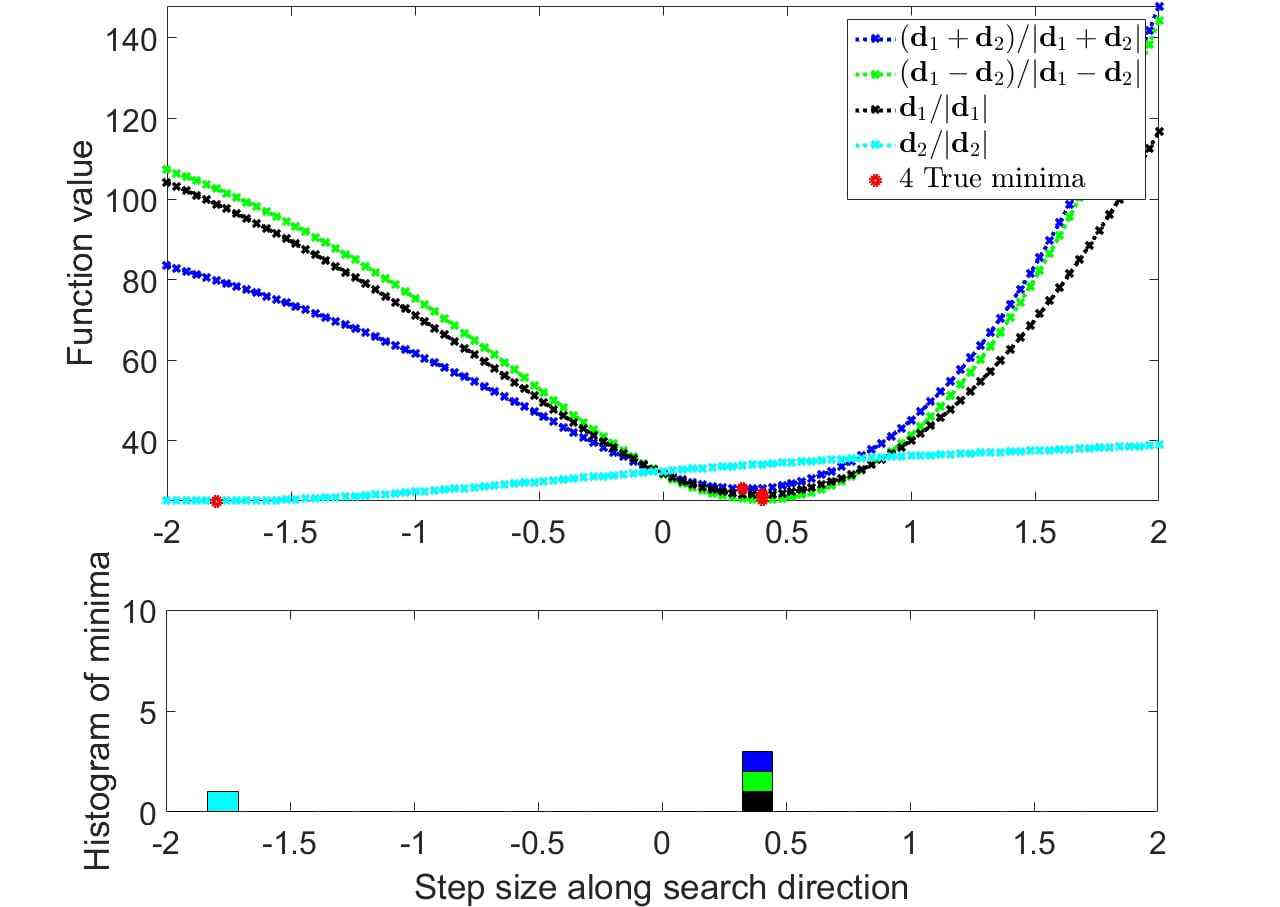}
		\caption{True minima along search directions}
		\label{fig_z_elu_fline_B}
	\end{subfigure}%
	\begin{subfigure}{.45\textwidth}
		\centering
		\includegraphics[width=0.9\linewidth]{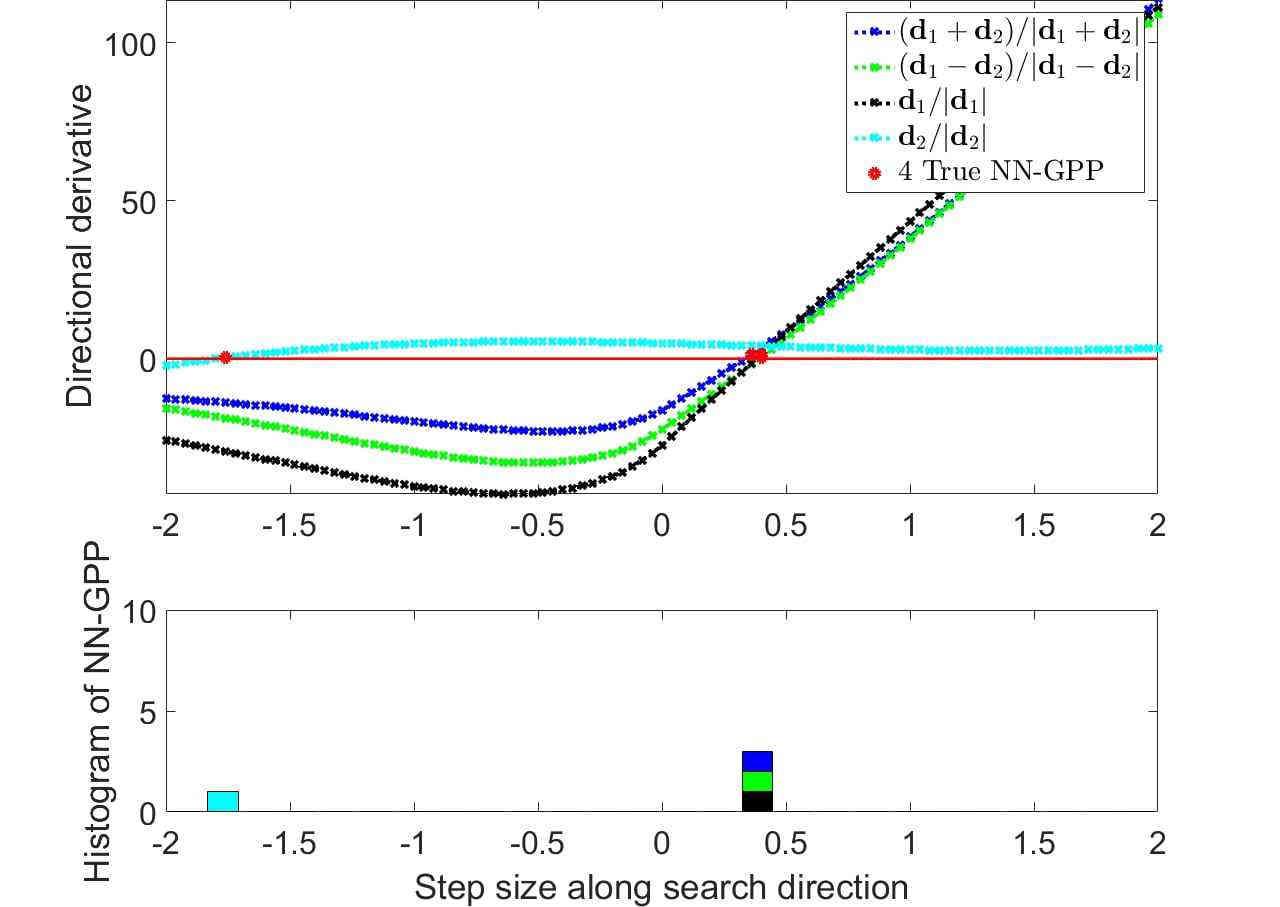}
		\caption{True NN-GPPs along search directions}
		\label{fig_z_elu_dline_B}
	\end{subfigure}%
	
	\begin{subfigure}{.45\textwidth}
		\centering 
		\includegraphics[width=0.9\linewidth]{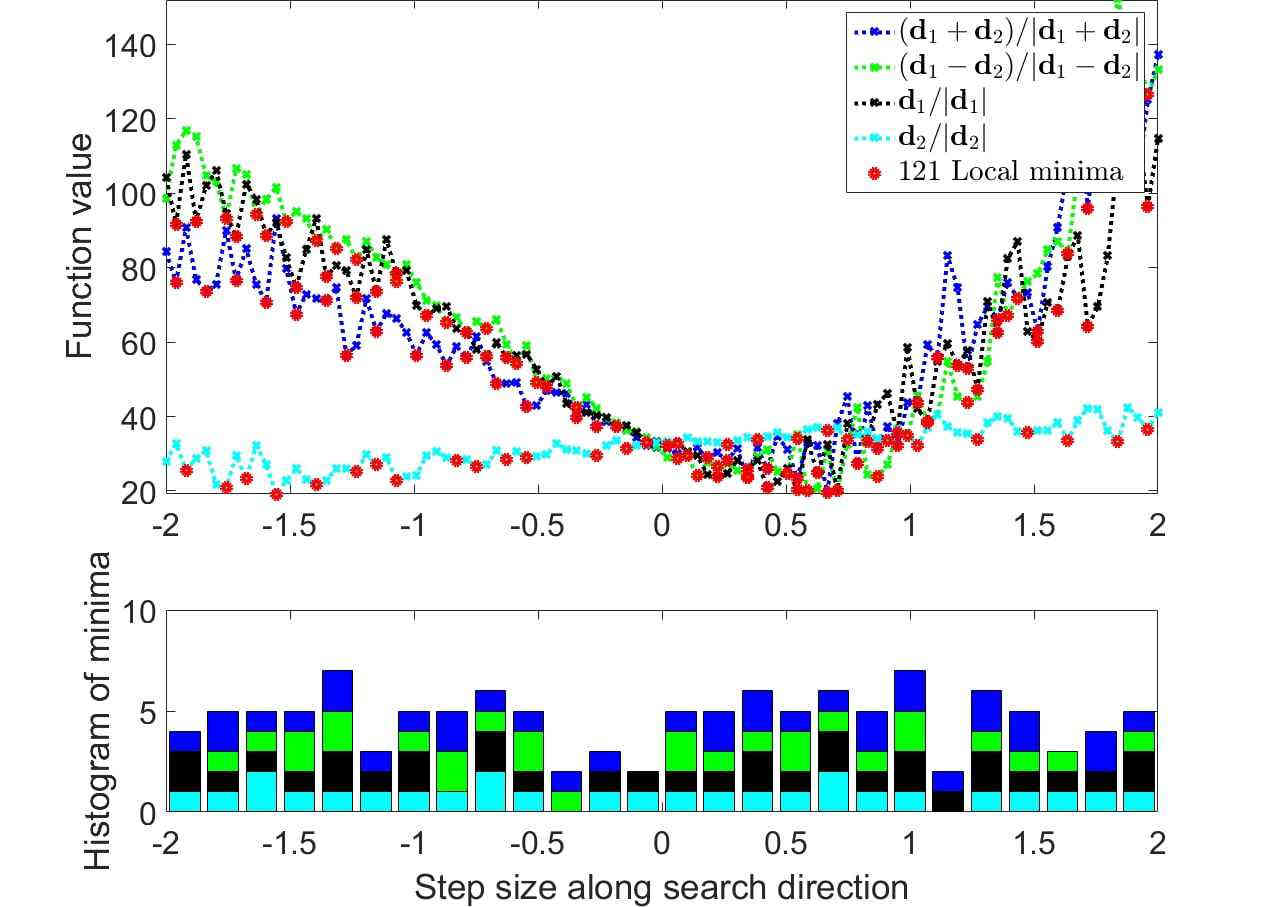}
		\caption{Local minima along search directions}
		\label{fig_z_elu_fline_M}
	\end{subfigure}%
	\begin{subfigure}{.45\textwidth}
		\centering
		\includegraphics[width=0.9\linewidth]{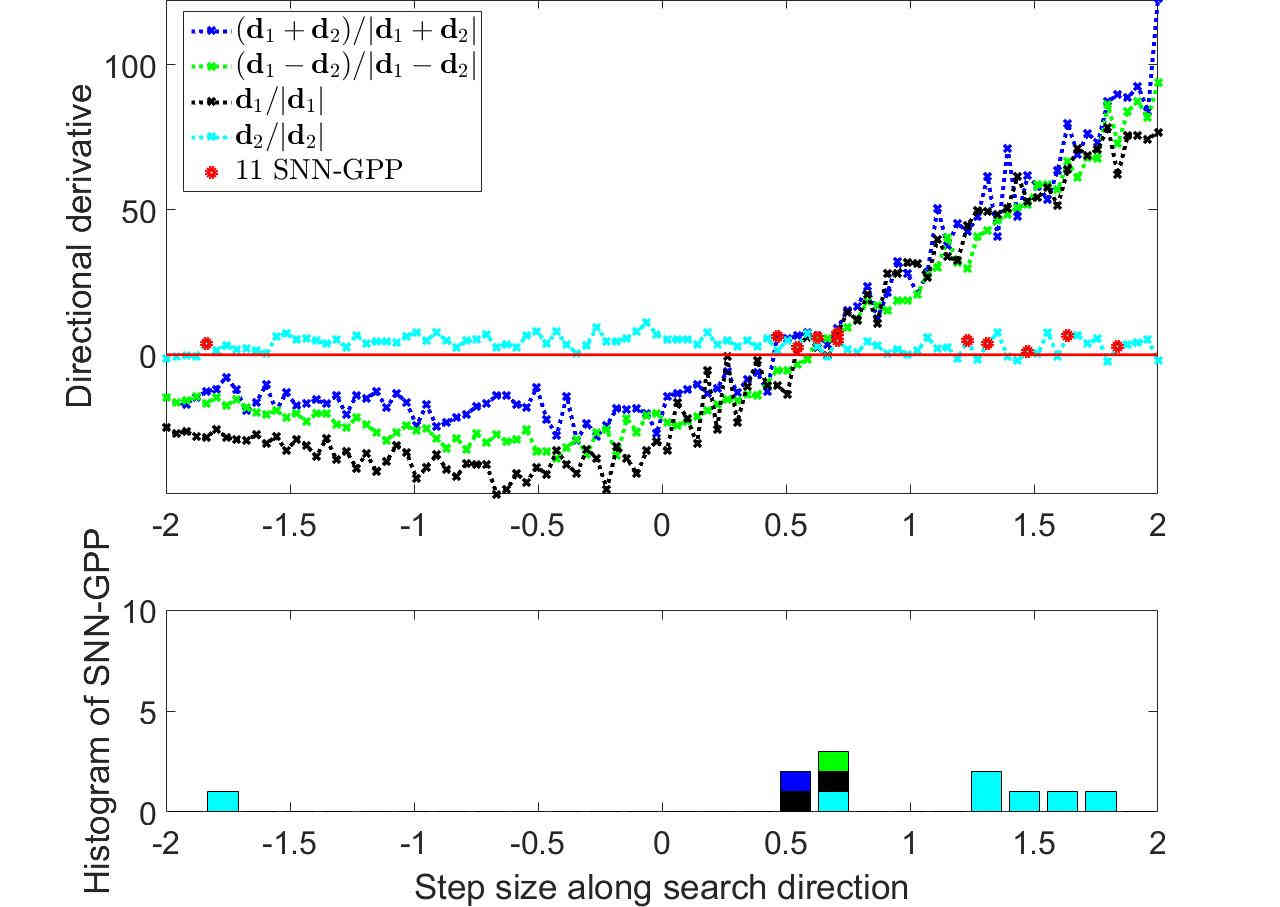}
		\caption{SNN-GPPs along search directions}
		\label{fig_z_elu_dline_M}
	\end{subfigure}%
	
	\caption{The ELU AF close-up: The smooth exponential derivative in the negative domain of the AF results in a greater magnitude in negative directional derivatives in the loss ((d) and (h)) compared to the remainder of the sparsity class. This in turn helps distance the directional derivatives from 0, benefiting localization of SNN-GPPs compared to leaky ReLU.}
	\label{fig_z_elu}
\end{figure}

\end{document}